\documentclass[letterpaper]{article} 
\usepackage{aaai25}  
\usepackage{times}  
\usepackage{helvet}  
\usepackage{courier}  
\usepackage[hyphens]{url}  
\usepackage{graphicx} 
\usepackage{natbib}  
\usepackage{caption} 
\usepackage{algorithm}
\usepackage{algorithmic}
\usepackage{amssymb}
\usepackage{amsmath}
\usepackage{amsthm}
\usepackage{centernot}
\newtheorem{definition}{Definition}
\newtheorem{proposition}{Proposition}
\newtheorem{theorem}{Theorem}
\usepackage{multirow}
\usepackage{subfig}
\usepackage{booktabs}
\usepackage{newfloat}
\usepackage{listings}
\DeclareCaptionStyle{ruled}{labelfont=normalfont,labelsep=colon,strut=off} 
\floatstyle{ruled}
\newfloat{listing}{tb}{lst}{}
\floatname{listing}{Listing}
\title{Towards Better Spherical Sliced-Wasserstein Distance Learning with Data-Adaptive Discriminative Projection Direction}
\author {
    Hongliang Zhang\textsuperscript{\rm 1},
    Shuo Chen\textsuperscript{\rm 2},
    Lei Luo\textsuperscript{\rm 1}\thanks{corresponding authors},
    Jian Yang\textsuperscript{\rm 1}\footnotemark[1]
}
\affiliations {
    \textsuperscript{\rm 1}PCA Lab, Key Lab of Intelligent Perception and Systems for High-Dimensional Information of Ministry of Education, School of Computer Science and Engineering, Nanjing University of Science and Technology, China\\
    \textsuperscript{\rm 2}School of Intelligence Science and Technology, Nanjing University, China\\
    \{zhang1hongliang, cslluo, csjyang\}@njust.edu.cn, shuo.chen@nju.edu.cn
}

\usepackage{bibentry}

\begin{document}

\maketitle

\begin{abstract}
Spherical Sliced-Wasserstein (SSW) has recently been proposed to measure the discrepancy between spherical data distributions in various fields, such as geology, medical domains, computer vision, and deep representation learning. However, in the original SSW, all projection directions are treated equally, which is too idealistic and cannot accurately reflect the importance of different projection directions for various data distributions. To address this issue, we propose a novel data-adaptive Discriminative Spherical Sliced-Wasserstein (DSSW) distance, which utilizes a projected energy function to determine the discriminative projection direction for SSW. In our new DSSW, we introduce two types of projected energy functions to generate the weights for projection directions with complete theoretical guarantees. The first type employs a non-parametric deterministic function that transforms the projected Wasserstein distance into its corresponding weight in each projection direction. This improves the performance of the original SSW distance with negligible additional computational overhead. The second type utilizes a neural network-induced function that learns the projection direction weight through a parameterized neural network based on data projections. This further enhances the performance of the original SSW distance with less extra computational overhead. Finally, we evaluate the performance of our proposed DSSW by comparing it with several state-of-the-art methods across a variety of machine learning tasks, including gradient flows, density estimation on real earth data, and self-supervised learning.
\end{abstract}

%

\section{Introduction}
In real-world scenarios, more and more tasks involve defining data distributions on the hypersphere, highlighting the importance and universality of spherical geometry. These tasks include characterizing the density distribution of geophysical \cite{di2014nonparametric, anSHaRPoseSparseHighResolution2024, hu2023transformer, hu2024towards} or meteorology data \cite{besombes2021producing}, magnetic imaging of the brain in the medical field \cite{vrba2001signal}, texture mapping in computer graphics \cite{dominitz2009texture}, etc, where the latent representation is mapped to a bounded space commonly known as a sphere \cite{wang2020understanding, pmlr-v119-chen20j}. 

The distribution analysis on the hypersphere is often focused on the statistical study of directions, orientations, and rotations. It is known as circle or sphere statistical analysis \cite{jammalamadaka2001topics}. Recently, there has been a growing interest in comparing probability measures on the hypersphere using Optimal Transport (OT) \cite{cui2019spherical}. This is driven by its appealing statistical, geometrical, and topological properties \cite{peyre2019computational}.

Two critical challenges in applying OT theory are the high computational complexity and the curse of dimensionality \cite{MAL-073}. These issues have led to a growing focus on developing faster solving tools \cite{NIPS2013_af21d0c9} and computationally efficient alternative distance metrics \cite{NEURIPS2019_f0935e4c}. One such metric is the Sliced-Wasserstein (SW) distance, which has lower computational complexity and does not suffer from the curse of dimensionality  \cite{NEURIPS2022_b4bc180b}. This distance metric inherits similar topological properties from the Wasserstein distance \cite{NEURIPS2020_eefc9e10} and is widely used as an alternative solution for comparing probability measures. The SW distance is defined as the expectation of the one-dimensional Wasserstein distance between two projected measures over a uniform distribution on the unit sphere. However, due to the intractability of this expectation, Monte Carlo estimation is often used to approximate the SW distance. Some SW variants aim to learn a discriminative or optimal projections distribution from training data, PAC-SW \cite{pmlr-v202-ohana23a} learns a discriminative projection direction distribution from training data based on the PAC-Bayesian theory. DSW \cite{nguyen2021distributional} learns optimal projections distribution from training data by the neural network.

Recently, SW distance has been employed to compare probability measures on the hypersphere due to its computational efficiency and simplicity of implementation. This has led to the development of spherical sliced OT approaches \cite{bonet2023spherical, Quellmalz_2023, tran2024stereographic}. The main challenge in developing these methods is extending the classical Radon transform to its spherical counterparts. Quellmalz et al. \cite{Quellmalz_2023} introduced two spherical extensions for the Radon transform to define sliced OT on the sphere: the vertical slice transform \cite{rubin2018verticalslicetransformspherical}, and the normalized semicircle transform \cite{groemer1998spherical}. Bonet et al. \cite{bonet2023spherical} proposed a new spherical Radon transform and leveraged the closed-form solution of the Wasserstein distance on the circle to define SSW for empirical probability measures. Both works \cite{bonet2023spherical, Quellmalz_2023} map a distribution defined on a hypersphere to its marginal distributions on a unit circle, and solve circular OT to compare these marginals. However, calculating the OT between two one-dimensional measures defined on a circle is more expensive. To address this issue, Tran et al. \cite{tran2024stereographic} first adopted the stereographic projection to transform the hypersphere into a hyperplane, and then utilized the classic Radon transform to define Stereographic Spherical Sliced-Wasserstein (S3W).

These existing works on SSW usually assume that all projection directions contribute equally when calculating the expectation of one-dimensional Wasserstein distance. However, this assumption is too idealistic and inconsistent with real-world situations, as this practice ignores the discriminative information from different projection directions in SSW. This paper aims to learn the better SSW distance to handle practical applications by considering the discrimination between projection directions. Specifically, we emphasize that different projection directions in SSW have varying degrees of importance and propose using projected weights adaptively learned from data to characterize them, where the weight is directly proportional to the Wasserstein distance of the corresponding projection direction. As weights can reflect the distribution of the data, this approach effectively captures valuable discriminative information hidden in various directions. This is beneficial in improving the accuracy of SSW distance. Towards this end, we propose two types of the projected energy function (\textit{i.e.}, non-parametric and parametric forms) to learn the weights of projection directions under the projection of supports, taking into account the efficiency and performance. In the non-parametric form, the weight is calculated from the projected Wasserstein distance using a non-parametric function, such as softmax, identity, or polynomial function. In the parametric form, a parametric neural network such as linear, nonlinear, or attention mechanisms is used to generate the weight based on the input of the projected supports. Our new method effectively characterizes the importance of each projection direction, allowing for a more precise computation of the discrepancy between real-world distributions. Our contributions are as follows:
\begin{itemize}
    \item We first propose learning discriminative projection direction for SSW distance which is implemented by the non-parametric function and the parametric neural network to consider specific data distributions.
    \item We provide the corresponding theoretical analysis and mathematical derivation to guarantee the topological and statistical properties of the novel DSSW distance.
    \item We apply our DSSW distance to several classical machine learning tasks, and the experimental results show that our DSSW is superior to the existing SSW, S3W, SW and Wasserstein distance.
\end{itemize}

\section{Background}
The goal of this work is to define a DSSW distance on the hypersphere $S^{d-1}=\left \{ x \mid x\in \mathbb{R}^{d}, \ \left \| x \right \| _{2} = 1 \right \} $. It is necessary to first review the definition of Wasserstein distance on manifolds: the SW distance on $\mathbb{R}^{d}$ and the SSW distance on the hypersphere $S^{d-1}$.

\textbf{Wasserstein Distance.} Let \emph{M} be a Riemannian manifold equipped with the distance $d\left ( \cdot ,\cdot \right ):M\times M\to \mathbb{R} _{+} $. For $1\le p< +\infty $, let two probability measures $\mu$ and $\nu\in \mathcal{P}_p(M)=\{\mu \mid \mu \in \mathcal{P}(M), \ \int_M d^p(x,x_0) \, \mathrm{d}\mu(x)<+\infty~\text{for any}~ x_0\in M\}$ be defined on manifold \emph{M} with \emph{p} finite moments, and $\mathcal{P}(M)$ means the set of all probability measures defined on $M$. The aim of OT is to transport the mass from $\mu$ to $\nu$ in a way that minimizing the expectation of transport distance. So the \emph{p}-Wasserstein distance \cite{MAL-073} can be defined as
\begin{align}
    W_p^p(\mu,\nu)=\inf_{\gamma\in \Pi(\mu,\nu)} \int_{M\times M} d\left ( x, y \right ) ^{p} \mathrm{d}\gamma(x,y),
    \label{eq:ot}
\end{align}
where $\Pi(\mu,\nu)$ is the set of couplings of $\mu$ and $\nu$.

Unfortunately, for discrete probability measures with \emph{n} samples, the Wasserstein distance can be calculated by linear programs with the computational complexity of $\mathcal{O} \left ( n^{3}\log{n} \right )$, so it is computationally expensive. Therefore, the alternative distance metrics with lower computational complexity are explored in Euclidean spaces. One of the widely adopted alternative distances is the SW distance.

\textbf{Sliced-Wasserstein Distance.} For one dimensional measures $\mu$, $\nu\in \mathcal{P}_p(\mathbb{R}^{d})$, the Wasserstein distance between $\mu$ and $\nu$ has the closed form as
\begin{align}
    W_p^p(\mu,\nu)=\int_{0}^1 \left | F_{\mu}^{-1}(t)-F_{\nu}^{-1}(t) \right |^p \mathrm{d}t,
\end{align}
where $F^{-1}_{\mu}$ and $F^{-1}_{\nu}$ are the quantile functions of $\mu$ and $\nu$. This property can be used to define the $p$-SW distance \cite{bonneel2015sliced} as
\begin{align}
    SW_p^p(\mu,\nu)=\int_{S^{d-1}}W_p^p(P^\theta\mu, P^\theta\nu) \mathrm{d}\lambda(\theta),
\end{align}
where $\lambda$ is the uniform distribution on the unit sphere $S^{d-1}=\left \{ \theta \mid \theta\in \mathbb{R}^{d}, \ \left \| \theta \right \| _{2} = 1  \right \}$. For any $x \in \mathbb{R}^{d}$, we define $P^\theta(x)=\left \langle x,\theta  \right \rangle$ termed as the projection of $x$. Since the expectation in the definition of the SW distance is intractable to calculate, the Monte-Carlo estimation is adopted to approximate the SW distance with the computational complexity of $\mathcal{O}\left ( Ln\left ( d+\log{n} \right )  \right ) $ as
\begin{align}
    \widehat{SW}_{p}^{p} \left ( \mu,\nu \right )=\frac{1}{L} \sum_{\ell=1}^LW_p^p(P^{\theta_{\ell} }\mu,  P^{\theta _{\ell} }\nu), \label{eq:sw}
\end{align}
where $\left \{\theta_\ell \right \}_{\ell=1}^{L}\stackrel{i.i.d.}{\sim}\mathcal{U}(\mathbb{S}^{d-1})$ are termed as projection directions sampled from the spherical uniform distribution $\mathcal{U}(\mathbb{S}^{d-1})$, and \emph{L} is the number of projections used for Monte-Carlo approximation.

\textbf{Spherical Sliced-Wasserstein Distance.} For $\mu$, $\nu\in \mathcal{P}_p(S^{d-1})$, we can define the SSW distance \cite{bonet2023spherical} between $\mu$ and $\nu$ as
\begin{align}
    SSW_p^p(\mu,\nu)&=\int_{\mathbb{V}_{d,2}}W_p^p(P_\#^U\mu, P_\#^U\nu) \mathrm{d}\sigma(U), \label{eq:ssw}
\end{align}
where
\begin{align}
    \quad P_\#^U(x)&=\frac{U^{T} x}{\left \| U^{T} x \right \| _{2} }, \label{eq:geo_proj}
\end{align}
and $\sigma$ is the uniform distribution over the Stiefel manifold \cite{bendokat2024grassmann} $\mathbb{V}_{d,2}=\{U \mid U\in\mathbb{R}^{d\times2}, \ U^TU=I_2\}$. $P_\#^U(x)$ denotes the geodesic projection on the circle determined by $U$. The $SSW_1$ can be computed by the binary search algorithm or the level median formulation, while $SSW_2$ can be calculated via Proposition 1 in \cite{bonet2023spherical}.

\section{Method}
It is well-known that computing the SSW distance involves averaging the projected Wasserstein distances across all projection directions. This means that each projection direction is given equal weight, resulting in a lack of discrimination. To improve the discriminative power of the projection directions, we propose assigning different weights to each direction. Our approach introduces formulating a novel SSW distance that incorporates these weights. Additionally, we present the projected energy function designed to generate these weights for the projection directions.

\begin{definition}[\textbf{DSSW Distance}]
     For $p \ge 1$, dimension $d \ge 1 $, two probability measures $\mu \in \mathcal{P}_p(S^{d-1})$ and $\nu\in \mathcal{P}_p(S^{d-1})$, and the projected energy function $f : \mathbb{S} ^{n} \times \mathbb{S} ^{n}\to \left (0, 1 \right) $, the DSSW distance between $\mu$ and $\nu$ is defined as follows:
    \begin{align}
    \begin{split}
        &DSSW_p^p(\mu,\nu; f)= \\
        &\int_{\mathbb{V}_{d,2}} f \left ( P_\#^U\mu, P_\#^U\nu \right ) \cdot W_p^p(P_\#^U\mu, P_\#^U\nu)\mathrm{d}\sigma(U),
        \label{eq:ebssw}        
    \end{split}
    \end{align}
\end{definition}
\noindent where $\sigma$ is the uniform distribution over the Stiefel manifold \cite{bendokat2024grassmann} $\mathbb{V}_{d,2}=\{U \mid U\in\mathbb{R}^{d\times2}, \ U^TU=I_2\}$. $P_\#^U(x)$ denotes the geodesic projection on the circle determined by $U$.

The projected energy function $f$ transforms the projection of two probability measures $\mu$ and $\nu$ into the weights of the projection directions. It can effectively learn the weights for the projection directions from the data distribution.

We now provide the detailed definition and formulation of the projected energy function $f$ as follows:

\begin{definition}[\textbf{Projected Energy Function}]
    For $p \ge 1$, dimension $d \ge 1$, two probability measures $\mu \in \mathcal{P}_p(S^{d-1})$ and $\nu\in \mathcal{P}_p(S^{d-1})$, and two transformation functions $g$ and $h$, the projected energy function $f$ used to calculate the weights for the $\ell$-th projection direction is defined as follows:
    \begin{align}
        f\left(P_\#^{U_\ell}\mu,P_\#^{U_\ell}\nu\right) :=\frac{g \left ( h(P_{\#}^{U_{\ell}}\mu, P_\#^{U_{\ell}} \nu) \right )}{ {\textstyle \sum_{k=1}^{L}} g \left ( h(P_{\#}^{U_{k}}\mu, P_\#^{U_{k}} \nu) \right )},
        \label{eq:energy}
    \end{align}
\end{definition}

\noindent where $P_\#^{U_\ell}\mu$ is the projection of $\mu$ on the $\ell$-th projection direction.

Taking into account the efficiency, we propose the non-parametric form of the projected energy function $f$. In this form, both the functions $h$ and $g$ are non-parametric. Specifically, $h$ is defined as Eq. (\ref{eq:ot}) 
to calculate the projected Wasserstein distance, while $g: \left [ 0, +\infty  \right ) \to \left (  0, +\infty \right )$ transforms the projected Wasserstein distance into the weights of the projection directions. Following the implementation in \cite{nguyen2024energy}, the non-parametric $g$ can be the exponential function ($i.e.$, $g(x)=e^x$), the identity function ($i.e.$, $g(x)=x$), or the polynomial function ($i.e.$, $g(x)=x^2$). Then, normalization is performed on the weights obtained from the projection directions. The non-parametric projected energy function shows that the weights are proportional to the Wasserstein distance for the respective projection direction. Given the above calculation steps, it can be seen that the only additional step in the original SSW calculation process is the calculation of the weights for the projection direction. Therefore, the added calculation time can be disregarded. This means that the proposed DSSW with the non-parametric projected energy function achieves better performance with minimal additional computing overhead compared to the original SSW. In the subsequent sections, DSSW with the exponential, identity, and polynomial functions will be referred to as DSSW (exp), DSSW (identity), and DSSW (poly), respectively.

Considering the accuracy, we present the parametric projected energy function $f$. In this case, $h$ is represented by a parameterized neural network $h_{\psi}$, where $\psi$ denotes the learnable weights of the neural network $h_{\psi}$. The network $h_{\psi }$ can be a linear neural network, a nonlinear neural network, or an attention mechanism. The detailed training configuration for the parameterized neural network $h_{\psi}$ is described in Algorithm 2 in Appendix Section B. Meanwhile, $g$ is a non-parametric function and is specialized as the exponential function ($i.e.$, $g(x)=e^x$). The normalization operation combined with the non-parametric function $g$ is equivalent to calculating the famous Softmax function. Using the Stochastic Gradient Descent (SGD) method to obtain more precise projection directions via the parameterized neural network $h_{\psi}$, our proposed DSSW with a parameterized projected energy function outperforms that with the non-parametric projected energy function and the original SSW. However, it does come with a higher computational cost compared to these two forms. In the subsequent sections, we will refer to DSSW with the linear neural network, nonlinear neural network, and attention mechanism as DSSW (linear), DSSW (nonlinear), and DSSW (attention), respectively.

\begin{proposition}
    \label{prop:metricity}
    For any $p \ge 1$ and the projected energy function $f$, the DSSW distance $DSSW_p$ is positive and symmetric.
\end{proposition}
The definition of the DSSW distance implies that it does not satisfy identity due to the case that the different points on the hypersphere $S^{d-1}$ may share the same projection on the circle determined by $U$. The proofs for the related propositions and theorems can be found in Appendix Section A.

\begin{proposition}
    \label{convergence:pro}
    For any $p \ge 1$ and the projected energy function $f$, let $\mu_{k}$, $\mu \in \mathcal{P}_p(S^{d-1})$. If ${\lim\limits_{k \to +\infty} \mu_{k} =\mu}$, then ${\lim\limits_{k \to +\infty} DSSW_p^p(\mu_{k},\mu; f) =0}$. 
\end{proposition}
Proposition~\ref{convergence:pro} indicates that $DSSW_p$ is asymptotically convergent. It implies that our DSSW distance also satisfies the property of weak convergence that is one of the most crucial requirements that a distance metric should satisfy \cite{NEURIPS2019_c9e1074f}.

\begin{proposition}
    \label{sample:com}
    For any $p \ge 1$, suppose that for $\mu$, $\nu \in P(S^{1})$, with empirical measures $\hat{\mu}=\frac{1}{n} {\textstyle \sum_{i=1}^{n}} \delta_{x_{i}}$, and $\hat{\nu}=\frac{1}{n}  {\textstyle \sum_{i=1}^{n}} \delta_{y_{i}}$, where $\left \{x_{i} \right \}_{i=1}^{n} \sim \mu$, $\left \{y_{i} \right \}_{i=1}^{n} \sim \nu$ are independent samples, we have
    \begin{align}
        \mathbb{E}[|W_p^p(\hat\mu_n, \hat\nu_n)-W_p^p(\mu, \nu)|] \leq\beta(p, n),
    \end{align}
    where $\beta(p, n)$ is independent of the dimensionality d and only depends on $p$ and $n$. Then, for the projected energy function $f$ and $\mu$, $\nu \in P(S^{d-1})$ with empirical measures $\hat{\mu}$ and $\hat{\nu}$, there exists a universal constant $C$ such that
    \small
    \begin{align}
    \mathbb{E}[|DSSW_p^p(\hat\mu_n, \hat\nu_n; f)\!-\!DSSW_p^p(\mu, \nu; f)|] \!\leq\! C\beta(p, n).
    \end{align}
\end{proposition}
Proposition~\ref{sample:com} demonstrates that the sample complexity of DSSW is independent of the dimension. This insight also verifies that our DSSW distance, akin to the SW distance, can avoid the curse of dimensionality.

\begin{theorem}
    \label{projection:com}
    For any $p \ge 1$, two probability measures $\mu$ and $\nu \in P(S^{1})$, and the projected energy function $f$, there exists a universal constant $C$ such that the error made with the Monte Carlo estimate of $DSSW_p^p$ can be bounded as
    \begin{align}
        & \mathbb{E}_U\left[\left | \widehat{DSSW}_{p,L}^p(\mu,\nu;f)\!-\! DSSW_p^p(\mu,\nu;f) \right | \right]^2 \nonumber \\
        & \leq\frac{C^{2}}{L} Var_U\left(W_p^p\left(P_\#^U\mu,P_\#^U\nu\right)\right),
    \end{align}
    where $\widehat{DSSW}_{p,L}^p(\mu,\nu; f) = \frac1L\sum_{\ell=1}^Lf\left(P_\#^{U_\ell}\mu,P_\#^{U_\ell}\nu\right) \cdot W_p^p\left(P_\#^{U_\ell}\mu,P_\#^{U_\ell}\nu\right)$ with $\left \{U_\ell \right \} _{\ell=1}^{L} \! \sim \! \sigma$ independent samples. $L$ is referred to as the number of projections.
\end{theorem}
Theorem~\ref{projection:com} highlights that the projection complexity of DSSW depends on the convergence rate of the Monte Carlo approximation towards the true integral that has been derived for sliced-based distances in \cite{nadjahi2020statistical}. This indicates that the estimation error in the Monte Carlo approximation is determined by the number of projections $L$ and the variance of the evaluations of the Wasserstein distance \cite{NEURIPS2020_eefc9e10}.

\section{Implementation Details}
Similar to the SSW distance, we adopt Monte-Carlo estimation to approximate the integral on $\mathbb{V}_{d,2}$ as in Eq. (\ref{eq:ssw}).

We begin by randomly sampling $L$ projections $\left \{U_\ell  \right \} _{\ell=1}^{L}$ from the uniform distribution $\sigma$ on the Stiefel manifold $\mathbb{V}_{d,2}$. Each projection is obtained by first constructing a matrix $Z\in \mathbb{R} ^{d\times 2}$ with each element drawn from the standard normal distribution $\mathcal{N} \left ( 0, 1 \right )$, followed by QR decomposition of each projection.

We then project the points on the circle $S^1$ according to Eq. (\ref{eq:geo_proj}) and calculate the coordinate of each point in each projection direction on this circle $S^1$ using the formula $\tilde{x}_{i}^{\ell}  = (\pi + \mathrm {atan2}(-x_{i,1}^{\ell}, -x_{i,2}^{\ell}))/(2\pi)$.

Then, we can compute the Wasserstein distance on the circle $S^1$ and determine the weights of the projection directions using Eq. (\ref{eq:energy}). The detailed computation procedure of computing the Wasserstein distance on the circle $S^1$ for $p=1$ and $p=2$ can be referred to \cite{bonet2023spherical}. Finally, we can calculate the DSSW distance using Eq. (\ref{eq:ebssw}). The pseudo-code for computing the DSSW distance is provided in Algorithms 1 and 2 in Appendix Section B.

\textbf{Computation Complexity.} Given $n$ samples from $\mu$ and $m$ samples from $\nu$, along with $L$ projections. Just as the work \cite{bonet2023spherical}, we can finish the QR factorization of $L$ matrices of size $d \times 2$ in $\mathcal{O}\left ( dL \right ) $. Projecting the points on the circle $S^1$ can be finished in $\mathcal{O}\left ( (m+n)dL \right )$. The complexity of computing the general $SSW_p$ can be written as $\mathcal{O}\left ( L\left ( m\!+\!n \right ) \left ( d\!+\!\log{\left ( \frac{1}{\epsilon} \right ) } \right ) \! + \! Ln\log{n}\! + \!Lm\log{m} \right )$, where $\epsilon$ denotes the desired accuracy. The complexity of calculating the weights of the projection directions is $\mathcal{O}\left ( L\right ) $ when using the non-parametric projected energy function $f$. When using the parametric projected energy function $f$, the complexity of calculating the weights of the projection directions is $\mathcal{O}\left ( TL \right )$, where $T$ is the maximum iterations for training the parameterized neural network. Therefore, the total complexity of computing the DSSW distance utilizing the non-parametric projected energy function $f$ is $\mathcal{O}\left ( L\left (m\!+\!n \right ) \left ( d\!+\!\log{\left ( \frac{1}{\epsilon} \right ) } \right ) \! + \! Ln\log{n}\! + \!Lm\log{m}\! + \! L \right )$. In contrast, for the parametric projected energy function $f$, the total complexity of our proposed method is $\! \mathcal{O}\left ( L\left (m\!+\!n \right ) \left ( d\!+\!\log{\left ( \frac{1}{\epsilon} \right ) } \right ) \! + \! Ln\log{n}\! + \! Lm\log{m} \! + \! TL \right ) \!$.

\textbf{Runtime Comparison.} We conducted runtime comparisons between various distances between the uniform distribution and the von Mises-Fisher distribution on $\mathbb{S}^{100}$. The results, shown in Figure \ref{fig:runtime_comparison}, are averaged over 50 iterations for varying sample sizes of each distribution. For all sliced approaches, we used $L=200$ projections. The results in Figure \ref{fig:runtime_comparison} include our DSSW with the non-parametric projected energy function variant (exp). It can be observed that the runtime curve of DSSW (exp) closely aligns with that of SSW, indicating that the additional computing overhead introduced by our DSSW  (exp) is negligible. Due to space limitations, runtime comparisons for DSSW with other non-parametric projected energy function variants (identity and poly) and DSSW with the parametric projected energy function variants (linear, nonlinear, and attention) are provided in Appendix Section C.1. Furthermore, we explore the evolution of our DSSW across varying dimensions, number of projections, and rotation numbers in Appendix Section C.2, along with the runtime analysis of the proposed DSSW in Appendix Section C.3.
\begin{figure}[h]
    \centering
    \includegraphics[width=0.47\textwidth]{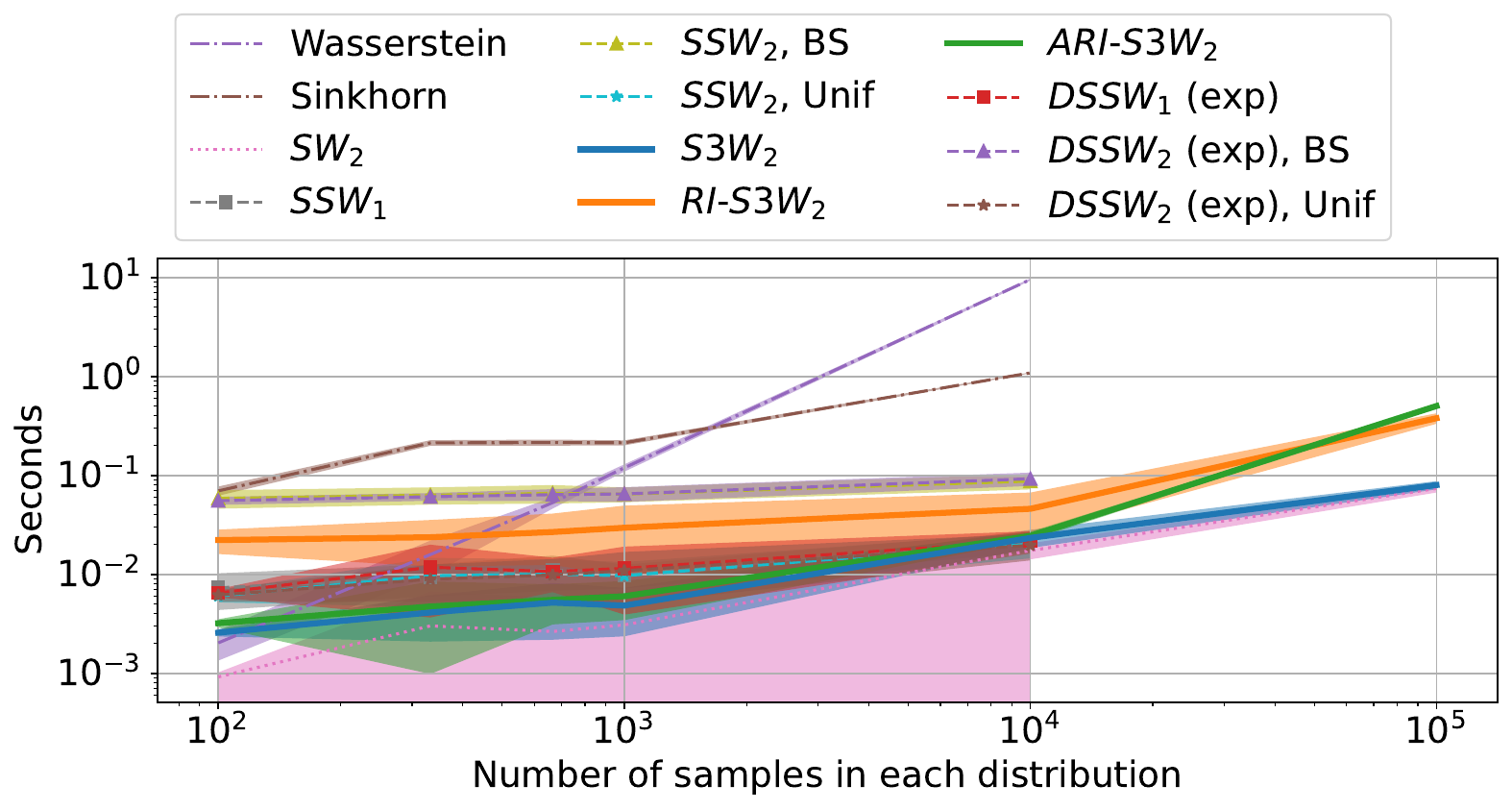}
    \caption{Runtime comparison for Wasserstein distance, Sinkhorn distance with geodesic distance as cost function, $SW_2$ (SW distance), $SSW_1$ distance with the level median, $SSW_2$ distance with binary search (BS), $SSW_2$ distance against a uniform distribution (Unif), $S3W_2$ distance, $RI$-$S3W_2$ (rotationally invariant extension of $S3W_2$) distance, $ARI$-$S3W_2$ (amortized rotationally invariant extension of $S3W_2$) distance, $DSSW_1$ (exp) (ours), $DSSW_2$ (exp), BS (ours), $DSSW_2$ (exp), Unif (ours).}
    \label{fig:runtime_comparison}
\end{figure}

\section{Experiments}

  In line with previous works 
  \cite{bonet2023spherical, tran2024stereographic}, we conducted five different numerical experiments to validate the effectiveness of our method in comparison to SW, SSW \cite{bonet2023spherical} and S3W distance \cite{tran2024stereographic}, where our DSSW distance serves as a loss to measure the distribution discrepancy on the sphere. The results of these experimental are detailed in this section and Appendix Section C. All our experiments are implemented by PyTorch \cite{paszke2019pytorch} on Ubuntu 20.04 and a single NVIDIA RTX 4090 GPU.

\subsection{Gradient Flows on The Sphere}
Suppose the explicit form of the target distribution is unknown and only samples $\left \{y_{j} \in \widehat{\nu_{n} }  \right \} _{j=1}^{N}$ are available, our goal is to iteratively minimize the objective function $\mathop{\arg\min}\limits_{\mu} d\left ( \widehat{\mu_{i} }, \widehat{\nu_{n_{i}}} \right )$, where $d$ is a distance metric such as SW, SSW, S3W, or DSSW. To achieve this goal, we employ the Projected Gradient Descent (PGD) algorithm \cite{madry2018towards} to estimate the target distribution with the update rule as follows:
\begin{align}
\left\{ 
	\begin{aligned}
    x_{i, k+1}^{'} \! &= \! x_{i, k}\!-\!\gamma \! \cdot \! \nabla_{x_{i, k}} \mathrm{DSSW}\left ( \hat{\mu}_k , \hat{\nu}_{n_{i}} \right )  \\
    x_{i, k+1} \! &= \! \frac{x_{i, k+1}^{'}}{\left \| x_{i, k+1}^{'} \right \| _{2} },
	\end{aligned}
\right.
\end{align}
where $\gamma$ is the learning rate for the update rule, $i$ denotes the index of the mini-batches, and $k$ is the gradient step.

\begin{figure*}[h]
    \centering
    \subfloat[Target Density]{
        \includegraphics[width=0.18\textwidth]{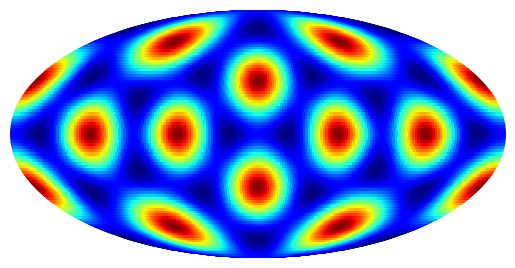}
    }
    \subfloat[SW]{
        \includegraphics[width=0.18\textwidth]{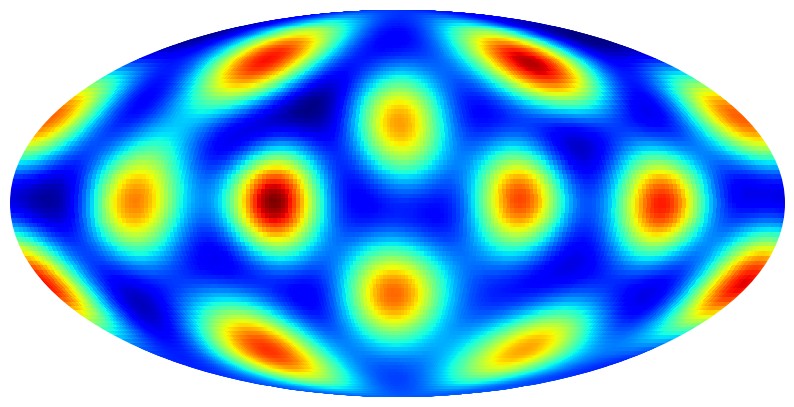}
    }
    \subfloat[SSW]{
        \includegraphics[width=0.18\textwidth]{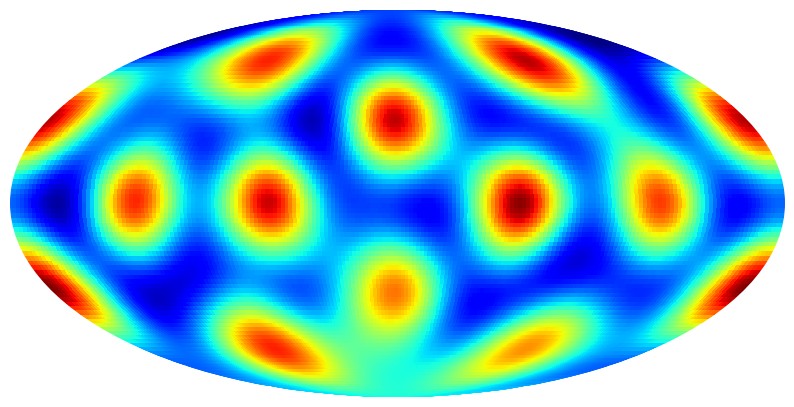}
    }
    \subfloat[S3W]{
        \includegraphics[width=0.18\textwidth]{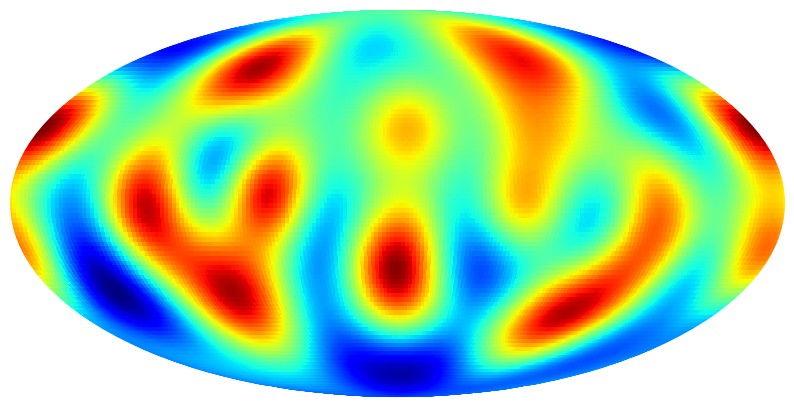}
    }
    \subfloat[RI-S3W (1)]{
        \includegraphics[width=0.18\textwidth]{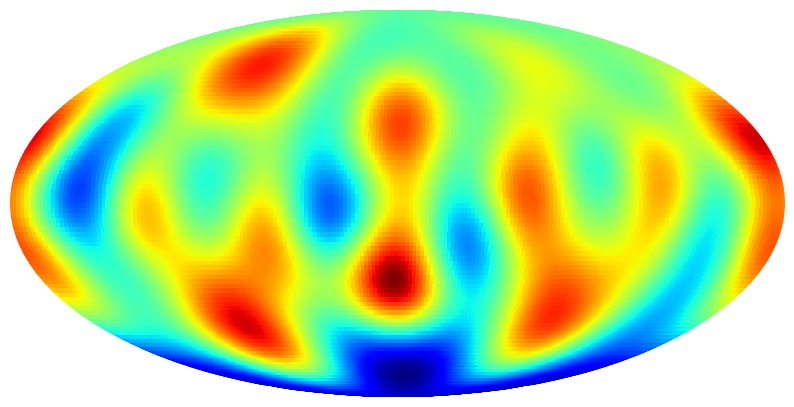}
    }\\
    \subfloat[RI-S3W (5)]{
        \includegraphics[width=0.18\textwidth]{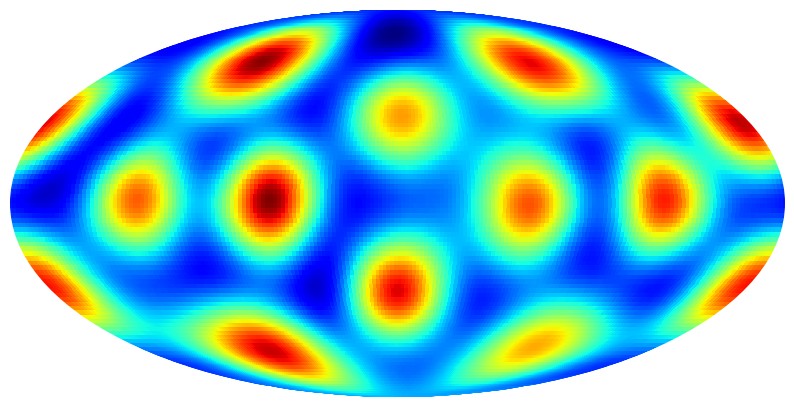}
    }
    \subfloat[RI-S3W (10)]{
        \includegraphics[width=0.18\textwidth]{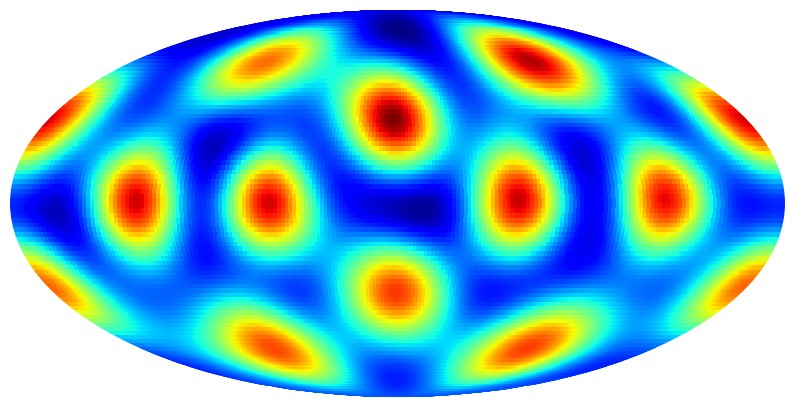}
    }
    \subfloat[ARI-S3W (30)]{
        \includegraphics[width=0.18\textwidth]{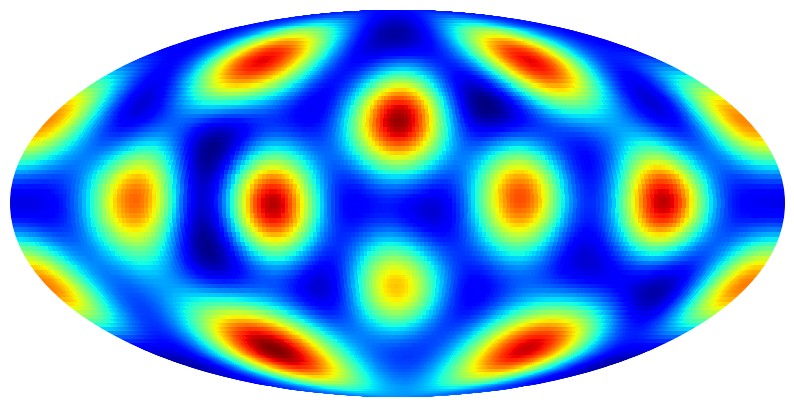}
    }
    \subfloat[DSSW (exp)]{
        \includegraphics[width=0.18\textwidth]{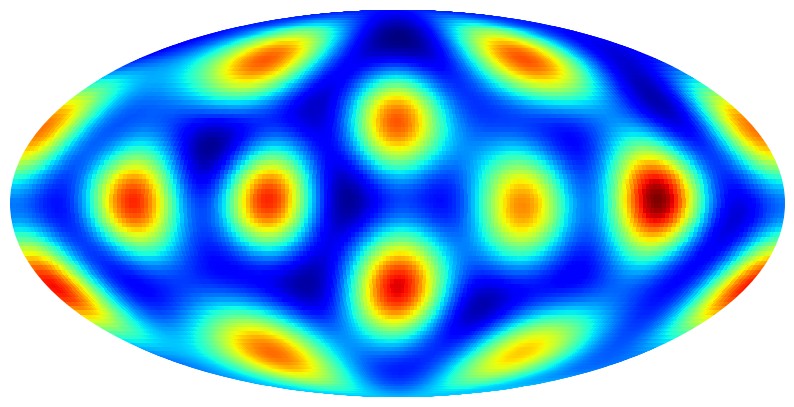}
    }
    \subfloat[DSSW (attention)]{
        \includegraphics[width=0.18\textwidth]{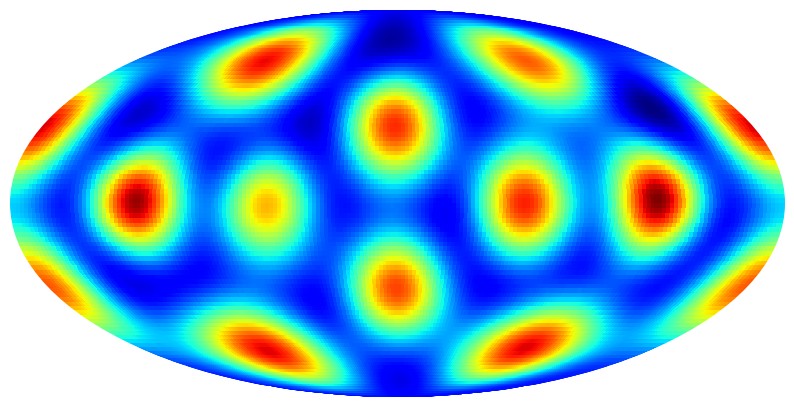}
    }
    \caption{The Mollweide projections for mini-batch projected gradient descent. We use 1, 5, and 30 rotations for RI-S3W (1), RI-S3W (5), and RI-S3W (10), respectively. We also use 30 rotations with a pool size of 1000 for ARI-S3W (30).}
    \label{fig:gradient_flows_mini}
\end{figure*}

We present both qualitative and quantitative results using the Negative Log-likelihood (NLL) and the logarithm of the 2-Wasserstein distance ($\log{W_{2}}$) as evaluation metrics, including mean and standard deviation for each. The mini-batch results for all distances are shown in Table \ref{tab:gradient_flows_mini}, and the Molleweide projections of the mini-batch are illustrated in Figure \ref{fig:gradient_flows_mini}. Table \ref{tab:gradient_flows_mini} shows that our DSSW performs on par or better than other baselines. Additionally, it is evident that our DSSW with a parametric projected energy function surpasses the non-parametric from. Overall, DSSW demonstrates superior performance in accurately learning the target distribution compared to other distances.

Full-batch results are reported in Appendix Section C.4. These results indicate that all distance measures perform well in learning the target distribution. In cases where the density of the target distribution is known up to a constant, we utilize the Sliced-Wasserstein Variational Inference (SWVI) framework \cite{pmlr-v189-yi23a}, optimized through MCMC \cite{DoucetFG01_ESS} methods. This approach does not require optimization or a tractable approximate posterior family, as detailed in Appendix C.8. The SWVI results further validate the accuracy of our DSSW method in approximating the target distribution compared to other competitors.

\begin{table}[h]
\centering
\small
\setlength{\tabcolsep}{1.0mm}
\begin{tabular}{@{}c@{}ccc@{}}
\toprule
& Distance          & NLL $\downarrow$          & $\log{W_{2}}$ $\downarrow$      \\ \midrule
\multirow{13}{*}{Mini-batch} & SW                & -282.48 ± 17.42  & -2.77 ± 0.10   \\
& SSW               & -287.11 ± 6.22   & -2.78 ± 0.08   \\
& S3W               & -181.38 ± 8.72   & -2.61 ± 0.07   \\
& RI-S3W (1)      & -213.63 ± 19.24  & -2.68 ± 0.09   \\
& RI-S3W (5)      & -256.23 ± 10.72  & -2.77 ± 0.13   \\
& RI-S3W (10)     & -285.20 ± 13.22  & -2.77 ± 0.11   \\
& ARI-S3W (30)    & \underline{-291.38 ± 16.48}  & \underline{-2.82 ± 0.12}   \\ \cmidrule{2-4} 
& DSSW (exp)       & -316.38 ± 6.90 $\ddagger$  & -2.92 ± 0.10 $\ddagger$  \\
& DSSW (identity)  & -310.45 ± 5.00 $\ddagger$  & -2.94 ± 0.12 $\ddagger$  \\
& DSSW (poly)     & -307.22 ± 7.50 $\ddagger$  & -2.94 ± 0.11 $\ddagger$  \\
& DSSW (linear)    & -319.41 ± 6.72 $\ddagger$  & -2.94 ± 0.10 $\ddagger$  \\
& DSSW (nonlinear)  & -319.96 ± 6.65 $\ddagger$  & \textbf{-2.97 ± 0.15} $\ddagger$ \\
& DSSW (attention) & \textbf{-320.16 ± 5.58} $\ddagger$ & -2.93 ± 0.10 $\ddagger$ \\ \bottomrule
\end{tabular}%
\caption{Mini-batch comparison between different distances as loss for gradient flows averaged over 10 training runs. Notation "$\ddagger$" indicates that DSSW variants are significantly better than the best baseline method using t-test when the significance level is 0.05.}
\label{tab:gradient_flows_mini}
\end{table}

\subsection{Earth Density Estimation}
We evaluate the performance of our proposed DSSW on the density estimation task using normalizing flows on $\mathbb{S}^{2}$. In alignment with \cite{bonet2023spherical, tran2024stereographic}, we adopt three datasets \cite{mathieu2020riemannian}: Earthquake \cite{earthquake}, Flood \cite{flood} and Fire \cite{fire}. The earth's surface is modeled as a spherical manifold. Following the implementation of \cite{bonet2023spherical, tran2024stereographic}, we utilize an exponential map normalizing flow model \cite{pmlr-v119-rezende20a}, which is optimized by $\min\limits_{T} DSSW\left (T_{\#} \mu, z \right )$. In this formulation, $T$ is the transformation introduced by the model, $\mu$ is the data distribution known by sampling samples $\left \{ x_{i}  \right \} _{i=1}^{N}$, and $z$ is a prior distribution on $\mathbb{S}^{2}$. The learned density $f_{\mu}$ can be obtained by
\begin{align}
    \ f_{\mu}\left ( x \right )=z\left ( T\left ( x \right )  \right )  \left | \mathrm{det}\ J_{T}\left ( x \right ) \right |,\ \forall x\in \mathbb{S}^{2},
\end{align}
where $J_{T}\left ( x \right )$ means the the Jacobian of $T$ at $x$.

The NLL values of density estimation on three earth datasets are demonstrated in Table \ref{tab:earth}. Stereo \cite{gemici2016normalizing} first projects samples from $\mathbb{S}^{2}$ to $\mathbb{R}^{2}$ and then applies the Real NVP \cite{dinh2017density} model in the projected space. The results show that our proposed DSSW outperforms all other baselines. Specifically, on the Earthquake dataset, DSSW (linear) achieves the best performance, while DSSW (nonlinear) performs best on the Flood and Fire datasets. These results indicate that DSSW is more suitable for fitting data on the sphere than other methods.

In addition, the density visualization using various distances on test data is shown in Appendix Section C.5. These visualizations further support that DSSW estimates a more accurate density distribution than other distances.

\begin{table}[h]
\centering
\small
\setlength{\tabcolsep}{0.8mm}
\begin{tabular}{@{}cccc@{}}
\toprule
\multirow{2}{*}{Method} & \multicolumn{3}{c}{Dataset}               \\ \cmidrule{2-4} 
                        & Earthquake $\downarrow$        & Flood $\downarrow$      & Fire $\downarrow$       \\ \midrule
Stereo                  & 2.04 ± 0.19   & 1.85 ± 0.03 & 1.34 ± 0.11 \\
SW                      & 1.12 ± 0.07   & 1.58 ± 0.02 & 0.55 ± 0.18 \\
SSW                     & 0.84 ± 0.05   & 1.26 ± 0.03 & 0.24 ± 0.18 \\
S3W                     & 0.88 ± 0.09   & 1.33 ± 0.05 & 0.36 ± 0.04 \\
RI-S3W (1)               & 0.79 ± 0.07   & 1.25 ± 0.02 & 0.15 ± 0.06 \\
ARI-S3W (50)             & \underline{0.78 ± 0.06}   & \underline{1.24 ± 0.04} & \underline{0.10 ± 0.04}  \\ \midrule
DSSW (exp)       & 0.70 ± 0.09 $\ddagger$ & 1.22 ± 0.04 $\ddagger$ & 0.05 ± 0.08 $\ddagger$ \\
DSSW (identity)  & 0.76 ± 0.08 $\ddagger$ & 1.23 ± 0.06 $\ddagger$ & 0.10 ± 0.13 \\
DSSW (poly)      & 0.74 ± 0.05 $\ddagger$ & 1.23 ± 0.08 $\ddagger$ & 0.22 ± 0.21 \\
DSSW (linear)    & \textbf{0.69 ± 0.04} $\ddagger$ & 1.21 ± 0.03 $\ddagger$ & 0.09 ± 0.04 $\ddagger$ \\
DSSW (nonlinear) & 0.71 ± 0.06 $\ddagger$ & \textbf{1.20 ± 0.03} $\ddagger$ & \textbf{0.05 ± 0.05} $\ddagger$ \\
DSSW (attention) & 0.70 ± 0.08 $\ddagger$ & 1.21 ± 0.03 $\ddagger$ & 0.08 ± 0.07 $\ddagger$ \\
 \bottomrule
\end{tabular}%
\caption{Earth datasets results. We evaluate the NLL on test data averaged over 5 training runs. We use 1 rotation for RI-S3W (1). We also use 50 rotations with a pool size of 1000 for ARI-S3W (50). The results of baselines are cited from \cite{tran2024stereographic}. The notation "$\ddagger$" indicates that DSSW variants are significantly better than the best baseline method using t-test when the significance level is 0.05.}
\label{tab:earth}
\end{table}

\subsection{Sliced-Wasserstein Autoencoder}
In this section, we employ the classical SWAE \cite{kolouri2018sliced} framework to evaluate the performance of various distances in generative modeling. Let $\alpha$ denote an encoder, and $\beta$ be a decoder in this framework. The goal of SWAE is to force the encoded embedding to follow a prior distribution in the latent space. For this experiment, we utilize a mixture of vMF distributions with 10 components on $\mathbb{S}^2$ as the prior distribution denoted as $z$. The training objective of the revised SWAE is
\begin{align}
    \min_{\alpha,\beta} \mathbb{E}_{x\sim \mu} \left [ c\left (x, \alpha\left ( \beta\left ( x \right )  \right )  \right )  \right ]\! +\! \eta \! \cdot \! DSSW_{2}^{2}\left ( \alpha_\# \mu, z \right ),
\end{align}
where $\mu$ is the unknown data distribution that we only have access to samples, $c$ refers to the reconstruction loss that is implemented as the standard Binary Cross Entropy (BCE) loss, and $\eta$ denotes the regularization coefficient.

The results on the CIFAR10 \cite{krizhevsky2009learning} benchmark for SWAE with vMF prior are shown in Table \ref{tab:swae_cifar10}. Our DSSW outperforms other methods in terms of $\log{W_{2}}$ and NLL. In terms of the reconstruction loss BCE, SW is better than other competitors. In a word, all the regularization priors are superior to the original supervised autoencoder (AE). This phenomenon indicates that a prior on the hypersphere can enhance the training performance of SWAE \cite{davidson2018hyperspherical, xu2018spherical}.

Moreover, we also demonstrate the details about the network architectures, training configurations, additional results and the latent space visualization on the MNIST \cite{LeCunBBH98_mnist} benchmark in Appendix Section C.6.

\begin{table*}[h]
\centering
\small
\begin{tabular}{@{}ccccc@{}}
\toprule
Method            & $\eta$     & $\log{W_{2}}$ $\downarrow$            & NLL $\downarrow$            & BCE $\downarrow$          \\ \midrule
Supervised AE     & 1     & -0.1313 ± 0.8101 & 0.0031 ± 0.0126  & 0.6329 ± 0.0021 \\
SSW               & 10    & -3.2368 ± 0.1836 & 0.0008 ± 0.0019  & 0.6323 ± 0.0017 \\
SW                & 0.001 & \underline{-3.2537 ± 0.1116} & \underline{-0.0004 ± 0.0030} & \textbf{0.6307 ± 0.0005} \\
S3W               & 0.001 & -3.0541 ± 0.2244 & -0.0000 ± 0.0027 & 0.6310 ± 0.0012 \\
RI-S3W (5)        & 0.001 & -2.8317 ± 0.8168 & 0.0004 ± 0.0034  & 0.6330 ± 0.0049 \\
ARI-S3W (5)       & 0.001 & -3.1639 ± 0.1744 & 0.0002 ± 0.0028  & 0.6315 ± 0.0016 \\ \midrule
DSSW (exp)       & 10    & -3.3607 ± 0.1349 $\ddagger$ & -0.0012 ± 0.0051 $\ddagger$ & 0.6321 ± 0.0006 \\
DSSW (identity)  & 10    & -3.4203 ± 0.0402 $\ddagger$ & -0.0011 ± 0.0025 $\ddagger$ & 0.6318 ± 0.0006 \\
DSSW (poly)      & 10    & -3.3454 ± 0.1117 $\ddagger$ & \textbf{-0.0018 ± 0.0021} $\ddagger$ & 0.6330   ± 0.0024 \\
DSSW (linear)    & 10    & -3.4027 ± 0.0480 $\ddagger$ & -0.0002 ± 0.0038 $\ddagger$ & 0.6314 ± 0.0004 \\
DSSW (nonlinear) & 10    & -3.4078 ± 0.0753 $\ddagger$  & -0.0014 ± 0.0045 $\ddagger$  & 0.6323 ± 0.0013   \\
DSSW (attention) & 10    & \textbf{-3.4242 ± 0.0337}  $\ddagger$ & -0.0002 ± 0.0044 $\ddagger$  & 0.6324 ± 0.0016   \\ \bottomrule
\end{tabular}%
\caption{CIFAR10 results for SWAE with vMF prior. We evaluate the latent regularization loss ($\log{W_{2}}$ and NLL), along with the BCE loss on the test data for $d = 3$. We use 5 rotations for RI-S3W (5). We also use 5 rotations with the pool size of 1000 for ARI-S3W (5). The notation "$\ddagger$" indicates that DSSW variants are significantly better than the best baseline method using a t-test when the significance level is 0.05.}
\label{tab:swae_cifar10}
\end{table*}

\subsection{Self-Supervised Representation Learning}
It has been proven that the contrastive objective can be decomposed into an alignment loss, which forces positive representations coming from the same image to be similar, and a uniformity loss, which preserves the maximum information of the feature distribution and thus avoids collapsing representations \cite{wang2020understanding, zhang2024few, zheng2023curricular}. Similar to SSW \cite{bonet2023spherical} and S3W \cite{tran2024stereographic}, we replace the Gaussian kernel uniformity loss with our DSSW. Therefore, the overall pre-trained objective of the self-supervised learning network can be defined as:
\begin{align}
\label{eq:ssl_ssw}
\begin{split}
    &\mathcal{L}_{\mathrm{DSSW-SSL}}=\frac{1}{n}\sum_{i=1}^{n}\lVert z_{i}^{A}-z_{i}^{B}\rVert_{2}^{2} \\
    &\!+\!\frac{\eta}{2}\big(DSSW_{2}^{2}(z^{A},\nu)\!+\!DSSW_{2}^{2}(z^{B},\nu)\big),
\end{split}
\end{align}
where $z^A$ and $z^B$ denote the hyperspherical projections of the representations from the network for two augmented versions of the same images, $\nu =\mathrm{Unif} \left ( \mathbb{S} ^{d-1}  \right ) $ is the uniform distribution on the hypersphere and $\eta > 0$ acts as a regularization coefficient to balance the alignment loss and the uniformity loss in Eq. (\ref{eq:ssl_ssw}).

\begin{table}[h]
	\centering
	\small
	\setlength{\tabcolsep}{1mm}
	\begin{tabular}{cccc}
		\toprule
		$d$                    & Method                                & E $\uparrow$ & $\mathbb{S}^{9}$   $\uparrow$    \\ \midrule
		\multirow{14}{*}{10} & Supervised                            & 92.38          & 91.77 \\
		& hypersphere                           & 79.76          & 74.57 \\
		& SimCLR                                & 79.69          & 72.78 \\
		& SW-SSL ($\eta$=1, L=200)                    & 74.45          & 68.35 \\
		& SSW-SSL ($\eta$=20, L=200)                  & 70.46          & 64.52 \\
		& S3W-SSL ($\eta$=0.5, L=200)                 & 78.54          & 73.84 \\
		& RI-S3W(5)-SSL ($\eta$=0.5, L=200)           & \underline{79.97}          & 74.27 \\
		& ARI-S3W(5)-SSL ($\eta$=0.5, L=200)          & 79.92          & \underline{75.07} \\ \cmidrule{2-4} 
		& DSSW-SSL (exp, $\eta$=100, L=200)         & 80.10           & 76.30  \\
		& DSSW-SSL (identity,   $\eta$=105, L=200)  & 79.65          & 75.14 \\
		& DSSW-SSL (poly,   $\eta$=105, L=200)      & 78.46          & 73.69 \\
		& DSSW-SSL (linear,   $\eta$=100, L=200)    & \textbf{80.15}          & \textbf{76.87} \\
		& DSSW-SSL (nonlinear,   $\eta$=100, L=200) & 79.73          & 76.61 \\
		& DSSW-SSL (attention,   $\eta$=100, L=200) & 79.66          & 75.98 \\ \midrule 
	\end{tabular}%
	\caption{Linear evaluation on CIFAR10 for $d=10$. E denotes the encoder output. We use 5 rotations for RI-S3W (5). We also use 5 rotations with the pool size of 1000 for ARI-S3W (5). The results are compared with methods cited from \cite{tran2024stereographic}.}
	\label{tab:ssl_cifa10_d10}
\end{table}

We conduct experiments on CIFAR10 \cite{krizhevsky2009learning} by adopting ResNet18 \cite{He_2016_CVPR} as the encoder. The results of the standard linear classifier evaluation for $d=10$ are reported in Table \ref{tab:ssl_cifa10_d10}. The results indicate that our DSSW (exp) and DSSW (linear) are superior to other self-supervised methods in terms of the accuracy for the encoder output and the projected features on $\mathbb{S}^{9}$. As expected, the supervised method achieves the highest precision due to the additional supervised signals.

\begin{figure}[h]
    \centering
    \subfloat[Supervised]{
        \includegraphics[width=0.2\textwidth]{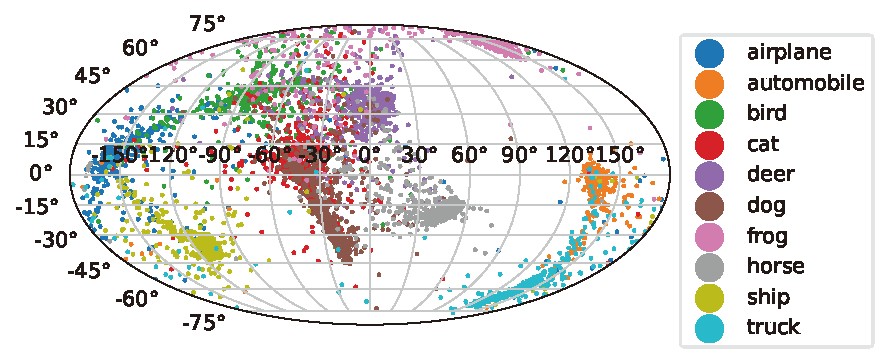}
    }
    \subfloat[SSW]{
        \includegraphics[width=0.2\textwidth]{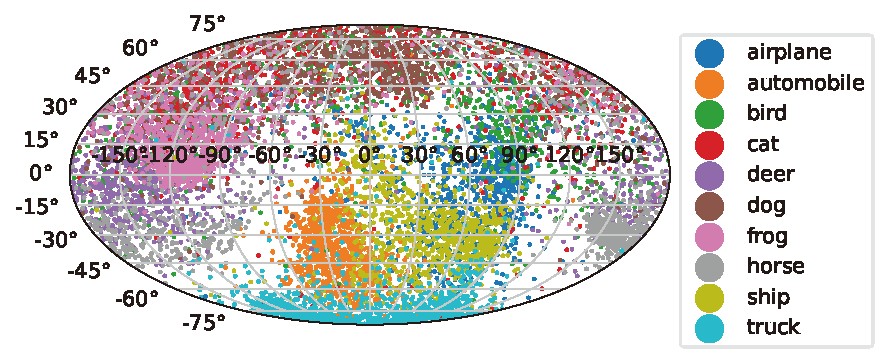}
    }\\
    \subfloat[RI-S3W (5)]{
        \includegraphics[width=0.2\textwidth]{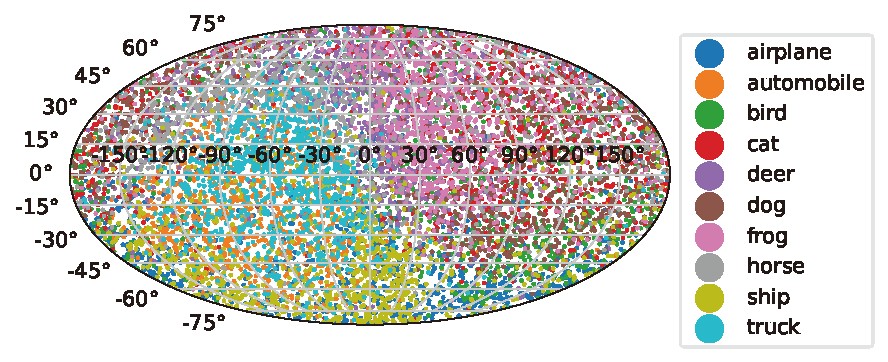}
    }
    \subfloat[ARI-S3W (5)]{
        \includegraphics[width=0.2\textwidth]{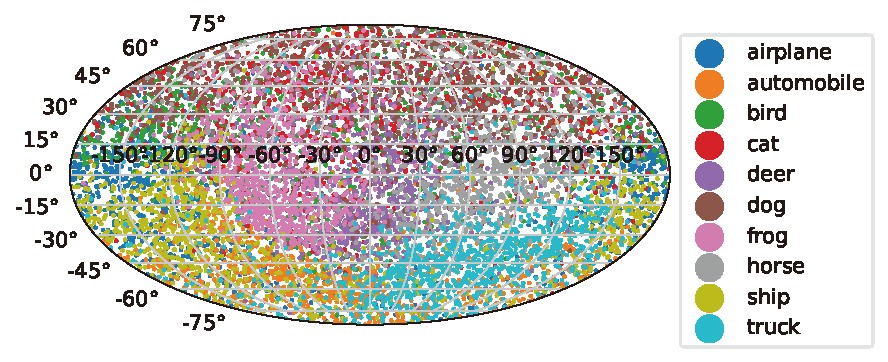}
    }\\
    \subfloat[DSSW (exp)]{
        \includegraphics[width=0.2\textwidth]{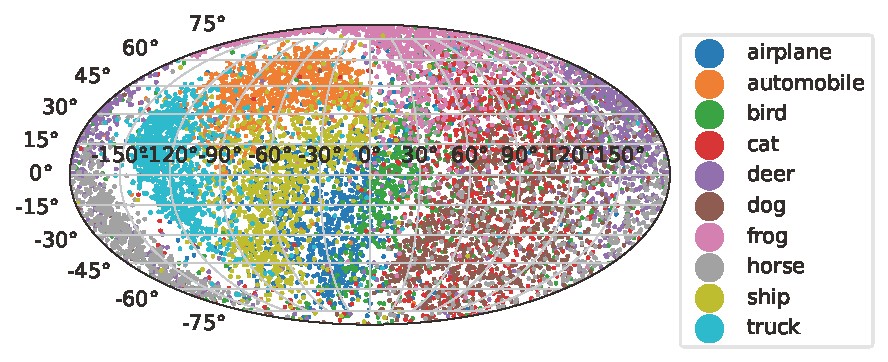}
    }
    \subfloat[DSSW (linear)]{
        \includegraphics[width=0.2\textwidth]{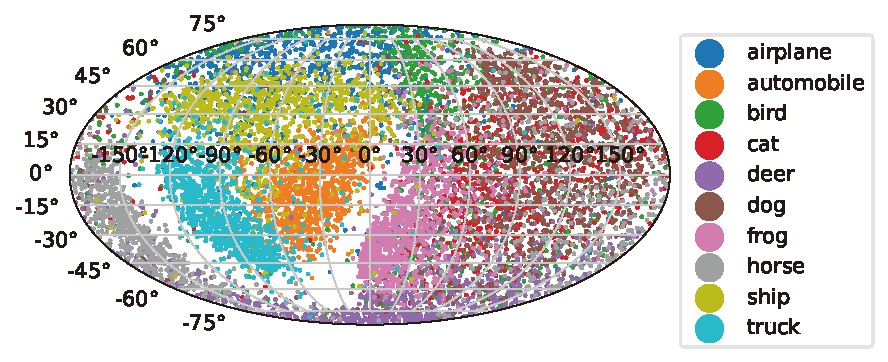}
    }
    \caption{Projected features on $\mathbb{S}^{2}$ for CIFAR10}
    \label{fig:ssl_cifar10_d3}
\end{figure}

Implementation details and the additional results of the standard linear classifier evaluation on $\mathbb{S}^{2}$ are reported in Appendix Section C.7. Furthermore, we visualize the projected features on $\mathbb{S}^{2}$ in Figure \ref{fig:ssl_cifar10_d3}, the visualization plot demonstrates that the cluster result of the projected features on $\mathbb{S}^{2}$ obtained by our DSSW is better than other methods. 

\section{Conclusion}
In this work, we propose a novel approach termed DSSW distance that emphasizes the importance of the projection direction. Our proposed DSSW employ a non-parametric projected energy function to learn a discriminative projection direction, considering both efficiency and accuracy. Our proposed DSSW has been proven to be effective and competitive in various applications. However, the issue of reducing the additional computing overhead caused by training the parametric neural network remains to be addressed in future research. Additionally, the idea of learning discriminative projection direction from the specific data distribution can also be extended to other non-Euclidean Sliced-Wasserstein methods.

\section{Acknowledgments}
This work was supported by the National Science Fund of China under Grant Nos. U24A20330, 62361166670, 62276135 and 62176124.

\bibliography{aaai25}

\appendix
\clearpage
\onecolumn
\renewcommand{\theequation}{A.\arabic{equation}}
\renewcommand{\thetable}{A\arabic{table}}
\renewcommand{\thefigure}{A\arabic{figure}}
\begin{center}
	{\bf{\Large{Appendix to ``Towards Better Spherical Sliced-Wasserstein Distance Learning with Data-Adaptive Discriminative Projection Direction"}}}
\end{center}
This appendix provides the technical proofs for the AAAI25 paper entitled Towards Better Spherical Sliced-Wasserstein Distance Learning with Data-Adaptive Discriminative Projection Direction in section A. We introduce the pseudo-code for computing two types of DSSW distance and the network structure of the parameterized neural network $h_{\psi}$ in section B. We also report the additional comprehensive experimental results to validate the the effectiveness and superiority of our DSSW distance in section C.
\section{Proofs}
\subsection{Proof of Proposition 1}
\begin{proposition}
	\label{prop:metricity}
	For any $p \ge 1$ and the projected energy function $f$, the DSSW distance $DSSW_p$ is positive and symmetric.
\end{proposition}
\noindent \emph{Proof.} Let $p \ge 1$. By the definition that the \emph{p}-Wasserstein distance is a distance metric, it is straightforward to see that for all $\mu, \nu \in \mathcal{P}_p(S^{d-1})$, $\int_{\mathbb{V}_{d,2}}W_p^p(P_\#^U\mu, P_\#^U\nu)\mathrm{d}\sigma(U) \ge 0$. At the same time by the Definition 2 it is obvious for us to conclude that $0 < f \left ( P_\#^U\mu, P_\#^U\nu \right ) < 1$. Then, we conclude that the DSSW distance $DSSW_p(\mu,\nu; f)= \left ( \int_{\mathbb{V}_{d,2}} f \left ( P_\#^U\mu, P_\#^U\nu \right ) \cdot W_p^p(P_\#^U\mu, P_\#^U\nu)\mathrm{d}\sigma(U) \right ) ^{\frac{1}{p}}$ is non-negative. For all $\mu, \nu \in \mathcal{P}_p(S^{d-1})$, according to the symmetry of the \emph{p}-Wasserstein distance we can derive that $\int_{\mathbb{V}_{d,2}}W_p^p(P_\#^U\mu, P_\#^U\nu)\mathrm{d}\sigma(U) = \int_{\mathbb{V}_{d,2}}W_p^p(P_\#^U\nu, P_\#^U\mu)\mathrm{d}\sigma(U)$. Similarly, it concludes that $f \left ( P_\#^U\mu, P_\#^U\nu \right ) = f \left ( P_\#^U\nu, P_\#^U\mu \right )$ by the Definition 2. So the DSSW distance $DSSW_p(\mu,\nu; f)= \left ( \int_{\mathbb{V}_{d,2}} f \left ( P_\#^U\mu, P_\#^U\nu \right ) \cdot W_p^p(P_\#^U\mu, P_\#^U\nu)\mathrm{d}\sigma(U) \right ) ^{\frac{1}{p}}$ is symmetric.


\subsection{Proof of Proposition 2}
\begin{proposition}
	\label{convergence:pro}
	For any $p \ge 1$ and the projected energy function $f$, let $\mu_{k}$, $\mu \in \mathcal{P}_p(S^{d-1})$. If ${\lim\limits_{k \to +\infty} \mu_{k} =\mu}$, then ${\lim\limits_{k \to +\infty} DSSW_p^p(\mu_{k},\mu; f) =0}$. 
\end{proposition}
\noindent \emph{Proof.} Since the Wasserstein distance satisfies the weak convergence, it exists ${\lim\limits_{k \to +\infty } P_\#^U\mu_k = P_\#^U\mu}$ by continuity, and ${\lim\limits_{k \to +\infty } W_p^p(P_\#^U\mu_k, P_\#^U\mu)=0}$. At the same time $0<f(P_\#^U\mu_k, P_\#^U\mu)<1$. So ${\lim\limits_{k \to +\infty } DSSW_p^p(\mu_{k},\mu; f)=\lim\limits_{k \to +\infty } \int_{\mathbb{V}_{d,2}} f \left ( P_\#^U\mu_k, P_\#^U\nu \right ) \cdot W_p^p(P_\#^U\mu_k, P_\#^U\nu)\mathrm{d}\sigma(U)=0}$ because of the dominated convergence theorem.

\subsection{Proof of Proposition 3}
\begin{proposition}
	\label{sample:com}
	For any $p \ge 1$, suppose that for $\mu$, $\nu \in P(S^{1})$, with empirical measures $\hat{\mu}=\frac{1}{n} {\textstyle \sum_{i=1}^{n}} \delta_{x_{i}}$, and $\hat{\nu}=\frac{1}{n}  {\textstyle \sum_{i=1}^{n}} \delta_{y_{i}}$, where $\left \{x_{i} \right \}_{i=1}^{n} \sim \mu$, $\left \{y_{i} \right \}_{i=1}^{n} \sim \nu$ are independent samples, we have
	\begin{align}
		\mathbb{E}[|W_p^p(\hat\mu_n, \hat\nu_n)-W_p^p(\mu, \nu)|] \leq\beta(p, n),
	\end{align}
	where $\beta(p, n)$ is independent of the dimensionality d and only depends on $p$ and $n$. Then, for the projected energy function $f$ and $\mu$, $\nu \in P(S^{d-1})$ with empirical measures $\hat{\mu}$ and $\hat{\nu}$, there exists a universal constant $C$ such that
	\small
	\begin{align}
		\mathbb{E}[|DSSW_p^p(\hat\mu_n, \hat\nu_n; f)\!-\!DSSW_p^p(\mu, \nu; f)|] \!\leq\! C\beta(p, n).
	\end{align}
\end{proposition}
\noindent \emph{Proof.} By utilizing the triangle inequality, Fubini-Tonelli, and the hypothesis on the sample complexity of $W_p^p$ on $S^{1}$, we can obtain the derivation as follows:
\begin{align}
	\begin{split}
		&\mathbb{E}[|DSSW_p^p(\hat\mu_n, \hat\nu_n; f)-DSSW_p^p(\mu, \nu; f)|] \\
		&=\mathbb{E}\left[\left | \int_{v_{d,2}}\left(f\left(P_\#^U\hat{\mu}_n,P_\#^U\hat{\nu}_n\right)\cdot W_p^p\left(P_\#^U\hat{\mu}_n,P_\#^U\hat{\nu}_n\right)-f\left(P_\#^U\mu,P_\#^U\nu\right) \cdot W_p^p\left(P_\#^U\mu,P_\#^U\nu\right)\right)\mathrm{d}\sigma(U)\right|\right] \\
		&\leq\mathbb{E}\left[\int_{\mathbf{v}_{d,2}}\left|f\left(P_\#^U\hat{\mu}_n,P_\#^U\hat{\nu}_n\right)\cdot W_p^p\left(P_\#^U\hat{\mu}_n,P_\#^U\hat{\nu}_n\right)-f\left(P_\#^U\mu,P_\#^U\nu\right) \cdot W_p^p\left(P_\#^U\mu,P_\#^U\nu\right)\right|\mathrm{d}\sigma(U)\right] \\
		&=\int_{\mathbf{v}_{d,2}}\mathbb{E}[\left|f\left(P_\#^U\hat{\mu}_n,P_\#^U\hat{\nu}_n\right) \cdot W_p^p\left(P_\#^U\hat{\mu}_n,P_\#^U\hat{\nu}_n\right)-f\left(P_\#^U\mu,P_\#^U\nu\right) \cdot W_p^p\left(P_\#^U\mu,P_\#^U\nu\right)\right|]\mathrm{d}\sigma(U).
	\end{split}
\end{align}
By the Definition 2 it is straightforward to see that that for all $\mu, \nu \in \mathcal{P}_p(S^{d-1})$, $0 < f \left ( P_\#^U\mu, P_\#^U\nu \right ) < 1$ and for all $\hat{\mu}_n, \hat{\nu}_n \in \mathcal{P}_p(S^{d-1})$, $0 < f \left ( P_\#^U\hat{\mu}_n,P_\#^U\hat{\nu}_n \right ) < 1$, so it must exist a universal constant $C$ that satisfies
\begin{align}
	\begin{split}
		&\left|f\left(P_\#^U\hat{\mu}_n,P_\#^U\hat{\nu}_n\right) \cdot W_p^p\left(P_\#^U\hat{\mu}_n,P_\#^U\hat{\nu}_n\right)-f\left(P_\#^U\mu,P_\#^U\nu\right) \cdot W_p^p\left(P_\#^U\mu,P_\#^U\nu\right)\right| \\
		&\le C\left|W_p^p\left(P_\#^U\hat{\mu}_n,P_\#^U\hat{\nu}_n\right)-W_p^p\left(P_\#^U\mu,P_\#^U\nu\right)\right|,
	\end{split}
\end{align}
then we can get the inequality as below:
\begin{align}
	\begin{split}
		&\leq C\int_{\mathbf{v}_{d,z}}\mathbb{E}[|W_p^p(P_\#^U\hat{\mu}_n,P_\#^U\hat{\nu}_n)-W_p^p(P_\#^U\mu,P_\#^U\nu)|]\mathrm{d}\sigma(U) \\
		&=\leq C\int_{\mathbf{v}_{d,2}}\beta(p,n)\mathrm{d}\sigma(U) \\
		&=C\beta(p,n).
	\end{split}
\end{align}

\subsection{Proof of Theorem 1}
\begin{theorem}
	\label{projection:com}
	For any $p \ge 1$, two probability measures $\mu$ and $\nu \in P(S^{1})$, and the projected energy function $f$, there exists a universal constant $C$ such that the error made with the Monte Carlo estimate of $DSSW_p^p$ can be bounded as
	\begin{align}
		& \mathbb{E}_U\left[\left | \widehat{DSSW}_{p,L}^p(\mu,\nu;f)\!-\! DSSW_p^p(\mu,\nu;f) \right | \right]^2 \nonumber \\
		& \leq\frac CLVar_U\left(W_p^p\left(P_\#^U\mu,P_\#^U\nu\right)\right),
	\end{align}
	where $\widehat{DSSW}_{p,L}^p(\mu,\nu; f) = \frac1L\sum_{\ell=1}^Lf\left(P_\#^{U_\ell}\mu,P_\#^{U_\ell}\nu\right) \cdot W_p^p\left(P_\#^{U_\ell}\mu,P_\#^{U_\ell}\nu\right)$ with $\left \{U_\ell \right \} _{\ell=1}^{L} \! \sim \! \sigma$ independent samples. $L$ is referred to as the number of projections.
\end{theorem}
\noindent \emph{Proof.} Let $\left ( U_i \right )_{i=1}^{L} \stackrel{i.i.d.}{\sim} \sigma$. Since the cost function is increasing convex function in the SSW distance, the SSW distance is convex. Then by first using Jensen inequality and then it has  $\mathbb{E}_{U}\left [ W_p^p(P_\#^U\mu,P_\#^U\nu ; f)\right ] = DSSW_p^p\left ( \mu, \nu; f \right )$, we have
\noindent
\begin{align}
	\begin{split}
		& \mathbb{E}_U\left[\left | \widehat{DSSW}_{p,L}^p(\mu,\nu;f)-DSSW_p^p(\mu,\nu;f) \right | \right]^2 \\
		& \le \mathbb{E}_U\left[\left | \widehat{DSSW}_{p,L}^p(\mu,\nu;f)-DSSW_p^p(\mu,\nu;f) \right |^2 \right] \\
		& \le \mathbb{E}_U\left[\left | \frac{1}{L} {\textstyle \sum_{i=1}^{L}} \left (f\left(P_\#^{U_i}\mu,P_\#^{U_i}\nu\right) \cdot W_p^p\left(P_\#^{U_i}\mu,P_\#^{U_i}\nu\right) - DSSW_p^p(\mu,\nu;f) \right ) \right |^2 \right] \\
		& \le \mathbb{E}_U\left[\left | \frac{1}{L} {\textstyle \sum_{i=1}^{L}} \left (f\left(P_\#^{U_i}\mu,P_\#^{U_i}\nu\right) \cdot W_p^p\left(P_\#^{U_i}\mu,P_\#^{U_i}\nu\right) -f\left(P_\#^U\mu,P_\#^U\nu\right) \cdot W_p^p\left(P_\#^U\mu,P_\#^U\nu\right) \right ) \right |^2 \right],
	\end{split}
\end{align}
By the Definition 2 it is straightforward to see that that for all $\mu, \nu \in \mathcal{P}_p(S^{d-1})$, $0 < f\left(P_\#^{U_i}\mu,P_\#^{U_i}\nu\right) < 1$ and $0 < f\left(P_\#^U\mu,P_\#^U\nu\right) < 1$, so it must exist a universal constant $C$ that satisfies
\noindent
\begin{align}
	\begin{split}
		& \mathbb{E}_U\left[\left | \frac{1}{L} {\textstyle \sum_{\ell=1}^{L}} \left (f\left(P_\#^{U_\ell}\mu,P_\#^{U_\ell}\nu\right) \cdot W_p^p\left(P_\#^{U_\ell}\mu,P_\#^{U_\ell}\nu\right) - \int_{\mathbb{V}_{d,2}} f \left ( P_\#^U\mu, P_\#^U\nu \right ) \cdot W_p^p(P_\#^U\mu, P_\#^U\nu)\mathrm{d}\sigma(U) \right ) \right |^2 \right] \\
		& \le \mathbb{E} _{U} \left [  \left | \frac{C}{L}  {\textstyle \sum_{i=\ell}^{L}} \left (  W_p^p\left(P_\#^{U_\ell}\mu,P_\#^{U_\ell}\nu\right) -W_p^p\left(P_\#^U\mu,P_\#^U\nu\right) \right )  \right |^2 \right ] \\
		& = Var\left ( \frac{C}{L} {\textstyle \sum_{i=\ell}^{L}} W_p^p\left(P_\#^{U_\ell}\mu,P_\#^{U_\ell}\nu\right) \right ) \\
		& = \frac{C^2}{L^2} Var\left ({\textstyle \sum_{i=\ell}^{L}} W_p^p\left(P_\#^{U_\ell}\mu,P_\#^{U_\ell}\nu\right) \right ) \\
		& = \frac{C^2}{L^2} {\textstyle \sum_{i=\ell}^{L}} Var\left ( W_p^p\left(P_\#^{U_\ell}\mu,P_\#^{U_\ell}\nu\right) \right ) \\
		& = \frac{C^2}{L} Var\left (  W_p^p\left(P_\#^{U}\mu,P_\#^{U}\nu\right) \right ) \\
		& = \frac{C^2}{L} \int_{\mathbb{V}_{d,2}} \left ( W_p^p(P_\#^U\mu, P_\#^U\nu)-W_p^p(\mu,\nu) \right ) ^{2} \mathrm{d}\sigma(U).
	\end{split}
\end{align}

\section{Additional Materials}
\subsection{DSSW Distance with Non-parametric Projected Energy Function}
The DSSW distance with non-parametric projected energy function can be computed by multiplying the Wasserstein distance on each direction with the corresponding direction weight, and the corresponding direction weight can be obtained by converting the corresponding Wasserstein distance into it by the non-parametric projected energy function. The non-parametric projected energy function can be the exponential function, the identity function or the polynomial function. We provide the pseudo-code for computing DSSW distance with non-parametric projected energy function in Algorithm \ref{alg:ebssw_non_parametric}.

\begin{algorithm*}[ht]
	\caption{DSSW Distance with Non-parametric Projected Energy Function}
	\label{alg:ebssw_non_parametric}
	\textbf{Input}: $\left \{ x_i \right \} _{i=1}^{n}  \sim \mu$, $\left \{ y_j \right \} _{i=1}^{m} \sim \nu$, $L$ the number of projections, $p$ the order, $f$ the non-parametric projected energy function, $g$ the exponential function, the identity function or the polynomial function.
	\begin{algorithmic}[1] 
		\FOR{$\ell=1$ {\bfseries to} $L$}
		\STATE Generate a random matrix $Z\in\mathbb{R}^{d\times 2}$ with each element following the standard normal distribution: $\forall i,j, \ Z_{i,j}\sim\mathcal{N}(0,1)$.
		\STATE Perform QR decomposition on matrix $Z$: $U=\mathrm{QR}(Z)$.
		\STATE Project the points to $S^1$: $\forall i,j,\ \hat{x}_i^{\ell}=\frac{U^T x_i}{\|U^Tx_i\|_2}$, $\hat{y}_j^\ell=\frac{U^T y_j}{\|U^T y_j\|_2}$.
		\STATE Compute the coordinates on the circle $S^1$: $\forall i,j,\ \Tilde{x}_{i}^\ell = (\pi+\mathrm{atan2}(-x_{i,2}, -x_{i,1}))/(2\pi)$, $\Tilde{y}_j^\ell = (\pi + \mathrm{atan2}(-y_{j,2}, -y_{j,1}))/(2\pi)$.
		\STATE Calculate $W_p^p(\frac{1}{n}\sum_{i=1}^n  \delta_{\Tilde{x}_i^\ell}, \frac{1}{m}\sum_{j=1}^m \delta_{\Tilde{y}_j^\ell})$ by the binary search algorithm or the level median formulation for $p=1$.
		\ENDFOR
		\FOR{$\ell=1$ {\bfseries to} $L$}
		\STATE Compute the weights of the corresponding projection directions: $f\left(\frac{1}{n}\sum_{i=1}^n  \delta_{\Tilde{x}_i^\ell}, \frac{1}{m}\sum_{j=1}^m \delta_{\Tilde{y}_j^\ell}\right) :=\frac{g \left ( W_p^p(\frac{1}{n}\sum_{i=1}^n  \delta_{\Tilde{x}_i^\ell}, \frac{1}{m}\sum_{j=1}^m \delta_{\Tilde{y}_j^\ell} \right )}{ {\textstyle \sum_{k=1}^L} g \left ( W_p^p(\frac{1}{n}\sum_{i=1}^n  \delta_{\Tilde{x}_i^k}, \frac{1}{m}\sum_{j=1}^m \delta_{\Tilde{y}_j^k} \right )}$.
		\ENDFOR
		\STATE \textbf{return} $\frac{1}{L}\sum_{\ell=1}^L f\left(\frac{1}{n}\sum_{i=1}^n  \delta_{\Tilde{x}_i^\ell}, \frac{1}{m}\sum_{j=1}^m \delta_{\Tilde{y}_j^\ell}\right) \cdot W_p^p(\frac{1}{n}\sum_{i=1}^n \delta_{\Tilde{x}_i^\ell}, \frac{1}{m}\sum_{j=1}^m \delta_{\Tilde{y}_j^\ell}) $.
	\end{algorithmic}
\end{algorithm*}

\subsection{DSSW Distance with Parametric Projected Energy Function}
The DSSW distance with parametric projected energy function can be computed by multiplying the Wasserstein distance on each direction with the corresponding direction weight, and the corresponding direction weight can be obtained by converting the corresponding direction projection into it by the parameterized neural network. The parameterized neural network can be the linear neural network, nonlinear neural network, or attention mechanism. We provide the pseudo-code for computing DSSW distance with parametric projected energy function in Algorithm \ref{alg:ebssw_parametric}.

\begin{algorithm*}[ht]
	\caption{DSSW Distance with Parametric Projected Energy Function}
	\label{alg:ebssw_parametric}
	\textbf{Input}: $\left \{ x_i \right \} _{i=1}^{n}  \sim \mu$, $\left \{ y_j \right \} _{i=1}^{m} \sim \nu$, $L$ the number of projections, $p$ the order, $h_{\psi}$ the parameterized neural network, $T$ the maximum epochs for training $h_{\psi}$, $\alpha$ the learning rate for training $h_{\psi}$, $g(x)=e^x$.
	\begin{algorithmic}[1] 
		\FOR{$\ell=1$ {\bfseries to} $L$}
		\STATE Generate a random matrix $Z\in\mathbb{R}^{d\times 2}$ with each element following the standard normal distribution: $\forall i,j, \ Z_{i,j}\sim\mathcal{N}(0,1)$.
		\STATE Perform QR decomposition on matrix $Z$: $U=\mathrm{QR}(Z)$.
		\STATE Project the points to $S^1$: $\forall i,j,\ \hat{x}_i^{\ell}=\frac{U^T x_i}{\|U^Tx_i\|_2}$, $\hat{y}_j^\ell=\frac{U^T y_j}{\|U^T y_j\|_2}$.
		\STATE Compute the coordinates on the circle $S^1$: $\forall i,j,\ \Tilde{x}_{i}^\ell = (\pi+\mathrm{atan2}(-x_{i,2}, -x_{i,1}))/(2\pi)$, $\Tilde{y}_j^\ell = (\pi + \mathrm{atan2}(-y_{j,2}, -y_{j,1}))/(2\pi)$.
		\STATE Calculate $W_p^p(\frac{1}{n}\sum_{i=1}^n  \delta_{\Tilde{x}_i^\ell}, \frac{1}{m}\sum_{j=1}^m \delta_{\Tilde{y}_j^\ell})$ by the binary search algorithm or the level median formulation for $p=1$.
		\ENDFOR
		
		\STATE Initialize the parameterized neural network $h_{\psi}$.
		\FOR{$i=1$ {\bfseries to} $T$}
		\STATE Compute the optimization objective of $h_{\psi}$: $\mathcal{L}\left ( \psi \right )  = \frac{1}{L}\sum_{\ell=1}^L \frac{g \left ( h_{\psi } \left ( \frac{1}{n}\sum_{i=1}^n  \delta_{\Tilde{x}_i^\ell}, \frac{1}{m}\sum_{j=1}^m \delta_{\Tilde{y}_j^\ell} \right )  \right )}{ {\textstyle \sum_{k=1}^L} g \left (h_{\psi }\left ( \frac{1}{n}\sum_{i=1}^n  \delta_{\Tilde{x}_i^k}, \frac{1}{m}\sum_{j=1}^m \delta_{\Tilde{y}_j^k} \right )  \right )}    \cdot W_p^p(\frac{1}{n}\sum_{i=1}^n  \delta_{\Tilde{x}_i^\ell}, \frac{1}{m}\sum_{j=1}^m \delta_{\Tilde{y}_j^\ell})$.
		\STATE $\psi = \psi - \alpha \frac{\partial \mathcal{L}\left ( \psi \right )}{\partial \psi }$.
		\ENDFOR
		
		\FOR{$\ell=1$ {\bfseries to} $L$}
		\STATE Compute the weights of the corresponding projection directions: $f\left(\frac{1}{n}\sum_{i=1}^n  \delta_{\Tilde{x}_i^\ell}, \frac{1}{m}\sum_{j=1}^m \delta_{\Tilde{y}_j^\ell}\right) :=\frac{g \left ( W_p^p(\frac{1}{n}\sum_{i=1}^n  \delta_{\Tilde{x}_i^\ell}, \frac{1}{m}\sum_{j=1}^m \delta_{\Tilde{y}_j^\ell}) \right )}{ {\textstyle \sum_{k=1}^L} g \left ( W_p^p(\frac{1}{n}\sum_{i=1}^n  \delta_{\Tilde{x}_i^k}, \frac{1}{m}\sum_{j=1}^m \delta_{\Tilde{y}_j^k}) \right )}$.
		\ENDFOR
		
		\STATE \textbf{return} $\frac{1}{L}\sum_{\ell=1}^L f\left(\frac{1}{n}\sum_{i=1}^n  \delta_{\Tilde{x}_i^\ell}, \frac{1}{m}\sum_{j=1}^m \delta_{\Tilde{y}_j^\ell}\right) \cdot W_p^p(\frac{1}{n}\sum_{i=1}^n \delta_{\Tilde{x}_i^\ell}, \frac{1}{m}\sum_{j=1}^m \delta_{\Tilde{y}_j^\ell}) $.
	\end{algorithmic}
\end{algorithm*}

\subsection{Network Architecture of $h_{\psi }$}
In this subsection we introduce three types of network structure used in our propsoed DSSW. The details of the three types of network structure are as follows:
\subsubsection{Linear Neural Network} The linear neural network used in our propsoed DSSW holds a linear layer. The structure of this linear neural network is shown in \ref{fig:network} \subref{linear}. It transforms the matrix with size of $L \times (m+n)$ into the vector with the size of $L$.
\subsubsection{Nonlinear Neural Network} The nonlinear neural network used in our propsoed DSSW holds a linear layer followd by two linear layers and an activation function of Sigmoid. The structure of this nonlinear neural network is shown in \ref{fig:network} \subref{nonlinear}. It is equivalent to append the nonlinearity to the aforesaid the type of the linear neural network, and this kind of nonlinearity is implemented by two linear layers and an activation function.
\subsubsection{Attention Mechanism} 
\begin{figure}[ht]
	\centering
	\subfloat[Linear]{
		\label{linear}
		\includegraphics[width=0.2\textwidth]{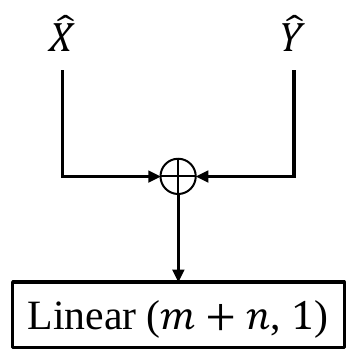}
	}
	\subfloat[Nonlinear]{
		\label{nonlinear}
		\includegraphics[width=0.2\textwidth]{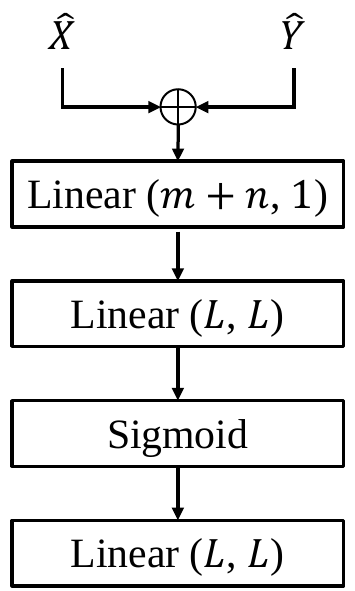}
	}
	\subfloat[Attention]{
		\label{attention}
		\includegraphics[width=0.5\textwidth]{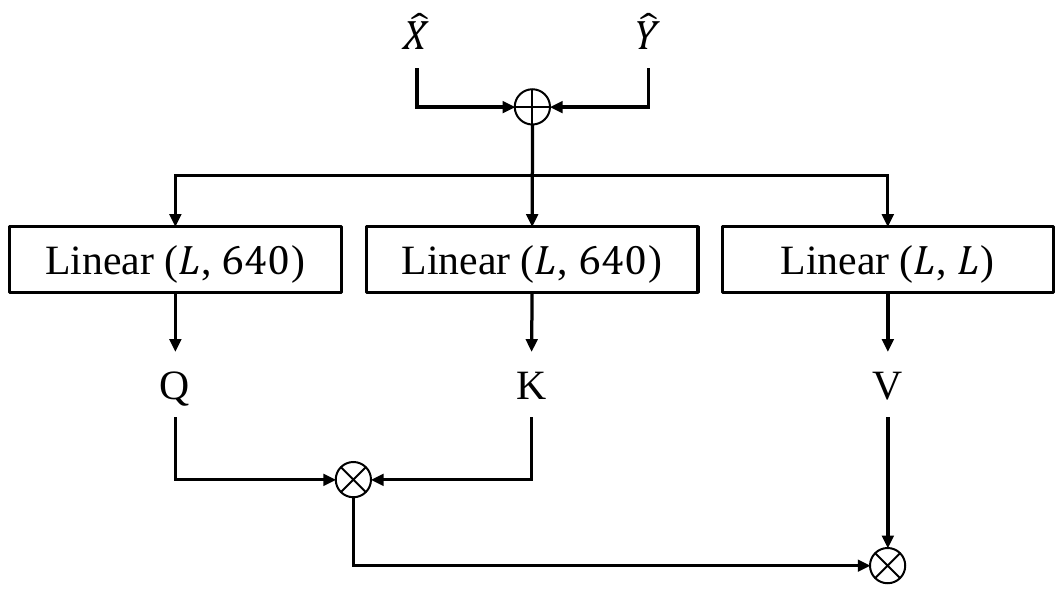}
	} 
	
	\caption{The network architecture of $h_{\psi }$ used in DSSW. $\hat{X}$ and $\hat{Y}$ denote the projection of $m$ samples from the source distribution $\mu$ with size of $L \times m$ and the projection of $n$ samples from the target distribution $\nu$ with size of $L \times n$, respectively. Notation "$\oplus$" indicates the concatenation operation, notation "$\otimes$" denotes the matrix multiplication.}
	\label{fig:network}
\end{figure}

The attention mechanism used in our propsoed DSSW is a simple version of attention mechanism. The structure of this attention mechanism is shown in \ref{fig:network} \subref{attention}. First $Q$, $K\in \mathbb{R}^{(m+n)\times 640}$, and $V\in \mathbb{R}^{(m+n)\times L}$ are obtained by a linear layer, respectively. Then the \textit{scaled dot-product} operator is performed as follows:
\begin{align}
	\mathrm{Attention}(Q,K,V)=\mathrm{softmax}_{\mathrm{row}}\left[\frac{QK^T}{\sqrt{640}}\right]V,
\end{align}
where $\mathrm{softmax}_{\mathrm{row}}$ denotes the row-wise softmax function.

\section{Additional Experiments}
In this section, we demonstrate the additional experimental results and analysis to reveal the effectiveness and superiority of our proposed method DSSW in comparison to SW, SSW and S3W distance.
\subsection{Runtime Comparison}
In this subsection we preform runtime comparisons on various distances between uniform distribution and a von Mises-Fisher distribution on $\mathbb{S}^{100}$. The comparison results are demonstrated in Figure \ref{fig:runtime_comparison} averaged over 50 iterations for changing sample sizes of each distribution, and $L=200$ projections are used for all sliced approaches. On Figure \ref{fig:runtime_comparison}, we can observe that the runtime curves of our DSSW (identity) and DSSW (poly) overlap with the original SSW, therefore, this phenomenon also verifies that the additional computing overhead introduced by our DSSW with the non-parametric projected energy function is negligible. Figure \ref{fig:runtime_comparison} also indicates that the computing overhead of our DSSW with the parametric projected energy function is more than the original SSW but it is less than Wasserstein distance. This is because the optimization of the parameterized neural network in DSSW needs less computational overhead.

\begin{figure}[ht]
	\centering
	\subfloat[DSSW (identity)]{
		\includegraphics[width=0.5\textwidth]{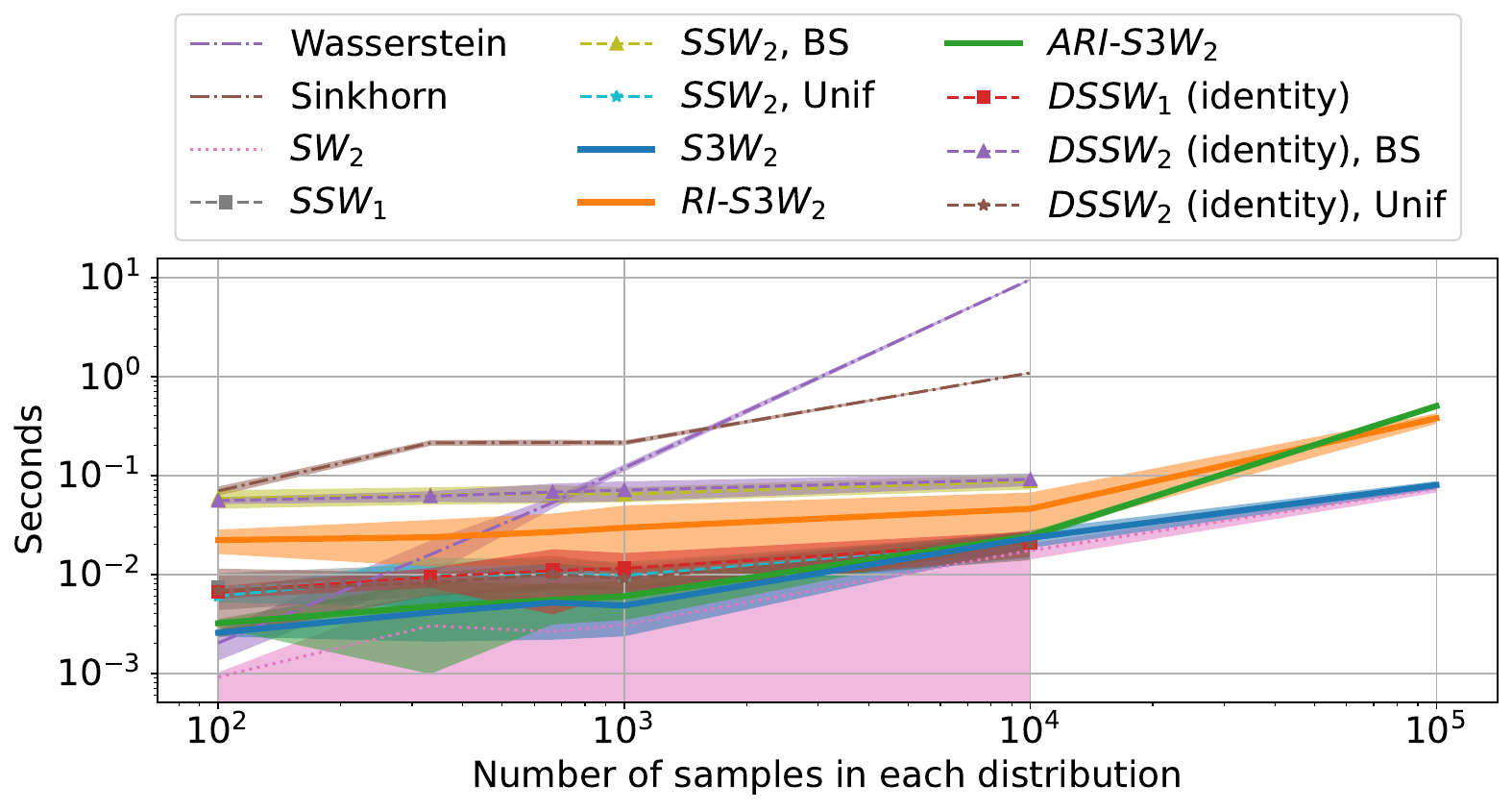}
	}
	\subfloat[DSSW (poly)]{
		\includegraphics[width=0.5\textwidth]{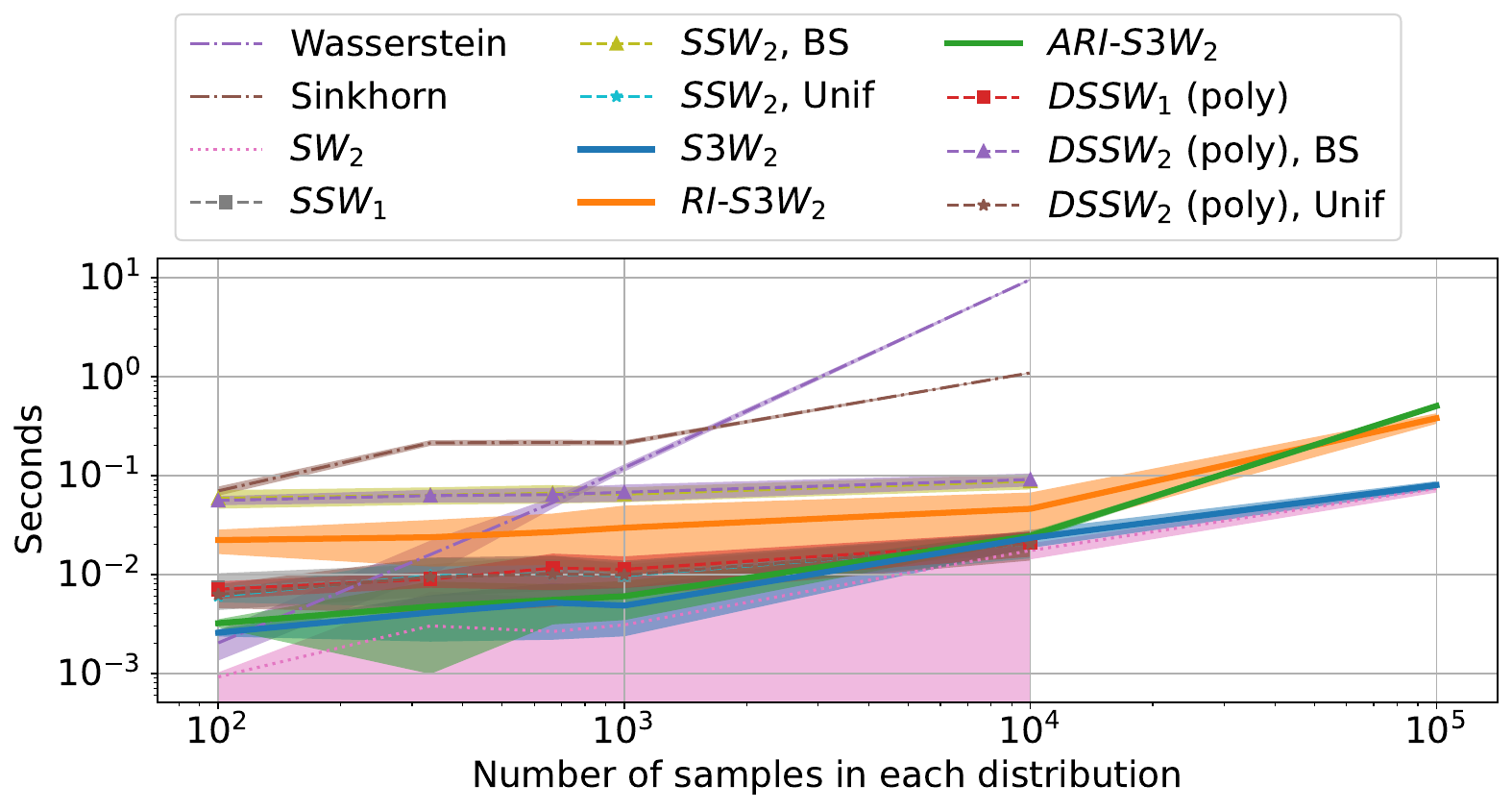}
	}\\ 
	\subfloat[DSSW (linear)]{
		\includegraphics[width=0.5\textwidth]{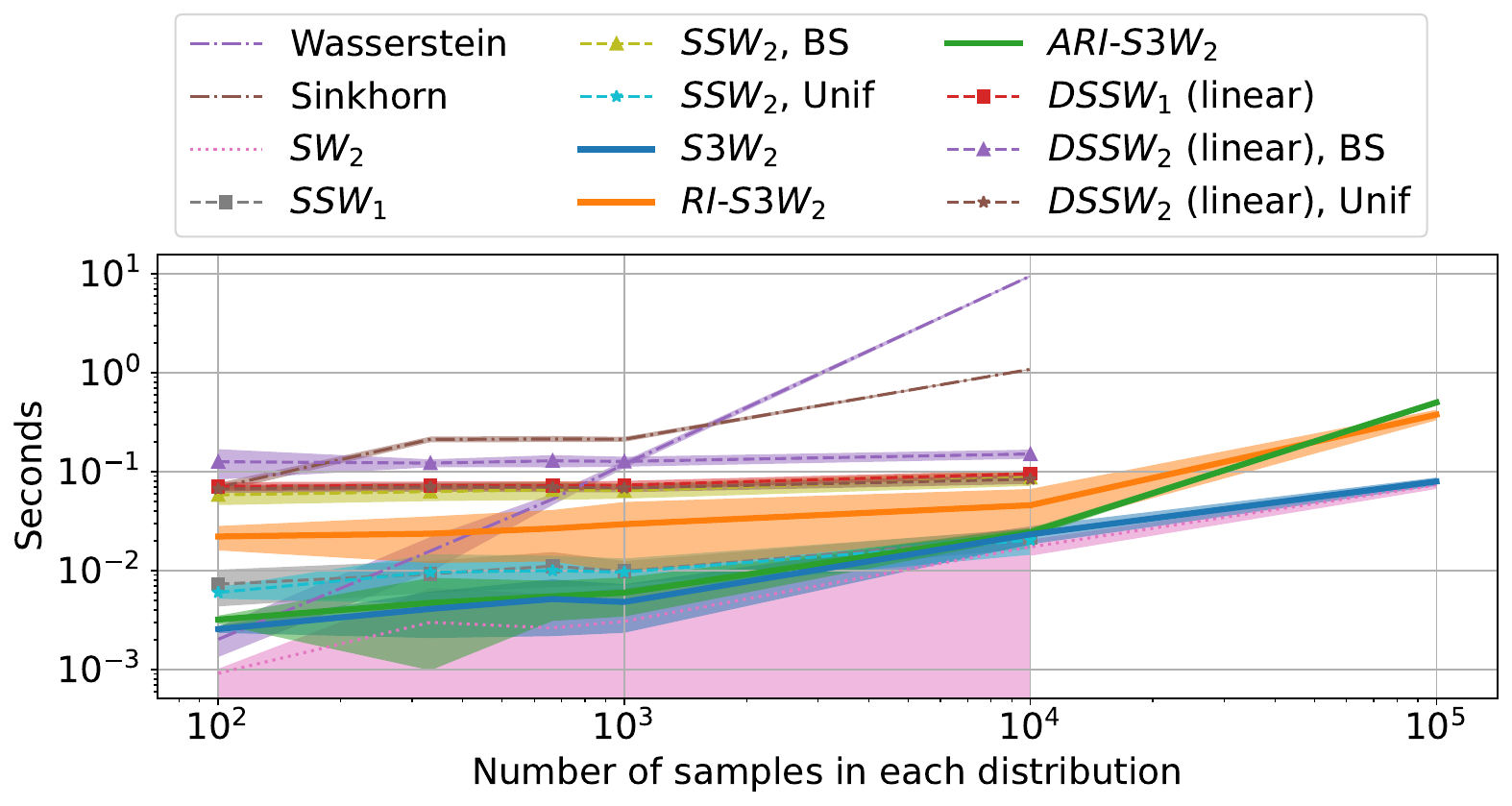}
	}
	\subfloat[DSSW (nonlinear)]{
		\includegraphics[width=0.5\textwidth]{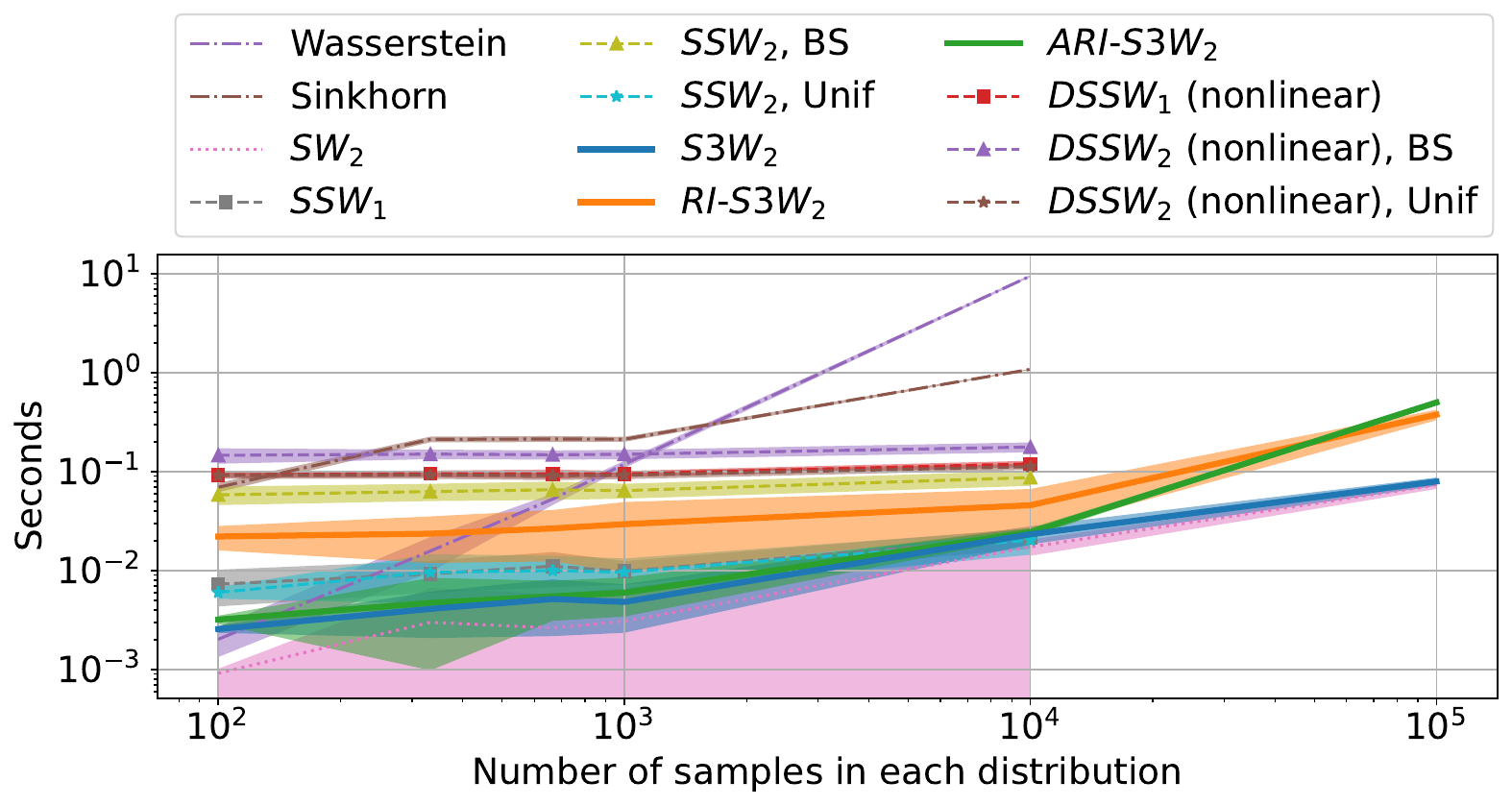}
	}\\ 
	\subfloat[DSSW (attention)]{
		\includegraphics[width=0.5\textwidth]{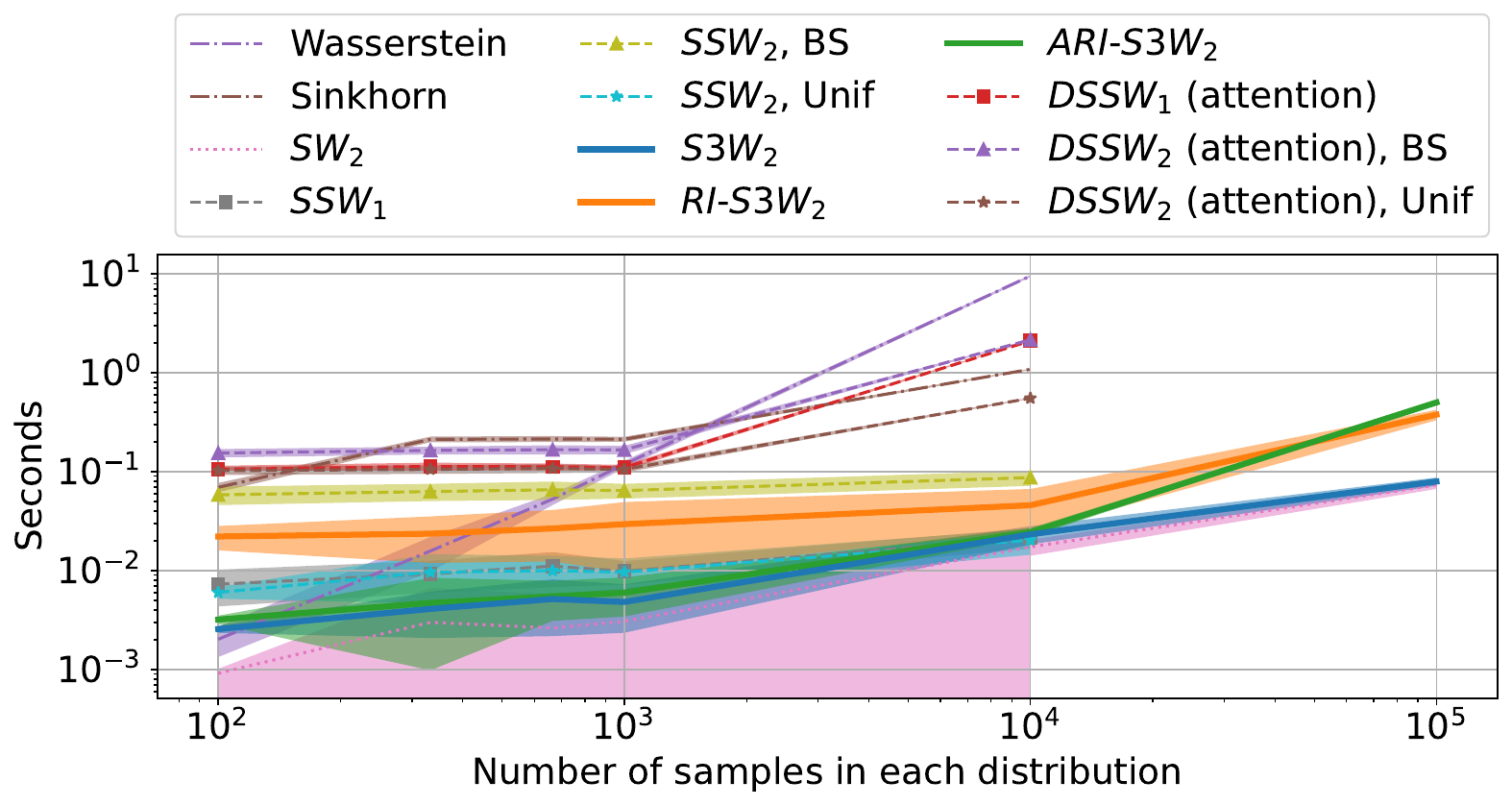}
	}
	
	\caption{Runtime comparison for Wasserstein distance, Sinkhorn distance with geodesic distance as cost function, $SW_2$ (SW distance), $SSW_1$ distance with the level median, $SSW_2$ distance with binary search (BS), $SSW_2$ distance against a uniform distribution (Unif), $S3W_2$ distance, $RI$-$S3W_2$ distance, $ARI$-$S3W_2$ distance, $DSSW_1$ (identity/poly/linear/nonlinear/attention) (ours), $DSSW_2$ (identity/poly/linear/nonlinear/attention), BS (ours), $DSSW_2$ (identity/poly/linear/nonlinear/attention), Unif (ours).}
	\label{fig:runtime_comparison}
\end{figure}

\subsection{Evolution of the DSSW Distance}
\subsubsection{Evolution of metrics w.r.t. $\kappa$ for Varying Dimensions} 
\begin{figure}[ht]
	\centering
	\subfloat[KL]{
		\includegraphics[width=0.3\textwidth]{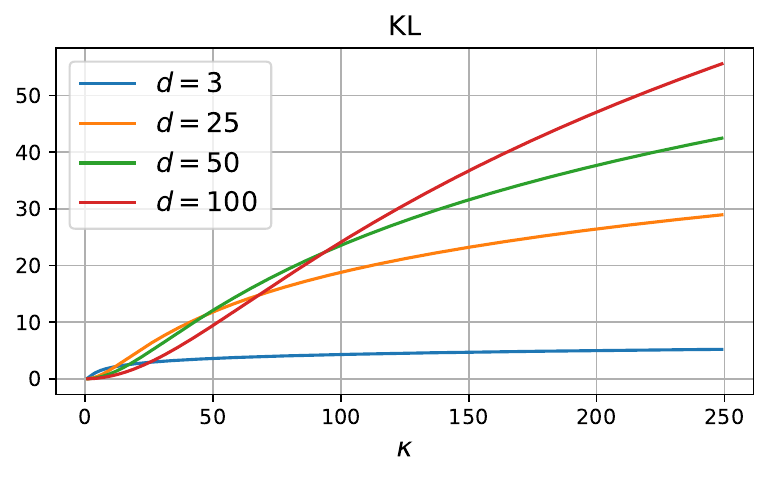}
	}
	\subfloat[SW]{
		\includegraphics[width=0.3\textwidth]{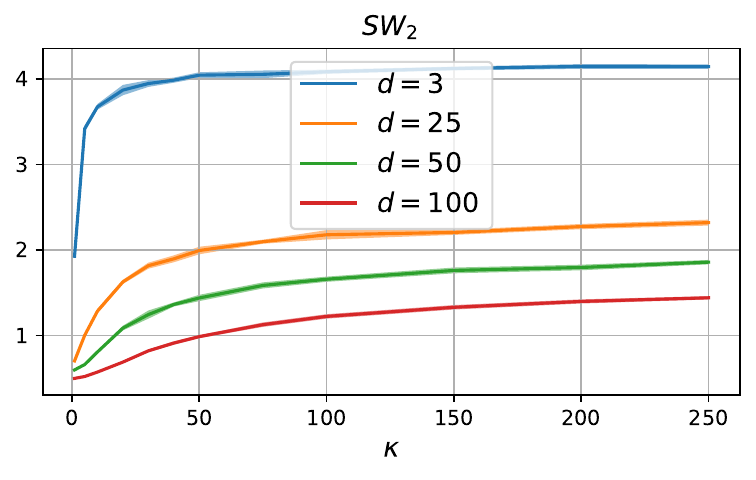}
	}
	\subfloat[SSW]{
		\includegraphics[width=0.3\textwidth]{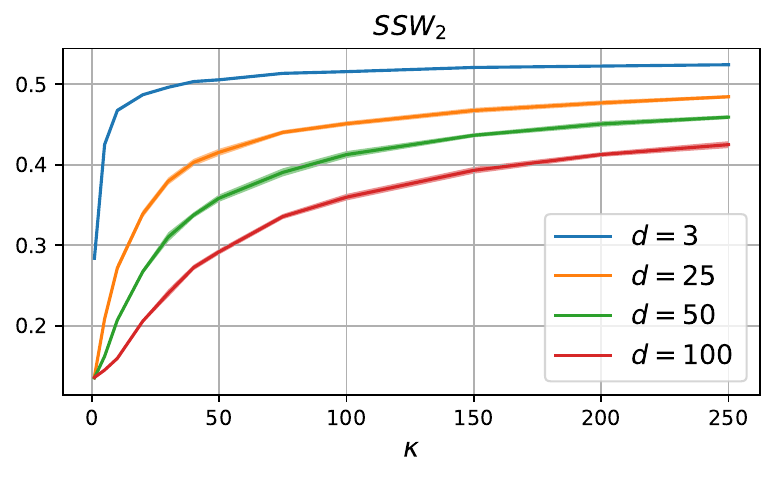}
	}\\ 
	\subfloat[S3W]{
		\includegraphics[width=0.3\textwidth]{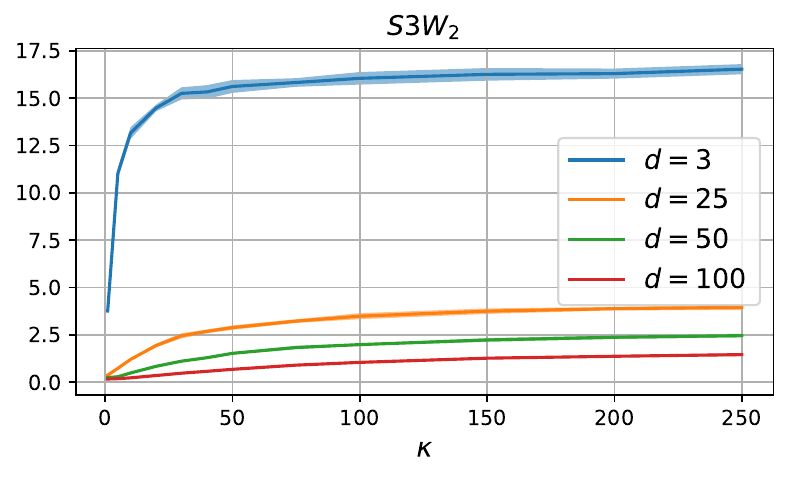}
	}
	\subfloat[RI-S3W]{
		\includegraphics[width=0.3\textwidth]{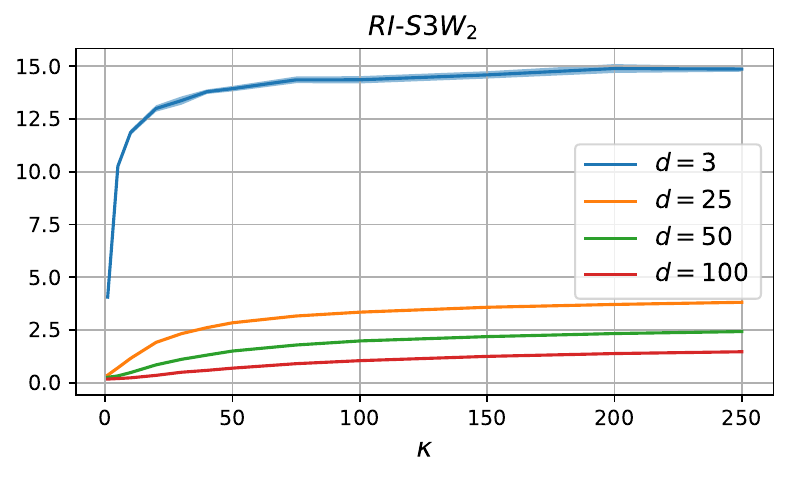}
	}
	\subfloat[ARI-S3W]{
		\includegraphics[width=0.3\textwidth]{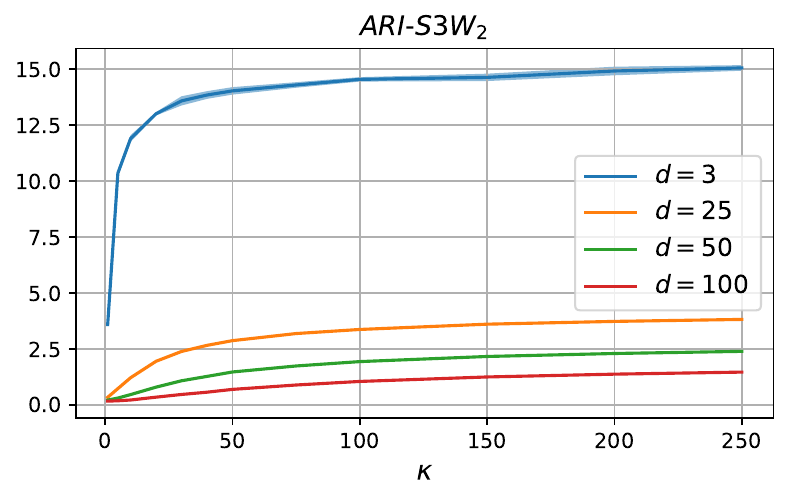}
	}\\ 
	\subfloat[DSSW (exp)]{
		\includegraphics[width=0.3\textwidth]{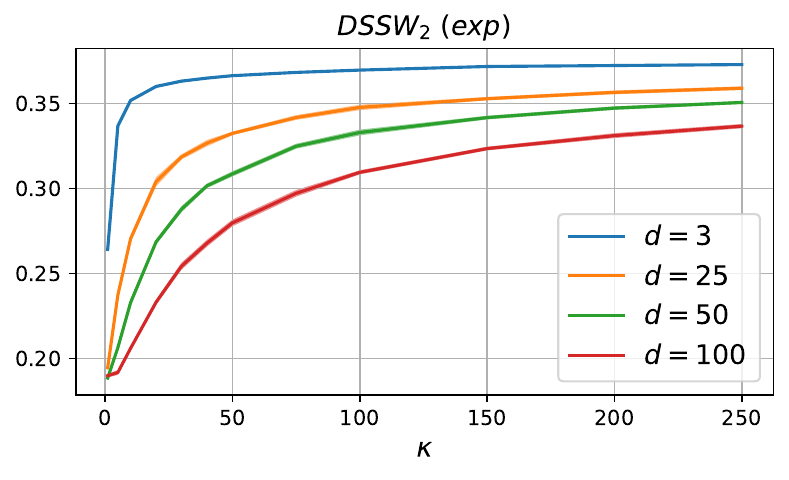}
	}
	\subfloat[DSSW (identity)]{
		\includegraphics[width=0.3\textwidth]{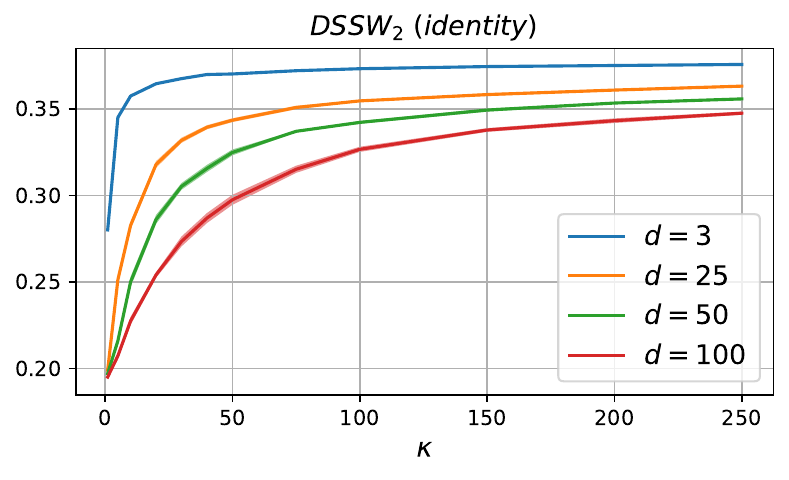}
	}
	\subfloat[DSSW (poly)]{
		\includegraphics[width=0.3\textwidth]{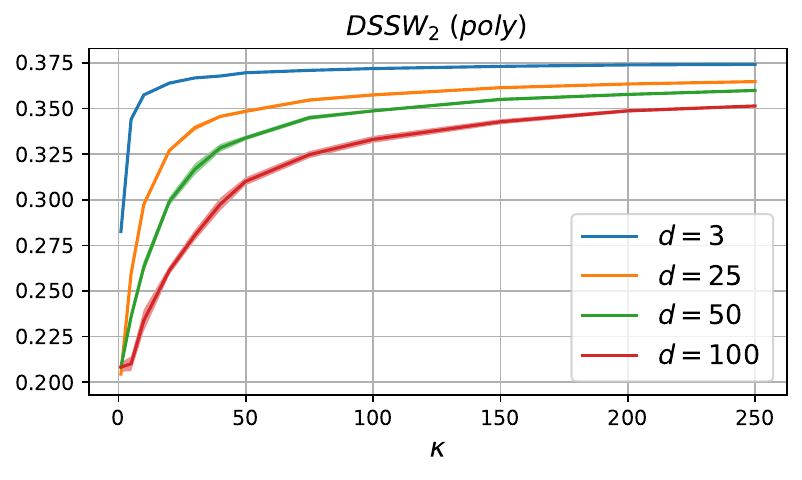}
	}\\ 
	\subfloat[DSSW (linear)]{
		\includegraphics[width=0.3\textwidth]{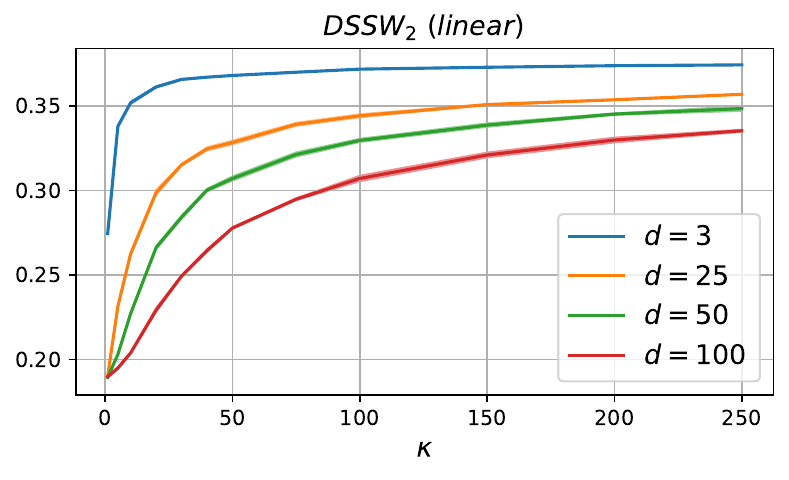}
	}
	\subfloat[DSSW (nonlinear)]{
		\includegraphics[width=0.3\textwidth]{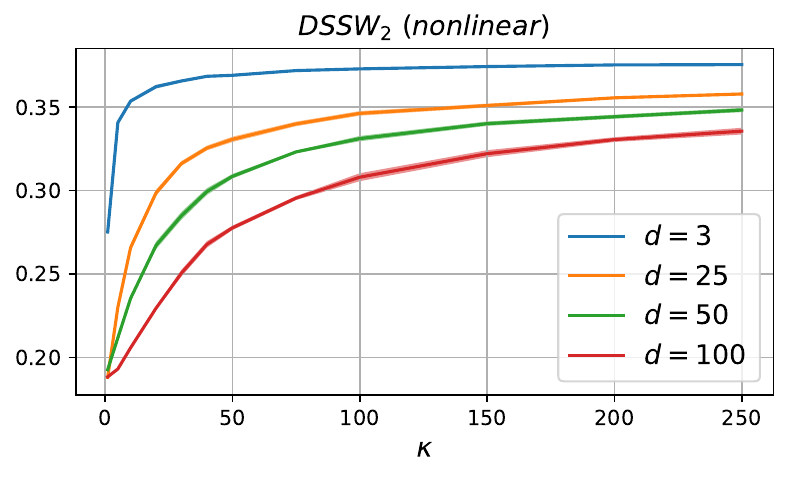}
	}
	\subfloat[DSSW (attention)]{
		\includegraphics[width=0.3\textwidth]{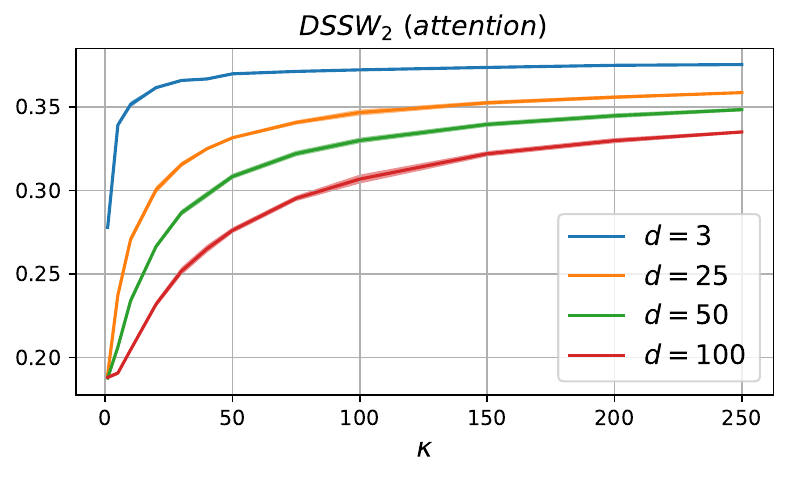}
	}
	
	\caption{Evolution between $\mathrm{vMF}\left ( \mu, \kappa  \right ) $ and $\mathrm{vMF}\left (\cdot , 0 \right ) $ w.r.t. $\kappa$ for varying dimensions. We employ $L=200$ projections for all distance metrics, $N_R=100$ rotations for RI-S3W and ARI-S3W, and pool size of 1000 for ARI-S3W. There are 500 samples in each distribution. We average each metric for $\kappa \in \left \{ 1, 5, 10, 20, 30, 40, 50, 75, 100, 150, 200, 250 \right \}$ over 10 iterations.}
	\label{fig:evolution_dimension_vMF}
\end{figure}

We demonstrate the evolution of KL divergence and various distances between $\mathrm{vMF}\left ( \mu, \kappa  \right ) $ and $\mathrm{vMF}\left (\cdot , 0 \right ) $ w.r.t. $\kappa$ for varying dimension in Figure \ref{fig:evolution_dimension_vMF}. It can be found that SSW gets lower with the dimension contrary to KL divergence, S3W, RI-S3W, ARI-S3W, and DSSW also follow the similar trend.

\subsubsection{Evolution of DSSW Distance w.r.t. Number of Projection} 
\begin{figure}[ht]
	\centering
	\subfloat[DSSW (exp)]{
		\includegraphics[width=0.45\textwidth]{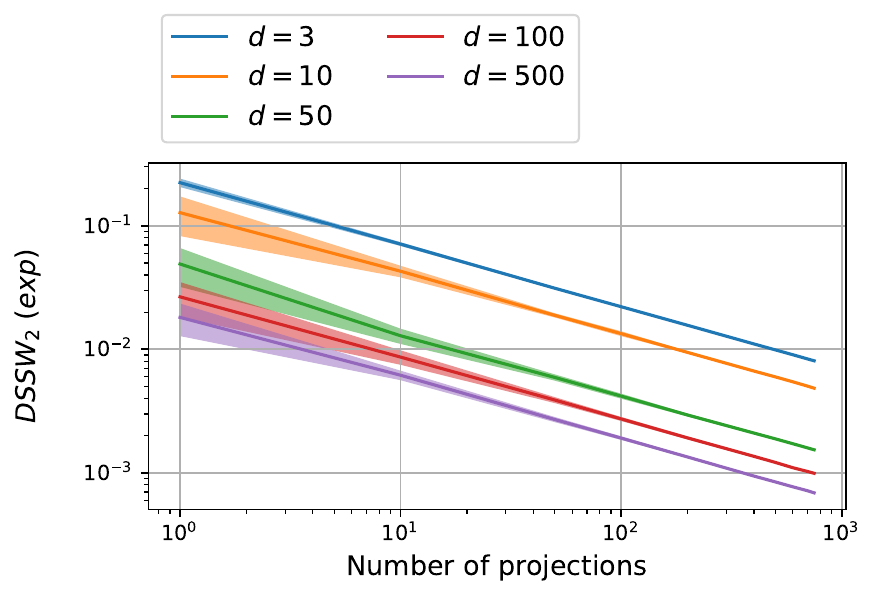}
		\includegraphics[width=0.45\textwidth]{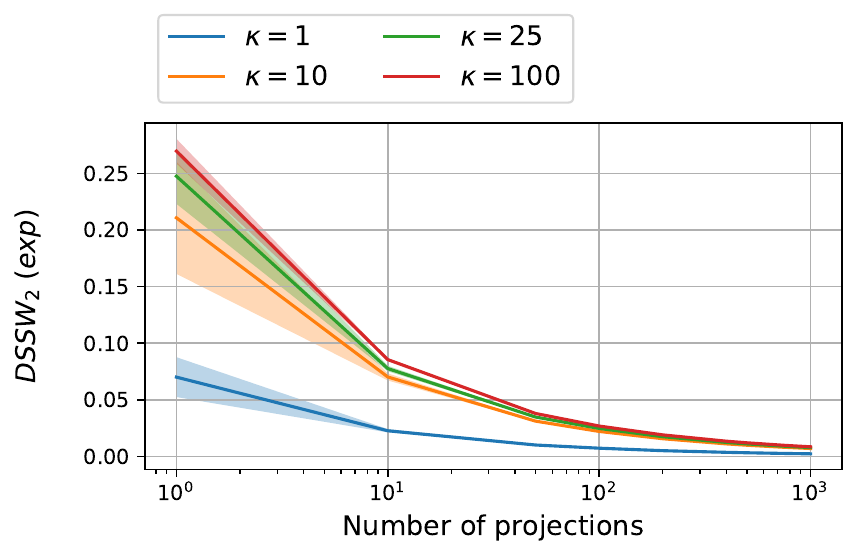}
	}\\ 
	\subfloat[DSSW (identity)]{
		\includegraphics[width=0.45\textwidth]{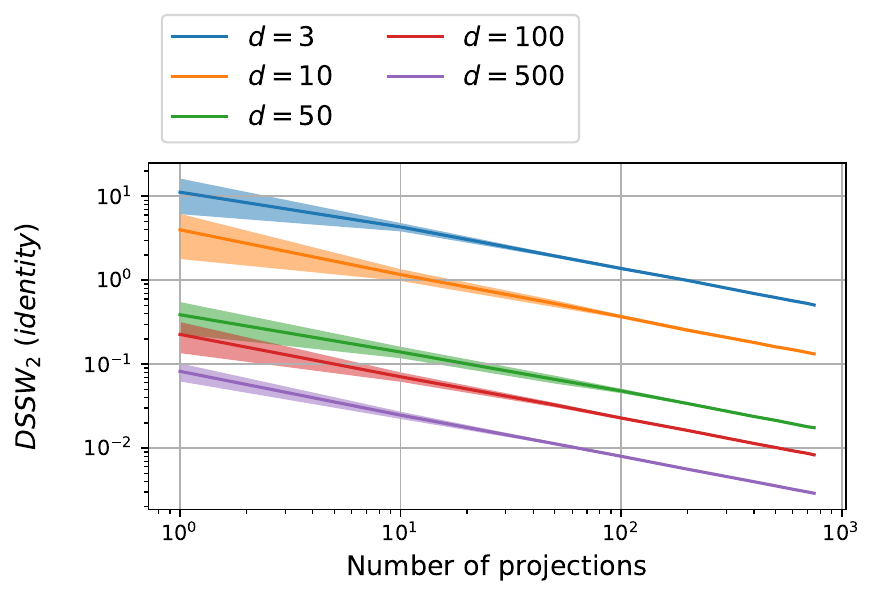}
		\includegraphics[width=0.45\textwidth]{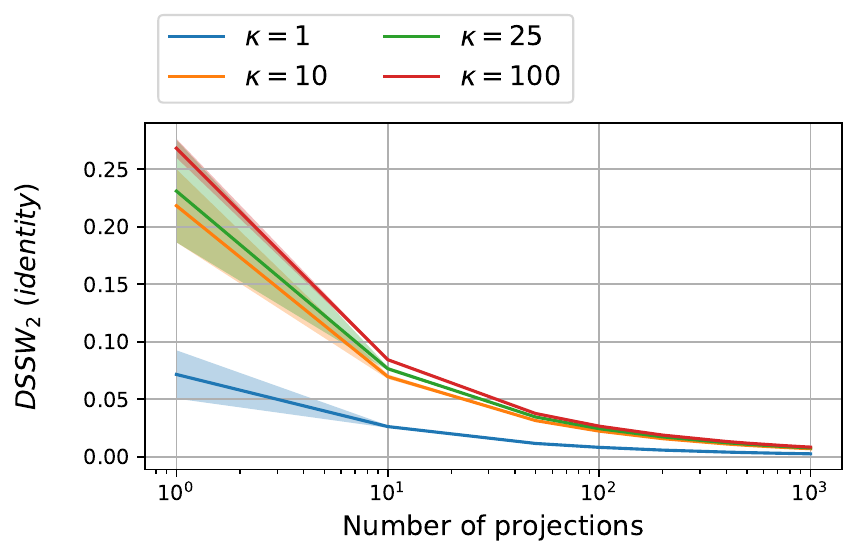}
	}\\ 
	\subfloat[DSSW (poly)]{
		\includegraphics[width=0.45\textwidth]{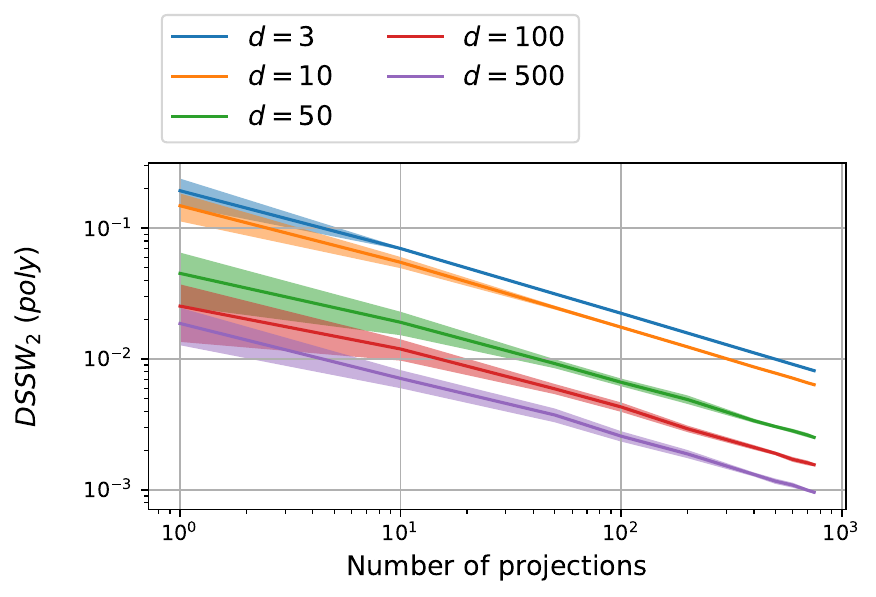}
		\includegraphics[width=0.45\textwidth]{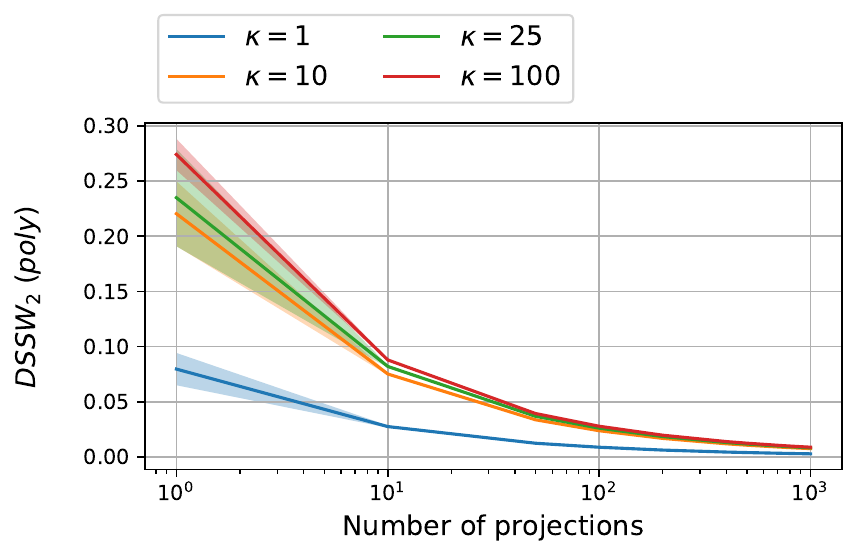}
	}
	
	\caption{Influence of the number of projections on computing DSSW with the non-parametric projected energy function. We calculate $SW_{2}^{2} \left ( \mathrm{vMF}\left ( \mu, \kappa  \right )\left |  \right | \mathrm{vMF}\left (\cdot , 0 \right ) \right )$ over 20 iterations with 500 samples. For the plot on the left column, we set $\kappa=10$, and for the plot on the right column we adopt $d=3$.}
	\label{fig:evolution_projection_ebssw_non_parametric}
\end{figure}

\begin{figure}[ht]
	\centering
	\subfloat[DSSW (linear)]{
		\includegraphics[width=0.45\textwidth]{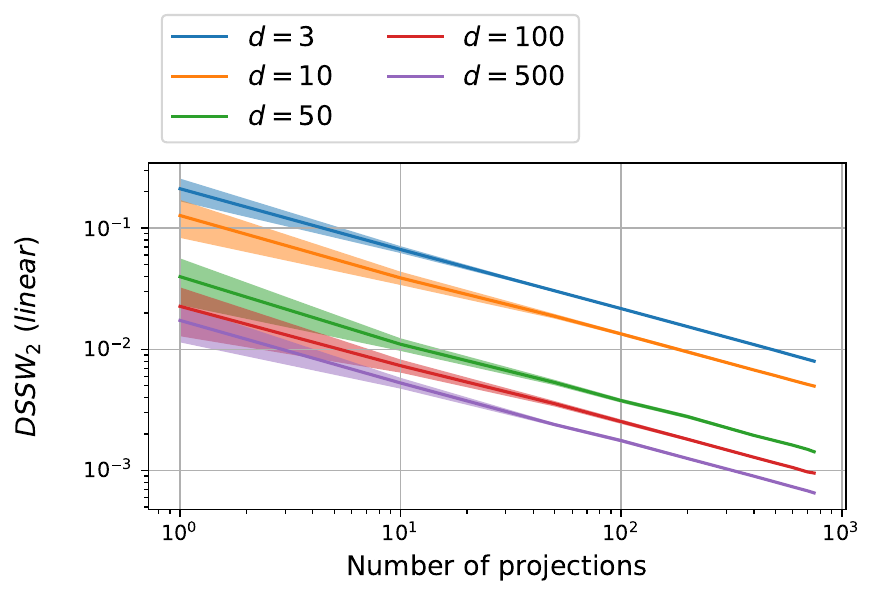}
		\includegraphics[width=0.45\textwidth]{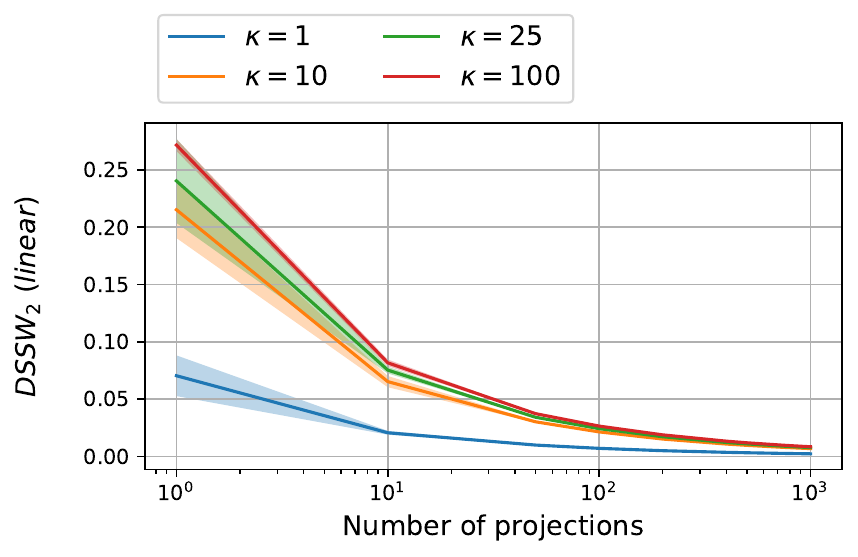}
	}\\ 
	\subfloat[DSSW (nonlinear)]{
		\includegraphics[width=0.45\textwidth]{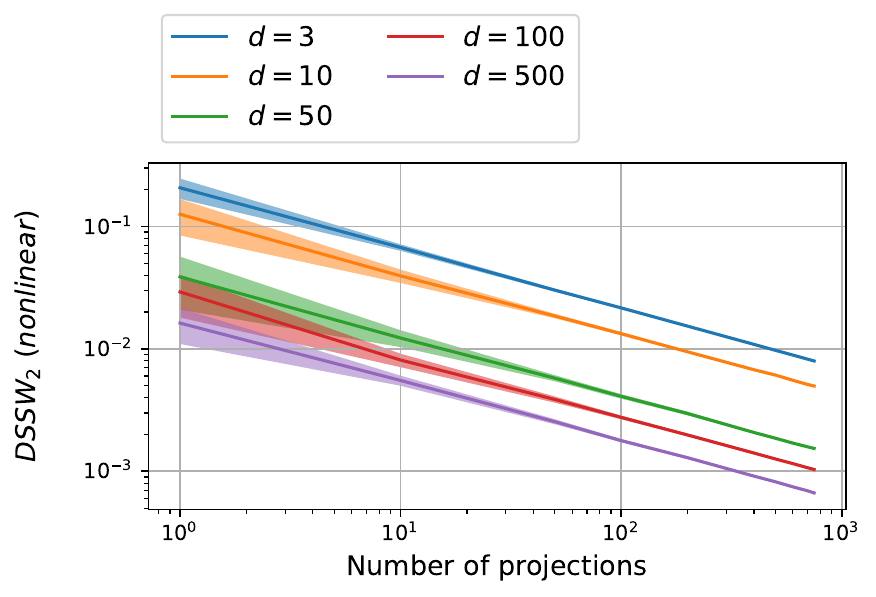}
		\includegraphics[width=0.45\textwidth]{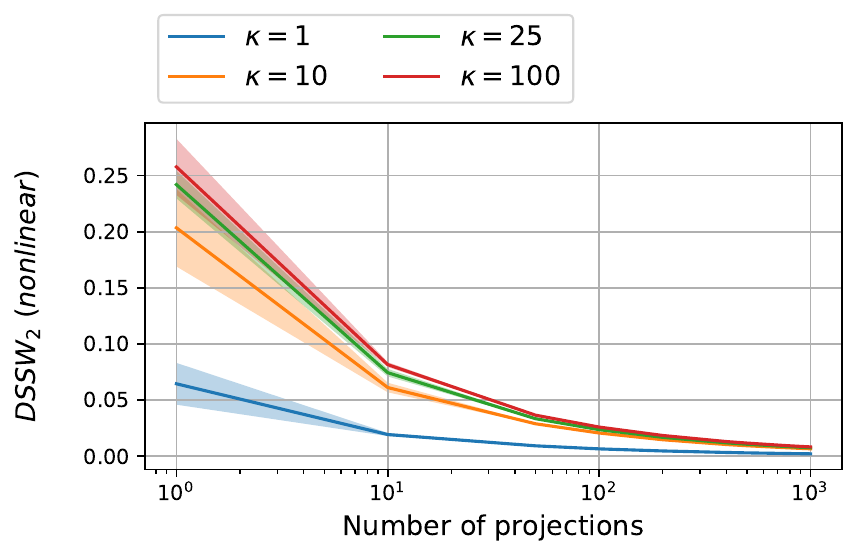}
	}\\ 
	\subfloat[DSSW (attention)]{
		\includegraphics[width=0.45\textwidth]{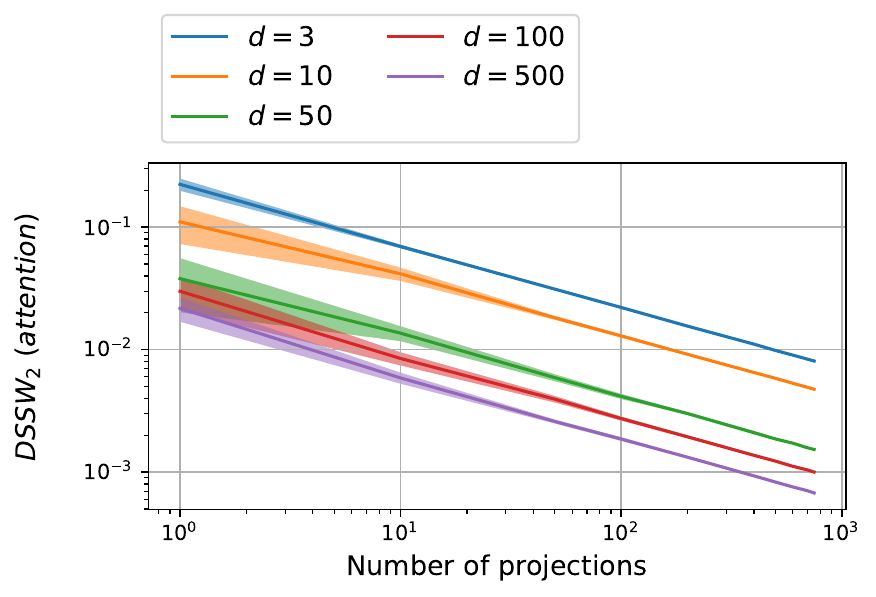}
		\includegraphics[width=0.45\textwidth]{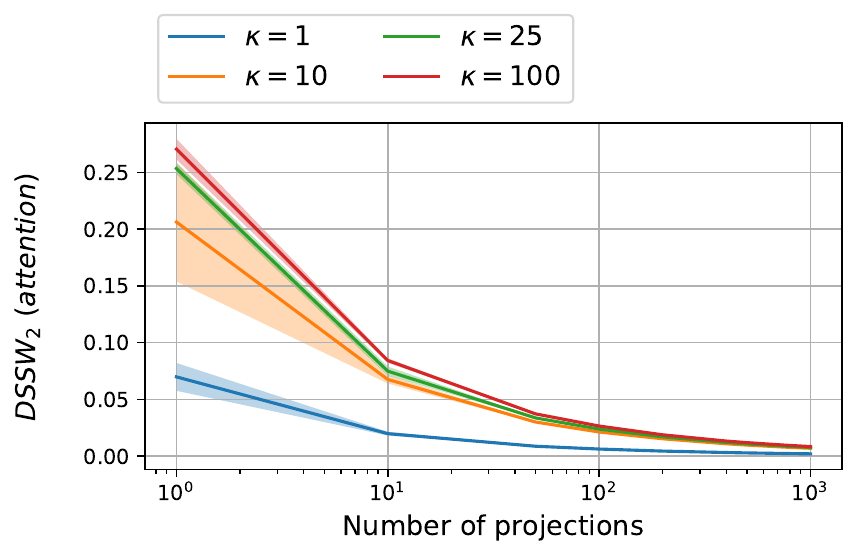}
	}
	
	\caption{Influence of the number of projections on computing DSSW with the parametric projected energy function. We calculate $SW_{2}^{2} \left ( \mathrm{vMF}\left ( \mu, \kappa  \right )\left |  \right | \mathrm{vMF}\left (\cdot , 0 \right ) \right )$ 20 times with 500 samples. For the plot on the left column, we set $\kappa=10$, and for the plot on the right column we adopt $d=3$.}
	\label{fig:evolution_projection_ebssw_parametric}
\end{figure}

We demonstrate  the evolution of DSSW between $\mathrm{vMF}\left ( \mu, \kappa  \right ) $ and $\mathrm{vMF}\left (\cdot , 0 \right ) $ w.r.t. varying projections in Figure \ref{fig:evolution_projection_ebssw_non_parametric} and \ref{fig:evolution_projection_ebssw_parametric}. It demonstrates that beyond $L=100$ projections, the variance of SSW, S3W, and DSSW becomes negligible across different dimensions and values of $\kappa$.

\subsubsection{Evolution of DSSW Distance w.r.t. Number of Rotations} We also perform evolution of various distance metrics between a fixed vMF distribution and the rotation of this distribution along a great circle. We observe similar behavior between all metrics, with each distance being maximal between $\mathrm{vMF}\left ( \mu_0, \kappa_0 \right )$ and $\mathrm{vMF}\left (-\mu_0, \kappa_0 \right )$ when $\theta=\pi$.

\begin{figure}[ht]
	\centering
	\subfloat[SW]{
		\includegraphics[width=0.3\textwidth]{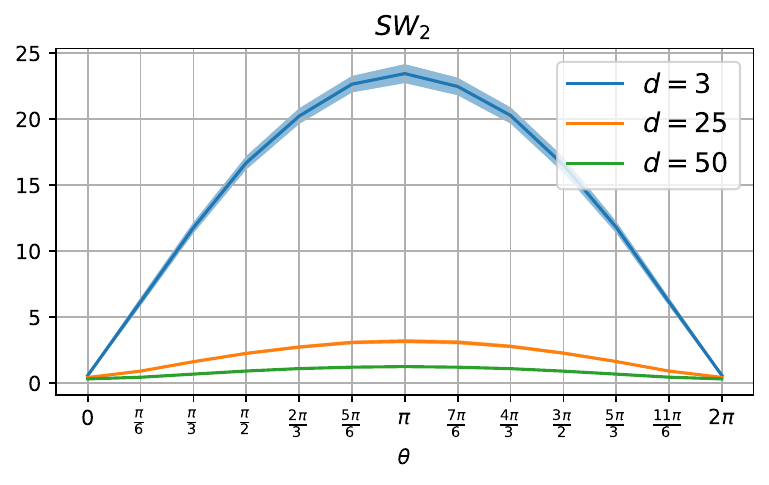}
	}
	\subfloat[SSW]{
		\includegraphics[width=0.3\textwidth]{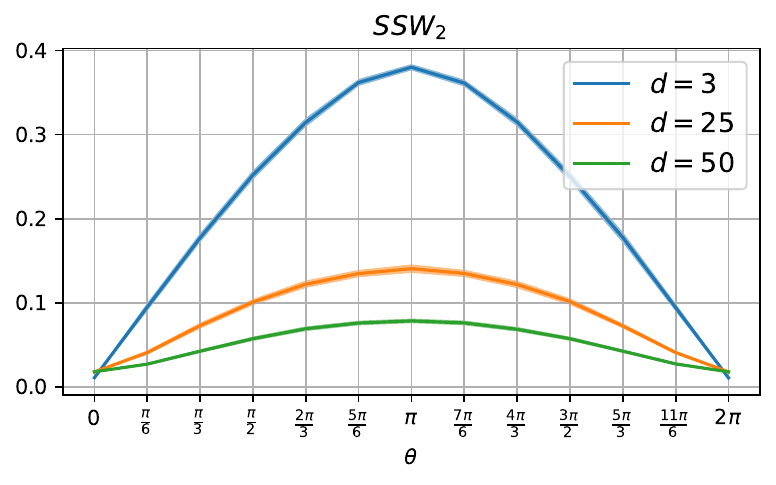}
	}
	\subfloat[S3W]{
		\includegraphics[width=0.3\textwidth]{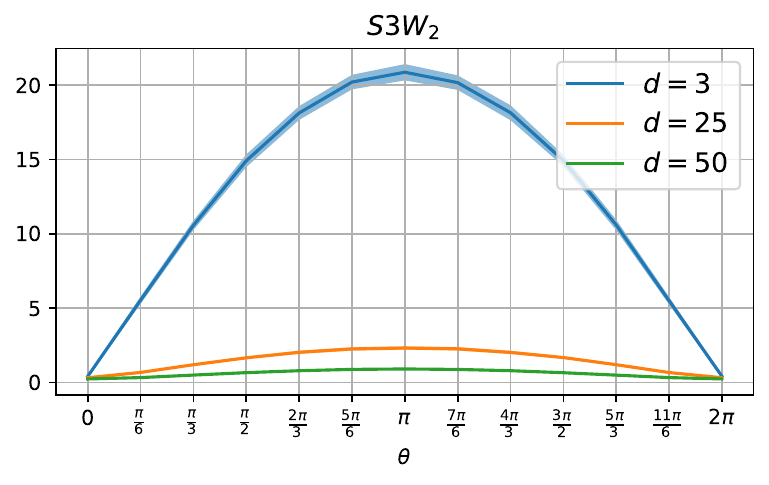}
	}\\ 
	\subfloat[DSSW (exp)]{
		\includegraphics[width=0.3\textwidth]{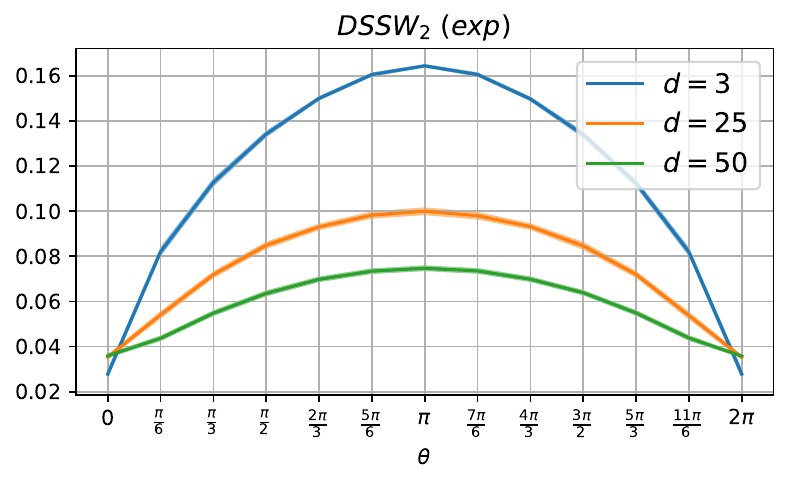}
	}
	\subfloat[DSSW (identity)]{
		\includegraphics[width=0.3\textwidth]{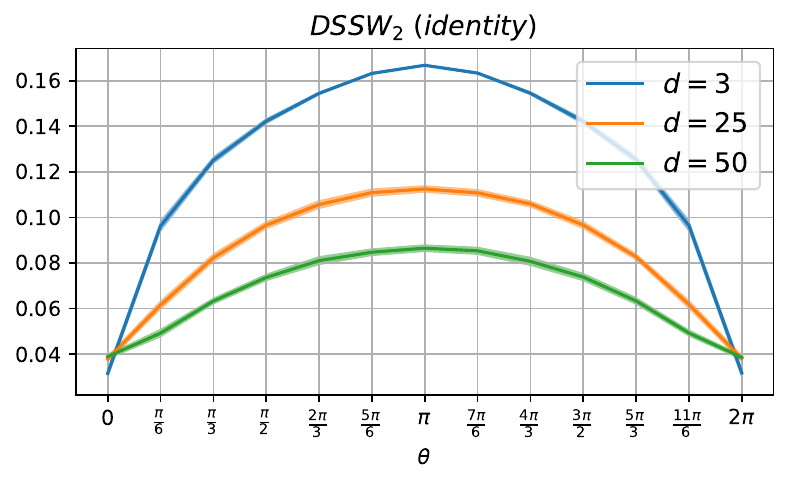}
	}
	\subfloat[DSSW (poly)]{
		\includegraphics[width=0.3\textwidth]{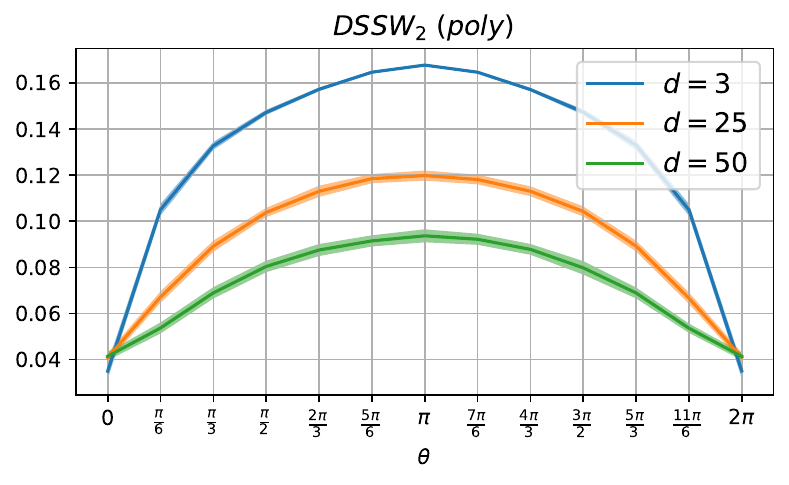}
	}\\ 
	\subfloat[DSSW (linear)]{
		\includegraphics[width=0.3\textwidth]{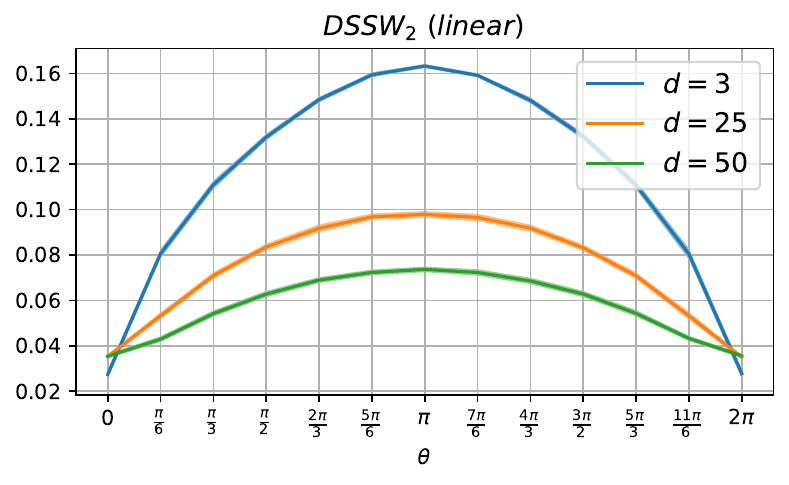}
	}
	\subfloat[DSSW (nonlinear)]{
		\includegraphics[width=0.3\textwidth]{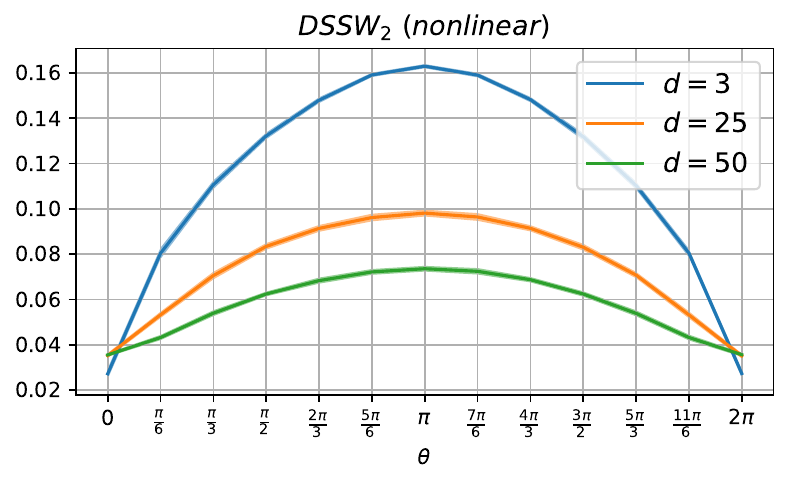}
	}
	\subfloat[DSSW (attention)]{
		\includegraphics[width=0.3\textwidth]{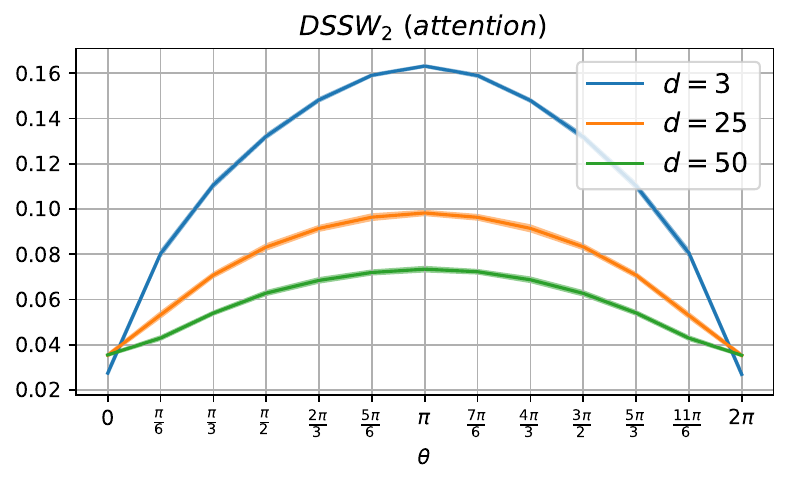}
	}
	
	\caption{Evolution between vMF distributions on $\mathbb{S}^{d-1}$. We employ $L=200$ projections for all distance metrics, $N_R=100$ rotations for RI-S3W and ARI-S3W, and pool size of 1000 for ARI-S3W. We calculate average each metric with 500 samples for each distribution over 100 iterations.}
	\label{fig:evolution_rotation_vMF}
\end{figure}

\subsection{Runtime Analysis}
\begin{figure}[ht]
	\centering
	\subfloat[DSSW (exp)]{
		\includegraphics[width=0.45\textwidth]{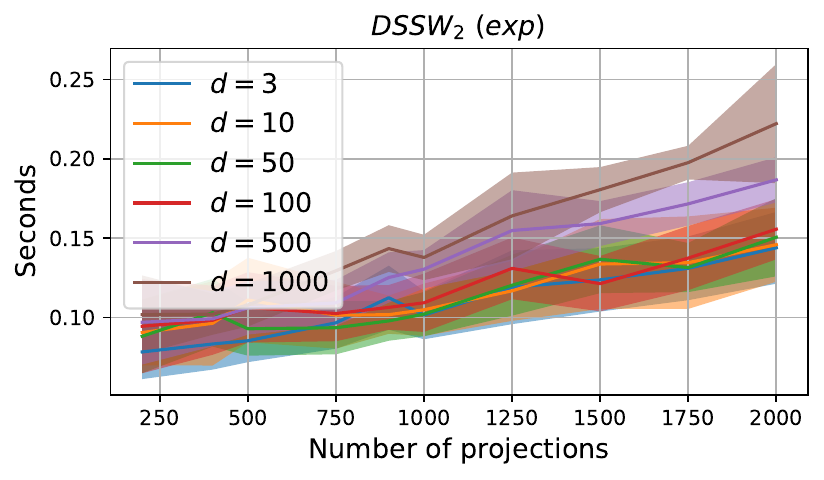}
		\includegraphics[width=0.45\textwidth]{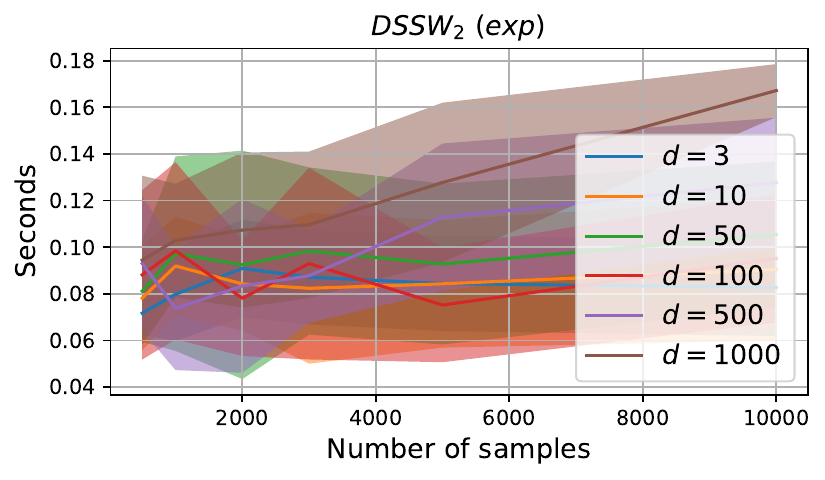}
	}\\
	\subfloat[DSSW (identity)]{
		\includegraphics[width=0.45\textwidth]{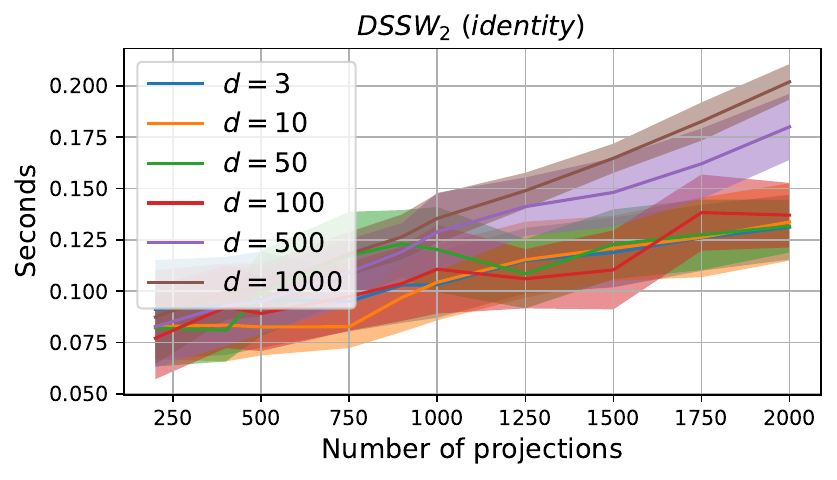}
		\includegraphics[width=0.45\textwidth]{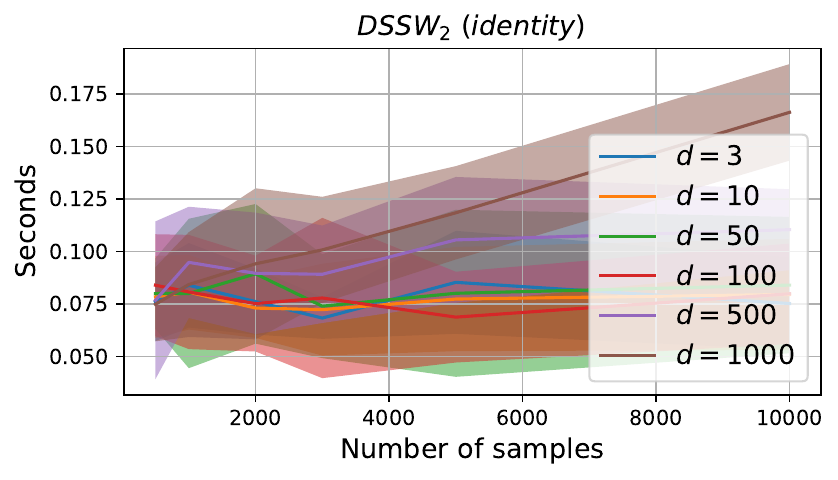}
	}\\
	\subfloat[DSSW (poly)]{
		\includegraphics[width=0.45\textwidth]{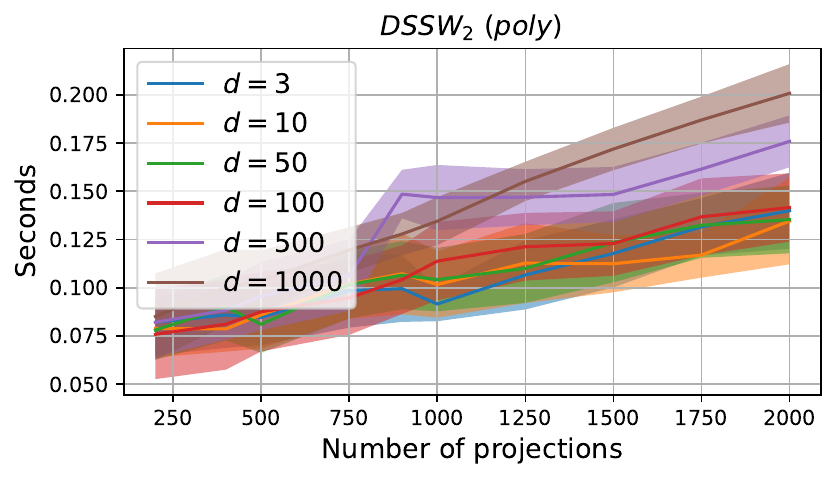}
		\includegraphics[width=0.45\textwidth]{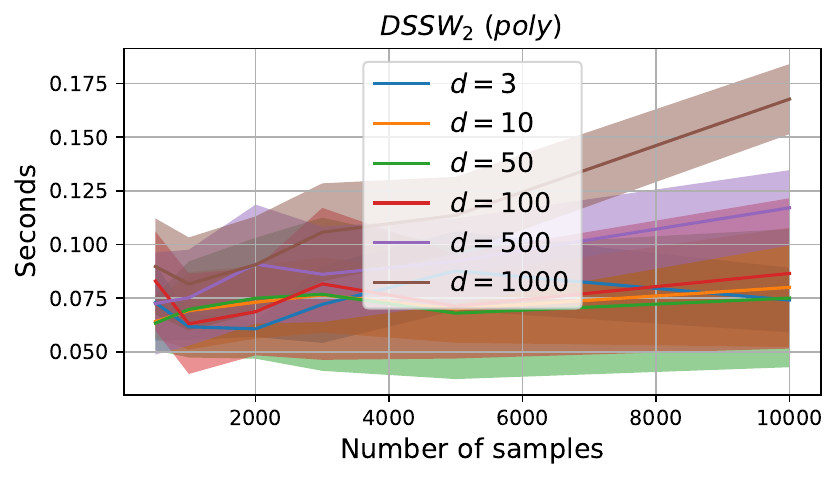}
	}
	\caption{Computation time of DSSW with the non-parametric projected energy function w.r.t. different number of projections and samples. For the plot on the left, we fix $N=500$ for each distribution, and the plot on the right we use $L=100$ projections.}
	\label{fig:runtime_ebssw_non_parametric}
\end{figure}

\begin{figure}[ht]
	\centering
	\subfloat[DSSW (linear)]{
		\includegraphics[width=0.45\textwidth]{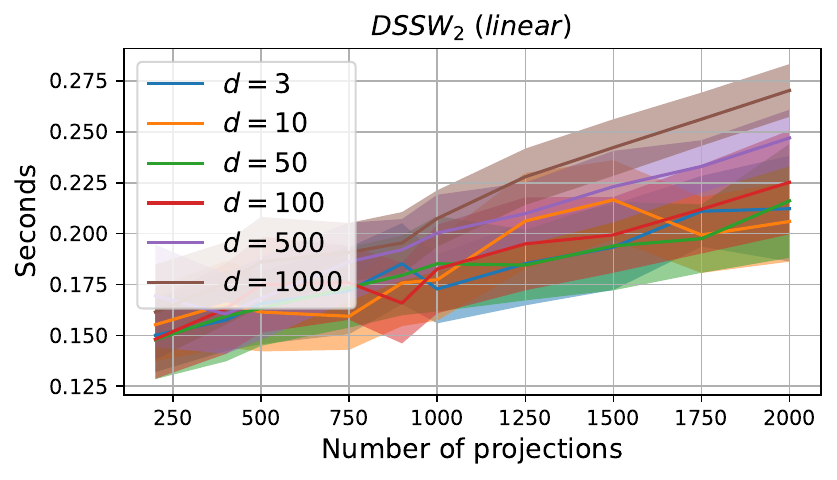}
		\includegraphics[width=0.45\textwidth]{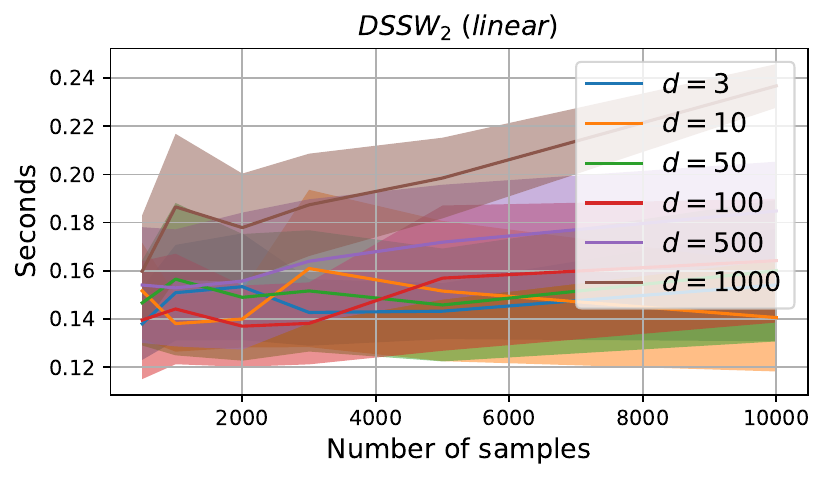}
	}\\ 
	\subfloat[DSSW (nonlinear)]{
		\includegraphics[width=0.45\textwidth]{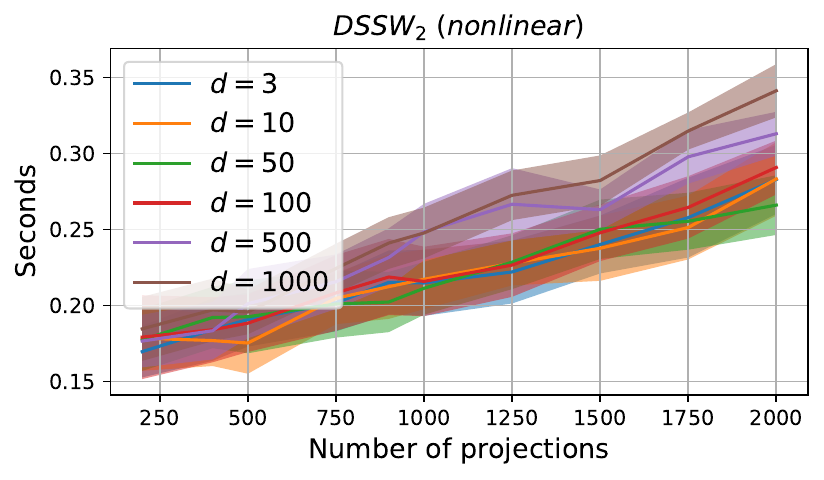}
		\includegraphics[width=0.45\textwidth]{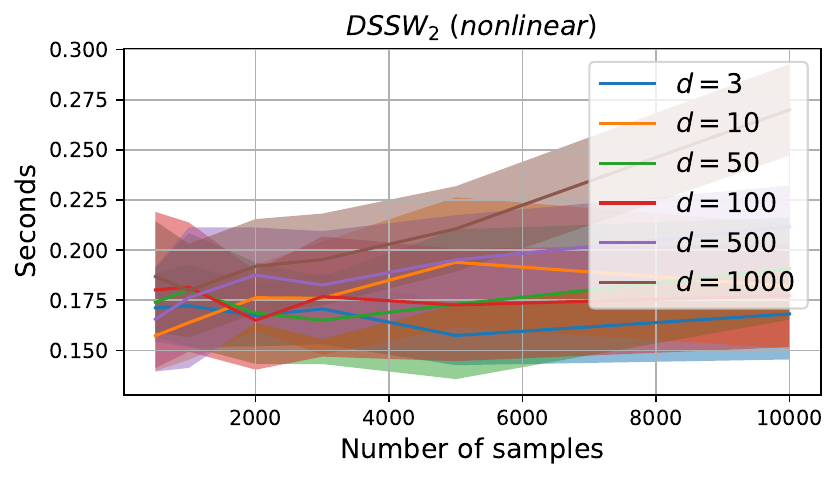}
	}\\ 
	\subfloat[DSSW (attention)]{
		\includegraphics[width=0.45\textwidth]{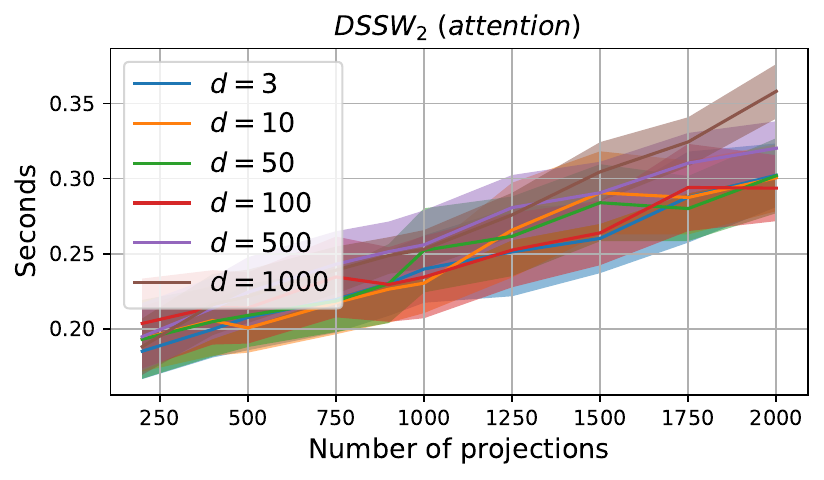}
		\includegraphics[width=0.45\textwidth]{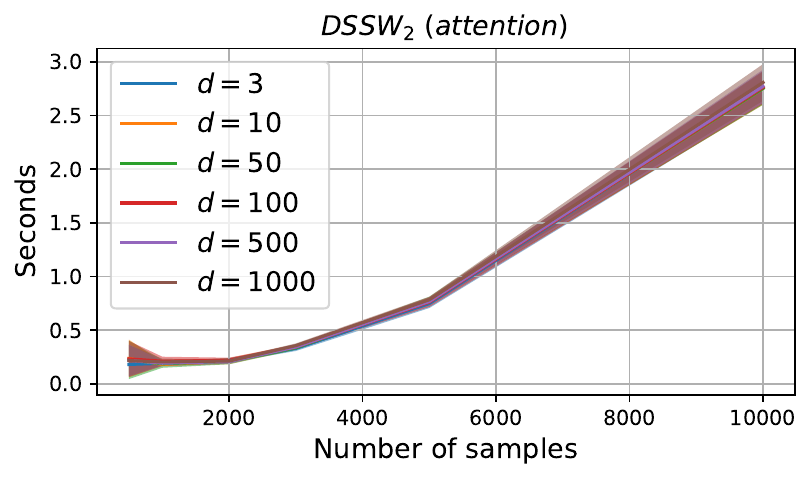}
	}
	
	\caption{Computation time of DSSW with the parametric projected energy function w.r.t. different number of projections and samples. For the plot on the left, we fix $N=500$ for each distribution, and the plot on the right we use $L=100$ projections.}
	\label{fig:runtime_ebssw_parametric}
\end{figure}

In this section, we perform additional runtime analysis of the proposed DSSW w.r.t. varying parameters and the corresponding experimental results is demonstrated in Figure \ref{fig:runtime_ebssw_non_parametric} and \ref{fig:runtime_ebssw_parametric}. It indicates a nonlinear relationship between the runtime and each parameter. 

\subsection{Gradient Flows On The Sphere}

\textbf{Implementation.} We deal with the target distribution on $\mathbb{S}^{2}$ of a mixture of 12 von Mises-Fisher distributions (vMFs), each vMF centered on the vertices of an icosahedron is determined by the golden ratio $\phi = \left ( 1+\sqrt{5} \right ) /2$. We pick 200 samples from each vMF with the total size of 2400 samples of the mixture, each vMF holds $\kappa =50$, the batch size $\left | n_{i}  \right |$ is set to 200 and 2400 for the schema of mini-batch and full-batch, respectively. We utilize 1000 projections for all distances. We adopt the Adam Optimizer with $\gamma=0.01$ for full-batch and $\gamma=0.001$ for mini-batch over 500 gradient steps to perform the optimization 10 times independently.

\begin{figure}[ht]
	\centering
	\subfloat[Target Density]{
		\includegraphics[width=0.18\textwidth]{imgs/Gradient_Flows/target_density.jpg}
	}
	\subfloat[SW]{
		\includegraphics[width=0.18\textwidth]{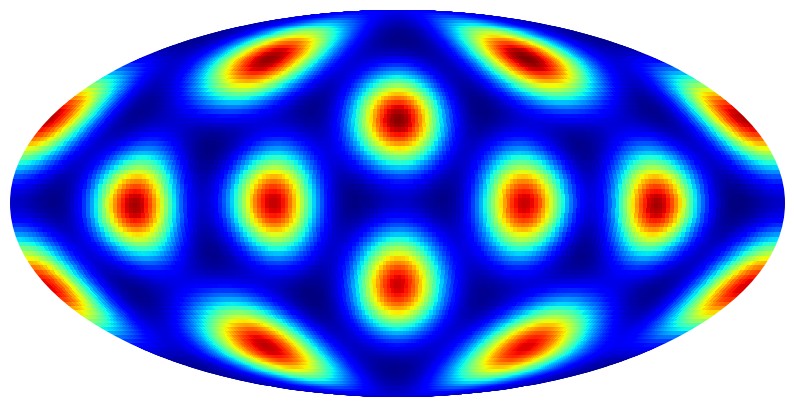}
	}
	\subfloat[SSW]{
		\includegraphics[width=0.18\textwidth]{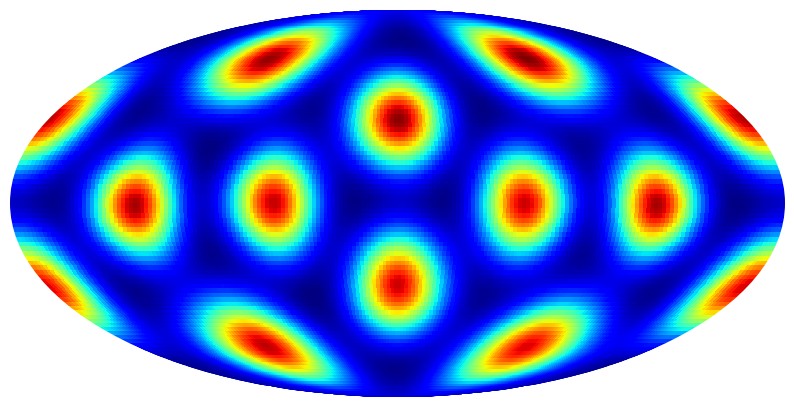}
	}
	\subfloat[S3W]{
		\includegraphics[width=0.18\textwidth]{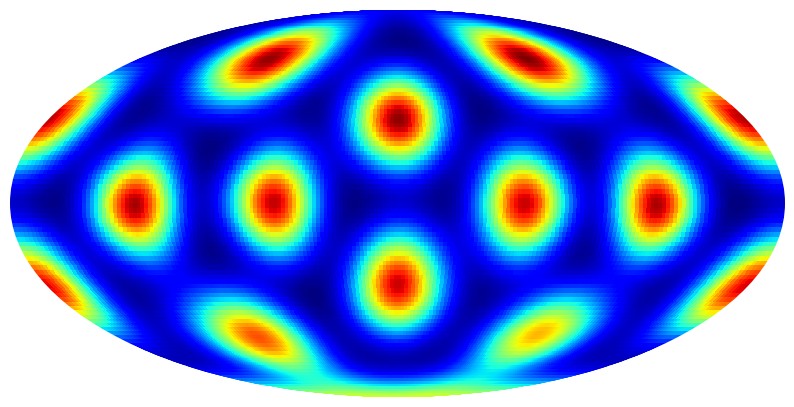}
	}
	\subfloat[RI-S3W (1)]{
		\includegraphics[width=0.18\textwidth]{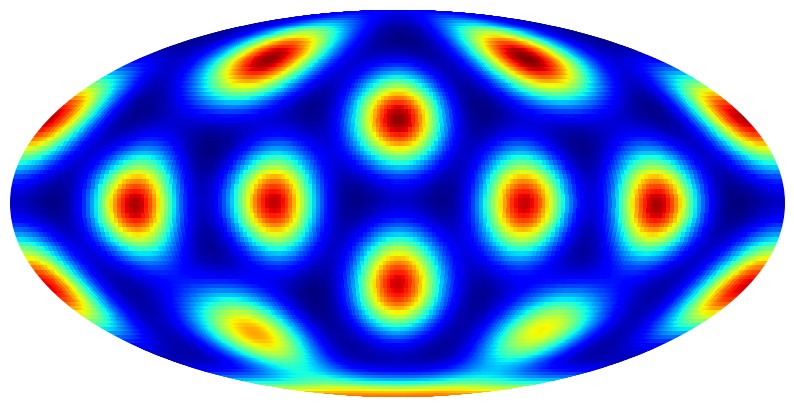}
	}\\ 
	\subfloat[RI-S3W (5)]{
		\includegraphics[width=0.18\textwidth]{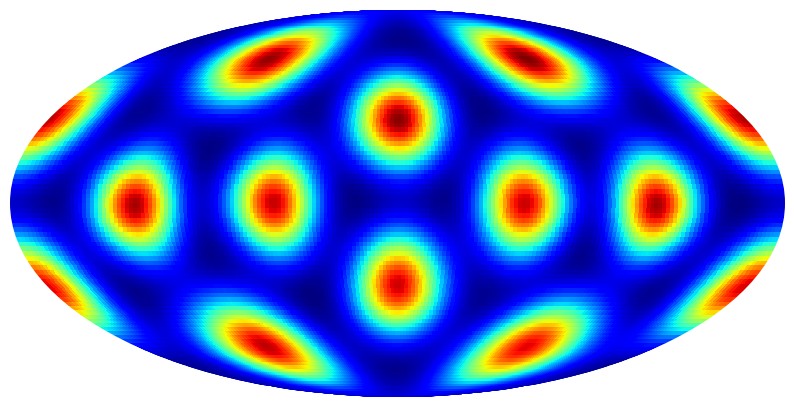}
	}
	\subfloat[RI-S3W (10)]{
		\includegraphics[width=0.18\textwidth]{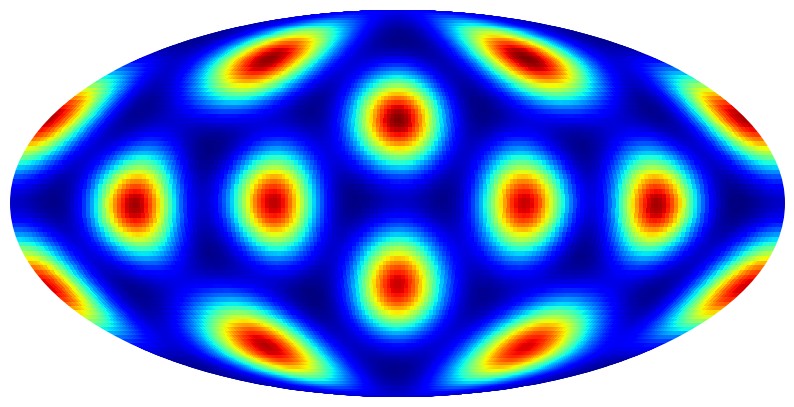}
	}
	\subfloat[ARI-S3W (30)]{
		\includegraphics[width=0.18\textwidth]{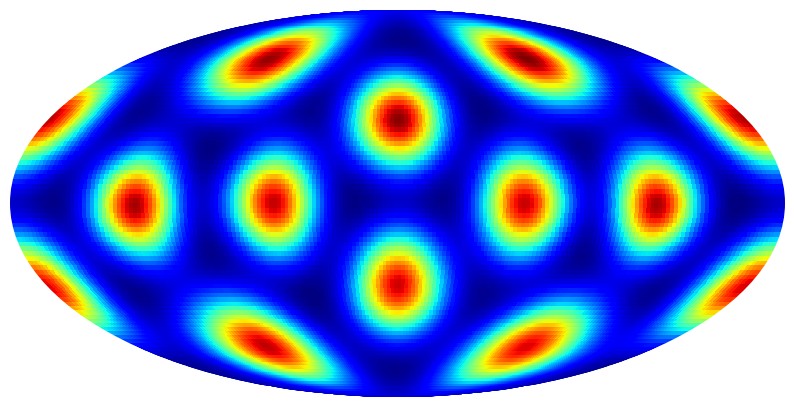}
	}
	\subfloat[DSSW(exp)]{
		\includegraphics[width=0.18\textwidth]{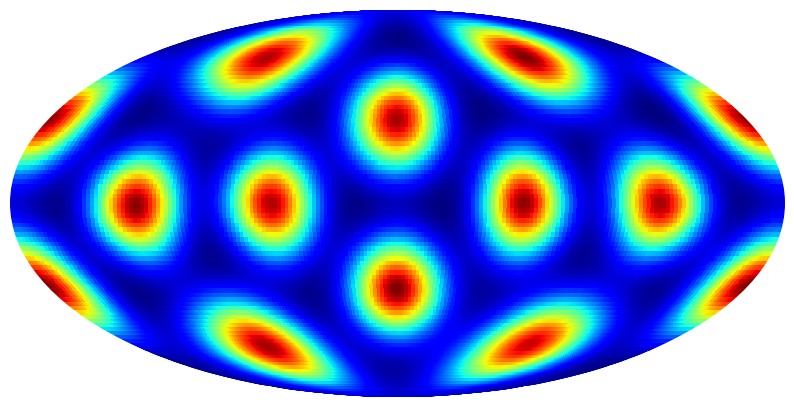}
	}
	\subfloat[DSSW(identity)]{
		\includegraphics[width=0.18\textwidth]{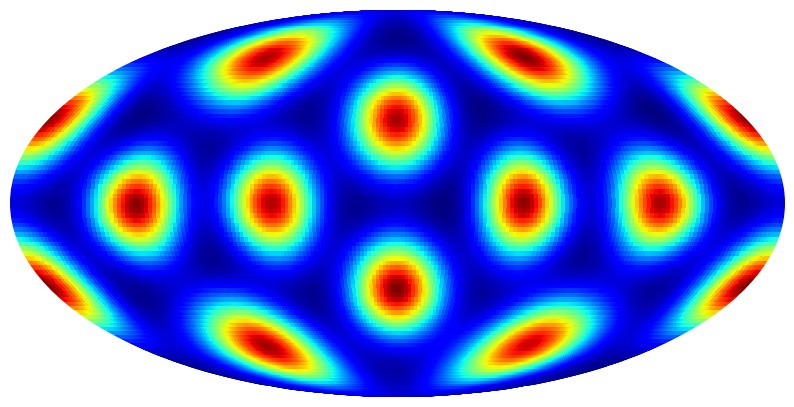}
	}\\ 
	\subfloat[DSSW(poly)]{
		\includegraphics[width=0.18\textwidth]{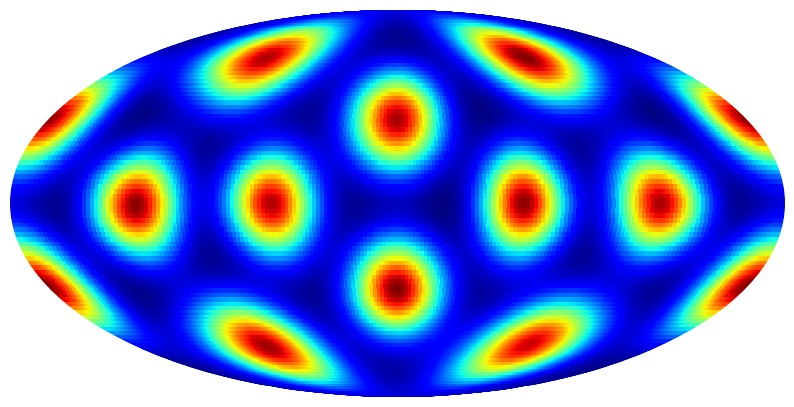}
	}
	\subfloat[DSSW(linear)]{
		\includegraphics[width=0.18\textwidth]{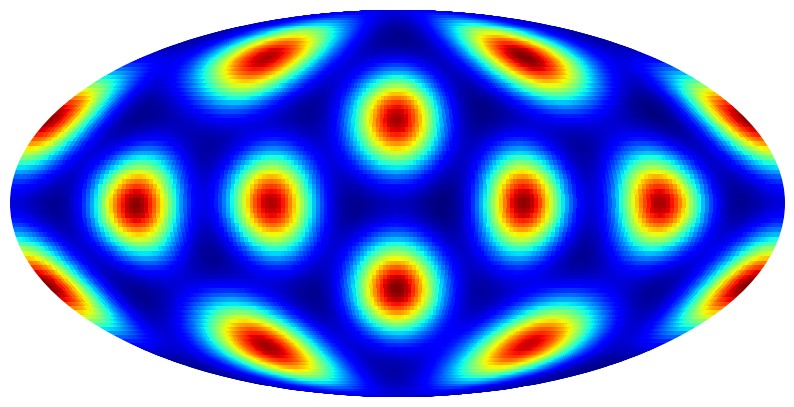}
	}
	\subfloat[DSSW(nonlinear)]{
		\includegraphics[width=0.18\textwidth]{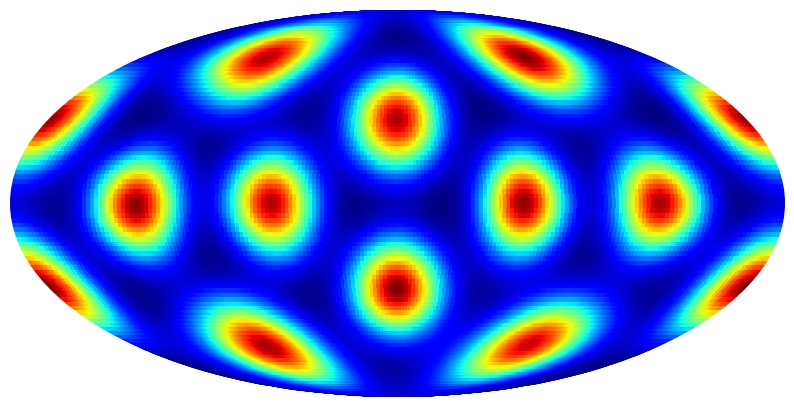}
	}
	\subfloat[DSSW(attention)]{
		\includegraphics[width=0.18\textwidth]{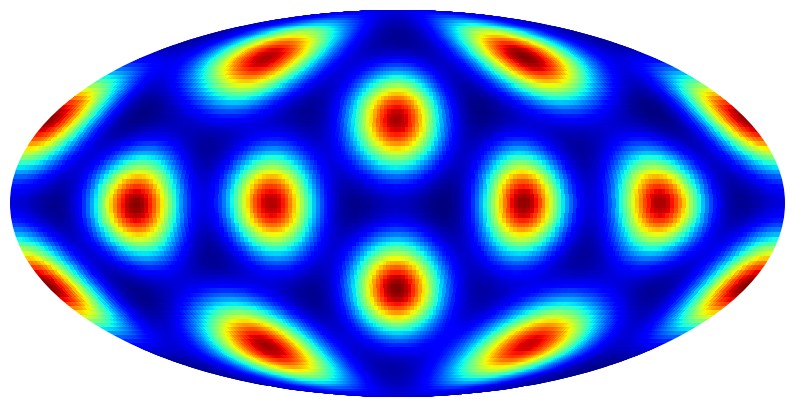}
	}
	
	\caption{The Mollweide projections for full-batch projected gradient descent. We use 1, 5, and 30 rotations for RI-S3W (1), RI-S3W (5), and RI-S3W (10), respectively. We also use 30 rotations with the pool size of 1000 for ARI-S3W (30).}
	\label{fig:gradient_flows_full}
\end{figure}

\textbf{Full-batch results.} We report the results of full-batch among various distance metrics in Table \ref{tab:gradient_flows_full}. We also provide the Mollweide projections for full-batch projected gradient descent in Figure \ref{fig:gradient_flows_full}. Figure \ref{fig:gradient_flows_full} shows that all distance metrics perform comparatively well to learn the target distribution.

\begin{table}[ht]
	\centering
	\begin{tabular}{cccc}
		\toprule
		& Distance          & NLL $\downarrow$          & $\log{W_{2}}$ $\downarrow$      \\ \midrule
		\multirow{13}{*}{Full-batch} & SW                & -4911.97  ± 0.00  & \textbf{-10.82   ± 0.00} \\
		& SSW               & \underline{-4916.66 ± 1.72}  & -7.42  ± 0.19  \\
		& S3W               & -4621.67   ± 63.38 & -5.01   ± 0.17  \\
		& RI-S3W   (1)      & -4896.81   ± 20.57 & -7.21   ± 0.07  \\
		& RI-S3W   (5)      & -4908.65   ± 7.65  & -7.84   ± 0.18  \\
		& RI-S3W   (10)     & -4911.32   ± 2.76  & -8.31   ± 0.26  \\
		& ARI-S3W   (30)    & -4912.45   ± 1.96  & -9.44   ± 0.55  \\ \cmidrule{2-4} 
		& DSSW (exp)       & -5004.65 ± 1.47 $\ddagger$ & -7.74   ± 0.33  \\
		& DSSW (identity)  & -5003.65 ± 2.69 $\ddagger$ & -7.75 ± 0.21  \\
		& DSSW (poly)      & -5001.16 ± 2.52 $\ddagger$ & -7.64 ± 0.16  \\
		& DSSW (linear)    & \textbf{-5006.57 ± 2.10} $\ddagger$ & -7.66 ± 0.12  \\
		& DSSW (nonlinear) & -5005.66 ± 1.89 $\ddagger$ & -7.89 ± 0.34  \\
		& DSSW (attention) & -5005.88 ± 2.33 $\ddagger$ & -7.56 ± 0.22  \\ \bottomrule
	\end{tabular}%
	\caption{Full-batch comparison between different distances as loss for gradient flows averaged over 10 training runs. RI-S3W (1), RI-S3W (5), and RI-S3W (10) have 1, 5, and 30 rotations, respectively. ARI-S3W (30) has 30 rotations with the pool size of 1000. Notation "$\ddagger$" indicates that DSSW variants are significantly better than the best baseline method using t-test when the significance level is 0.05.}
	\label{tab:gradient_flows_full}
\end{table}

\subsection{Earth Density Estimation}
\textbf{Implementation.} For exponential map normalizing flows, we construct the model using 48 radial blocks, each consisting of 100 components. For the training process, we run 20000 epochs with full batch size. We adopt the optimizer of Adam with the learning rate of 0.1 for SW, SSW and DSSW with various projected energy functions, and 0.05 for S3W and its variants. For Stereo, we adopt a learning rate of 0.001.

\begin{figure*}[ht]
	\centering
	\subfloat[Earthquake]{
		\includegraphics[width=0.33\textwidth]{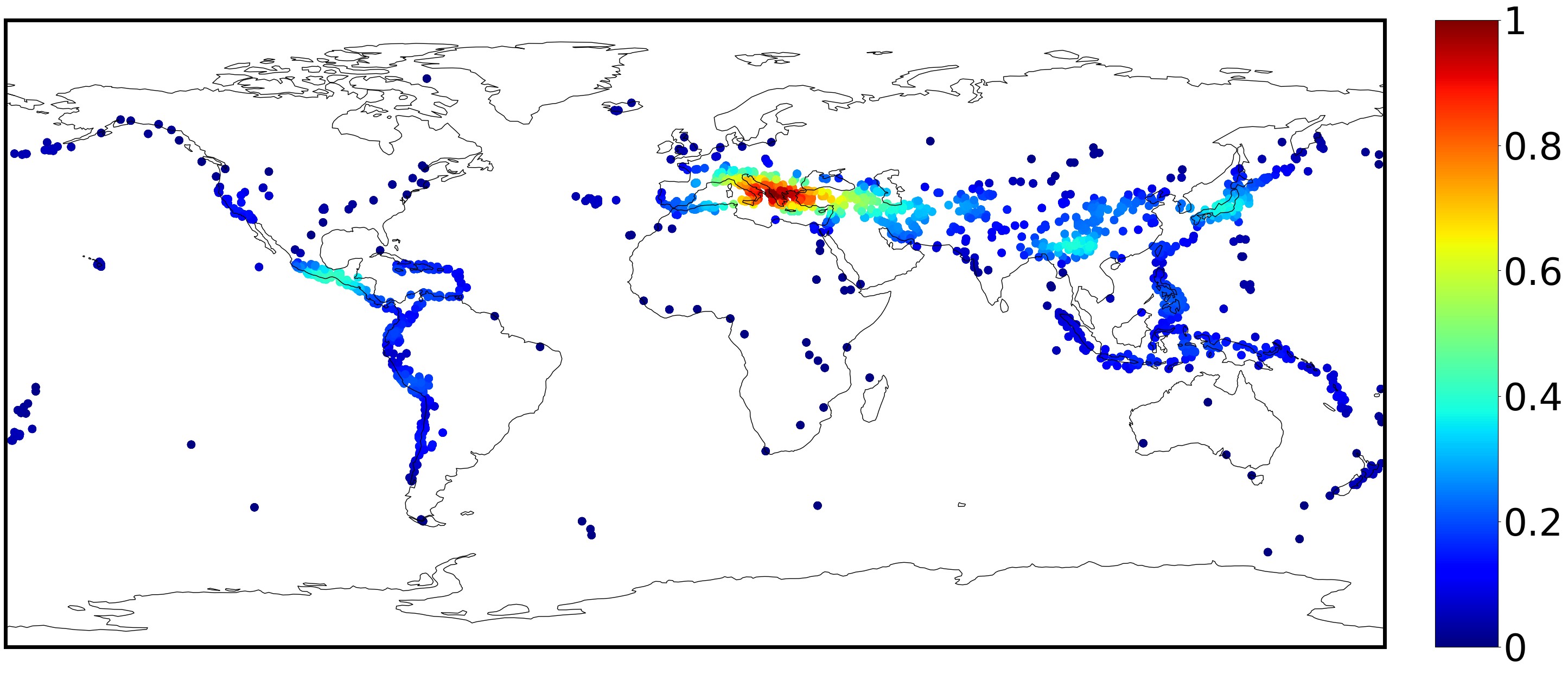}
	}
	\subfloat[Flood]{
		\includegraphics[width=0.33\textwidth]{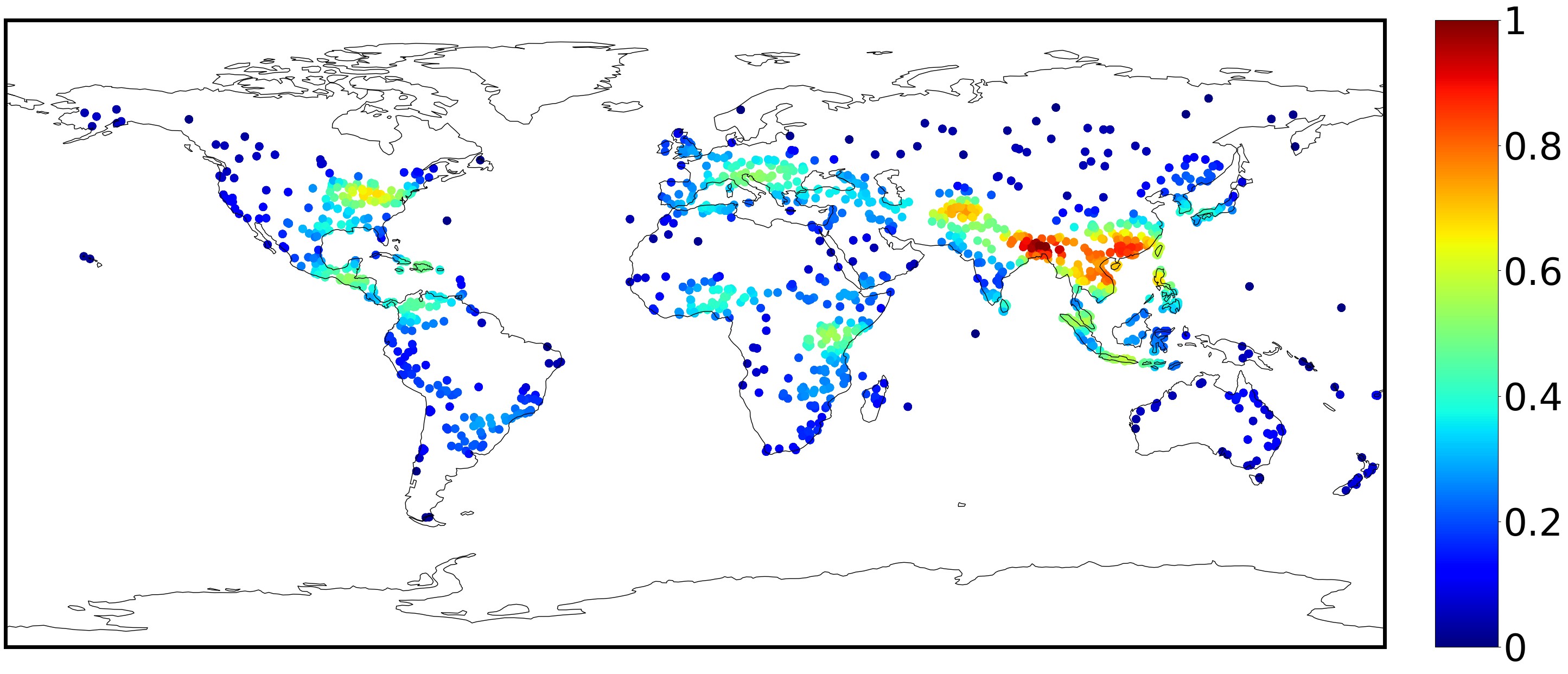}
	}
	\subfloat[Fire]{
		\includegraphics[width=0.33\textwidth]{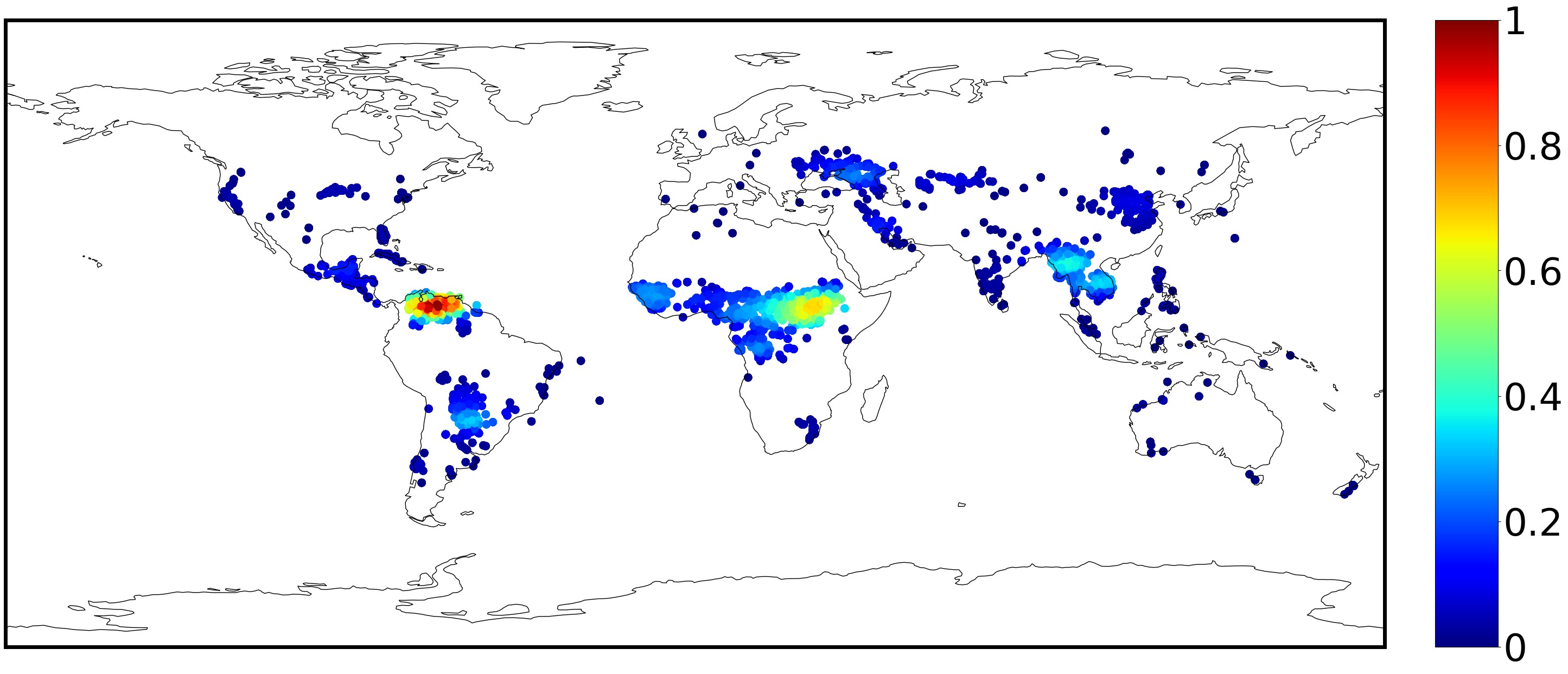}
	}
	
	\caption{Ground truth as estimated with KDE using test data.}
	\label{fig:density_estimation_ground_truth}
\end{figure*}

\begin{figure*}[ht]
	\centering
	\subfloat[Earthquake]{
		\includegraphics[width=0.3\textwidth]{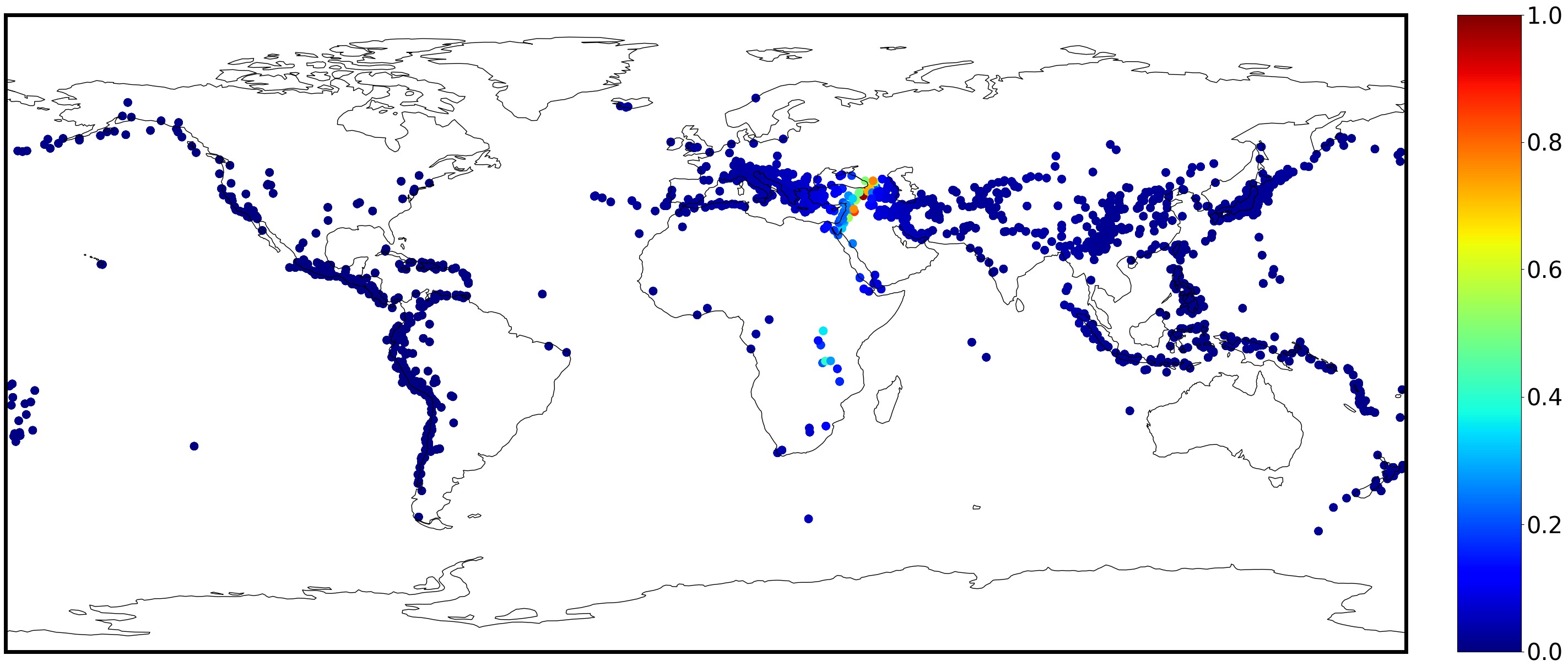}
	}
	\subfloat[Flood]{
		\includegraphics[width=0.3\textwidth]{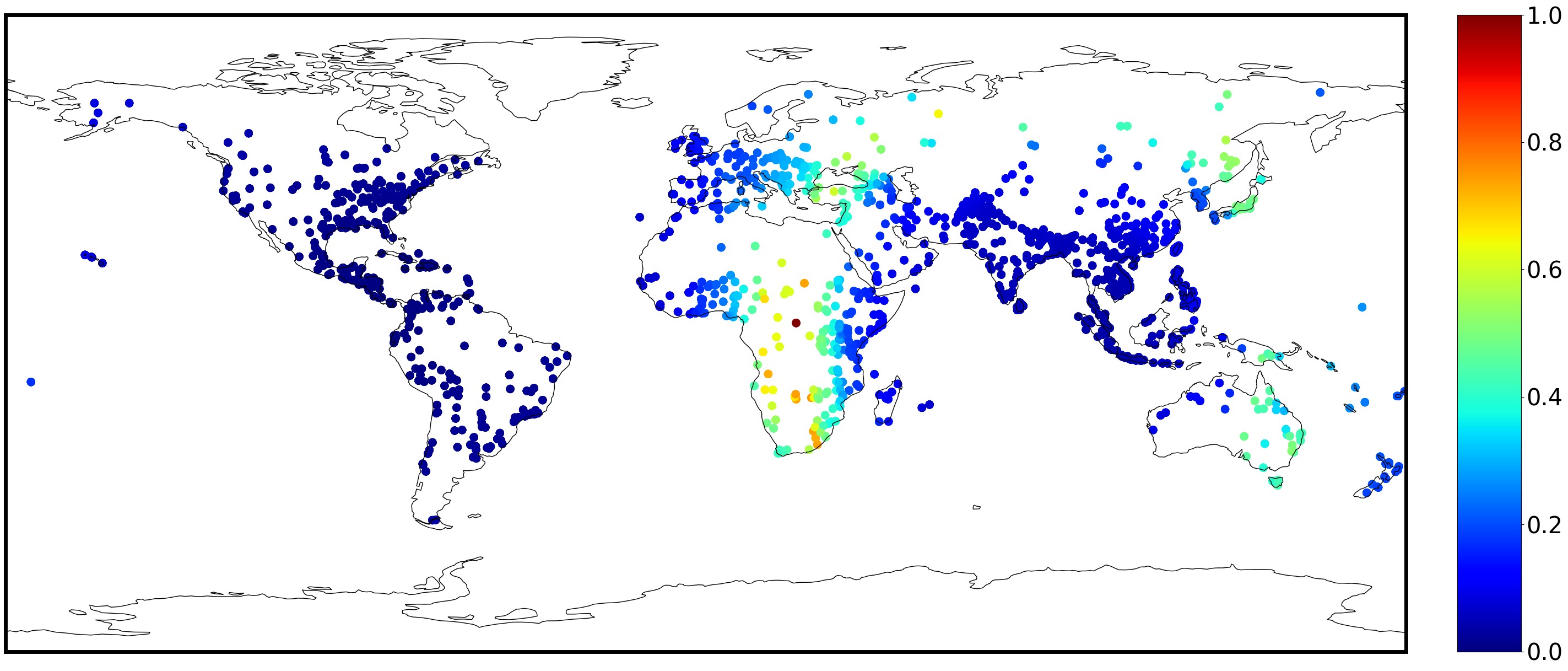}
	}
	\subfloat[Fire]{
		\includegraphics[width=0.3\textwidth]{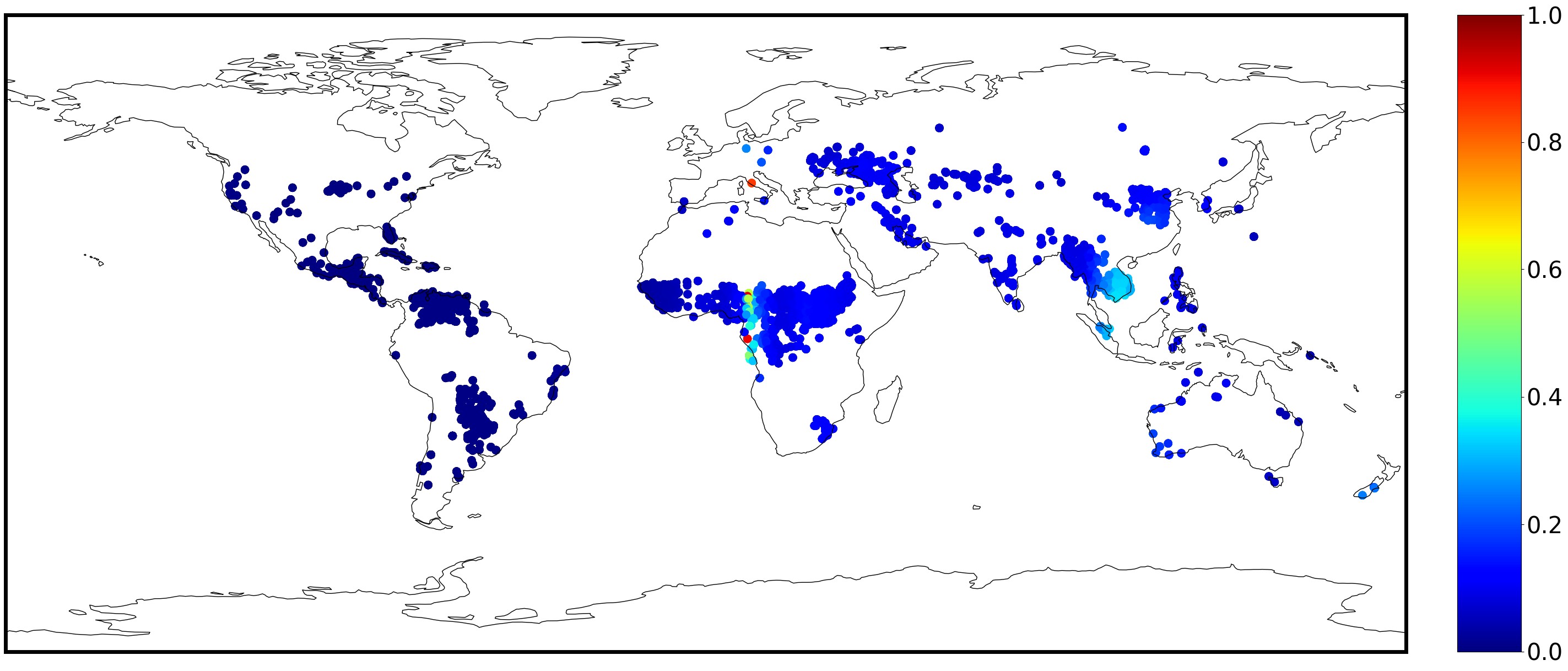}
	}
	
	\caption{Stereo + RealNVP}
	\label{fig:density_estimation_stereo}
\end{figure*}

\begin{figure*}[ht]
	\centering
	\subfloat[Earthquake]{
		\includegraphics[width=0.3\textwidth]{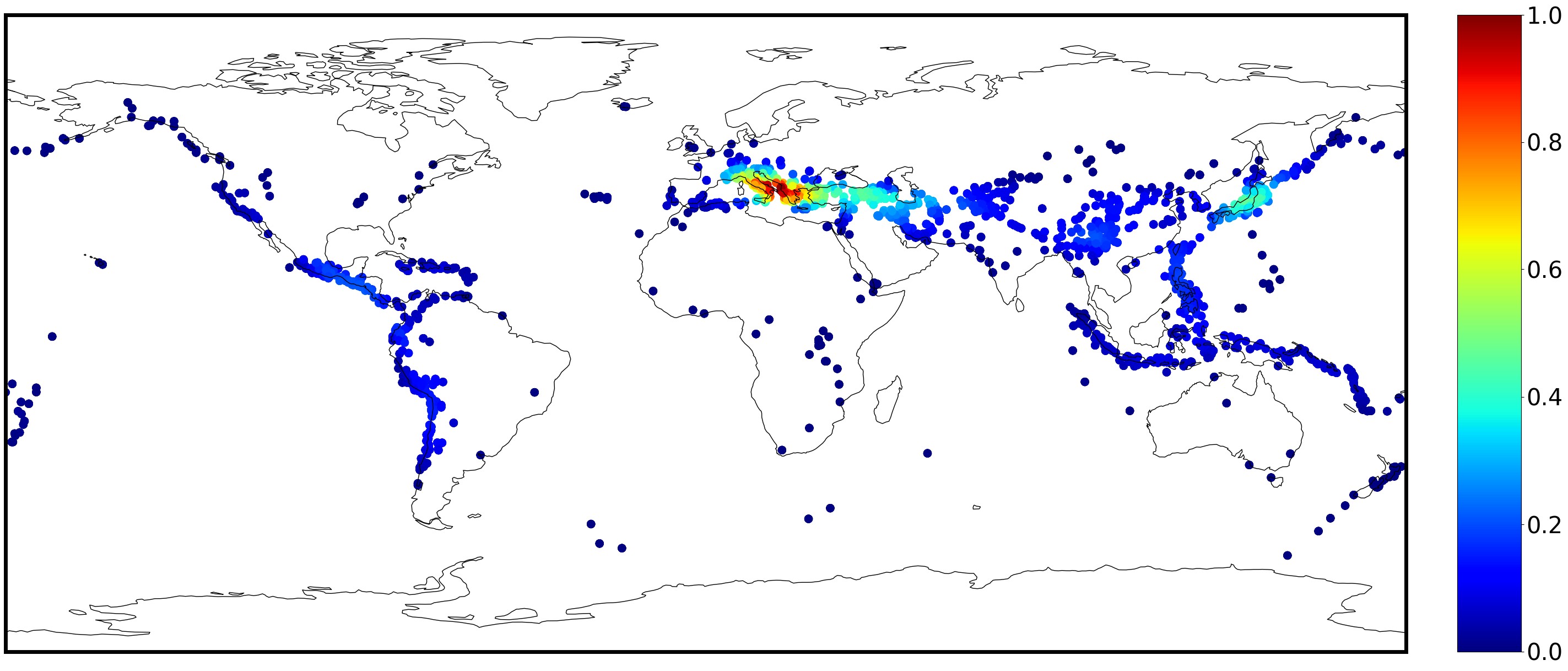}
	}
	\subfloat[Flood]{
		\includegraphics[width=0.3\textwidth]{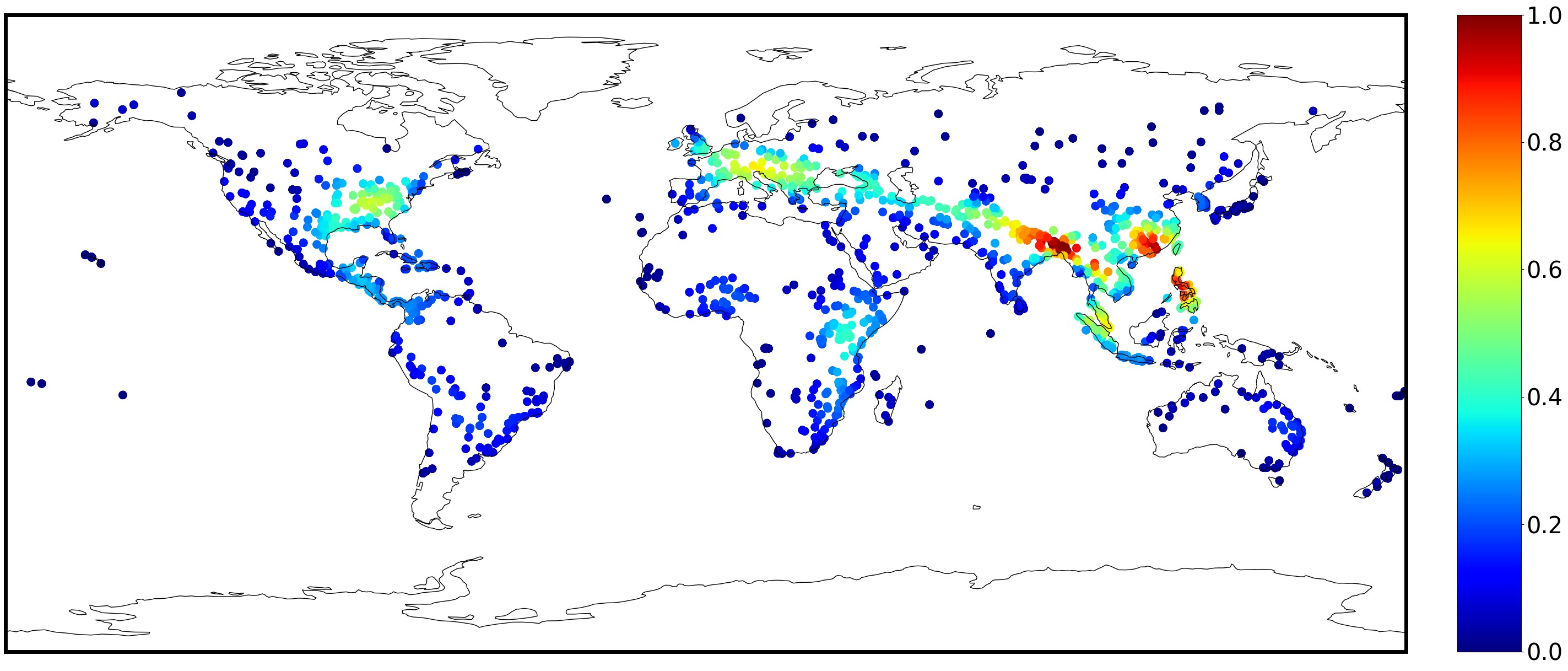}
	}
	\subfloat[Fire]{
		\includegraphics[width=0.3\textwidth]{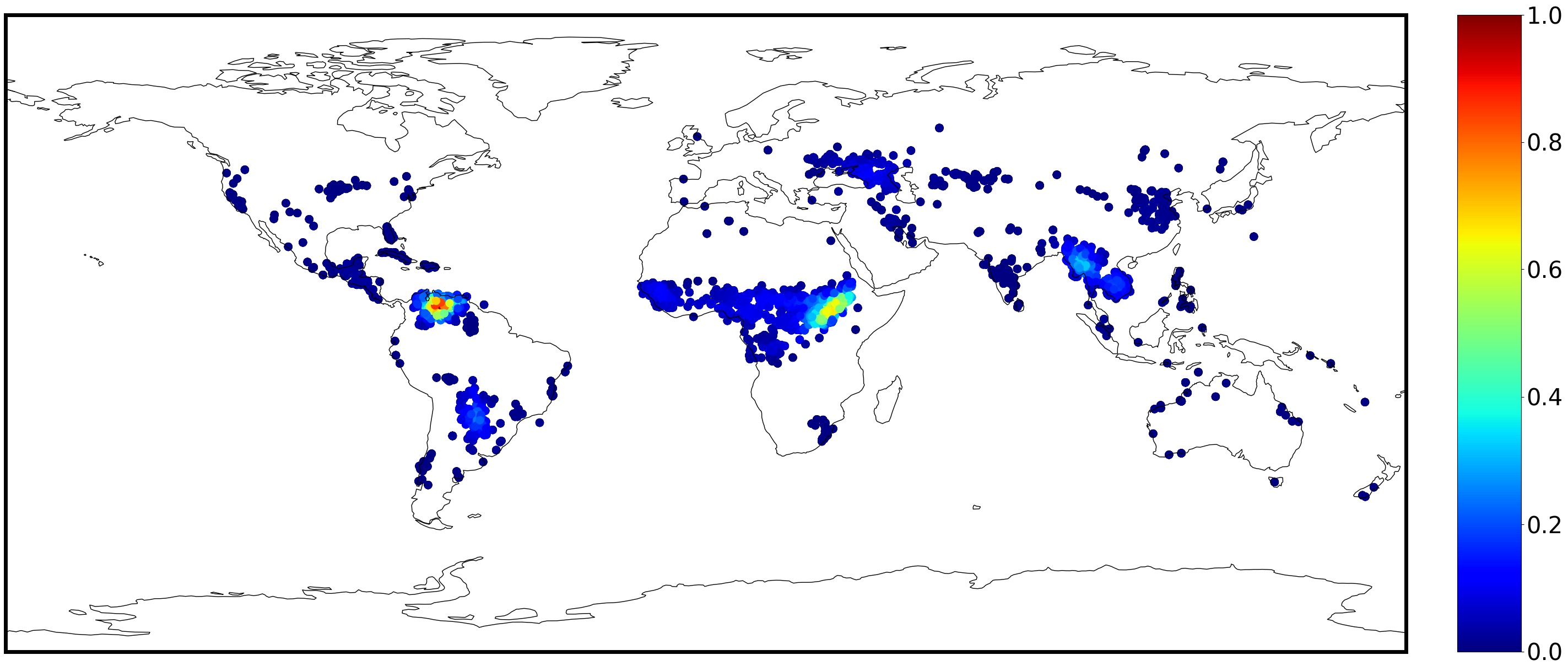}
	}
	
	\caption{SW}
	\label{fig:density_estimation_sw}
\end{figure*}

\begin{figure*}[ht]
	\centering
	\subfloat[Earthquake]{
		\includegraphics[width=0.3\textwidth]{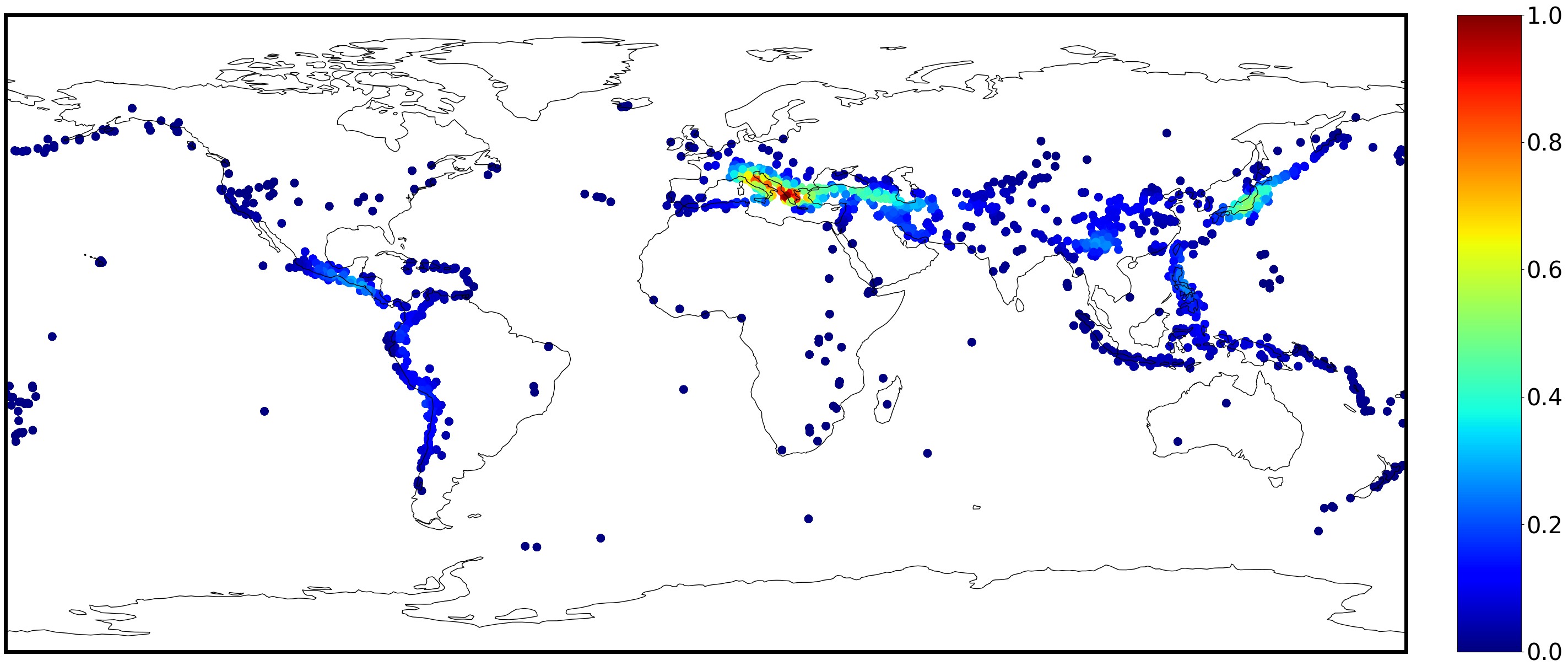}
	}
	\subfloat[Flood]{
		\includegraphics[width=0.3\textwidth]{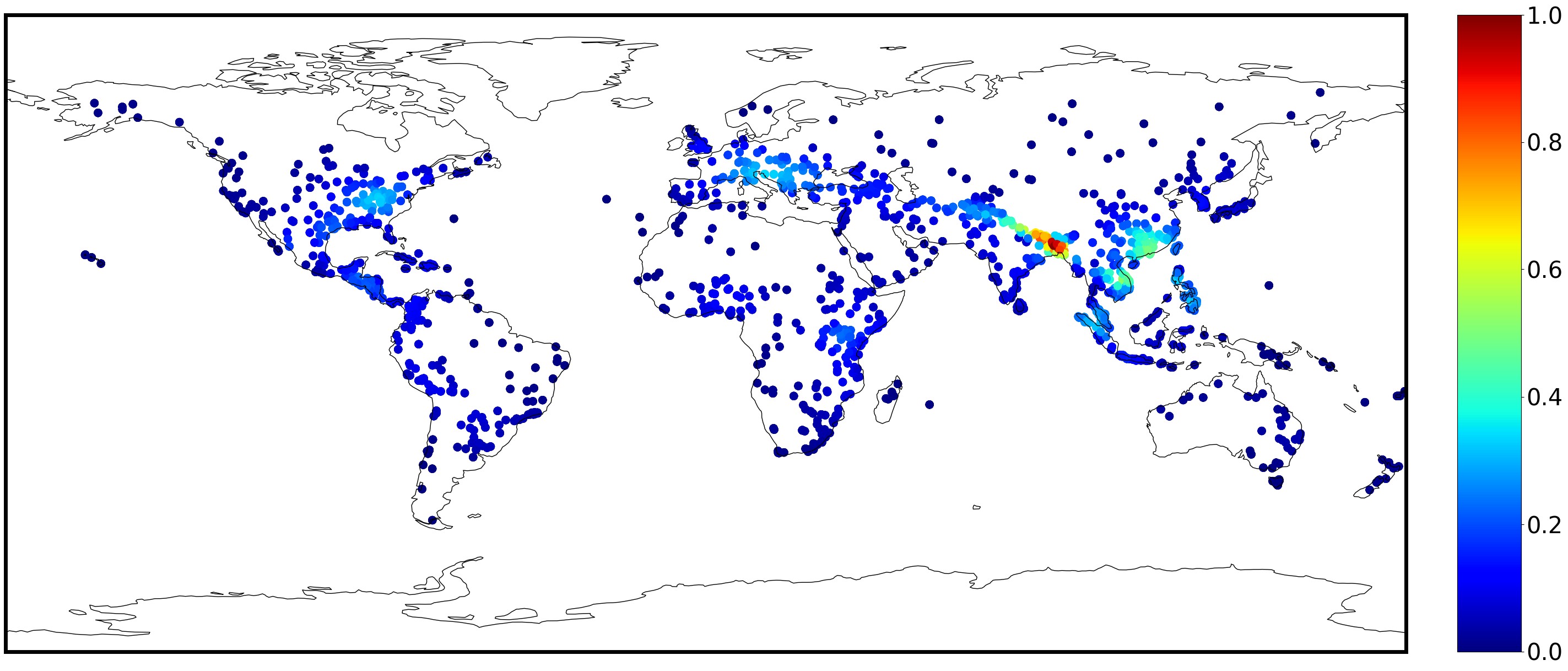}
	}
	\subfloat[Fire]{
		\includegraphics[width=0.3\textwidth]{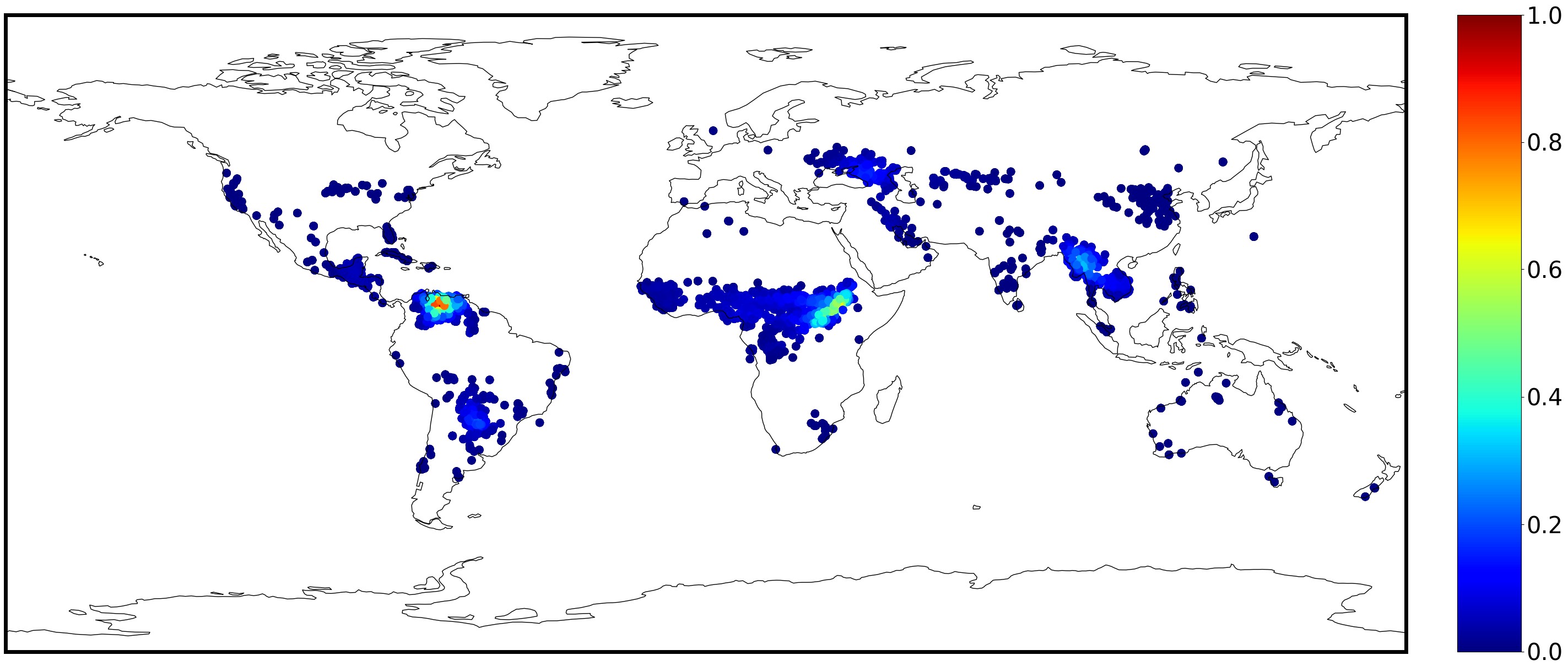}
	}
	
	\caption{SSW}
	\label{fig:density_estimation_ssw}
\end{figure*}

\begin{figure*}[ht]
	\centering
	\subfloat[Earthquake]{
		\includegraphics[width=0.3\textwidth]{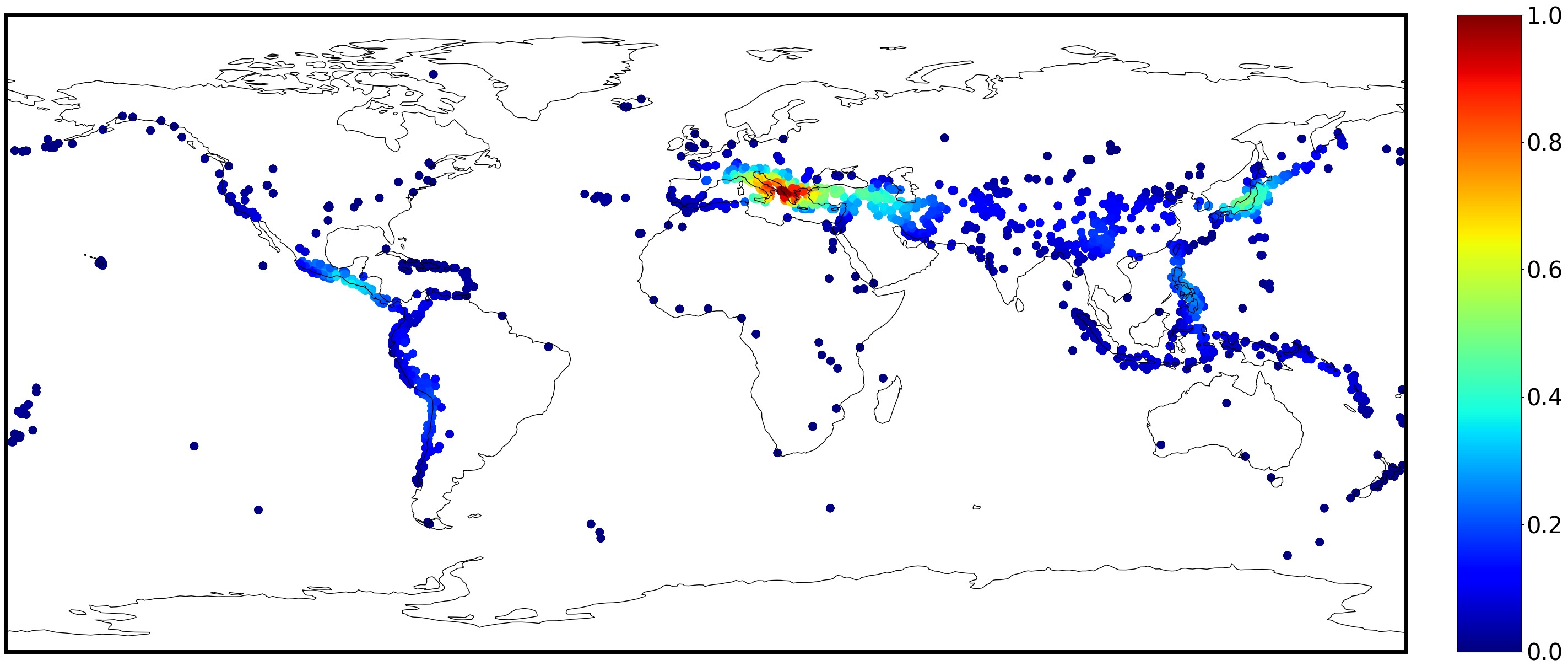}
	}
	\subfloat[Flood]{
		\includegraphics[width=0.3\textwidth]{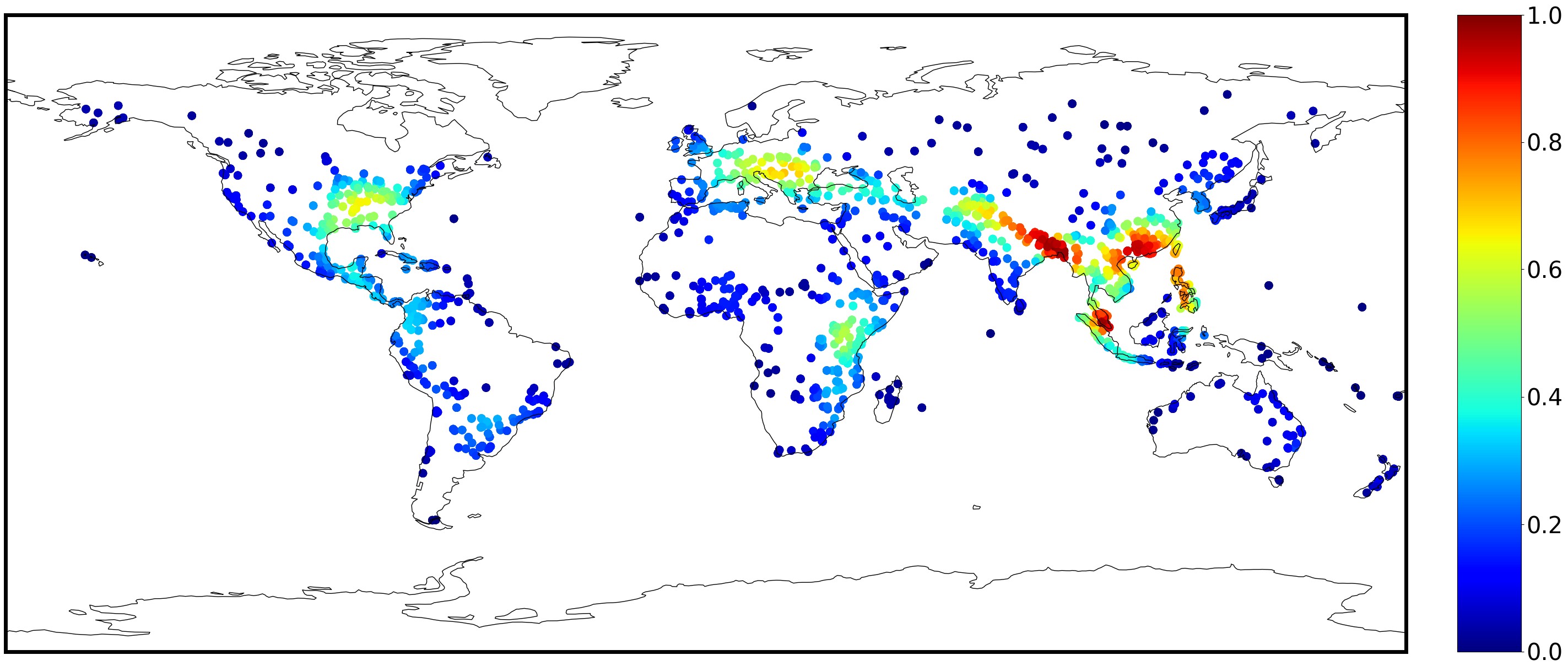}
	}
	\subfloat[Fire]{
		\includegraphics[width=0.3\textwidth]{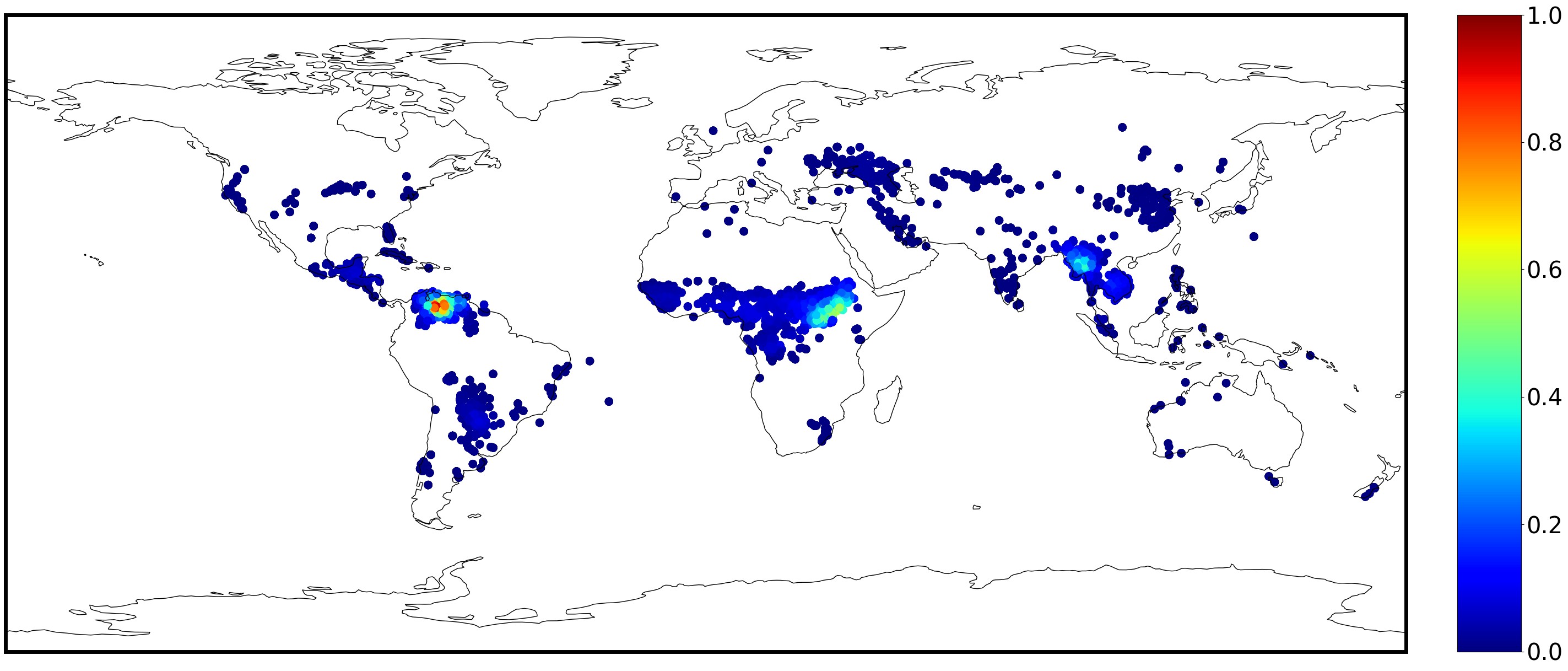}
	}
	
	\caption{S3W}
	\label{fig:density_estimation_s3w}
\end{figure*}

\begin{figure*}[ht]
	\centering
	\subfloat[Earthquake]{
		\includegraphics[width=0.3\textwidth]{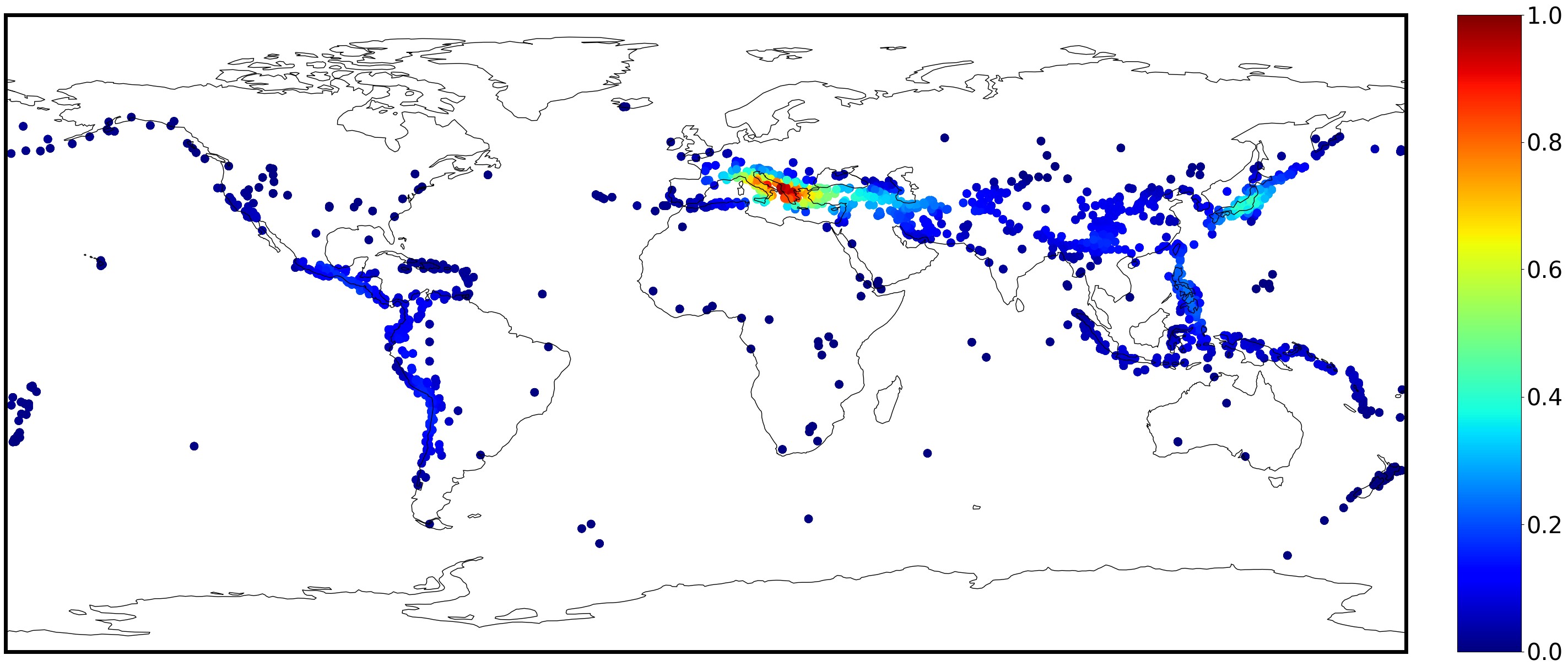}
	}
	\subfloat[Flood]{
		\includegraphics[width=0.3\textwidth]{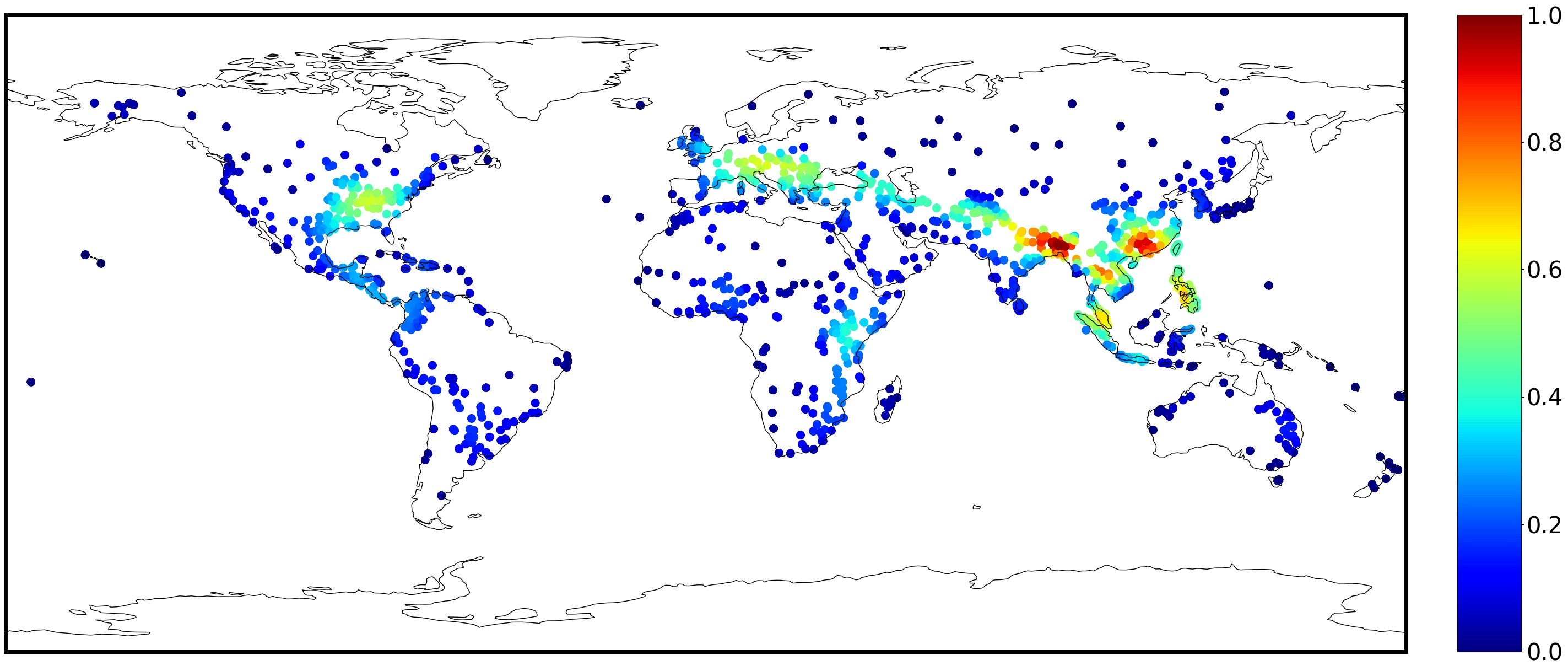}
	}
	\subfloat[Fire]{
		\includegraphics[width=0.3\textwidth]{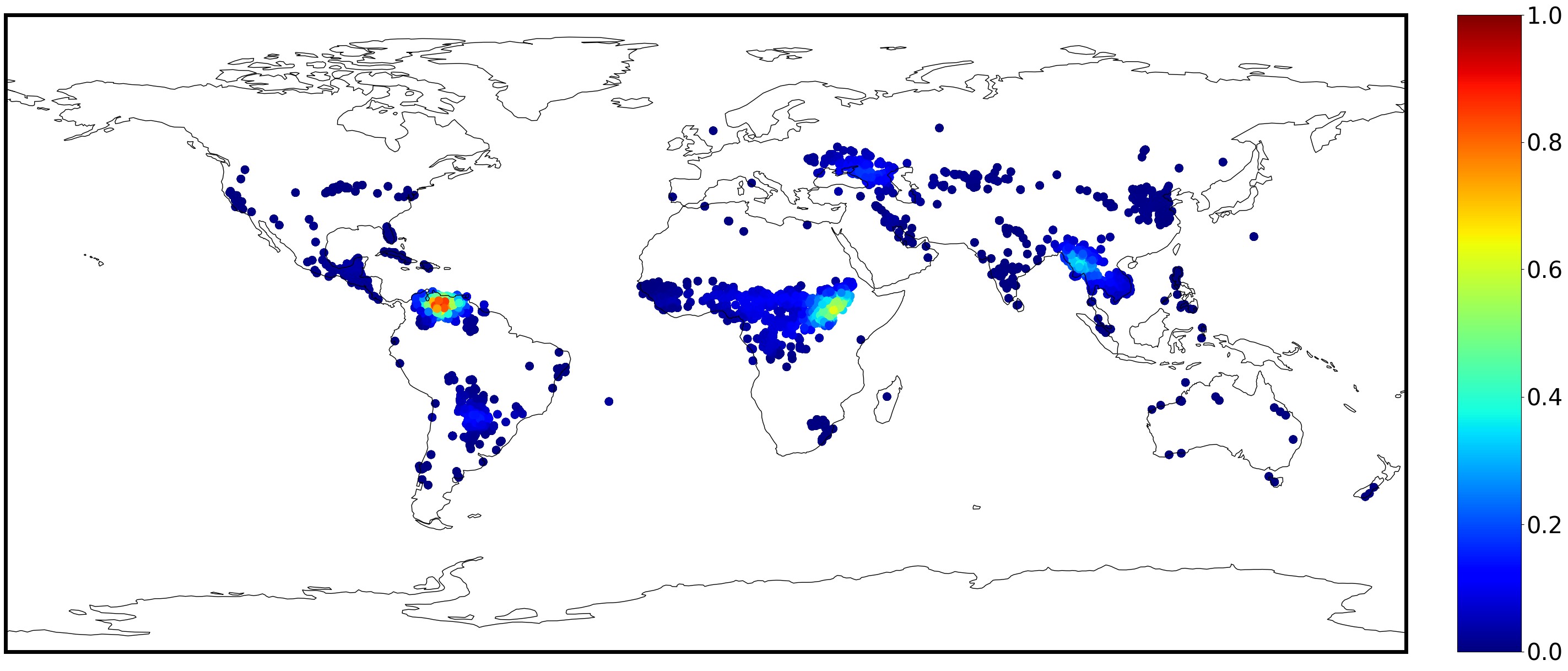}
	}
	
	\caption{RI-S3W (1)}
	\label{fig:density_estimation_r1}
\end{figure*}

\begin{figure*}[ht]
	\centering
	\subfloat[Earthquake]{
		\includegraphics[width=0.3\textwidth]{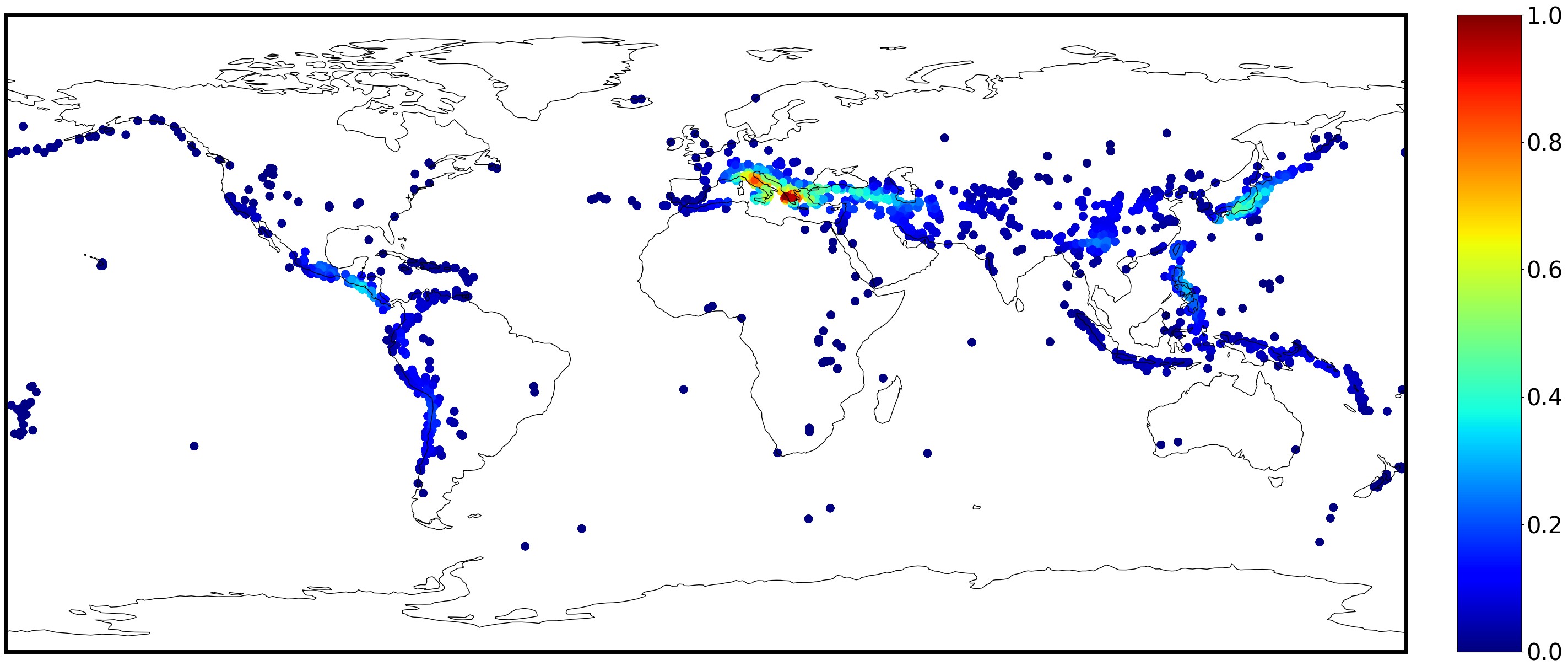}
	}
	\subfloat[Flood]{
		\includegraphics[width=0.3\textwidth]{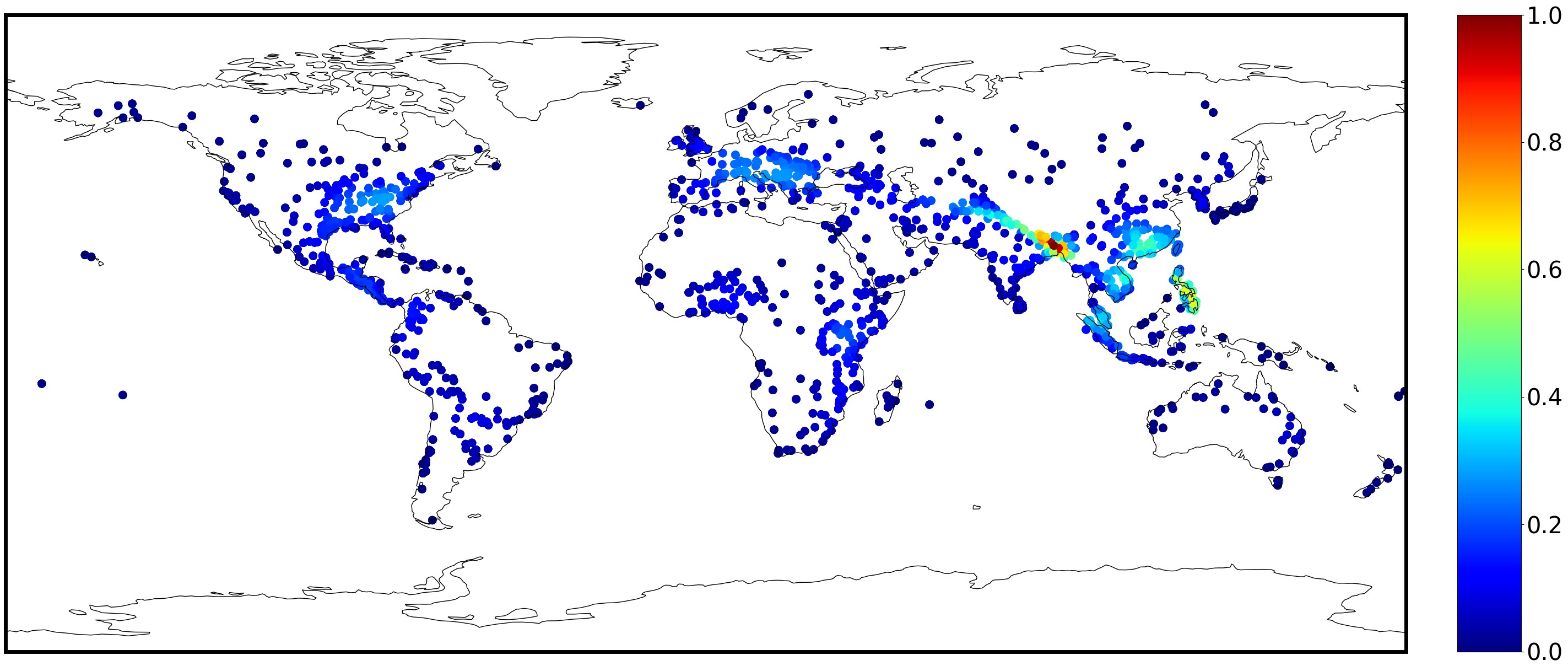}
	}
	\subfloat[Fire]{
		\includegraphics[width=0.3\textwidth]{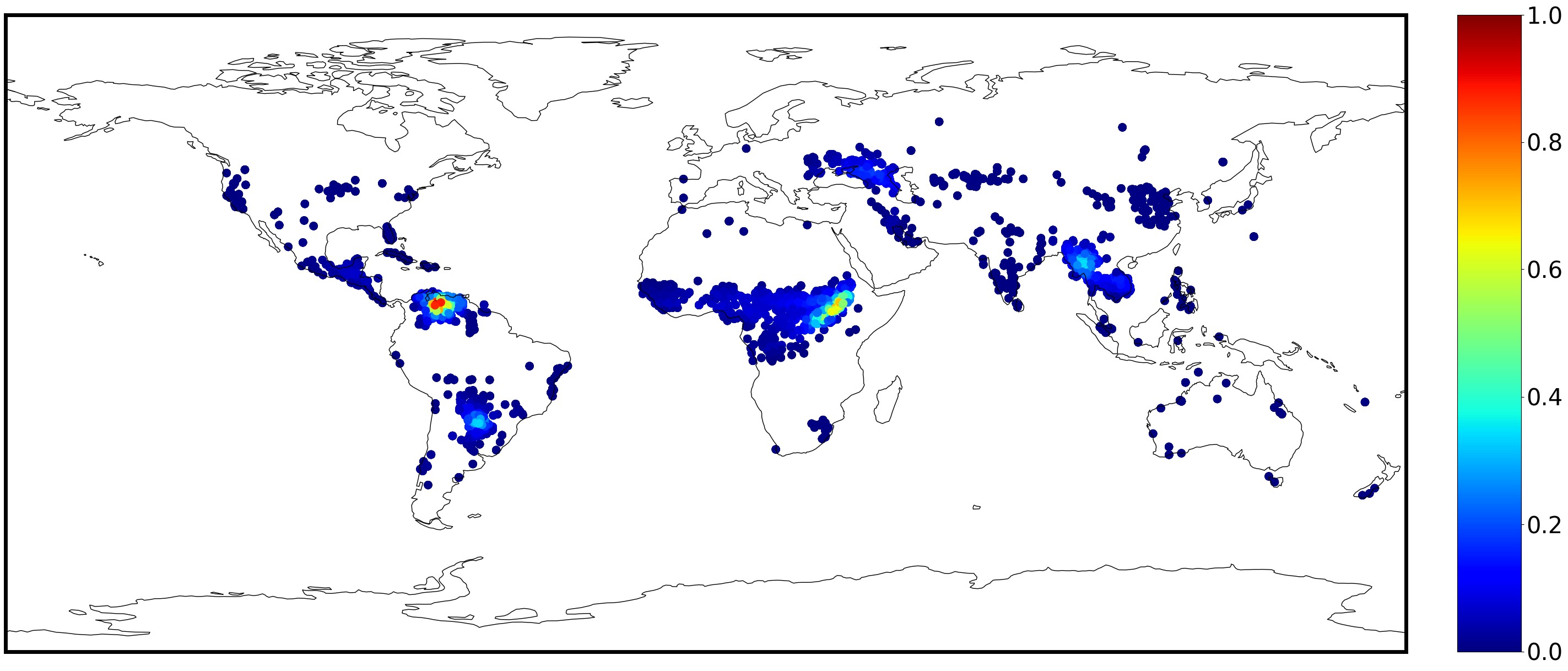}
	}
	
	\caption{DSSW (exp)}
	\label{fig:density_estimation_exp}
\end{figure*}

\begin{figure*}[ht]
	\centering
	\subfloat[Earthquake]{
		\includegraphics[width=0.3\textwidth]{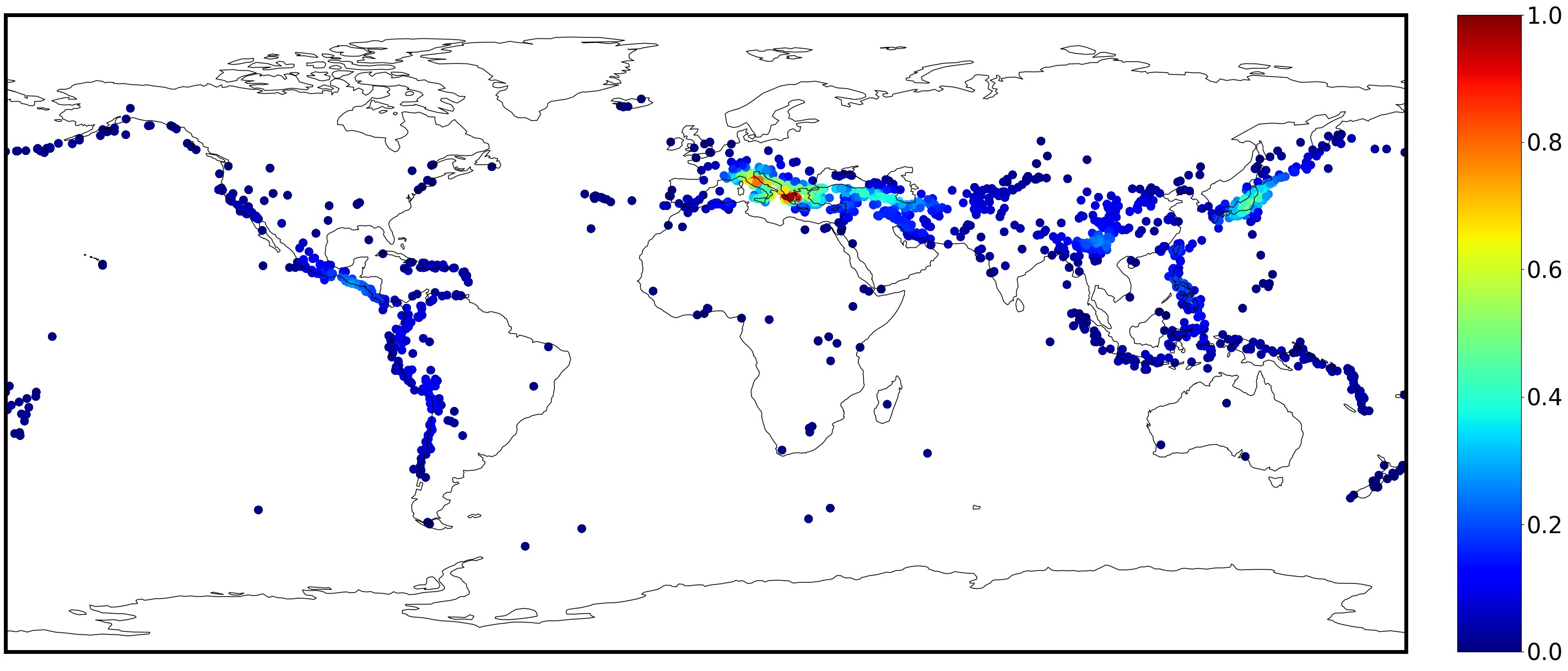}
	}
	\subfloat[Flood]{
		\includegraphics[width=0.3\textwidth]{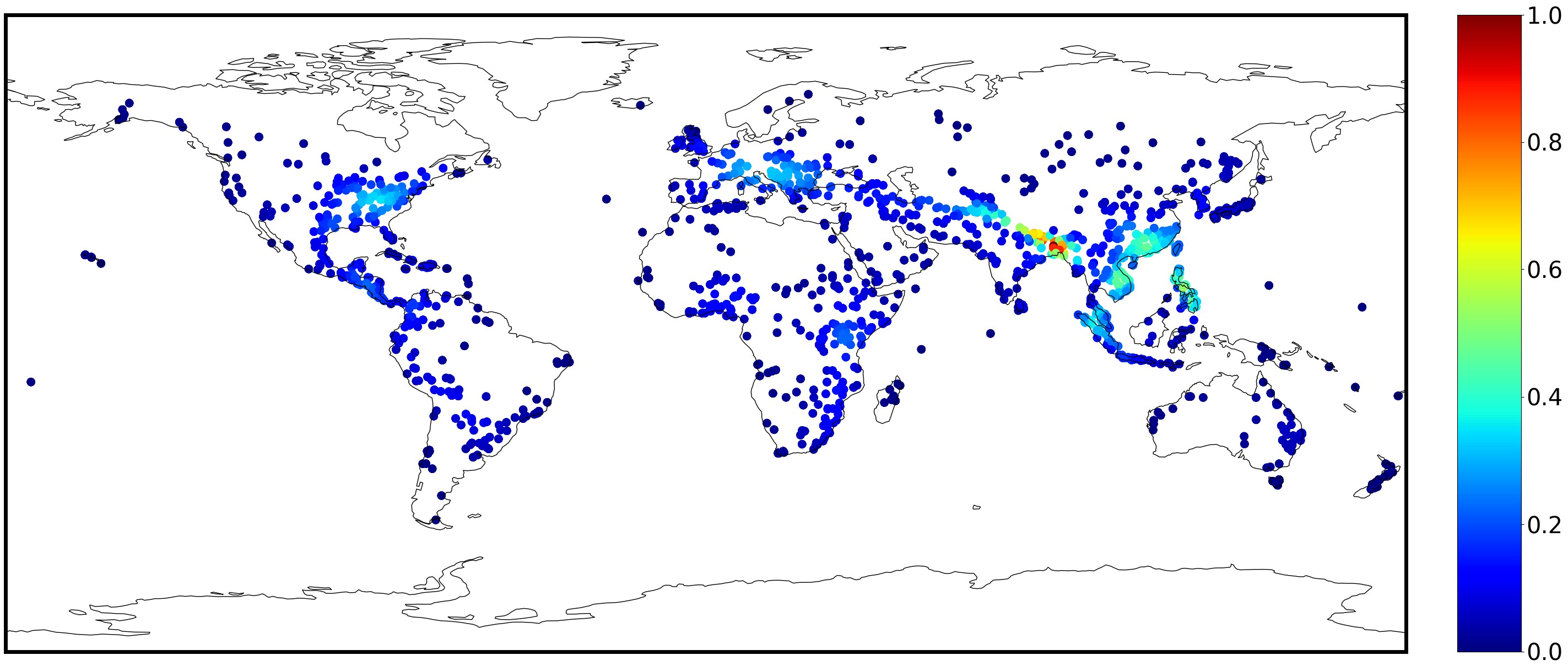}
	}
	\subfloat[Fire]{
		\includegraphics[width=0.3\textwidth]{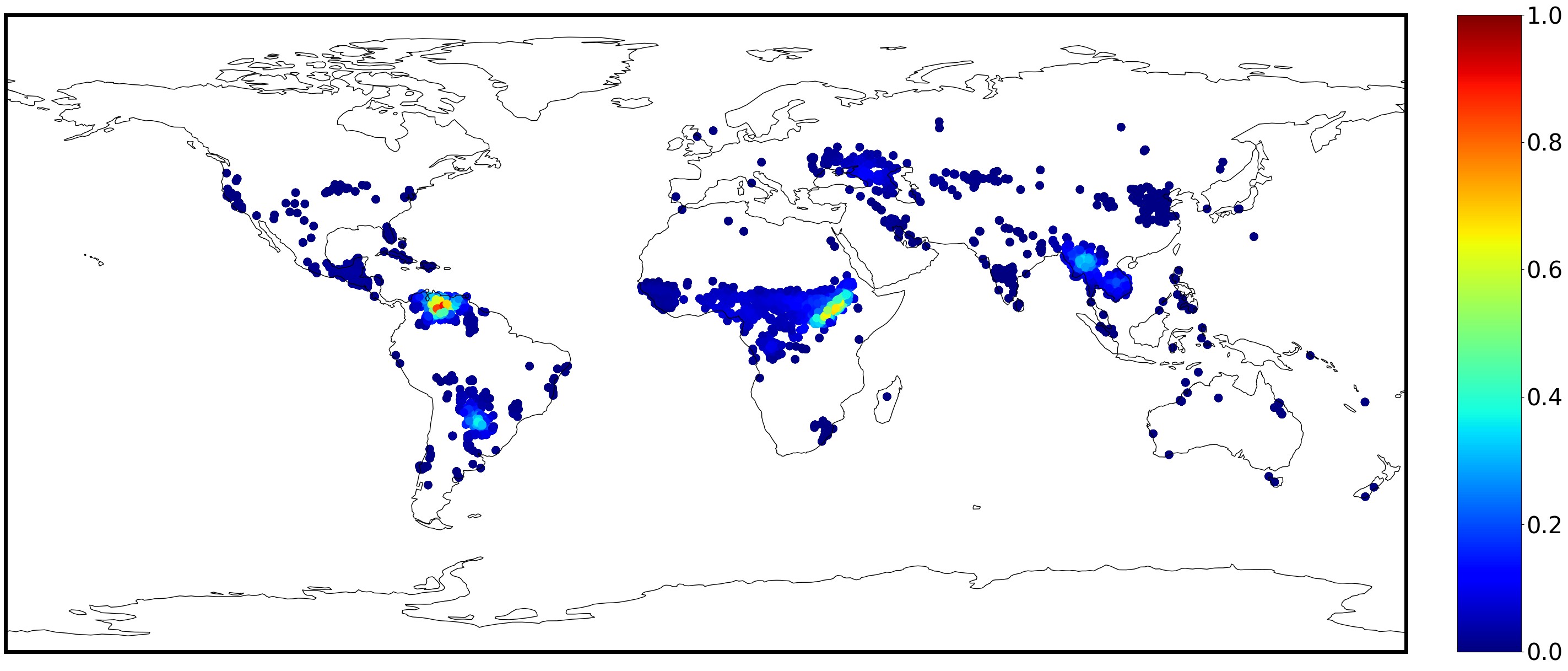}
	}
	
	\caption{DSSW (identity)}
	\label{fig:density_estimation_identity}
\end{figure*}

\begin{figure*}[ht]
	\centering
	\subfloat[Earthquake]{
		\includegraphics[width=0.3\textwidth]{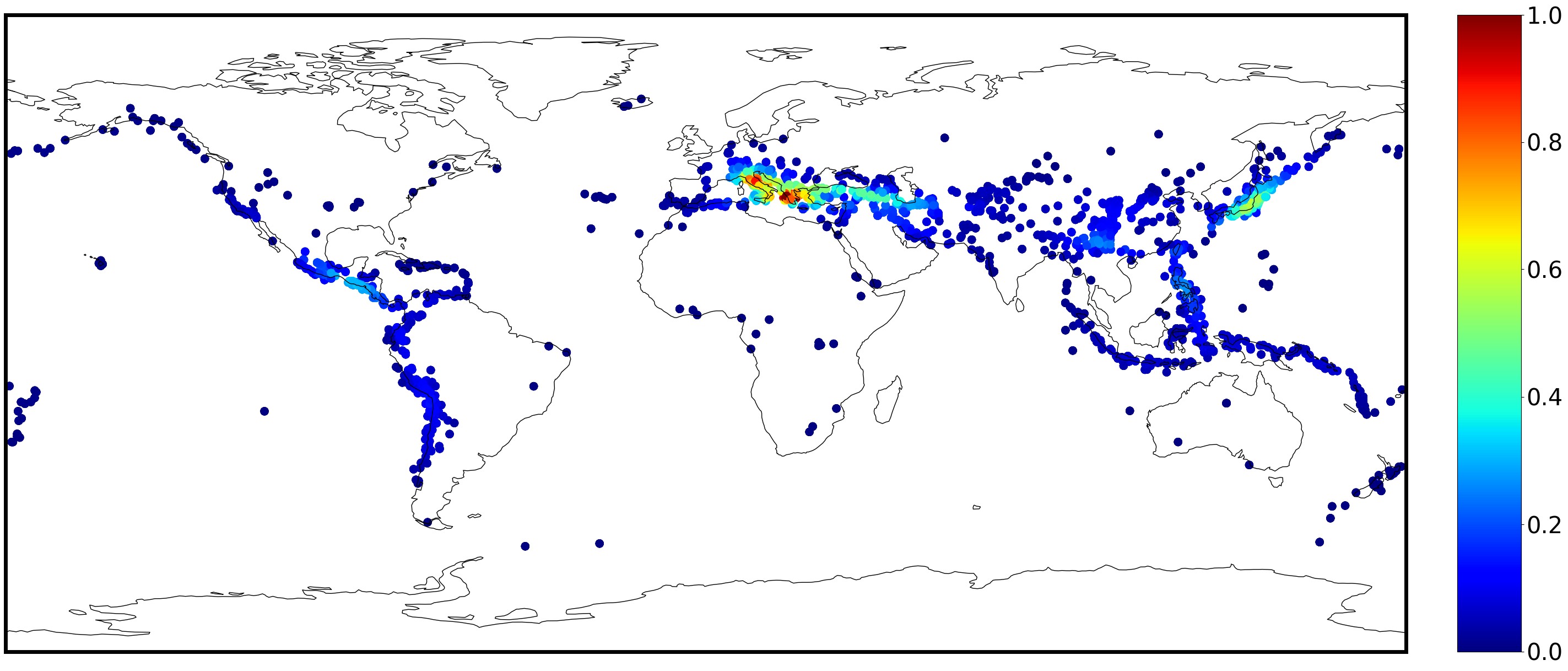}
	}
	\subfloat[Flood]{
		\includegraphics[width=0.3\textwidth]{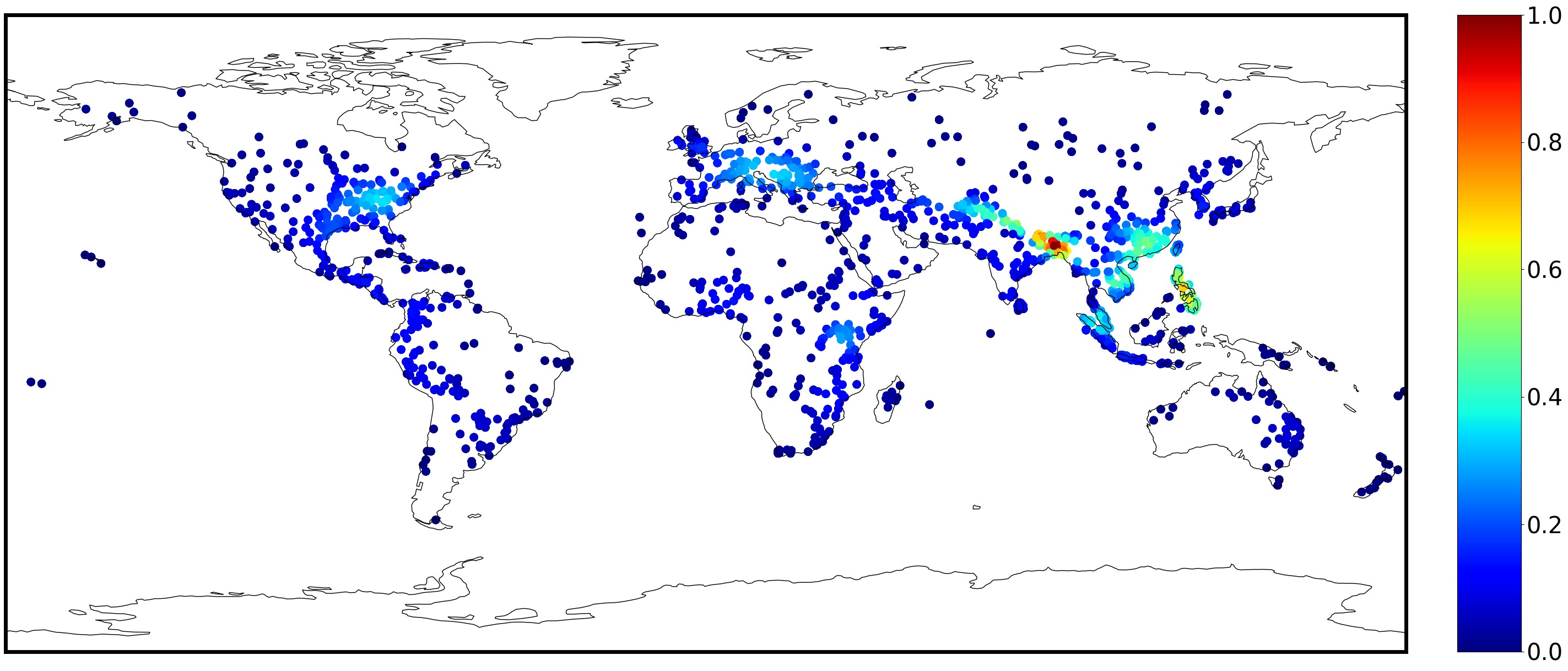}
	}
	\subfloat[Fire]{
		\includegraphics[width=0.3\textwidth]{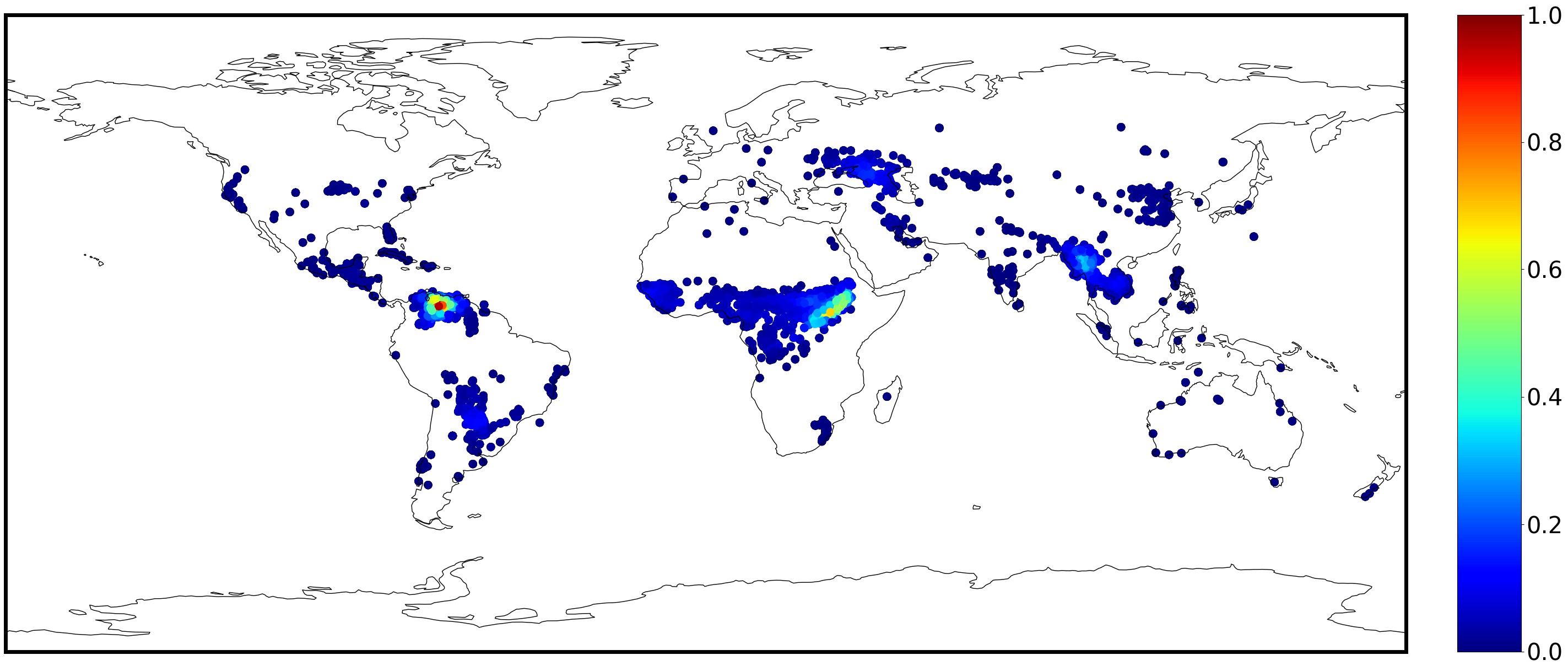}
	}
	
	\caption{DSSW (poly)}
	\label{fig:density_estimation_poly}
\end{figure*}

\begin{figure*}[ht]
	\centering
	\subfloat[Earthquake]{
		\includegraphics[width=0.3\textwidth]{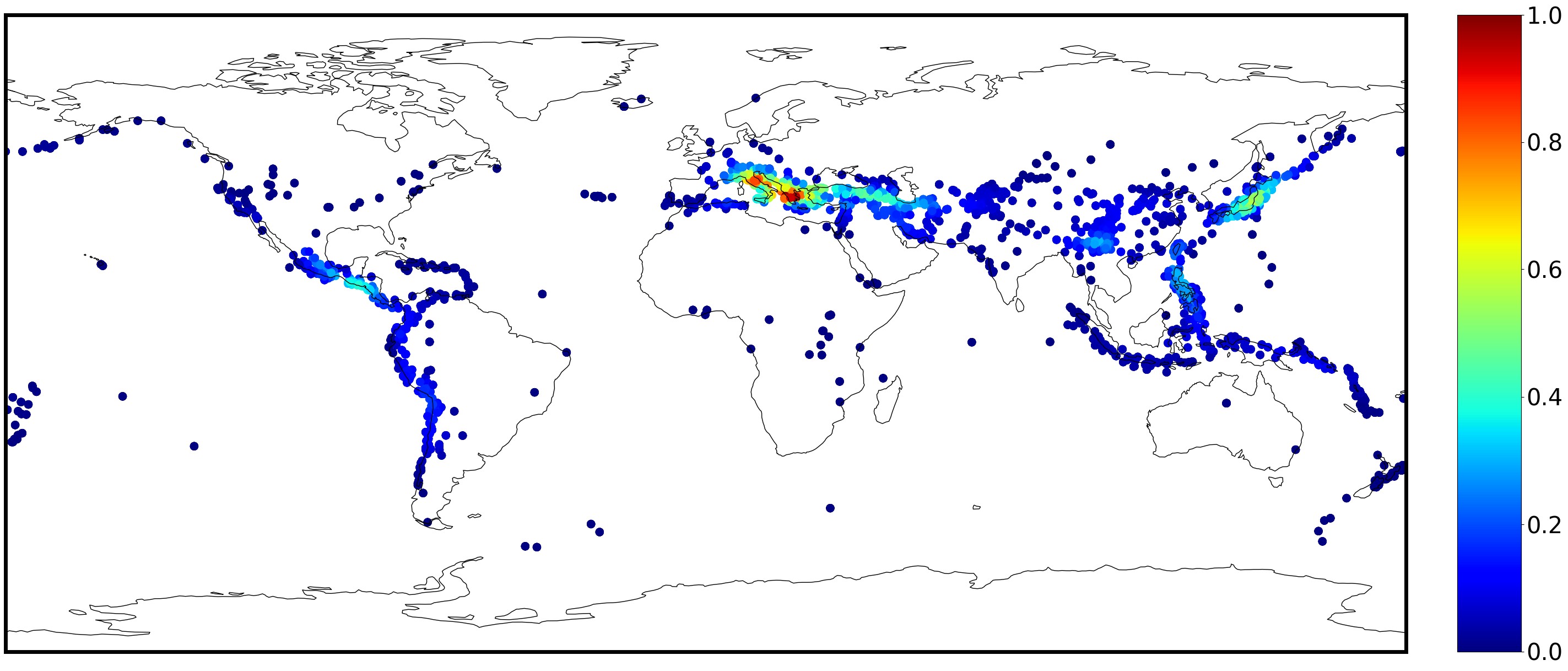}
	}
	\subfloat[Flood]{
		\includegraphics[width=0.3\textwidth]{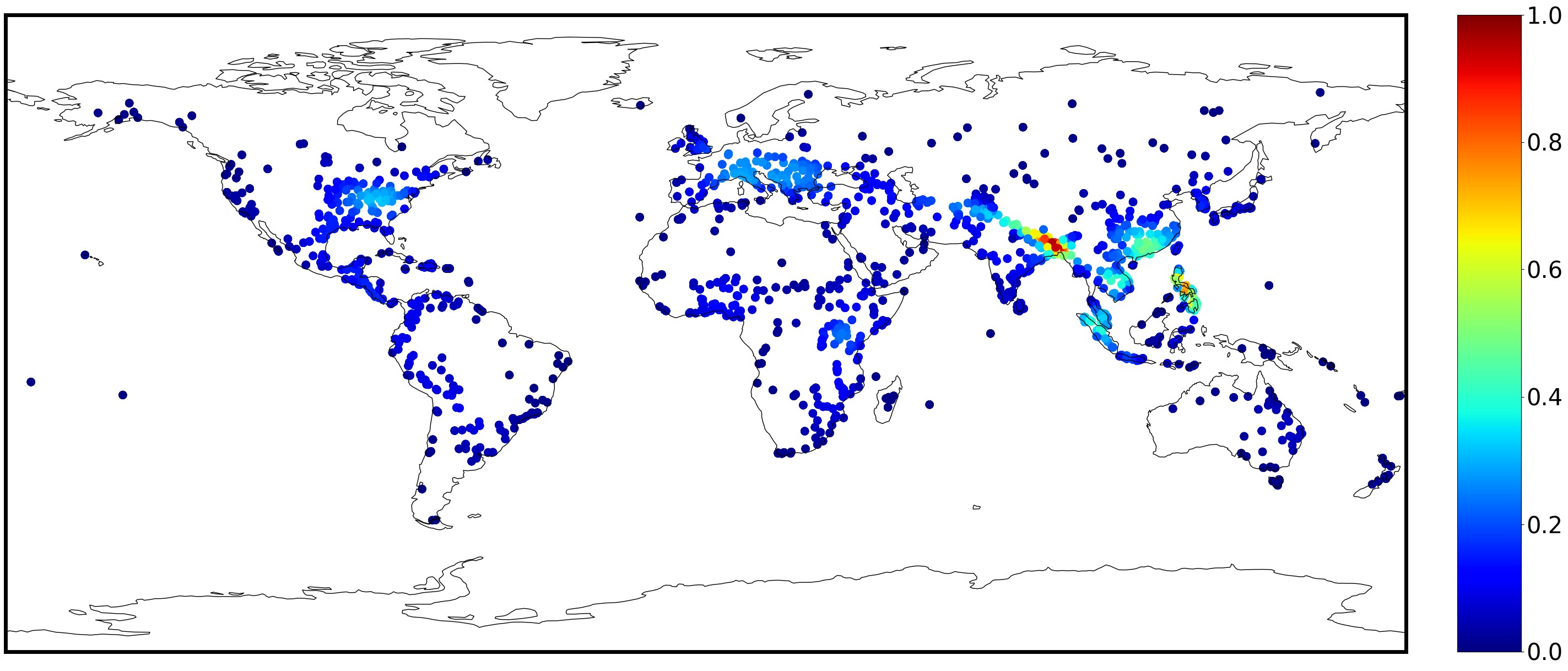}
	}
	\subfloat[Fire]{
		\includegraphics[width=0.3\textwidth]{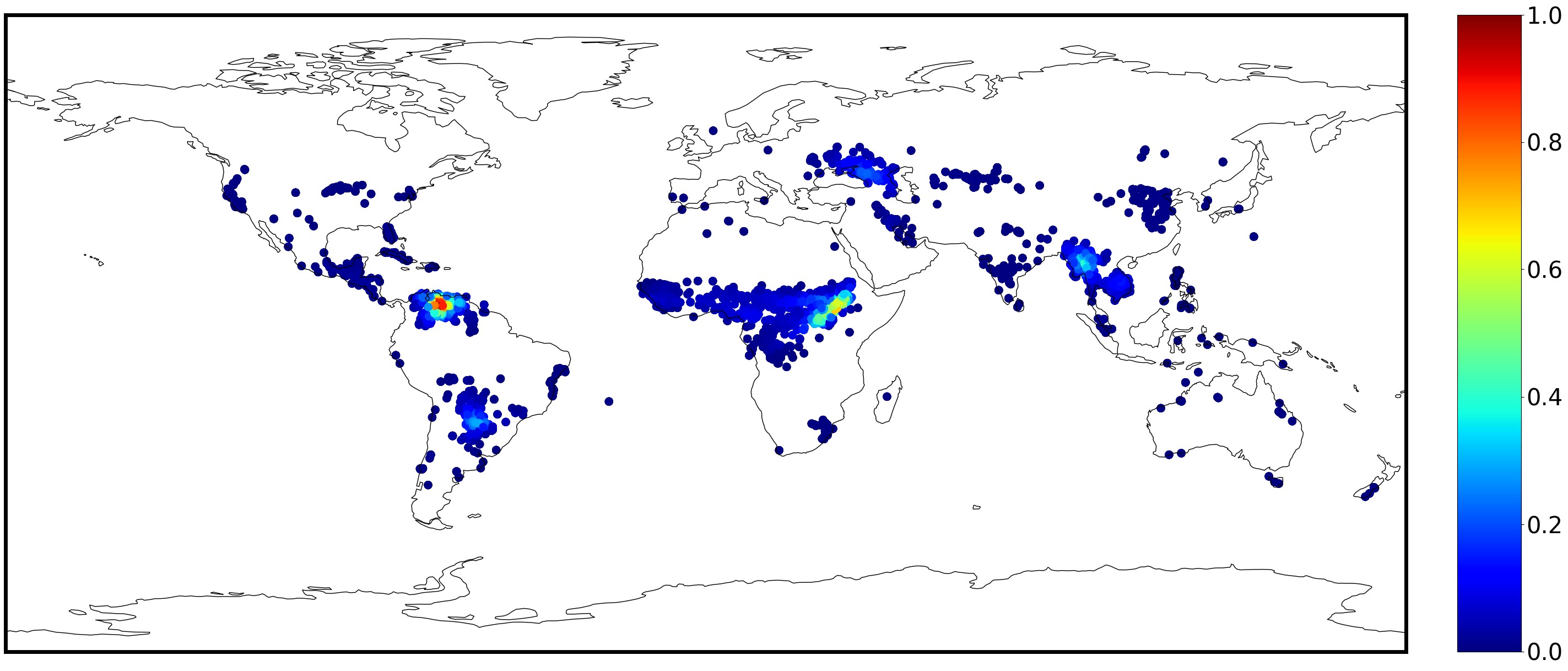}
	}
	
	\caption{DSSW (linear)}
	\label{fig:density_estimation_linear}
\end{figure*}

\begin{figure*}[ht]
	\centering
	\subfloat[Earthquake]{
		\includegraphics[width=0.3\textwidth]{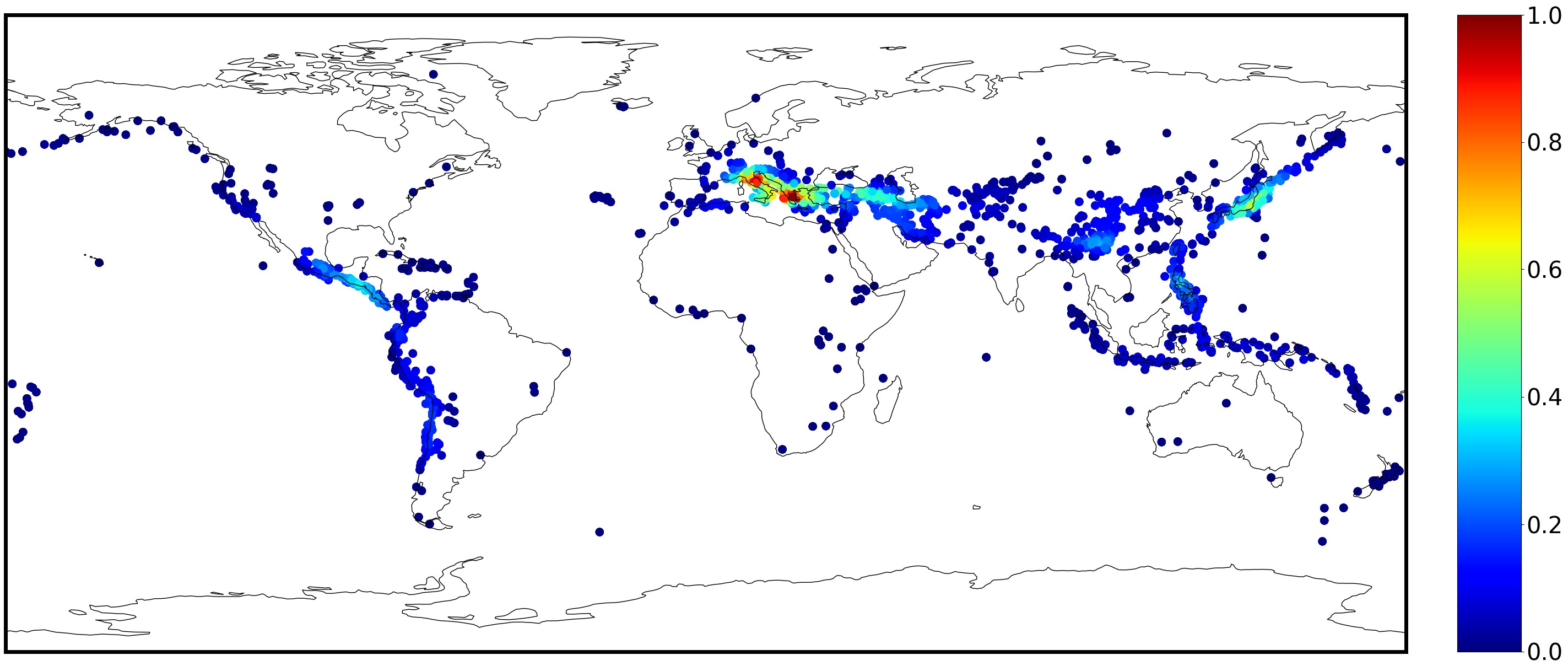}
	}
	\subfloat[Flood]{
		\includegraphics[width=0.3\textwidth]{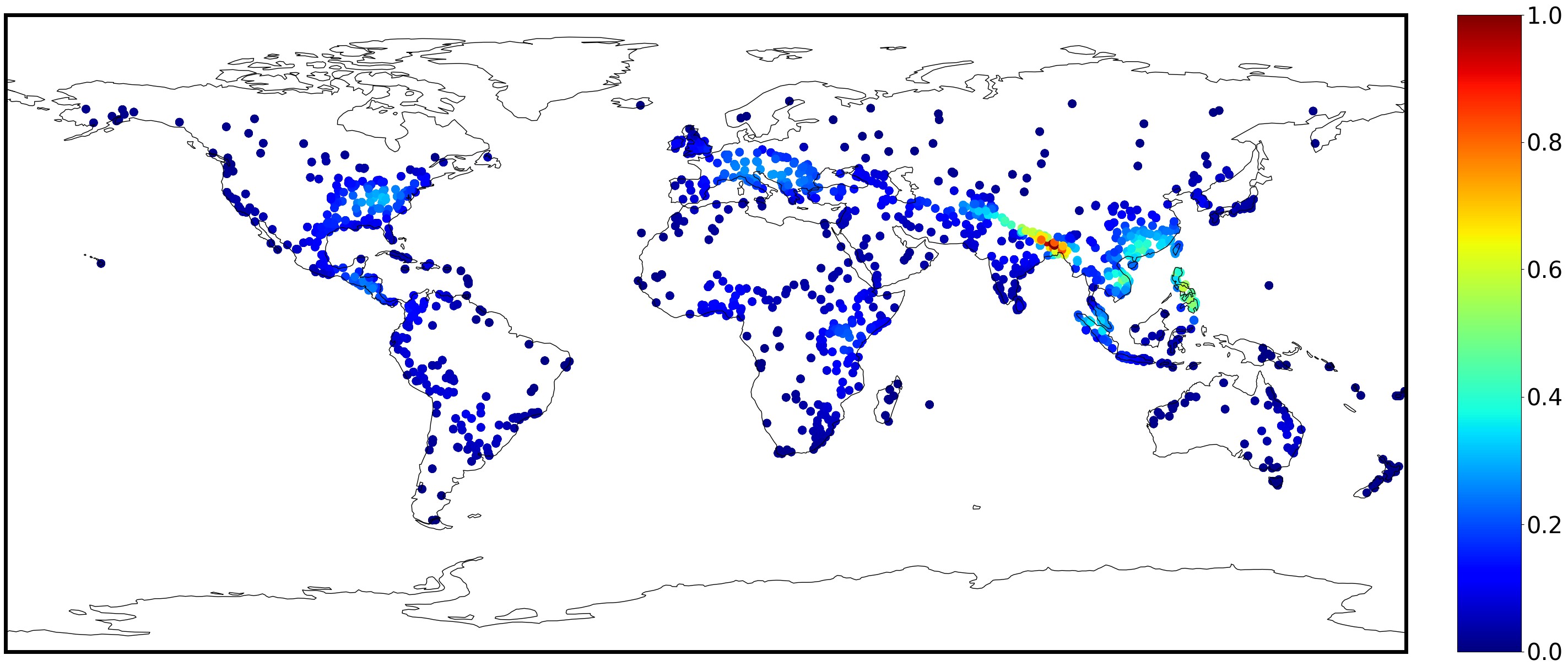}
	}
	\subfloat[Fire]{
		\includegraphics[width=0.3\textwidth]{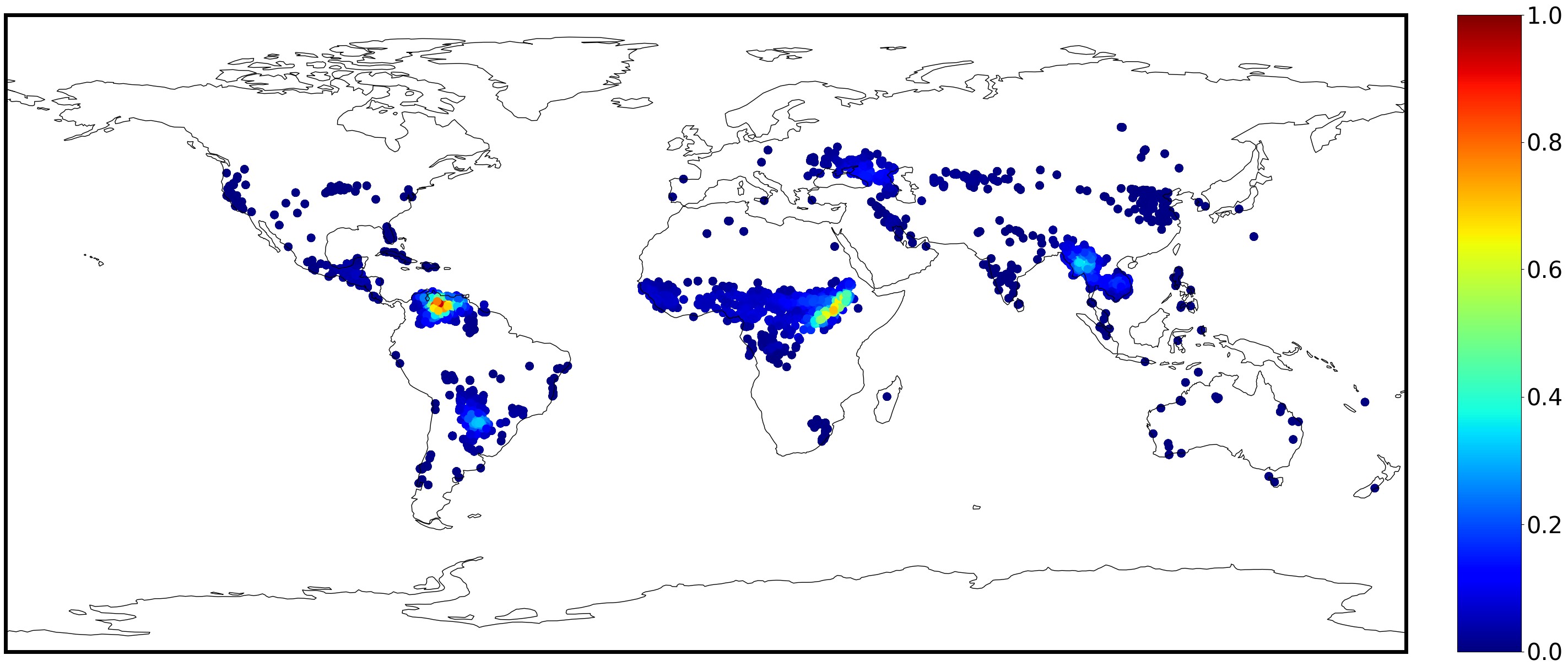}
	}
	
	\caption{DSSW (nonlinear)}
	\label{fig:density_estimation_nonlinear}
\end{figure*}

\begin{figure*}[ht]
	\centering
	\subfloat[Earthquake]{
		\includegraphics[width=0.3\textwidth]{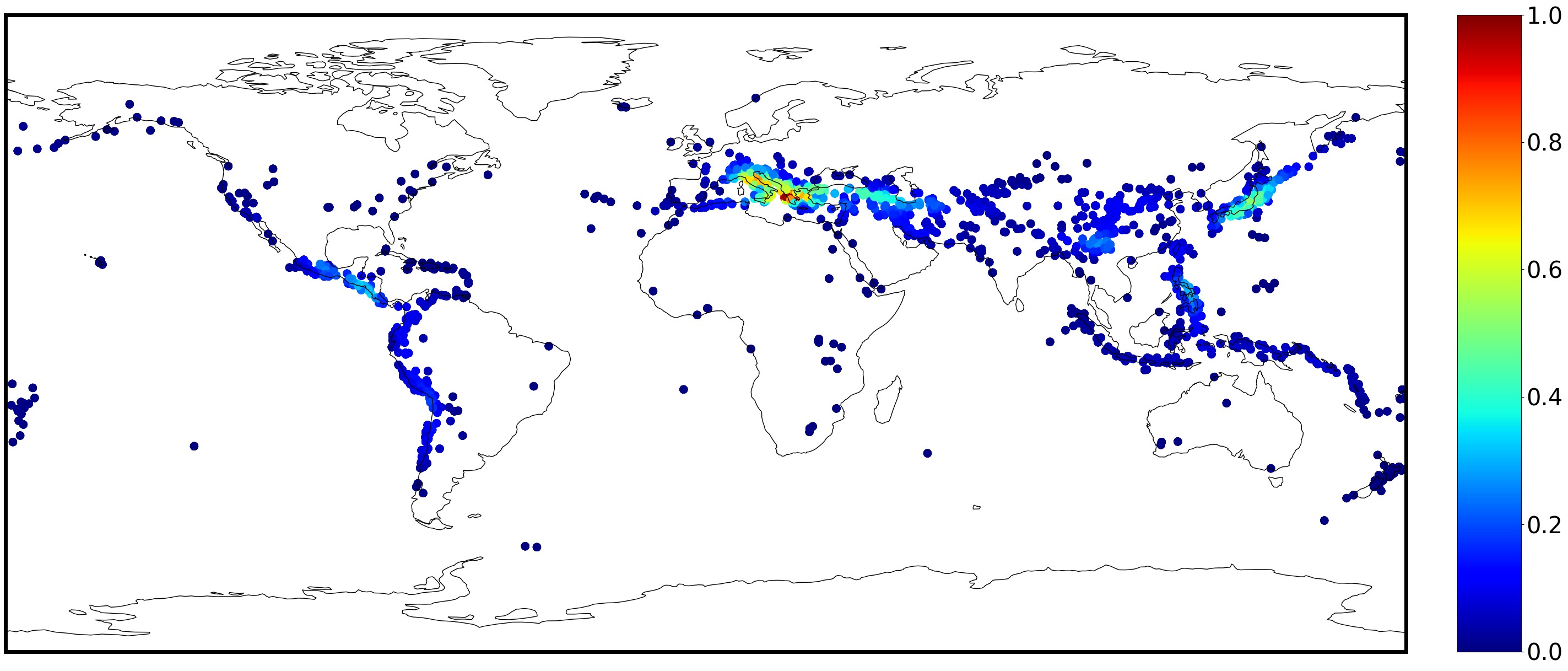}
	}
	\subfloat[Flood]{
		\includegraphics[width=0.3\textwidth]{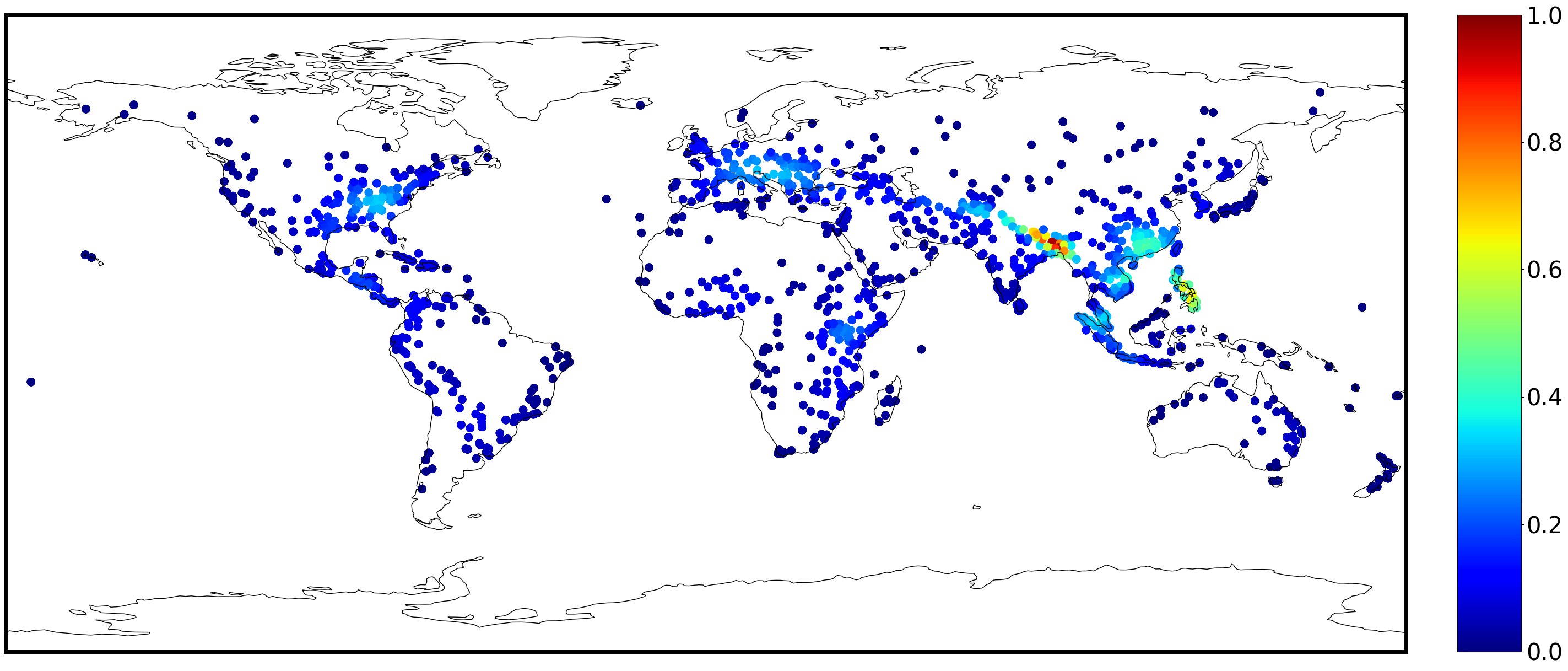}
	}
	\subfloat[Fire]{
		\includegraphics[width=0.3\textwidth]{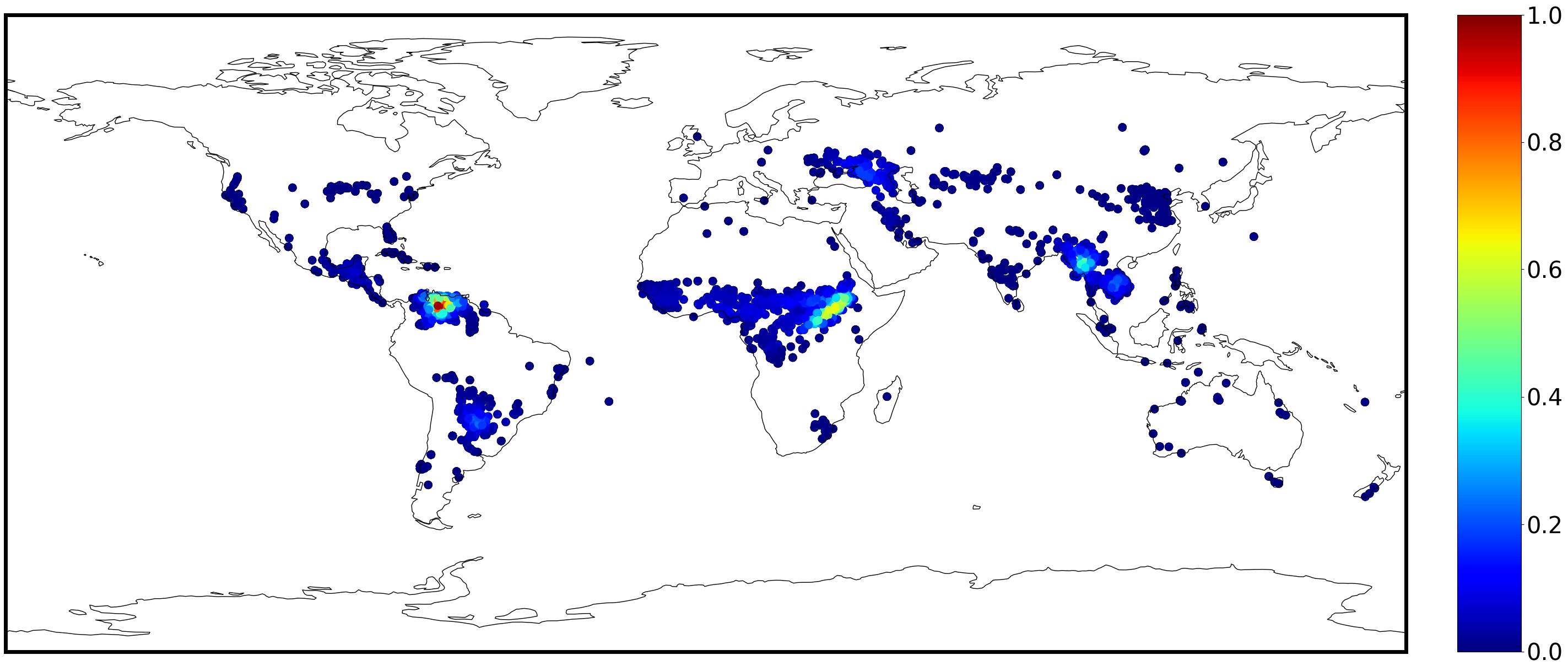}
	}
	
	\caption{DSSW (attention)}
	\label{fig:density_estimation_attention}
\end{figure*}

The ground truth as estimated with KDE using test data is shown in Figure \ref{fig:density_estimation_ground_truth}. We also demonstrate the learned density of test data by various metrics in Figure \ref{fig:density_estimation_stereo}-\ref{fig:density_estimation_attention}. We observe on the density visualization plot that the normalizing flows (NFs) put mass where most data points lie. The density visualization plot also indicates that our proposed DSSW learns the more accurate density distribution, this conclusion is consistent with the phenomenon that our method DSSW obtains the minimum Negative Log-likelihood value on all three datasets.

\subsection{Sliced-Wasserstein Autoencoder}
\textbf{Implementation.} In our experiments we adopt the same architectures as S3W, so the details of the architectures can be referred in S3W. In all experiments, we train the networks using the Adam optimizer with a learning rate of $10^{-3}$ and a batch size of 500 over 100 epochs. For CIFAR10, we set $\eta=0.001$ for SW, RI-S3W, and ARI-S3W, and $\eta=10$ for ssw and DSSW. For MNIST, we provide a comprehensive evaluation of the comparison methods w.r.t. varying regularization coefficients. for SW, SSW and DSSW we set $\eta \in \left \{ 1, 10, 100, 1000 \right \}$, and for S3W, RI-S3W and ARI-S3W we set $\eta \in \left \{ 0.001, 0.01, 0.1, 1.0 \right \}$. For all experiments we employ $L=100$ projections for all distance metrics, we set $N_R=5$ for RI-S3W and a pool size of 100 for ARI-S3W.

\begin{figure}[ht]
	\centering
	\subfloat[AE]{
		\includegraphics[width=0.45\textwidth]{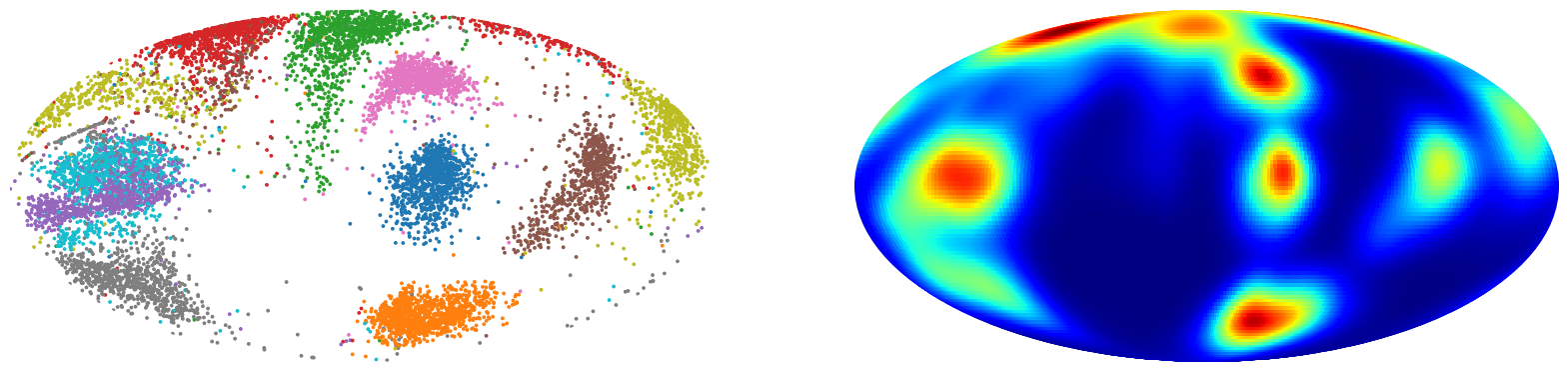}
	}
	\subfloat[SW]{
		\includegraphics[width=0.45\textwidth]{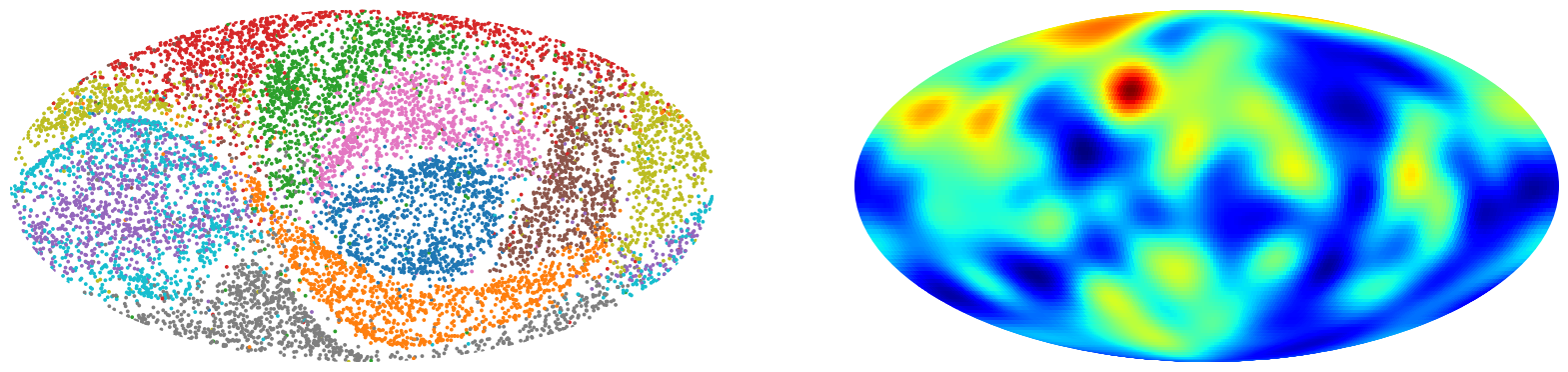}
	}\\
	\subfloat[SSW]{
		\includegraphics[width=0.45\textwidth]{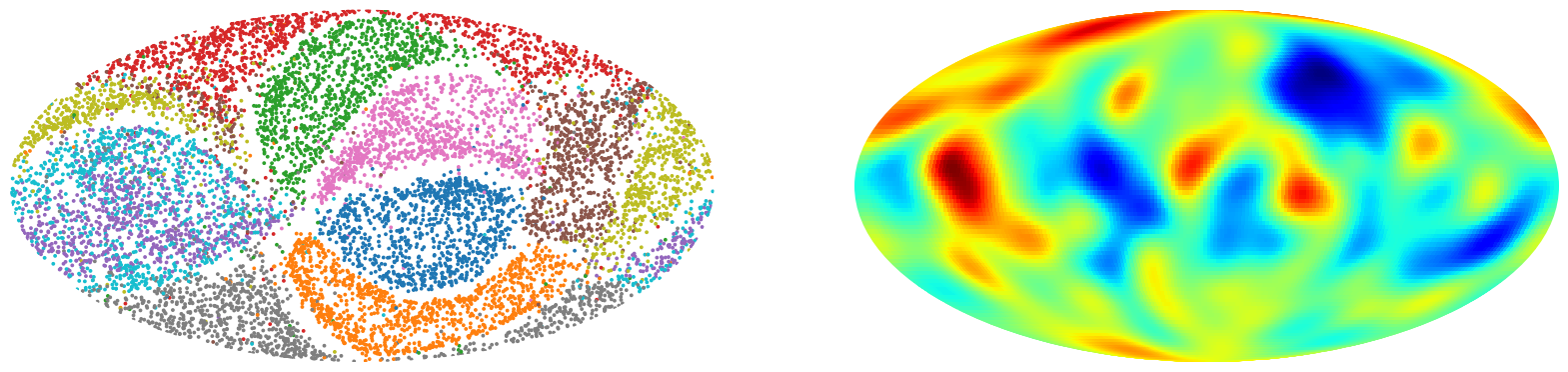}
	}
	\subfloat[S3W]{
		\includegraphics[width=0.45\textwidth]{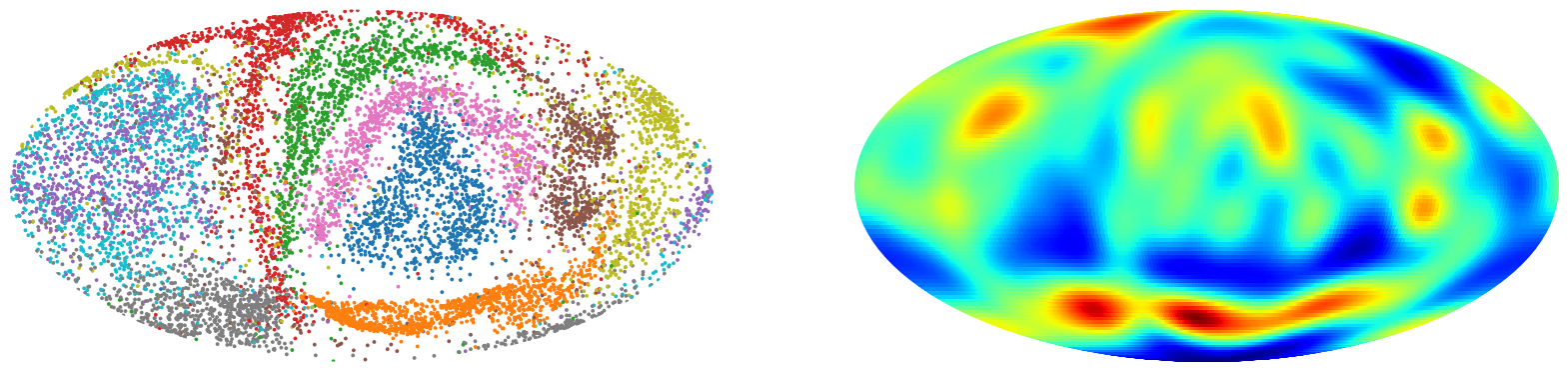}
	}\\
	\subfloat[RI-S3W (1)]{
		\includegraphics[width=0.45\textwidth]{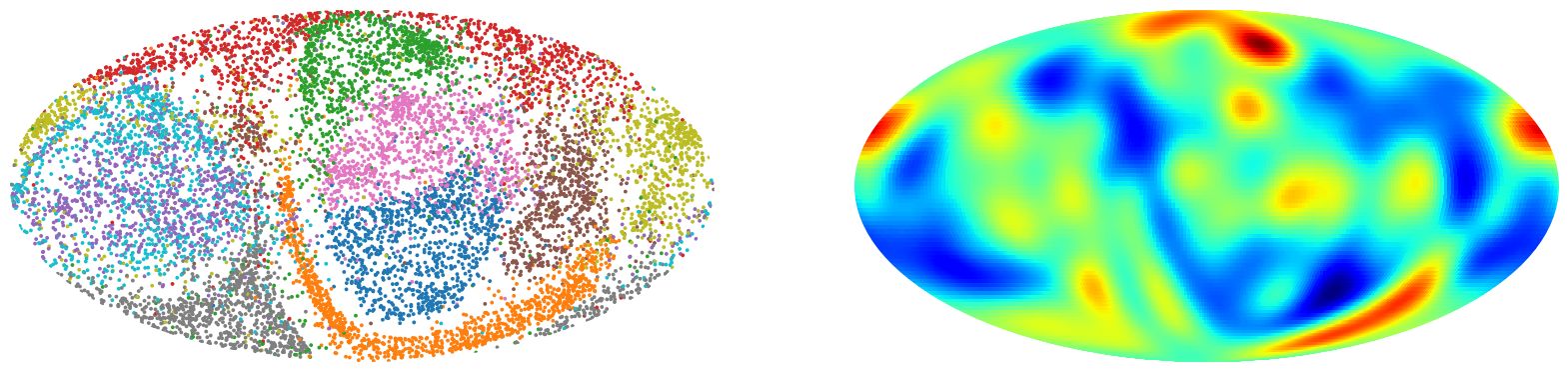}
	}
	\subfloat[RI-S3W (10)]{
		\includegraphics[width=0.45\textwidth]{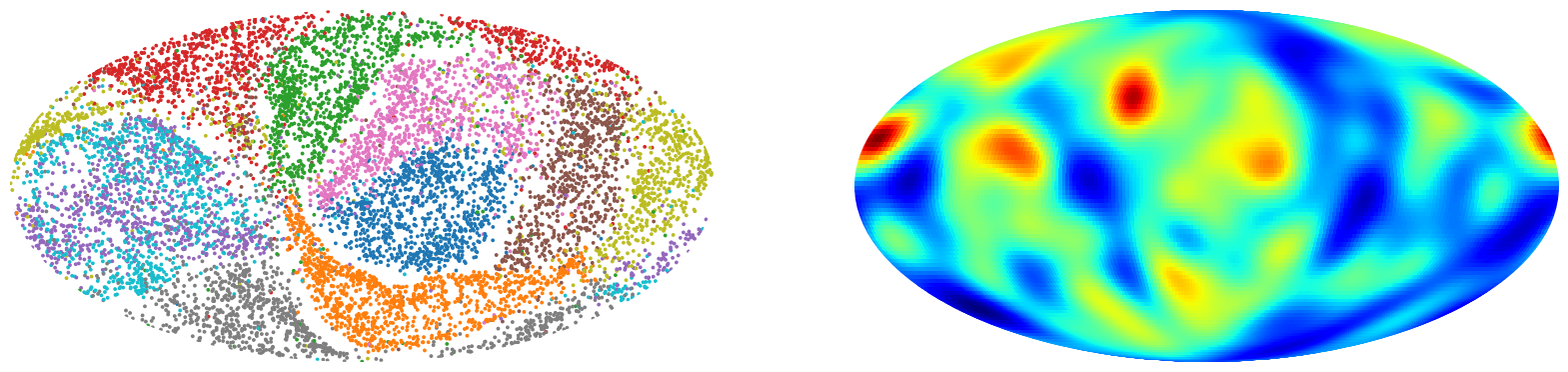}
	}\\
	\subfloat[ARI-S3W (10)]{
		\includegraphics[width=0.45\textwidth]{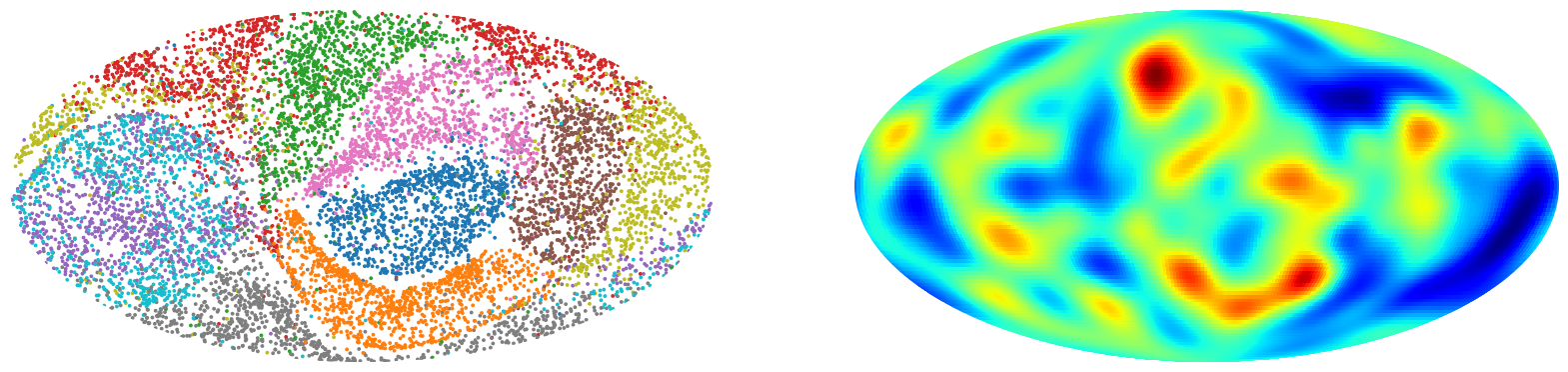}
	}
	\subfloat[ARI-S3W (30)]{
		\includegraphics[width=0.45\textwidth]{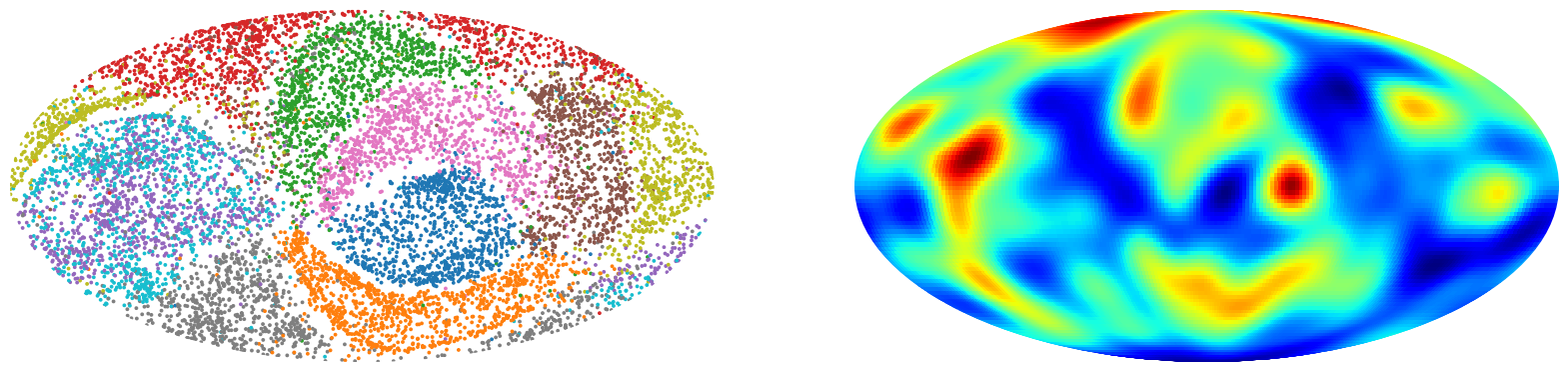}
	}\\
	\subfloat[DSSW (exp)]{
		\includegraphics[width=0.45\textwidth]{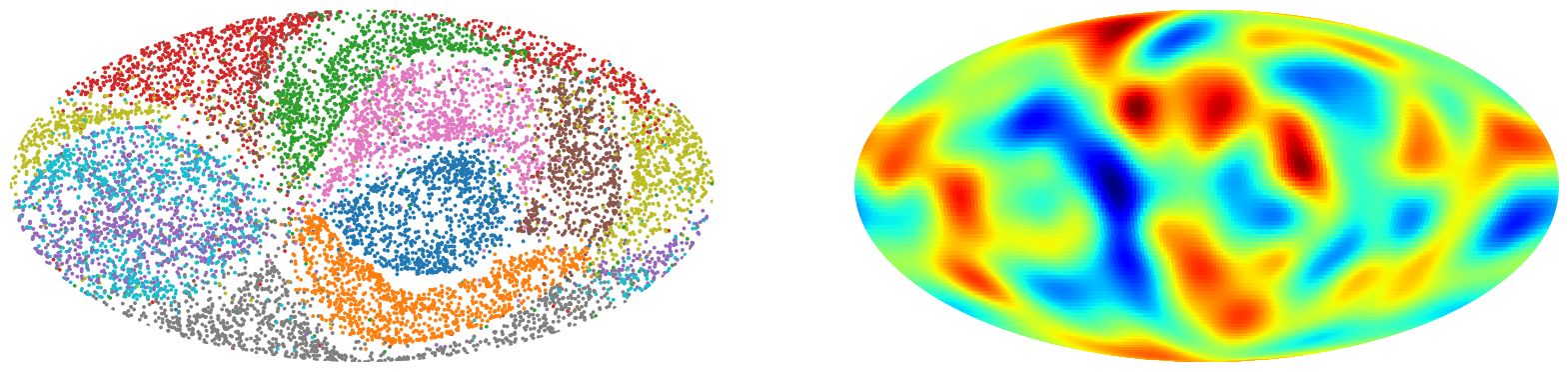}
	}
	\subfloat[DSSW (identity)]{
		\includegraphics[width=0.45\textwidth]{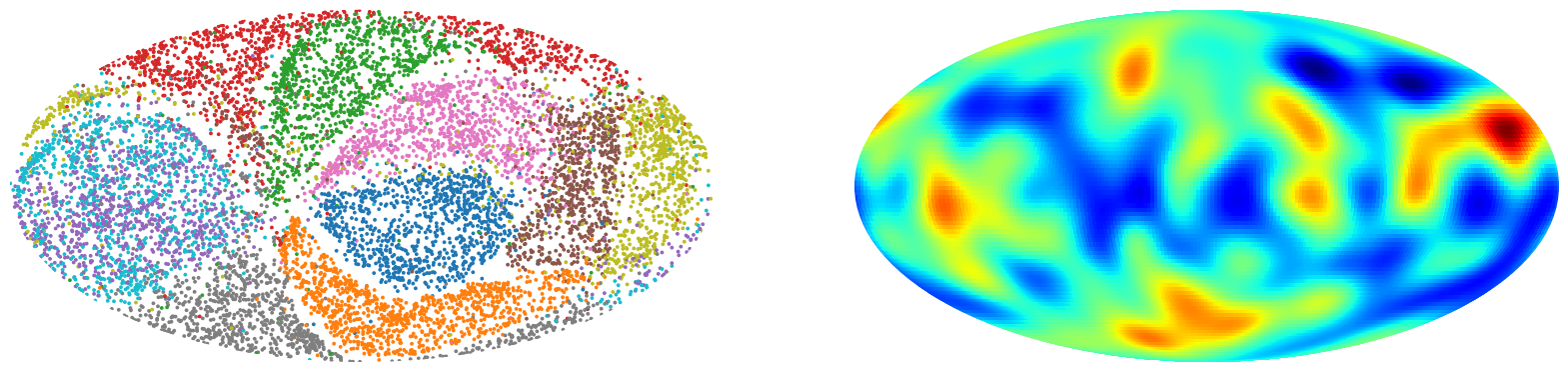}
	}\\
	\subfloat[DSSW (poly)]{
		\includegraphics[width=0.45\textwidth]{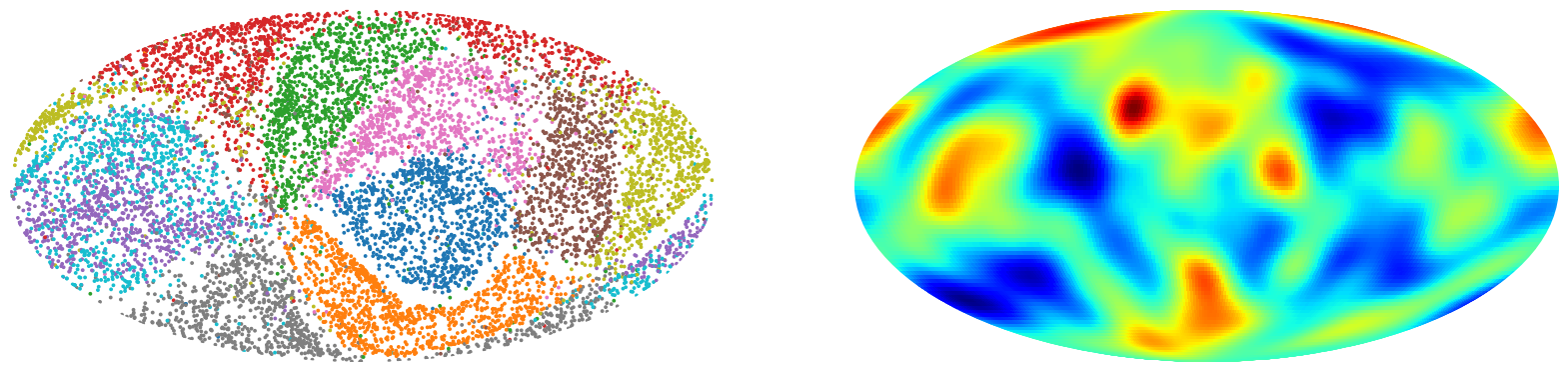}
	}
	\subfloat[DSSW (linear)]{
		\includegraphics[width=0.45\textwidth]{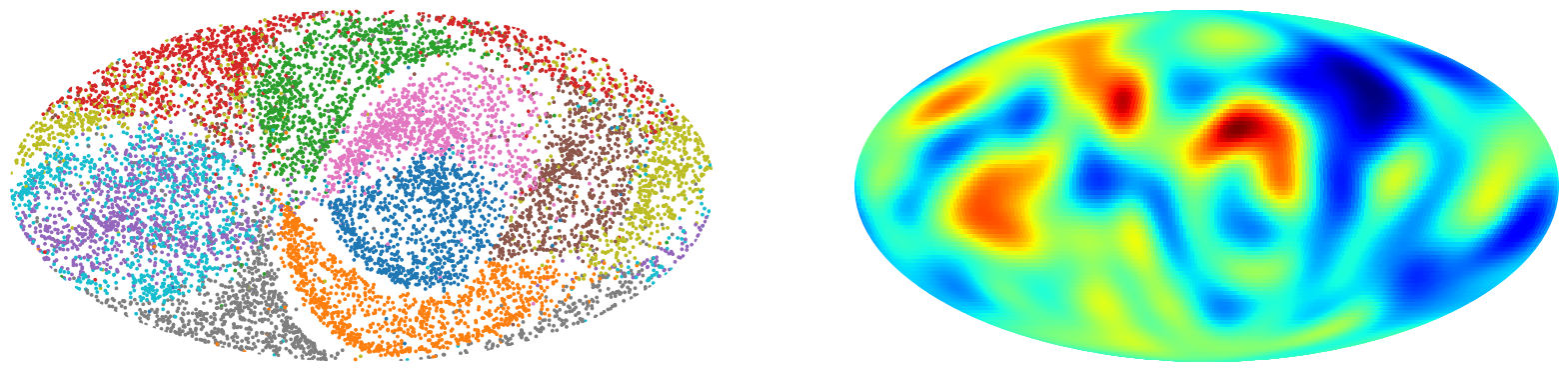}
	}\\
	\subfloat[DSSW (nonlinear)]{
		\includegraphics[width=0.45\textwidth]{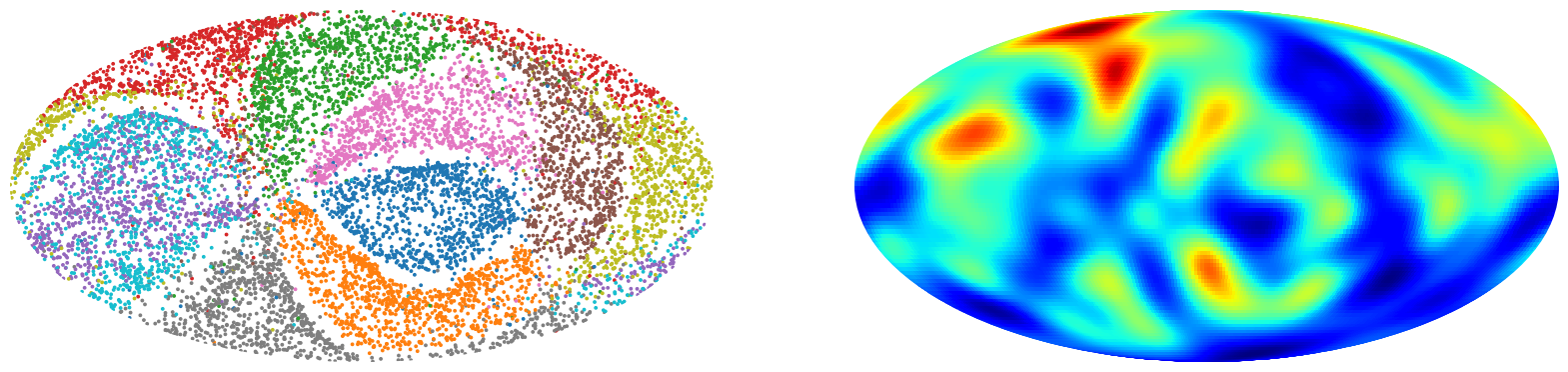}
	}
	\subfloat[DSSW (attention)]{
		\includegraphics[width=0.45\textwidth]{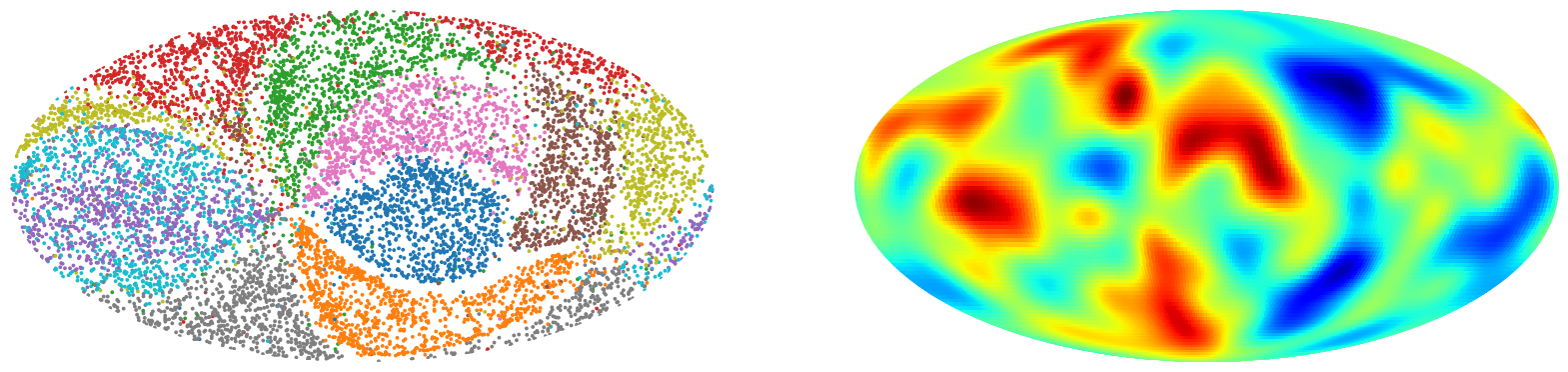}
	}
	\caption{The latent space visualization of MNIST}
	\label{fig:swae_mnist}
\end{figure}

\begin{figure}[ht]
	\centering
	\subfloat[SW]{
		\includegraphics[width=0.25\textwidth]{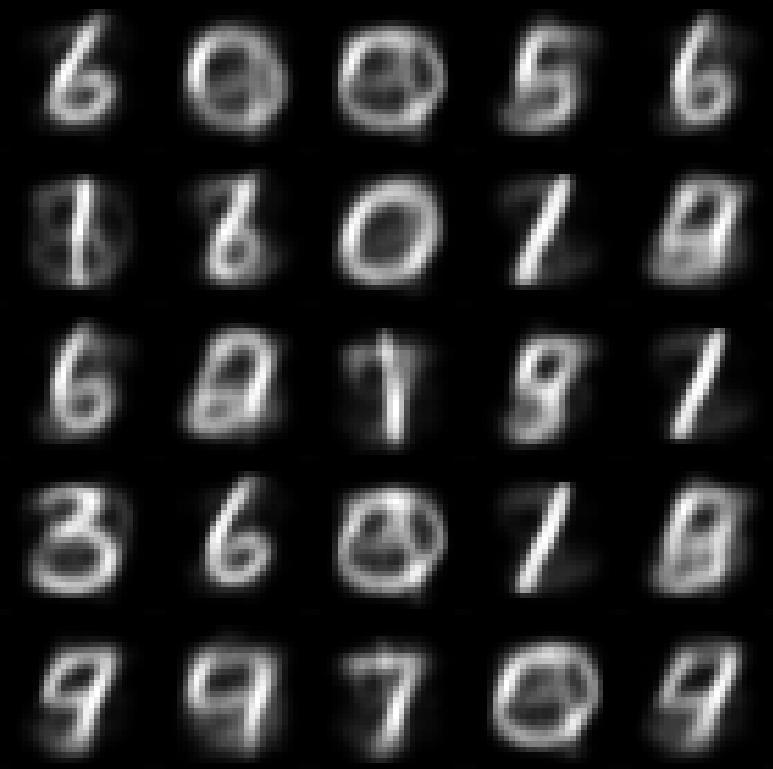}
	}
	\subfloat[SSW]{
		\includegraphics[width=0.25\textwidth]{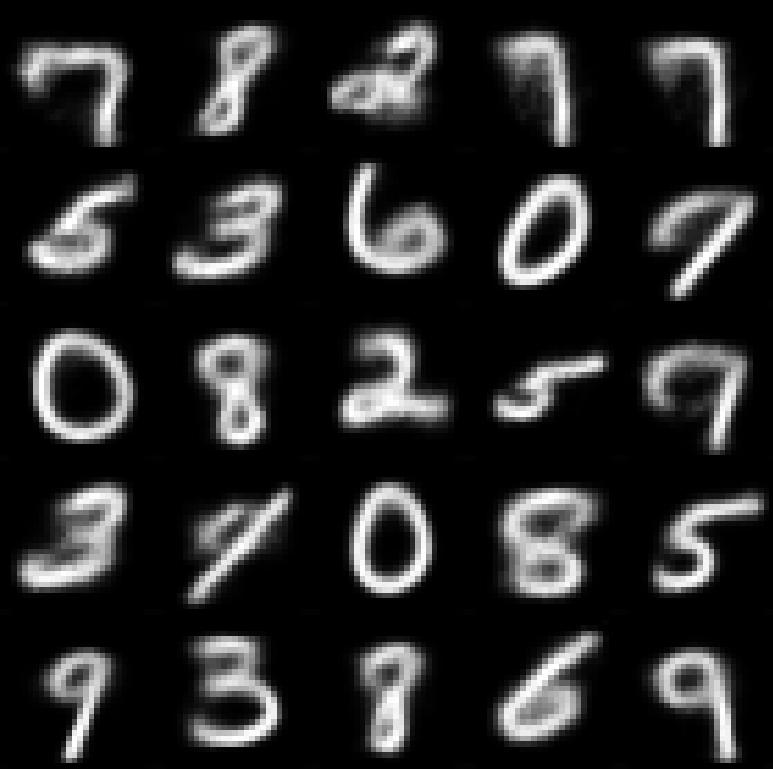}
	}
	\subfloat[S3W]{
		\includegraphics[width=0.25\textwidth]{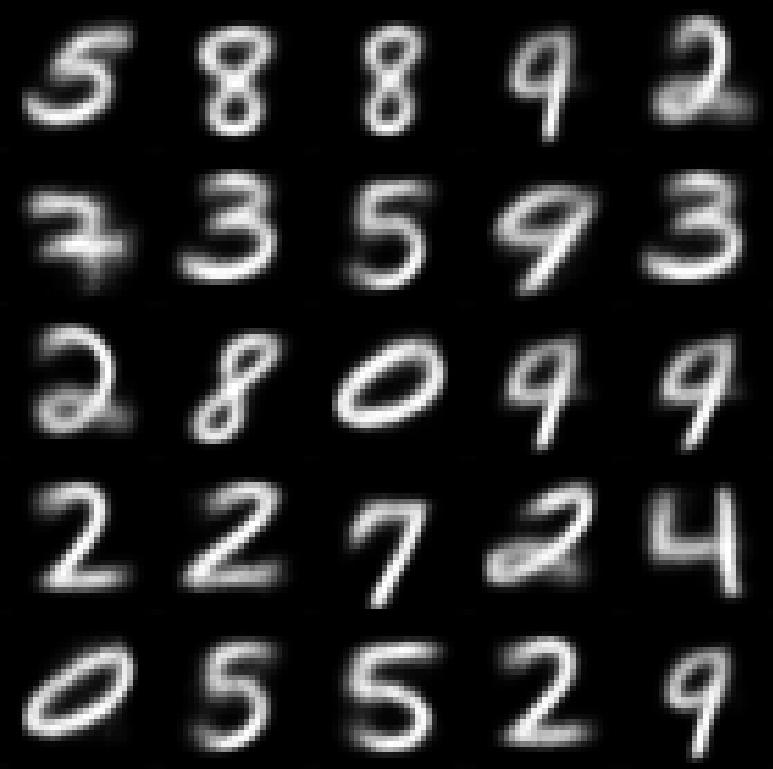}
	}\\
	\subfloat[RI-S3W (1)]{
		\includegraphics[width=0.25\textwidth]{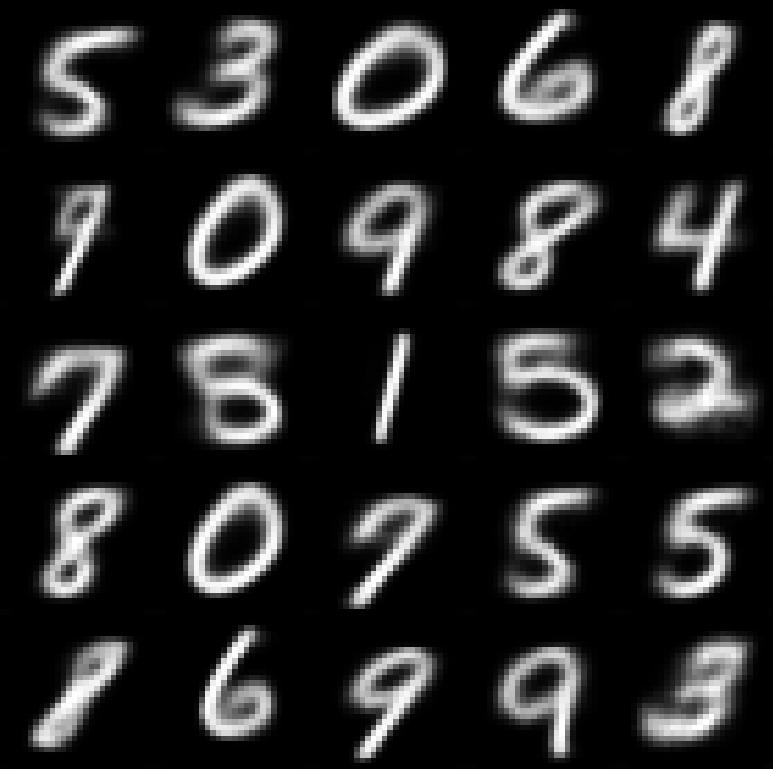}
	}
	\subfloat[RI-S3W (10)]{
		\includegraphics[width=0.25\textwidth]{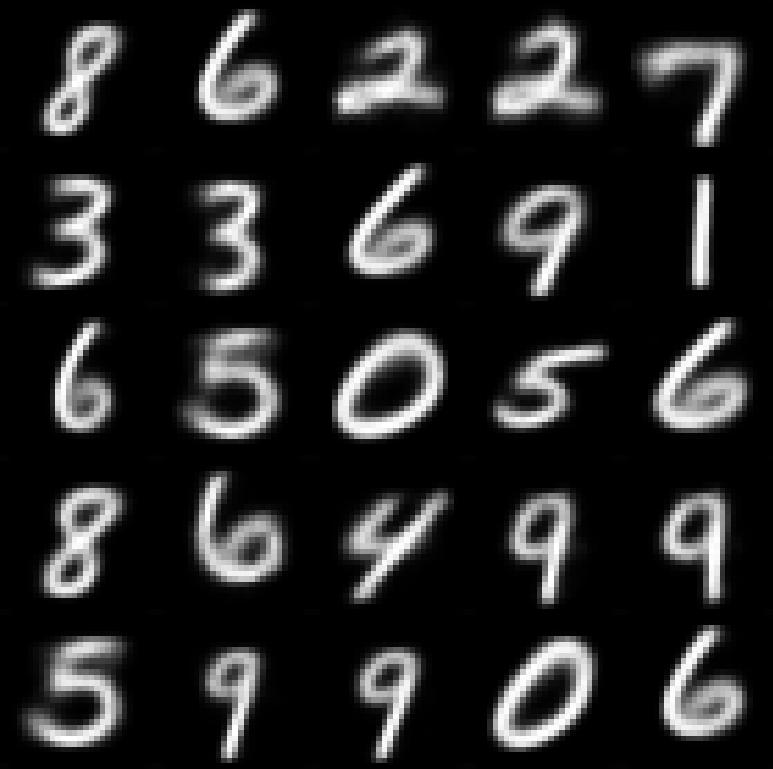}
	}
	\subfloat[ARI-S3W (30)]{
		\includegraphics[width=0.25\textwidth]{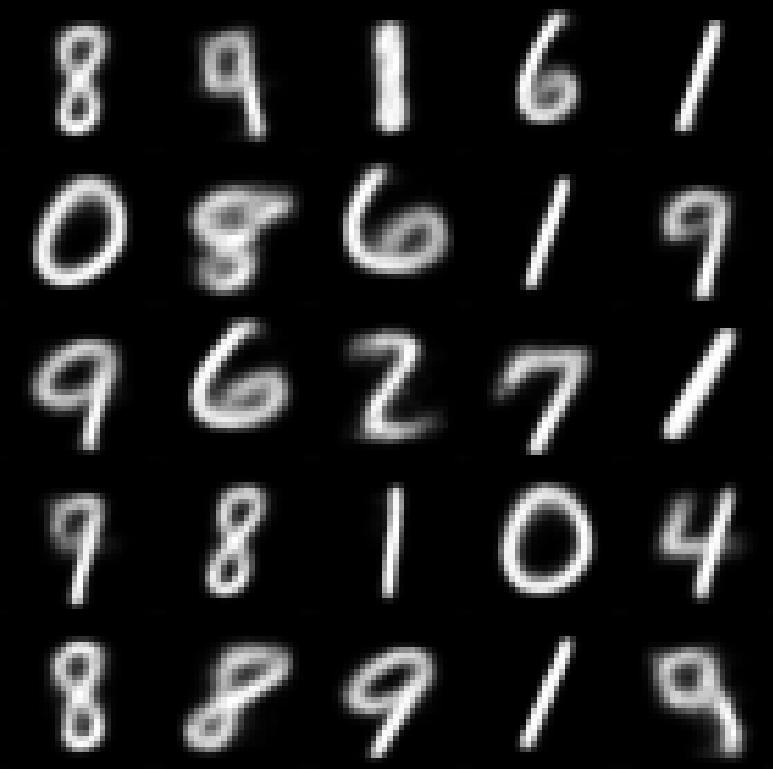}
	}\\
	\subfloat[DSSW (exp)]{
		\includegraphics[width=0.25\textwidth]{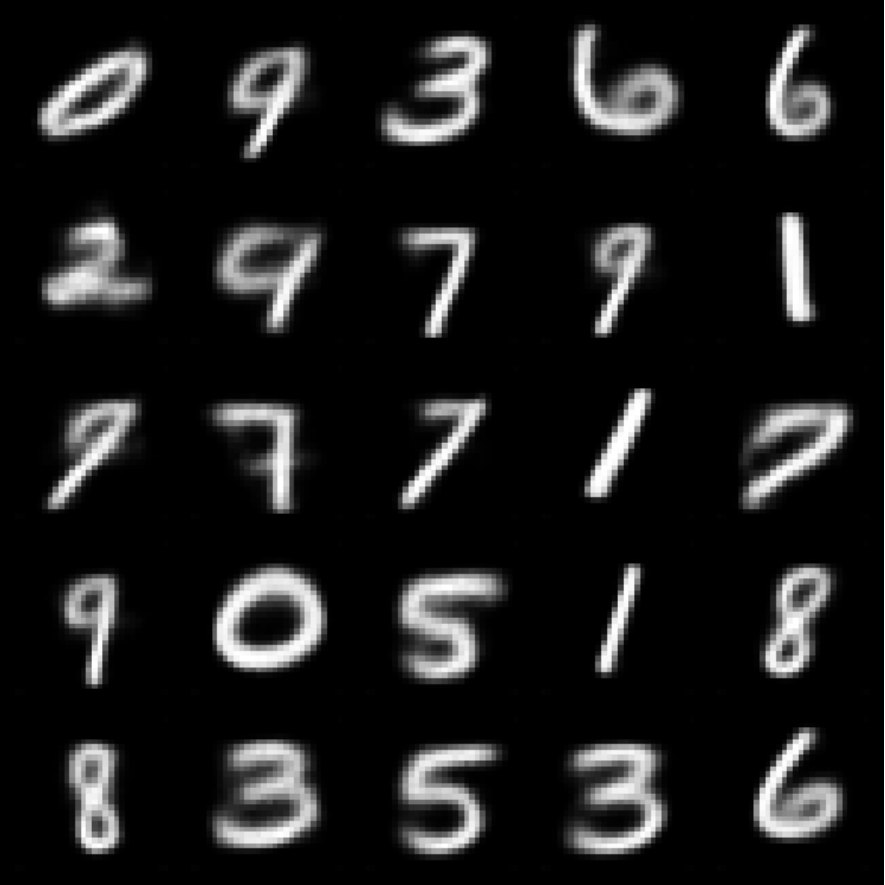}
	}
	\subfloat[DSSW (identity)]{
		\includegraphics[width=0.25\textwidth]{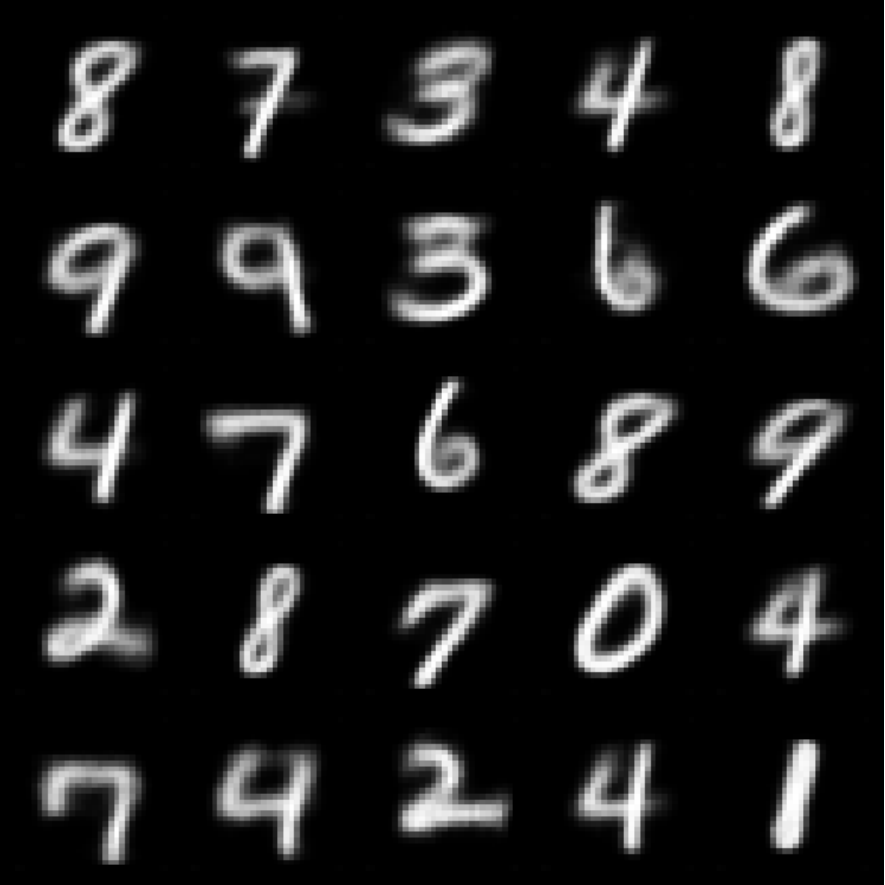}
	}
	\subfloat[DSSW (poly)]{
		\includegraphics[width=0.25\textwidth]{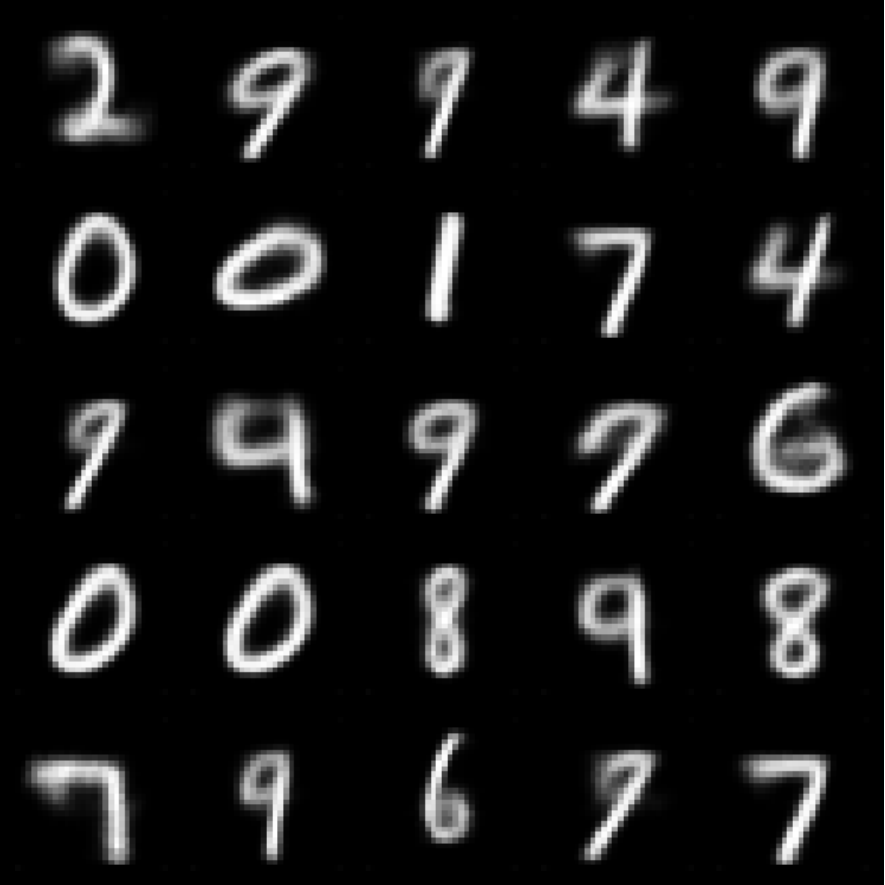}
	}\\
	\subfloat[DSSW (linear)]{
		\includegraphics[width=0.25\textwidth]{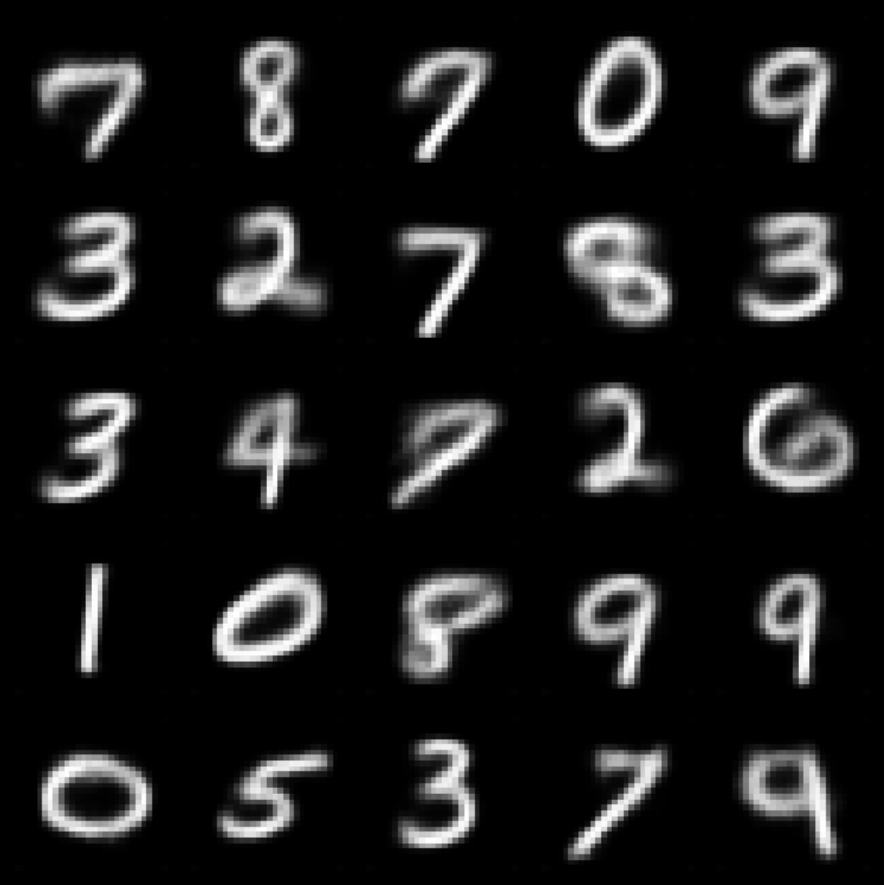}
	}
	\subfloat[DSSW (nonlinear)]{
		\includegraphics[width=0.25\textwidth]{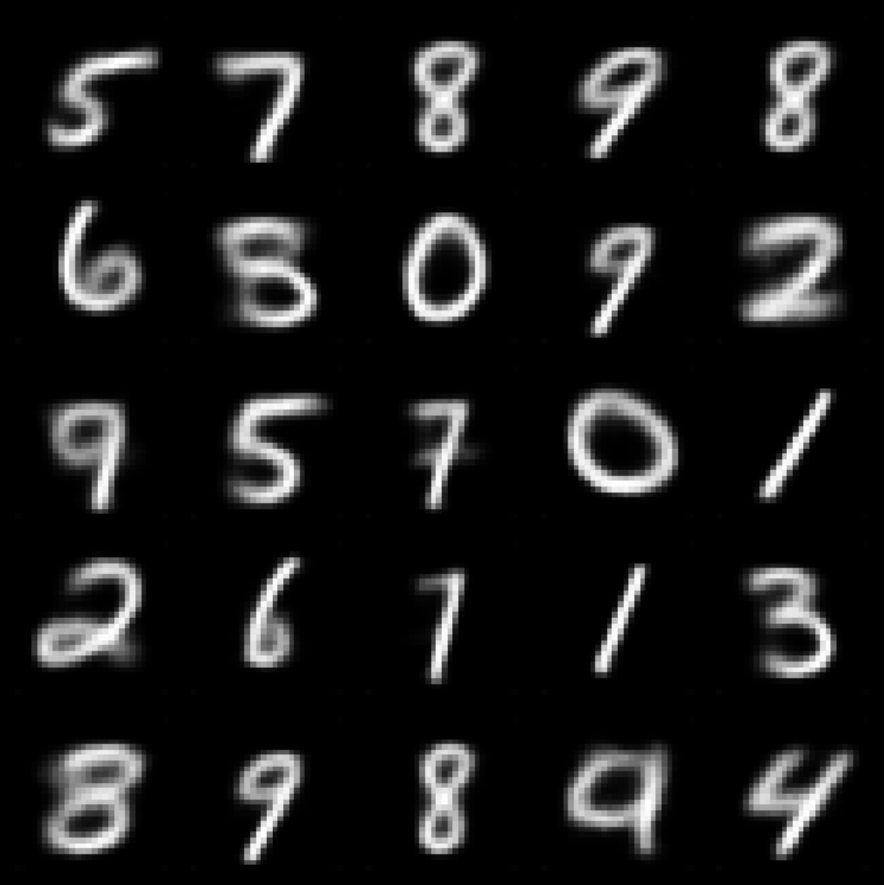}
	}
	\subfloat[DSSW (attention)]{
		\includegraphics[width=0.25\textwidth]{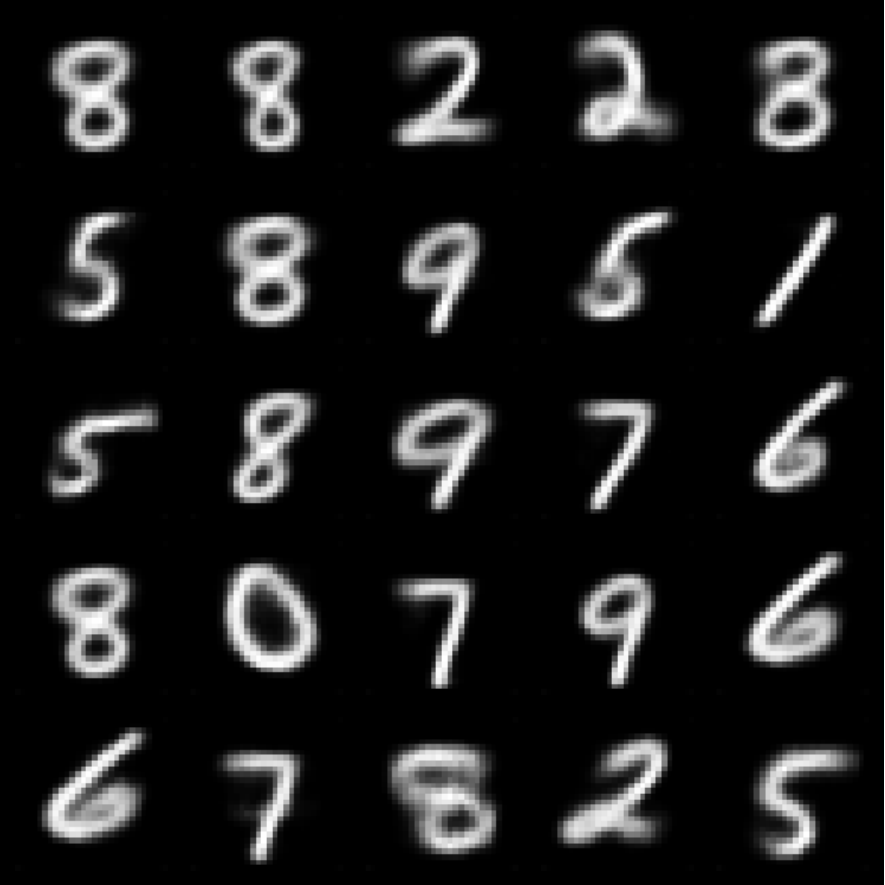}
	}
	\caption{Samples generated by SWAE models with a uniform prior on $\mathbb{S}^{2}$. The figures of baselines are cited from S3W.}
	\label{fig:samples_mnist}
\end{figure}

The results on MNIST can be seen in Table \ref{tab:swae_mnist}. These results indicate that DSSW with the projected energy function of linear neural network obtains the smallest $\log{W_{2}}$ and DSSW with the projected energy function of attention mechanism obtains the smallest BCE loss. These results reveal that the regularization coefficient $\eta$ can adjust the impact of the reconstruction loss (BCE) and the regularizer. Specifically, the bigger value of $\eta$ leads to  the lower value of $\log{W_{2}}$, the smaller value of $\eta$ results in the lower value of BCE. In this experiment we adopt the uniform distribution $\mathrm{Unif} \left ( \mathbb{S} ^{2} \right)$ as the latent prior. We report the mean and standard deviation of the FID scores using 10000 samples over 5 independent runs in Table \ref{tab:mnist_fid}. Moreover, we visualize samples generated by each model in Figure \ref{fig:samples_mnist}. We also show the latent space visualization about this experiment in Figure \ref{fig:swae_mnist}. The visualization plot indicates that the cluster result of embedding obtained by our method DSSW is better than other methods in the latent space.

\begin{table}[ht]
	\centering
	\begin{tabular}{cccc}
		\toprule
		Method                                & $\eta$     & $\log{W_{2}}$ $\downarrow$            & BCE $\downarrow$              \\ \midrule
		Supervised AE                         & 1     & -1.8946 ± 0.1492 & 0.1615 ± 0.0026 \\ \midrule
		\multirow{4}{*}{SW}                   & 1     & -2.9277 ± 0.1504 & 0.2217 ± 0.0022 \\
		& 10    & -2.9745 ± 0.2389   & 0.2288 ± 0.0037   \\
		& 100   & -2.9610 ± 0.1310   & 0.2284 ± 0.0043   \\
		& 1000  & \underline{-3.0618 ± 0.0776}   & 0.2275 ± 0.0069   \\ \midrule
		\multirow{4}{*}{SSW}                  & 1     & -2.3619 ± 0.1245   & 0.1620 ± 0.0017 \\
		& 10    & -2.5087 ± 0.1754   & 0.1625 ± 0.0027   \\
		& 100   & -2.6118 ± 0.1233   & 0.1659 ± 0.0004   \\
		& 1000  & -2.8418 ± 0.0678   & 0.2096 ± 0.0049   \\ \midrule
		\multirow{4}{*}{S3W}                  & 0.001 & -2.4723 ± 0.0720 & 0.1627 ± 0.0022 \\
		& 0.01  & -2.5265 ± 0.1035   & 0.1632 ± 0.0011   \\
		& 0.1   & -2.6112 ± 0.1866   & 0.1926 ± 0.0067   \\
		& 1     & -2.9694 ± 0.0998   & 0.2232 ± 0.0056   \\ \midrule
		\multirow{4}{*}{RI-S3W (1)}           & 0.001 & -2.3256 ± 0.0742 & 0.1622 ± 0.0009 \\
		& 0.01  & -2.5146 ± 0.0613   & 0.1651 ± 0.0006   \\
		& 0.1   & -2.6851 ± 0.1704   & 0.1982 ± 0.0019   \\
		& 1     & -2.8612 ± 0.1945   & 0.2264 ± 0.0057   \\ \midrule
		\multirow{4}{*}{RI-S3W (10)}          & 0.001 & -2.4326 ± 0.1191 & 0.1624 ± 0.0014 \\
		& 0.01  & -2.5731 ± 0.1018   & 0.1631 ± 0.0006   \\
		& 0.1   & -2.7127 ± 0.1353   & 0.1858 ± 0.0025   \\
		& 1     & -2.7635 ± 0.1923   & 0.2229 ± 0.0022   \\ \midrule
		\multirow{4}{*}{ARI-S3W (10)}         & 0.001 & -2.4106 ± 0.1232 & \underline{0.1619 ± 0.0018} \\
		& 0.01  & -2.5441 ± 0.0736   & 0.1627 ± 0.0010   \\
		& 0.1   & -2.6902 ± 0.1491   & 0.1856 ± 0.0044   \\
		& 1     & -2.9553 ± 0.1338   & 0.2248 ± 0.0066   \\ \midrule
		\multirow{4}{*}{ARI-S3W (30)}         & 0.001 & -2.3879 ± 0.1630 & 0.1623 ± 0.0015 \\
		& 0.01  & -2.5286 ± 0.0646   & 0.1625 ± 0.0009   \\
		& 0.1   & -2.6305 ± 0.0997   & 0.1835 ± 0.0055   \\
		& 1     & -2.8889 ± 0.1171   & 0.2207 ± 0.0032   \\ \midrule
		\multirow{4}{*}{DSSW (exp)}         & 1     & -2.3547 ± 0.0977 & 0.1615 ± 0.0005 $\ddagger$ \\
		& 10    & -2.5247 ± 0.1343   & 0.1627 ± 0.0007   \\
		& 100   & -2.7222 ± 0.0962   & 0.1776 ± 0.0006   \\
		& 1000  & -2.9398 ± 0.2408   & 0.2221 ± 0.0026   \\ \midrule
		\multirow{4}{*}{DSSW   (identity)}  & 1     & -2.3673 ± 0.0893 & 0.1620 ± 0.0012 \\
		& 10    & -2.5048 ± 0.1437   & 0.1623 ± 0.0015   \\
		& 100   & -2.6176 ± 0.1942   & 0.1810 ± 0.0026   \\
		& 1000  & -3.0093 ± 0.1765   & 0.2205 ± 0.0054   \\ \midrule
		\multirow{4}{*}{DSSW (poly)}        & 1     & -2.3536 ± 0.0968 & 0.1617 ± 0.0011 $\ddagger$ \\
		& 10    & -2.5051 ± 0.1489   & 0.1628 ± 0.0014   \\
		& 100   & -2.7369 ± 0.1043   & 0.1840 ± 0.0028   \\
		& 1000  & -2.9645 ± 0.2028   & 0.2193 ± 0.0035   \\ \midrule
		\multirow{4}{*}{DSSW (linear)}      & 1     & -2.3462 ± 0.0759 & 0.1623 ± 0.0017 \\
		& 10    & -2.5765 ± 0.1029   & 0.1641 ± 0.0020   \\
		& 100   & -2.5083 ± 0.1713   & 0.1788 ± 0.0032   \\
		& 1000  & \textbf{-3.0748 ± 0.0781} $\ddagger$  & 0.2197 ± 0.0054   \\ \midrule
		\multirow{4}{*}{DSSW   (nonlinear)} & 1     & -2.3143 ± 0.1372 & 0.1619 ± 0.0014 \\
		& 10    & -2.5428 ± 0.1604   & 0.1625 ± 0.0014   \\
		& 100   & -2.7383 ± 0.1642   & 0.1780 ± 0.0018   \\
		& 1000  & -3.0664 ± 0.1276 $\ddagger$  & 0.2289 ± 0.0081   \\ \midrule
		\multirow{4}{*}{DSSW   (attention)} & 1     & -2.3463 ± 0.0826 & 0.1638 ± 0.0071 \\
		& 10    & -2.6247 ± 0.1215   & \textbf{0.1606 ± 0.0007} $\ddagger$ \\
		& 100   & -2.7549 ± 0.1024   & 0.1786 ± 0.0023   \\
		& 1000  & -3.0032 ± 0.0880   & 0.2189 ± 0.0039   \\ \bottomrule
	\end{tabular}%
	\caption{SWAE with different regularizations on the MNIST dataset with uniform prior for $d = 3$ averaged over 5 training runs. We use 1 and 10 rotations for RIS3W (1) and RIS3W (10), respectively. We also use the pool size of 1000 for ARI-S3W. Notation "$\ddagger$" indicates that DSSW variants are significantly better than the best baseline method using t-test when the significance level is 0.05.}
	\label{tab:swae_mnist}
\end{table}

\begin{table}[ht]
	\centering
	\begin{tabular}{ccc}
		\toprule
		Method            & $\eta$     & FID $\downarrow$   \\ \midrule
		SW                & 0.001 & 74.2301 ± 2.7727 \\
		SSW               & 0.001 & 75.0354 ± 2.8609 \\
		S3W               & 0.001 & 75.0717 ± 2.6026 \\
		RI-S3W (1)        & 0.001 & 73.5115 ± 3.4937 \\
		RI-S3W (10)       & 0.001 & 70.2262 ± 3.4730 \\
		ARI-S3W (30)      & 0.001 & \underline{69.5562 ± 1.5386} \\ \midrule
		DSSW (exp)       & 1.0 & 69.6907 ± 0.3758 \\
		DSSW (identity)  & 1.0 & 70.0464 ± 0.3136 \\
		DSSW (poly)      & 1.0 & 69.8649 ± 0.0692 \\
		DSSW (linear)    & 1.0 & \textbf{69.1176 ± 0.4021} $\ddagger$ \\
		DSSW (nonlinear) & 1.0 & 71.8204 ± 0.5264 \\
		DSSW (attention) & 1.0 & 70.4635 ± 0.1998 \\ \bottomrule
	\end{tabular}%
	\caption{FID scores for SWAE models on MNIST for $d=3$. We use 1 and 10 rotations for RIS3W (1) and RIS3W (10), respectively. We also use the pool size of 1000 for ARI-S3W. The results of baselines are cited from S3W. Notation "$\ddagger$" indicates that DSSW variants are significantly better than the best baseline method using t-test when the significance level is 0.05.}
	\label{tab:mnist_fid}
\end{table}

\subsection{Self-Supervised Representation Learning}
\textbf{Implementation.} Following the setting of SSW and S3W, we adopt a ResNet18 encoder with 1024 dimensional features that are then projected onto the sphere $\mathbb{S}^{d-1}$ using a last fully connected layer followed by a $\ell_{2}$ normalization on CIFAR10. The models are pretrained for 200 epochs by using minibatch SGD optimizer with a momentum of 0.9 and a weight decay of 0.001, the initial learning rate is 0.05. We adopt a batch size of 512 samples and a standard random augmentation set including random crop, horizontal flipping, color jittering, and gray scale transformation that be common used for self-supervised learning (SSL).

In order to evaluate the performance of representations obtained by the pretained model, we adopt the common linear evaluation protocol that a linear classifier is fitted on top of the pre-trained representations and on the projected features in $\mathbb{S}^{d-1}$, finally the best validation accuracy of the two scenarios is reported. We train linear classifiers for 100 epochs by using the Adam optimizer with a learning rate of 0.001 with a decay of 0.2 at epoch 60 and 80.

Similar to S3W, for the experiment on $\mathbb{S}^{9}$, we use $L=200$ for all distance metrics, we let $\eta=1$ for SW, $\eta=20$ for SSW, $\eta=0.5$ for S3W, RI-S3W, and ARI-S3W, $\eta=100$ for DSSW with the projected energy function of exponential function, linear neural network, nonlinear neural network, and attention mechanism, and $\eta=105$ for DSSW with the projected energy function of identity function and polynomial function.

Similar to SSW, for the experiment on $\mathbb{S}^{2}$, we use $L=10$ for all distance metrics, the regularization coefficient $\eta$ is set as 1 for SW and 20 for SSW, we let $\eta=5$ for DSSW with the projected energy function of exponential function, identity function, polynomial function, and linear neural network. The best validation accuracy of linear evaluation on CIFAR10 for $d=3$ is reported in Table \ref{tab:ssl_cifar10_d3}. From Table \ref{tab:ssl_cifar10_d3} we can observe that DSSW with the projected energy function of linear neural network obtains the best validation accuracy on the two scenarios, all the DSSW variants are superior to all the competitors when projecting the features in $\mathbb{S}^{2}$.

\begin{table}[ht]
	\centering
	\begin{tabular}{cccc}
		\toprule
		$d$                   & Method                             & E $\uparrow$ & $\mathbb{S}^{2}$   $\uparrow$         \\ \midrule
		\multirow{11}{*}{3} & Supervised                         & 82.26 & 81.43 \\
		& hypersphere                        & 60.53   & 55.86 \\
		& SimCLR                             & 66.55   & 59.09 \\
		& SW-SSL ($\eta$=1, L=10)                  & 62.65   & 57.77 \\
		& SSW-SSL ($\eta$=20, L=10)                & 64.89   & 58.91 \\ \cmidrule(lr){2-4} 
		& DSSW-SSL (exp, $\eta$=5, L=10)         & 64.75           & 59.24         \\
		& DSSW-SSL (identity, $\eta$=5, L=10)  & 64.26           & 59.16         \\
		& DSSW-SSL (poly, $\eta$=5, L=10)      & 63.86           & 59.05         \\
		& DSSW-SSL (linear, $\eta$=5, L=10)    & \textbf{65.04}           & \textbf{59.87}         \\
		& DSSW-SSL (nonlinear, $\eta$=6, L=10) & 64.40           & 59.31         \\
		& DSSW-SSL (attention, $\eta$=6, L=10) & 64.41           & 59.14         \\ \bottomrule
	\end{tabular}%
	\caption{Linear evaluation on CIFAR10 for $d=3$. E denotes the encoder output. The results of baselines are cited from SSW.}
	\label{tab:ssl_cifar10_d3}
\end{table}

\subsection{Sliced-Wasserstein Variational Inference}
Sliced-Wasserstein variational inference is optimized via MCMC without optimization or requiring a tractable approximate posterior family. Following the recommendation of SSW and S3W , in this experiment, we replace SW with DSSW and adopt the Geodesic Langevin Algorithm (GLA) with normalization to update the approximate target distribution as follows,

\begin{align}
	\label{eq:swvi}
	x_{t+1}=\frac{x_t-\gamma\left(\nabla V(x_t)-\langle\nabla V(x_t),x_t\rangle x_t\right)+\sqrt{2\gamma}Z}{\|x_t-\gamma\left(\nabla V(x_t)-\langle\nabla V(x_t),x_t\rangle x_t\right)+\sqrt{2\gamma}Z\|_{2}},
\end{align}
where $Z\sim \mathcal{N} \left(0, I \right) $ denotes the Gaussian noise.

\textbf{Implementation.} Similar to SSW and S3W, for 4 von Mises-Fisher distributions, we adopt the Geodesic Langevin Algorithm (GLA) to produce samples that track the target distribution and used to guide our variational model, it is an exponential map normalizing flows $f_{\mu_T} \left ( x; T_{exp} \right )$ with 6 blocks and 5 components. The goal of the exponential map normalizing flows $f_{\mu_T}$ is to transform uniform noise on the sphere into a close approximation of the target distribution that is a mixture of 4 von Mises-Fisher distributions. The parameters of target distributions are initialized as $\mu_{1} =\left (1.5, 0.7+\frac{\pi}{2}\right)$, $\mu_{2} =\left (1.0, -1.0+\frac{\pi}{2}\right)$, $\mu_{3} =\left (5.0, 0.6+\frac{\pi}{2}\right)$, $\mu_{4} =\left (4.0, -0.7+\frac{\pi}{2}\right)$ and $\kappa=10$, these 4 von Mises-Fisher distributions are equally weighted. We run 10000 iterations with 20 GLA steps in each iteration by adopting the Adam optimizer with the learning rate of 0.01 for all considered methods.

\begin{figure*}[ht]
	\centering
	\subfloat[Target Density]{
		\includegraphics[width=0.18\textwidth]{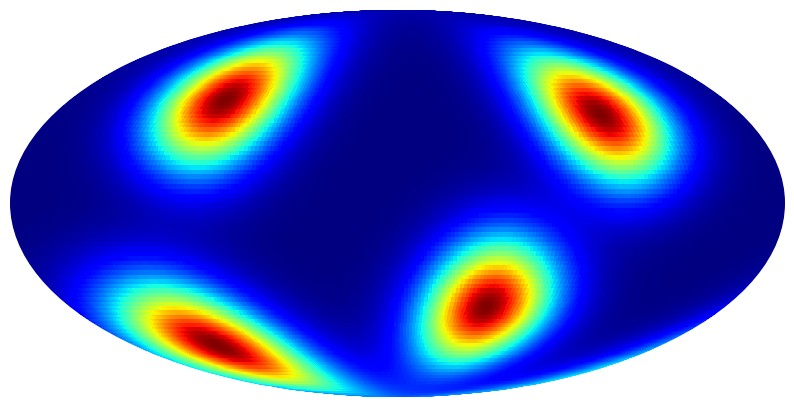}
	}\\ \vspace{-5pt}
	\subfloat[SSW]{
		\includegraphics[width=0.18\textwidth]{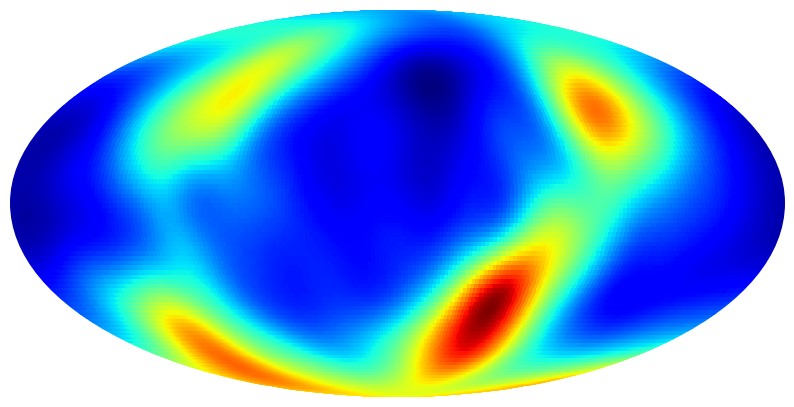}
	}
	\subfloat[S3W]{
		\includegraphics[width=0.18\textwidth]{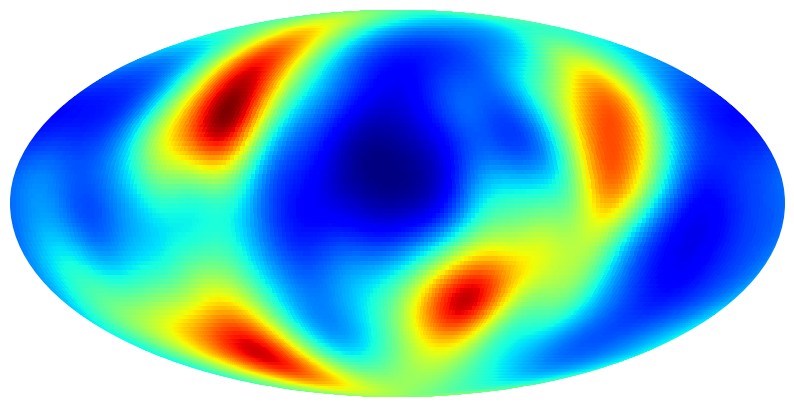}
	}
	\subfloat[RI-S3W (10)]{
		\includegraphics[width=0.18\textwidth]{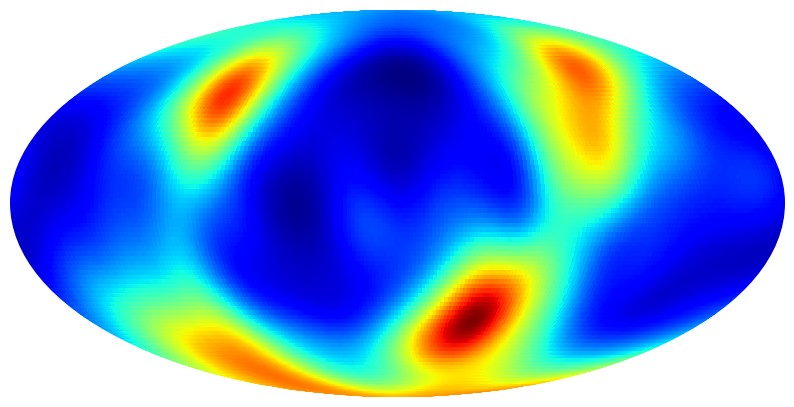}
	}
	\subfloat[ARI-S3W (30)]{
		\includegraphics[width=0.18\textwidth]{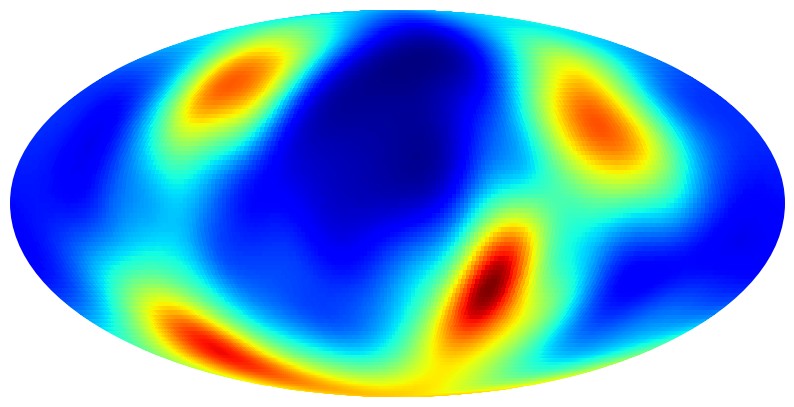}
	}
	\subfloat[DSSW (exp)]{
		\includegraphics[width=0.18\textwidth]{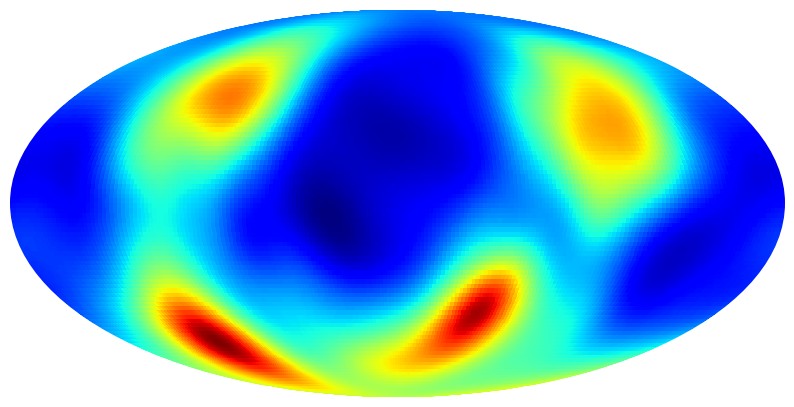}
	}\\ \vspace{-5pt}
	\subfloat[DSSW (identity)]{
		\includegraphics[width=0.18\textwidth]{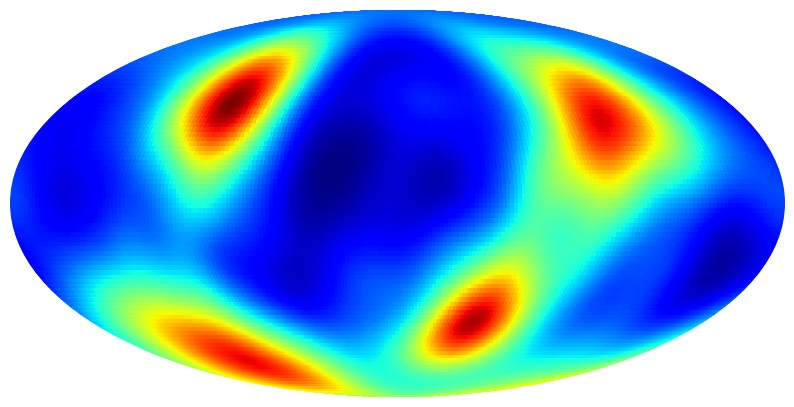}
	}
	\subfloat[DSSW (poly)]{
		\includegraphics[width=0.18\textwidth]{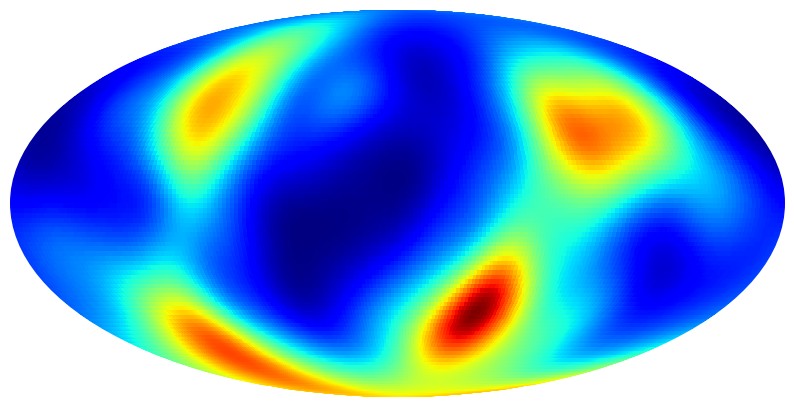}
	}
	\subfloat[DSSW (linear)]{
		\includegraphics[width=0.18\textwidth]{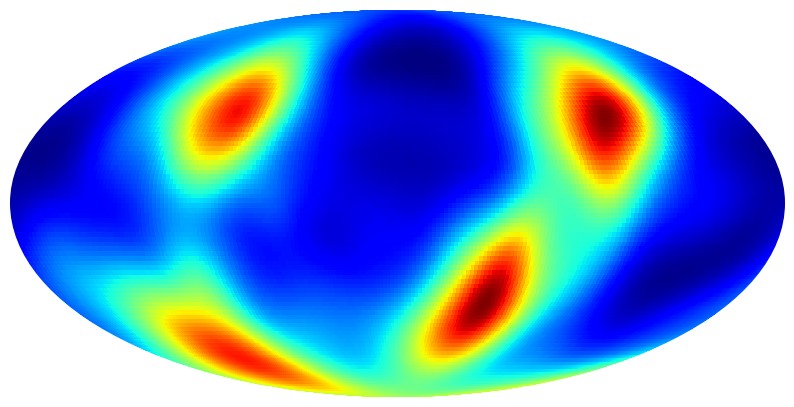}
	}
	\subfloat[DSSW (nonlinear)]{
		\includegraphics[width=0.18\textwidth]{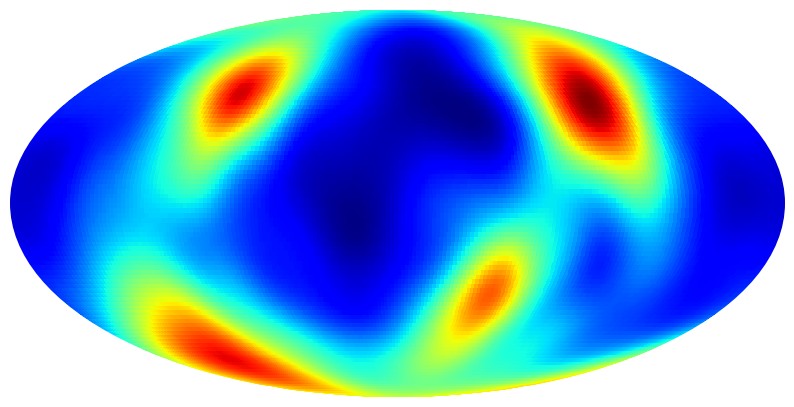}
	}
	\subfloat[DSSW (attention)]{
		\includegraphics[width=0.18\textwidth]{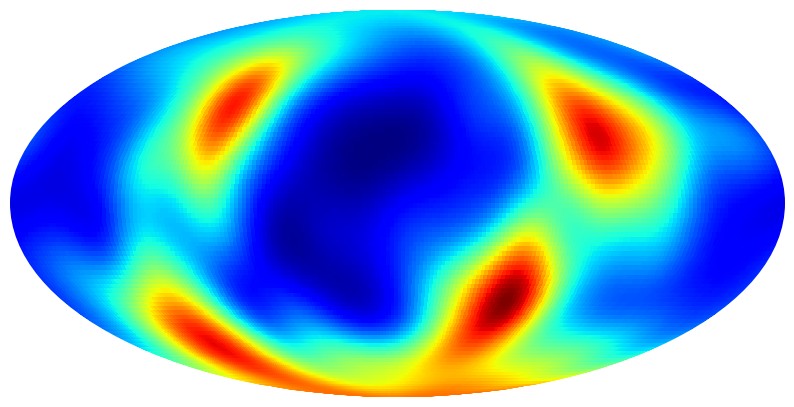}
	}
	\caption{Learning a mixture of 4 von Mises-Fisher distributions. We perform Kernel Density Estimation with the Scott adaptive bandwidth to get the Mollweide projections. We use $L=1000$ projections and 10 rotations for RI-S3W (10), and 30 rotations with the pool size of 1000 for ARI-S3W (30).}
	\label{fig:swvi_vmf}
\end{figure*}

Figure \ref{fig:swvi_vmf} demonstrates the Mollweide projection of the learned density learned by various distances for four von Mises-Fisher distributions. We learn to transform uniform noise on the sphere into a close approximation of the target distribution. We do not have direct access to the density, and we report a kernel density estimate with a Gaussian kernel using the implementation of Scipy.

From Figure \ref{fig:swvi_vmf} we can observe that all the comparison methods have the capacity to match the target distribution closely, but our proposed DSSW obviously outperforms other baselines. It appears that DSSW with the parametric projected energy function is obviously better than SSW, S3W,  RI-S3W, ARI-S3W, and DSSW with the non-parametric projected energy function.

\begin{figure*}[ht]
	\centering
	\subfloat[KL]{
		\includegraphics[width=0.45\textwidth]{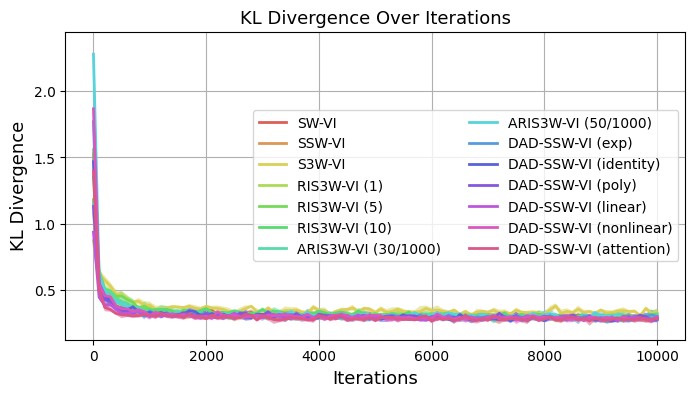}
	}\hspace{5mm}
	\subfloat[ESS]{
		\includegraphics[width=0.45\textwidth]{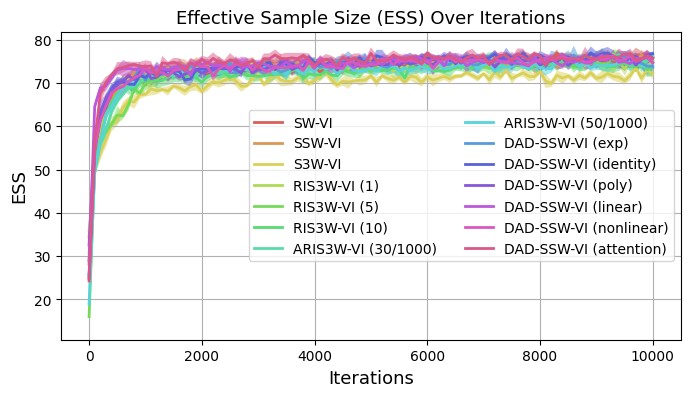}
	}
	\caption{Evolution between the source and target vMFs, 500 samples for each distribution. We use $L=200$ projections for all distance metrics, and use 1, 5, 10 rotations for RIS3W (1), RIS3W (5), and RIS3W (10), respectively. We also use the pool size of 1000 for ARI-S3W.}
	\label{fig:swvi_vmf_kl_ess}
\end{figure*}

Similar to SSW and S3W , we also employ the metric of the KL Divergence and the Effective Sample Size (ESS) to further compare the performance of various distances. The results of KL Divergence and ESS over iterations are demonstrated on Figure \ref{fig:swvi_vmf_kl_ess}. The figure reveals a convergence trend across different distances. The KL Divergence plot indicates that most models with different distances can fast converge to the target distribution. The KL Divergence plot reveals that our DSSW with the parametric projected energy function is better than other comparison methods. The ESS plot demonstrates sampling efficacy. Larger ESS values indicate greater independence between samples. The plot of ESS also indicates that DSSW with the parametric projected energy function is superior to other methods.

\begin{figure}[ht]
	\centering
	\subfloat[Epoch 0]{
		\includegraphics[width=0.25\textwidth]{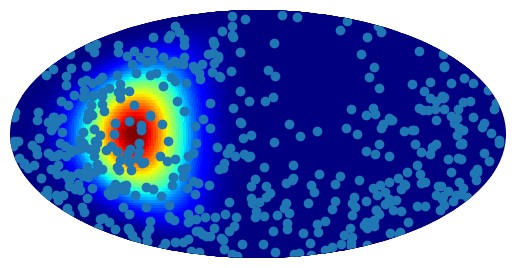}
	}
	\subfloat[Epoch 1000]{
		\includegraphics[width=0.25\textwidth]{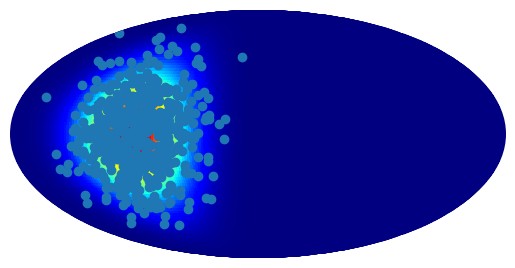}
	}
	\subfloat[Epoch 2000]{
		\includegraphics[width=0.25\textwidth]{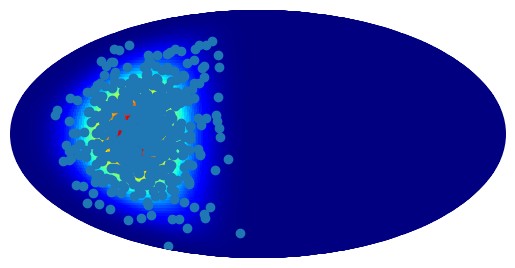}
	}
	\caption{SSW}
	\label{fig:swvi_ps_ssw}
\end{figure}

\begin{figure}[ht]
	\centering
	\subfloat[Epoch 0]{
		\includegraphics[width=0.25\textwidth]{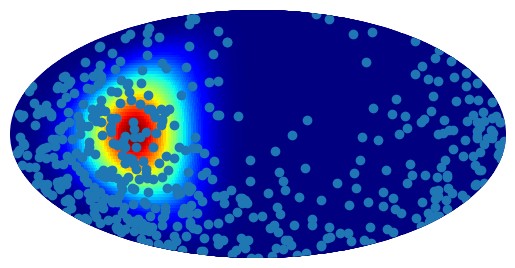}
	}
	\subfloat[Epoch 1000]{
		\includegraphics[width=0.25\textwidth]{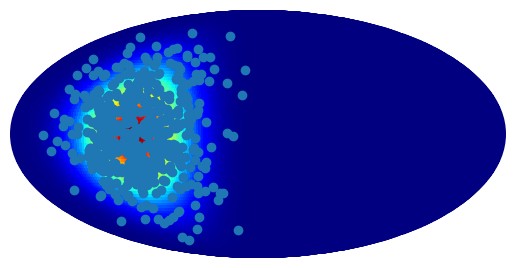}
	}
	\subfloat[Epoch 2000]{
		\includegraphics[width=0.25\textwidth]{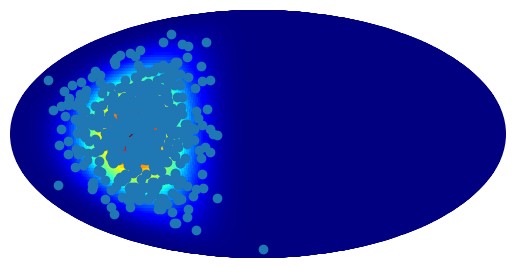}
	}
	\caption{S3W}
	\label{fig:swvi_ps_s3w}
\end{figure}

\begin{figure}[ht]
	\centering
	\subfloat[Epoch 0]{
		\includegraphics[width=0.25\textwidth]{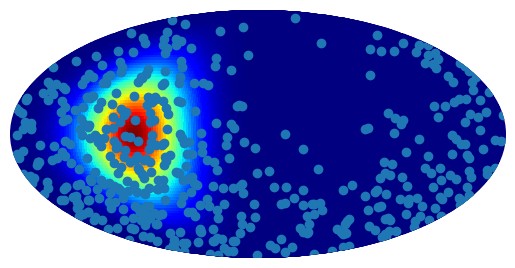}
	}
	\subfloat[Epoch 1000]{
		\includegraphics[width=0.25\textwidth]{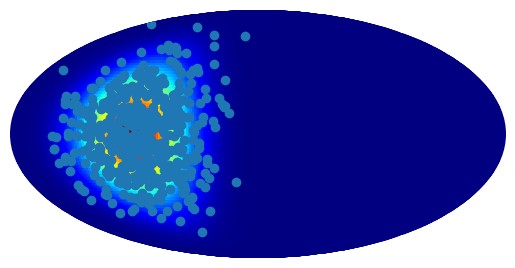}
	}
	\subfloat[Epoch 2000]{
		\includegraphics[width=0.25\textwidth]{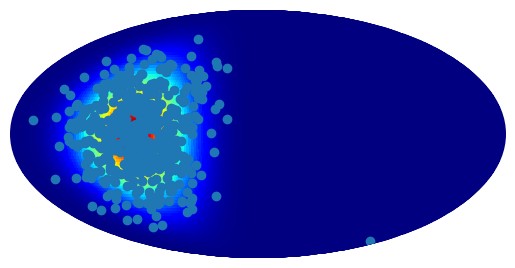}
	}
	\caption{RI-S3W (10)}
	\label{fig:swvi_ps_ri10}
\end{figure}

\begin{figure}[ht]
	\centering
	\subfloat[Epoch 0]{
		\includegraphics[width=0.25\textwidth]{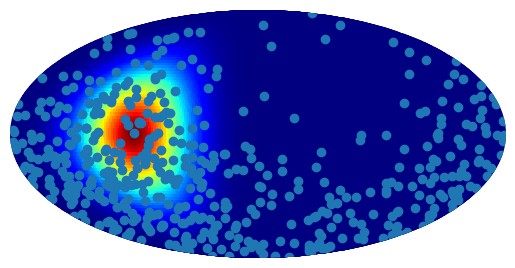}
	}
	\subfloat[Epoch 1000]{
		\includegraphics[width=0.25\textwidth]{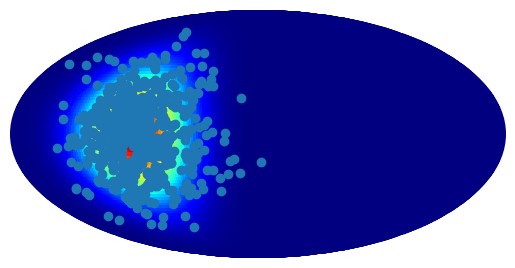}
	}
	\subfloat[Epoch 2000]{
		\includegraphics[width=0.25\textwidth]{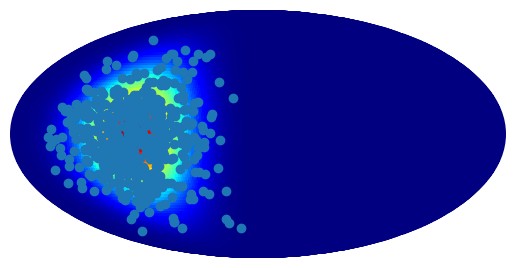}
	}
	\caption{ARI-S3W (10)}
	\label{fig:swvi_ps_ari10}
\end{figure}

\begin{figure}[ht]
	\centering
	\subfloat[Epoch 0]{
		\includegraphics[width=0.25\textwidth]{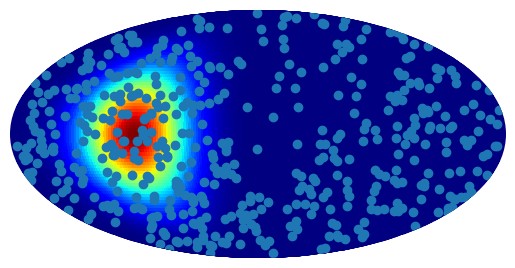}
	}
	\subfloat[Epoch 1000]{
		\includegraphics[width=0.25\textwidth]{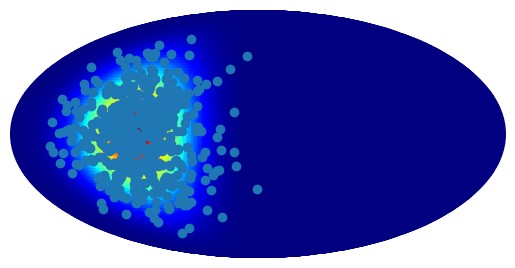}
	}
	\subfloat[Epoch 2000]{
		\includegraphics[width=0.25\textwidth]{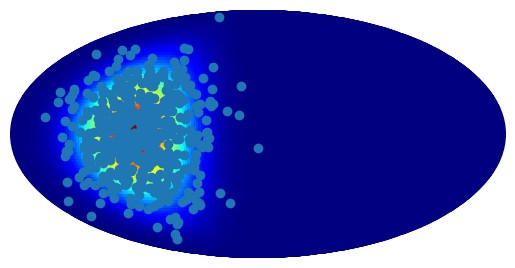}
	}
	\caption{DSSW (exp)}
	\label{fig:swvi_ps_exp}
\end{figure}

\begin{figure}[ht]
	\centering
	\subfloat[Epoch 0]{
		\includegraphics[width=0.25\textwidth]{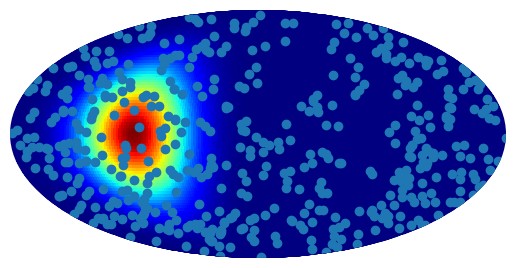}
	}
	\subfloat[Epoch 1000]{
		\includegraphics[width=0.25\textwidth]{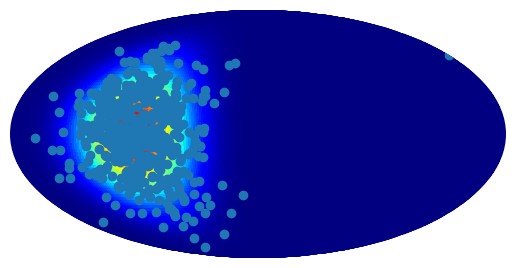}
	}
	\subfloat[Epoch 2000]{
		\includegraphics[width=0.25\textwidth]{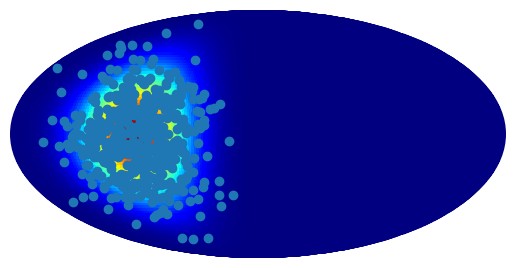}
	}
	\caption{DSSW (identity)}
	\label{fig:swvi_ps_identity}
\end{figure}

\begin{figure}[ht]
	\centering
	\subfloat[Epoch 0]{
		\includegraphics[width=0.25\textwidth]{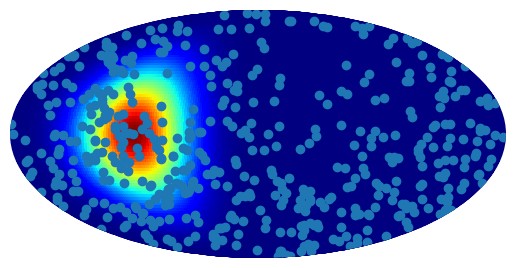}
	}
	\subfloat[Epoch 1000]{
		\includegraphics[width=0.25\textwidth]{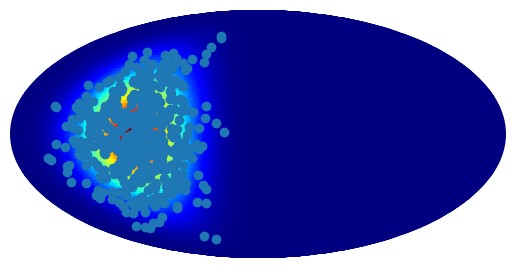}
	}
	\subfloat[Epoch 2000]{
		\includegraphics[width=0.25\textwidth]{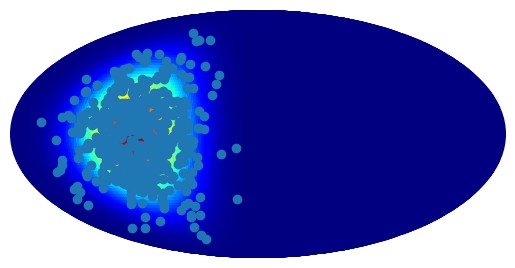}
	}
	\caption{DSSW (poly)}
	\label{fig:swvi_ps_poly}
\end{figure}

\begin{figure}[ht]
	\centering
	\subfloat[Epoch 0]{
		\includegraphics[width=0.25\textwidth]{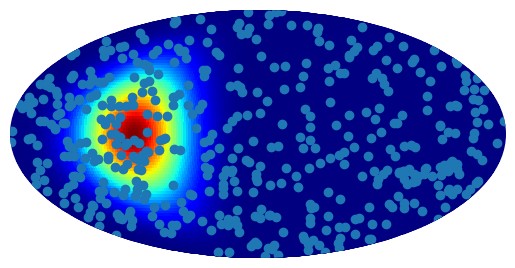}
	}
	\subfloat[Epoch 1000]{
		\includegraphics[width=0.25\textwidth]{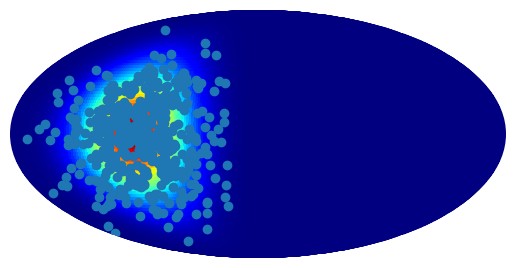}
	}
	\subfloat[Epoch 2000]{
		\includegraphics[width=0.25\textwidth]{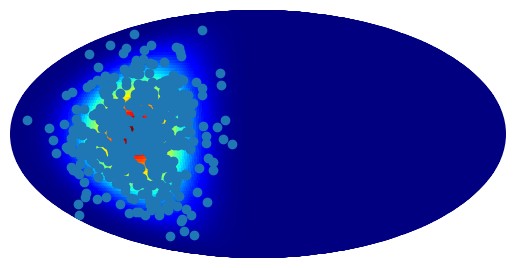}
	}
	\caption{DSSW (linear)}
	\label{fig:swvi_ps_linear}
\end{figure}

\begin{figure}[ht]
	\centering
	\subfloat[Epoch 0]{
		\includegraphics[width=0.25\textwidth]{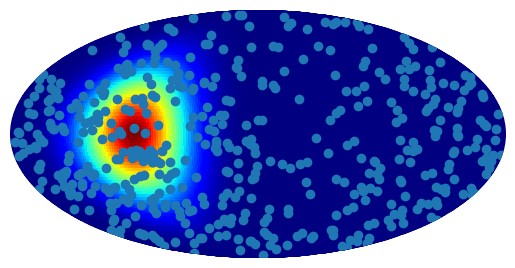}
	}
	\subfloat[Epoch 1000]{
		\includegraphics[width=0.25\textwidth]{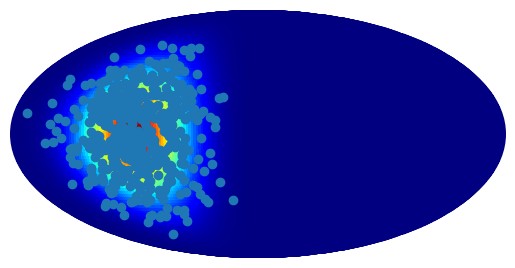}
	}
	\subfloat[Epoch 2000]{
		\includegraphics[width=0.25\textwidth]{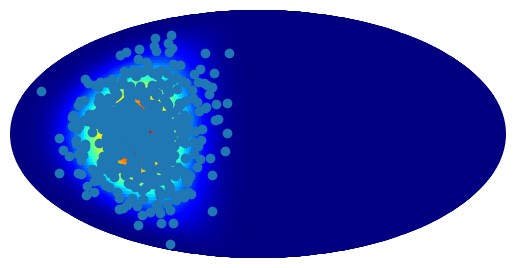}
	}
	\caption{DSSW (nonlinear)}
	\label{fig:swvi_ps_nonlinear}
\end{figure}

\begin{figure}[ht]
	\centering
	\subfloat[Epoch 0]{
		\includegraphics[width=0.25\textwidth]{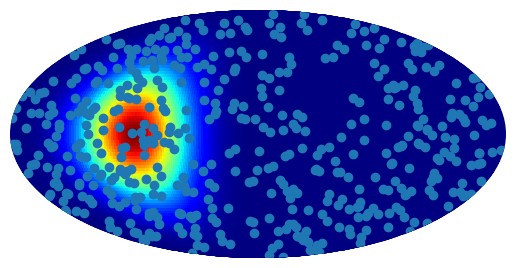}
	}
	\subfloat[Epoch 1000]{
		\includegraphics[width=0.25\textwidth]{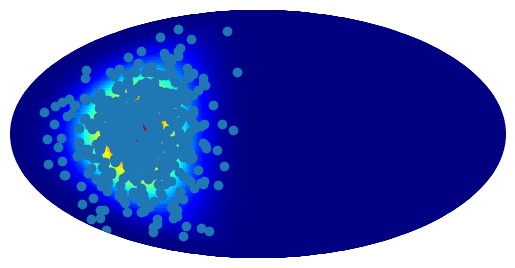}
	}
	\subfloat[Epoch 2000]{
		\includegraphics[width=0.25\textwidth]{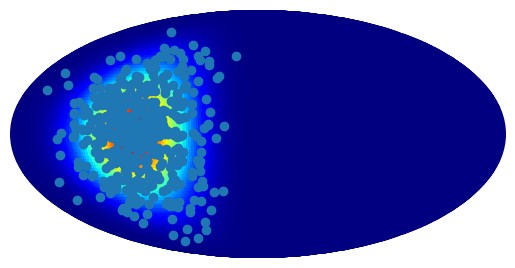}
	}
	\caption{DSSW (attention)}
	\label{fig:swvi_ps_attention}
\end{figure}

For Power Spherical distribution, we initialize its parameters as $\mu=(1, 1, 1)$ and $\kappa=0.1$ for source distribution, and $\mu=(0, 1, 0)$ and $\kappa=10$ for target distribution. We perform 2000 Riemannian Gradient Descent steps and then 20 steps of GLA with the step size of 0.001, 1000 projections, the learning rate of 2. We use $K=2000$ steps with $N=500$ particles. We perform Kernel Density Estimation with the Scott adaptive bandwidth to get the Mollweide projections. The Mollweide projections of the target distribution learned by the considered methods on epoch=0, 1000, 2000 are demonstrated in Figure \ref{fig:swvi_ps_ssw}-\ref{fig:swvi_ps_attention}. From the Mollweide projections visualization plot we can observe that all the considered methods perform comparatively well to learn the target distribution.

\end{document}